\documentclass[runningheads]{llncs}
\usepackage{graphicx}

\usepackage{comment}
\usepackage{amsmath,amssymb} %
\usepackage{color}
\usepackage[font=small,labelfont=bf]{caption}
\usepackage{microtype}

\usepackage[accsupp]{axessibility}  %

\usepackage{subcaption}
\usepackage{float}
\usepackage{lscape}                                         %

\usepackage[lined,ruled,linesnumbered]{algorithm2e}

\usepackage{booktabs}                   %
\usepackage{multirow}

\usepackage{paralist}
\usepackage{enumitem}

\usepackage{bm}                          %
\usepackage{epsfig}                      %
\usepackage{graphicx}                  %
\usepackage{mathtools}
\usepackage{amssymb,amsmath}   %

\usepackage{units}
\usepackage{color}

\usepackage{comment}

\usepackage{url}  %
\usepackage[pagebackref=true,breaklinks=true,letterpaper=true,colorlinks,bookmarks=false,citecolor=blue]{hyperref}

\usepackage{xspace}
\usepackage[table]{xcolor}
\usepackage{setspace}

\def\etal{et~al.}			  %
\def\eg{e.g.,~}               %
\def\ie{i.e.,~}               %

\newlength\paramarginsize
\newlength\figmarginsize
\newlength\secmarginsize
\newlength\figcapmarginsize
\newlength\tabcapmarginsize

\newcommand{\secmargin}{\vspace{\secmarginsize}}

\newcommand{\figcapmargin}{\vspace{\figcapmarginsize}}
\newcommand{\tabcapmargin}{\vspace{\tabcapmarginsize}}

\newcommand{\red}{\textcolor{red}}
\newcommand{\blue}{\textcolor{blue}}

\newcommand{\mpage}[2]
{
\begin{minipage}{#1\linewidth}\centering
#2
\end{minipage}
}

\newcommand{\topic}[1]
{
\vspace{0.9mm}
\noindent \textbf{#1}
}

\newcommand{\figcaption}[2]
{
\caption{
\textbf{#1.}  %
#2            %
}
}

\newcommand{\secref}[1]{Section~\ref{sec:#1}}
\newcommand{\figref}[1]{Figure~\ref{fig:#1}} 
\newcommand{\tabref}[1]{Table~\ref{tab:#1}}
\newcommand{\eqnref}[1]{\eqref{eq:#1}}

\long\def\ignorethis#1{}

\newcommand{\tb}[1]{\textbf{#1}}

\newcommand{\best}[1]{\textbf{\red{#1}}}
\newcommand{\second}[1]{\underline{\blue{#1}}}

\def\xi{\mathbf{x}_i}

\graphicspath{{figure}, {images}, {example}}

\usepackage[width=122mm,left=12mm,paperwidth=146mm,height=193mm,top=12mm,paperheight=217mm]{geometry}

\usepackage{amsmath,amsfonts,bm}

\def\secref#1{section~\ref{#1}}

\def\eqref#1{equation~\ref{#1}}

\def\1{\bm{1}}

\newcommand{\test}{\mathcal{D_{\mathrm{test}}}}

\DeclareMathAlphabet{\mathsfit}{\encodingdefault}{\sfdefault}{m}{sl}
\SetMathAlphabet{\mathsfit}{bold}{\encodingdefault}{\sfdefault}{bx}{n}

\usepackage{url}
\usepackage{xspace}
\usepackage{soul,color}

\begin{document}

\pagestyle{headings}
\mainmatter
\def\ECCVSubNumber{1035}  %

\title{
Learning Instance-Specific Adaptation\\
for Cross-Domain Segmentation
}

\authorrunning{Y. Zou et al.}
\author{
Yuliang Zou$^{1}$
\;
Zizhao Zhang$^{2}$
\;
Chun-Liang Li$^{2}$
\;
Han Zhang$^{3}$
\\
Tomas Pfister$^{2}$
\;
Jia-Bin Huang$^{4}$
\\
\institute{$^{1}$Virginia Tech\; $^{2}$Google Cloud AI\\ $^{3}$Google Brain\;$^{4}$University of Maryland, College Park}
}

\maketitle
\begin{center}
\centering

\mpage{0.01}{\rotatebox[origin=c]{90}{Cross-domain}}\hfill
\mpage{0.01}{\rotatebox[origin=c]{90}{segmentation}}\hfill
\mpage{0.93}{
\includegraphics[width=\linewidth]{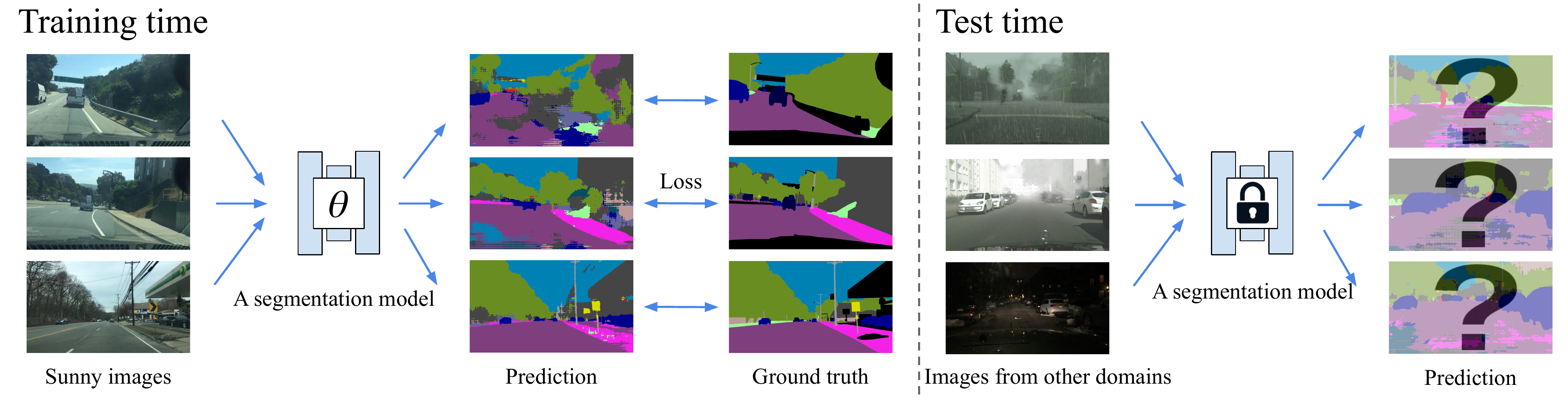}
}
\\
\vspace{-4mm}

\mpage{0.01}{\rotatebox[origin=c]{90}{Semantic}}\hfill
\mpage{0.01}{\rotatebox[origin=c]{90}{(syn$\rightarrow$real)}}\hfill
\mpage{0.93}{
\mpage{0.32}{
\includegraphics[width=\linewidth]{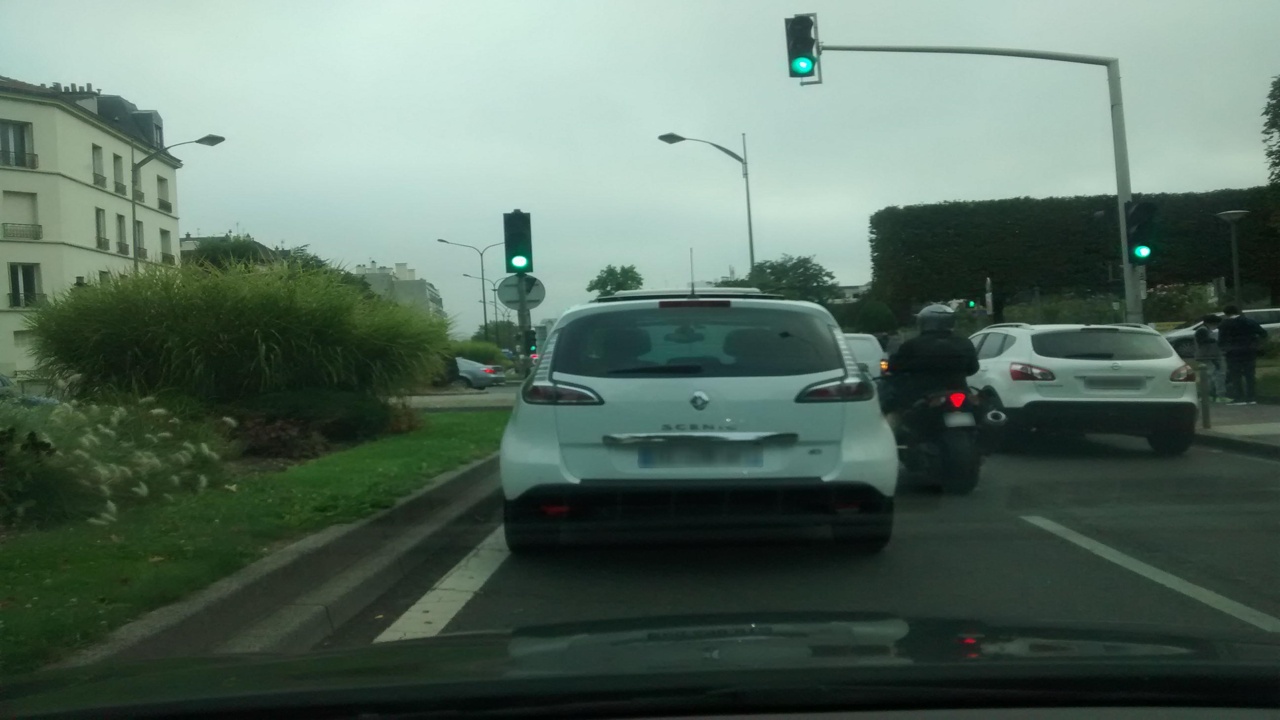}
}\hfill
\mpage{0.32}{
\includegraphics[width=\linewidth]{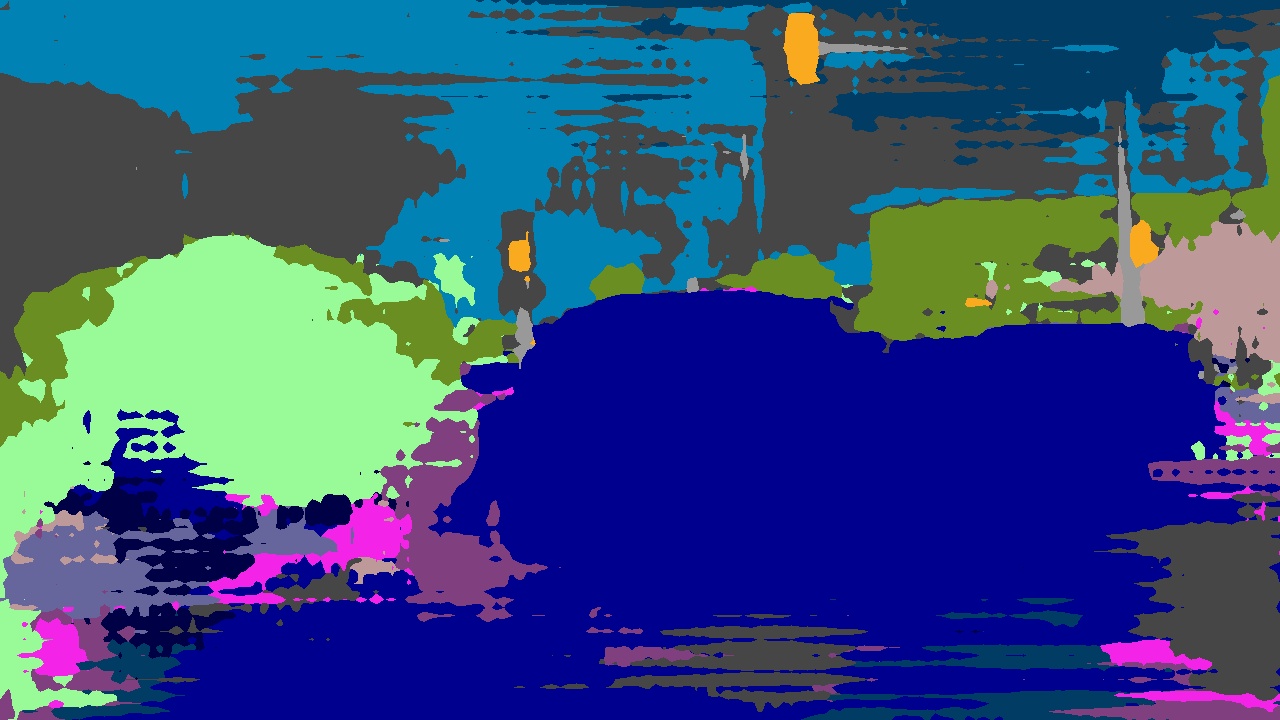}
}\hfill
\mpage{0.32}{
\includegraphics[width=\linewidth]{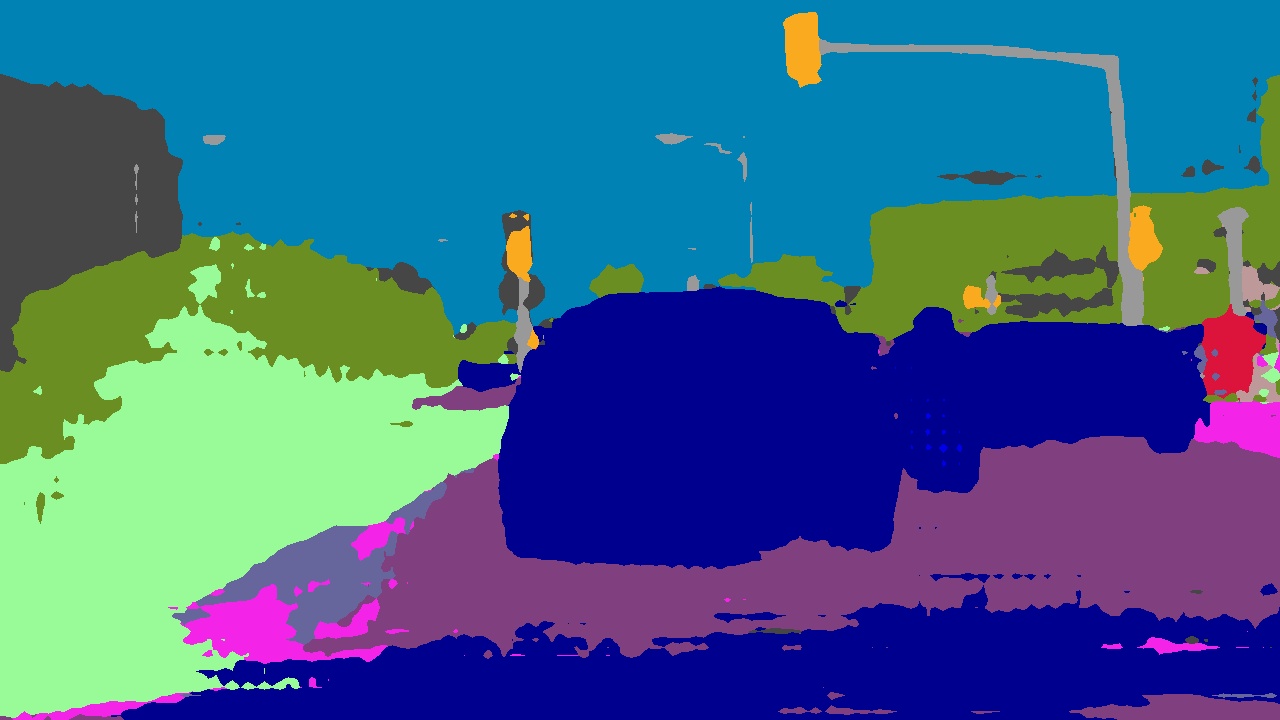}
}\hfill
\\
}
\vspace{-3mm}

\mpage{0.01}{\rotatebox[origin=l]{90}{Panoptic}}\hfill
\mpage{0.01}{\rotatebox[origin=l]{90}{(sun$\rightarrow$fog)}}\hfill
\mpage{0.93}{
\mpage{0.32}{
\includegraphics[width=\linewidth]{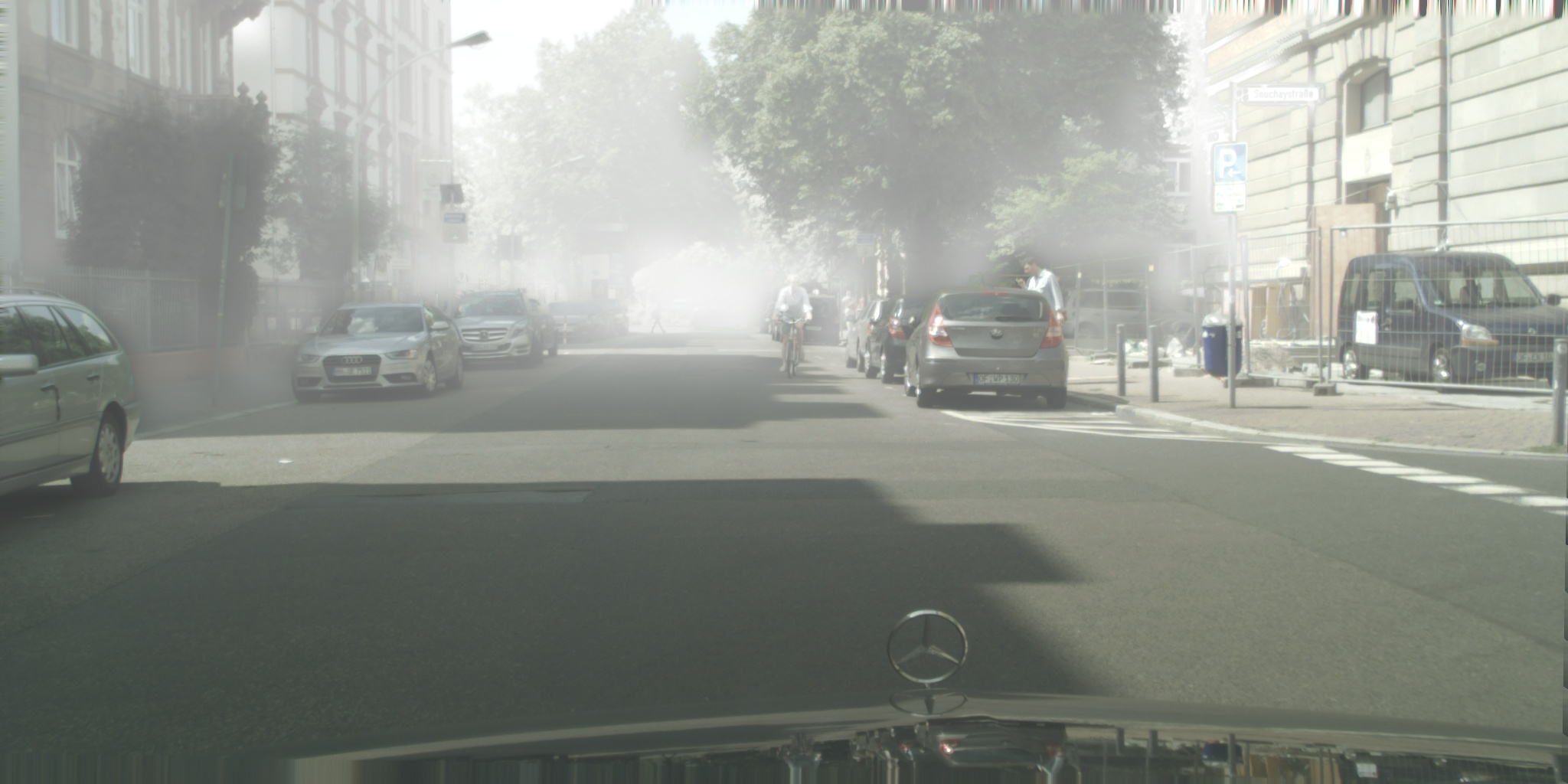}
}\hfill
\mpage{0.32}{
\includegraphics[width=\linewidth]{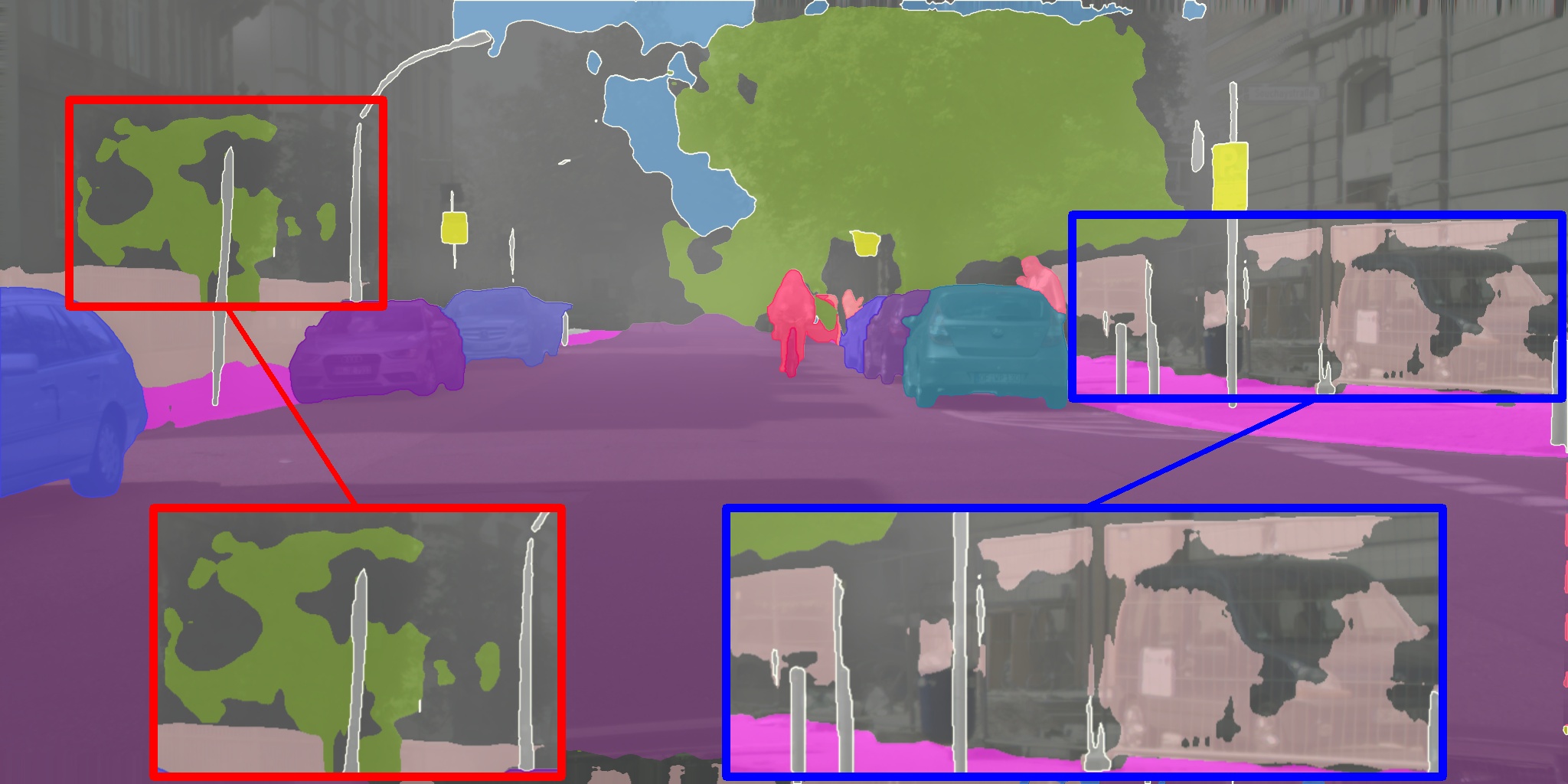}
}\hfill
\mpage{0.32}{
\includegraphics[width=\linewidth]{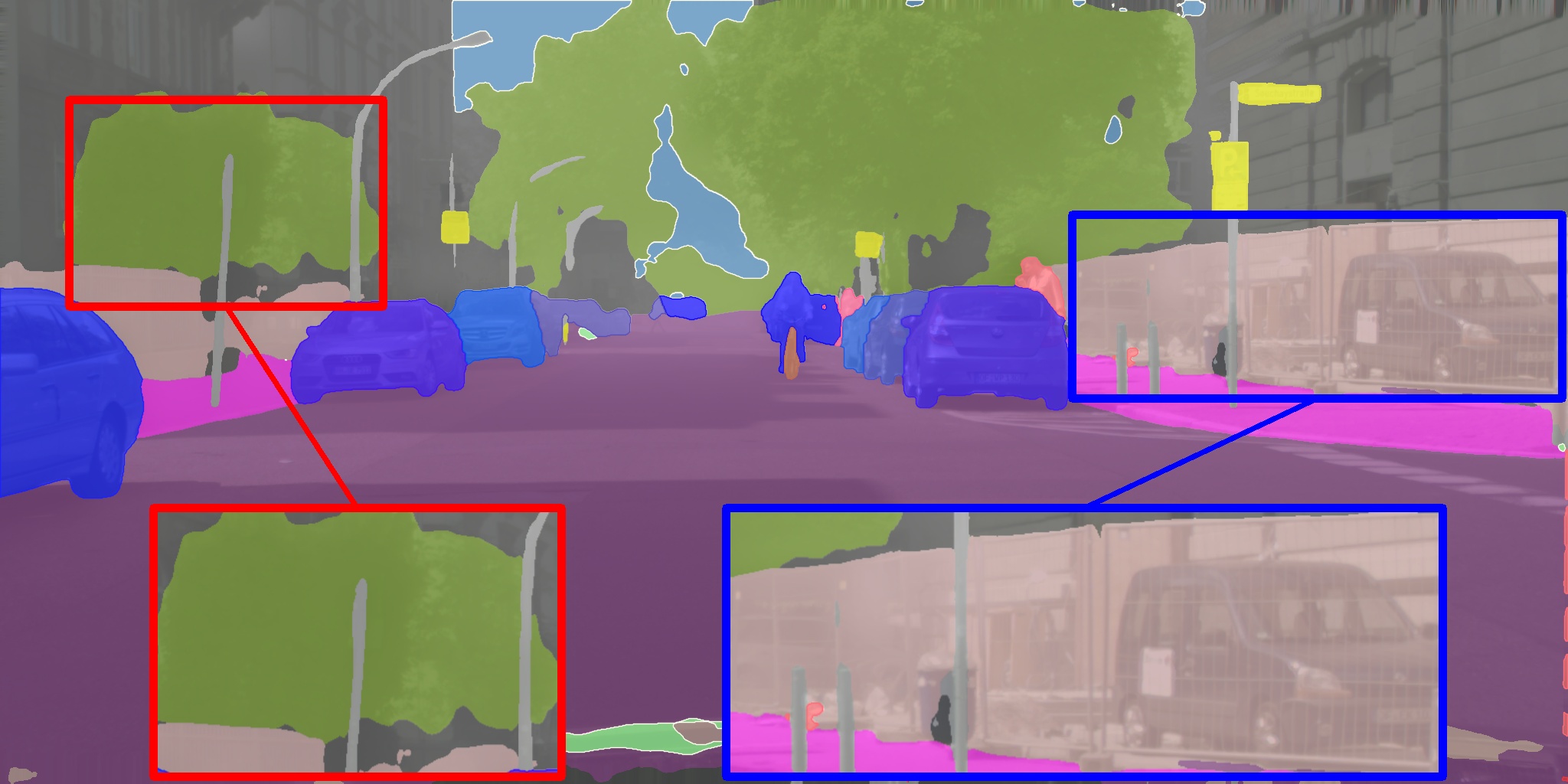}
}\hfill
\\
}
\vspace{-3.5mm}

\hfill
\mpage{0.94}{

\mpage{0.32}{Target domain input}\hfill
\mpage{0.32}{Pre-trained}\hfill
\mpage{0.32}{Ours}\hfill
}
\\

\vspace{-5mm}
\captionof{figure}
{\textbf{Cross-domain segmentation.}
Models trained with the standard recipe on source domain data perform poorly on unseen target domains. On the contrary, the proposed method significantly improves upon off-the-shelf pre-trained models, without accessing the target domain at training time or parameter optimization at test-time.}
\label{fig:teaser}
\end{center}

\begin{abstract}

We propose a test-time adaptation method for cross-domain image segmentation. Our method is simple: Given a new unseen instance at test time, we adapt a pre-trained model by conducting instance-specific BatchNorm (statistics) calibration.
Our approach has two core components.
First, we replace the manually designed BatchNorm calibration rule with a learnable module. 
Second, we leverage strong data augmentation to simulate random domain shifts for learning the calibration rule.
In contrast to existing domain adaptation methods, our method does not require accessing the target domain data at training time or conducting computationally expensive test-time model training/optimization.
Equipping our method with models trained by standard recipes achieves significant improvement, comparing favorably with several state-of-the-art domain generalization and one-shot unsupervised domain adaptation approaches.
Combining our method with the domain generalization methods further improves performance, reaching a new state of the art.

\end{abstract}

\section{Introduction}
\label{sec:intro}
\secmargin

Deep neural networks have shown impressive results in many computer vision applications.
However, these models suffer from inevitable performance drops when deployed in out-of-distribution environments due to domain shift~\cite{ben2010theory}.
For example, segmentation models trained on sunny images may perform poorly on foggy or rainy scenes~\cite{choi2021robustnet}.
Improving the cross-domain performance of deep vision models has thus received considerable attention in recent years.

\topic{Domain adaptation.} One straightforward approach for reducing the domain shift is to collect diverse labeled data in the target domain of interest for supervised fine-tuning.
However, collecting sufficient annotated data in the target domain could be expensive or infeasible (e.g., in continuously changing environments).
This is particularly challenging for many dense prediction tasks such as image segmentation as it requires \emph{dense} (pixel-wise) labels.
Unsupervised domain adaptation (UDA)~\cite{ganin2016domain,luo2019taking,tzeng2017adversarial,vu2019advent} is an alternative route for reducing the domain gap by using \emph{unlabeled} target data.
However, UDA methods require accessing target domain data for model training \emph{before} deployment.
Such assumptions may not hold as we are not able to anticipate what scenarios the model would encounter (e.g., different weather conditions) and therefore cannot collect the unlabeled data accordingly. 
One-shot UDA~\cite{benaim2018one,luo2020adversarial} relaxes the constraint by requiring only one target example for model training.
The model can thus use the first example encountered in the unseen target domain as the training example.
However, the adaptation procedure often requires thousands of training steps~\cite{luo2020adversarial}, hindering its applicability as a plug-and-play module when deployed on new target domains.
These UDA methods also require access to source data \emph{during} the adaptation process, which may be unrealistic at test time.

\topic{Domain generalization (DG)} ~\cite{balaji2018metareg,Li_2019_ICCV,matsuura2020domain,zhao2020domain} overcomes the above limitations by learning invariant representations using multiple source domains to improve model robustness on unseen or continuously changing environments.
Recent approaches~\cite{qiao2020learning,zhao2020maximum} relax the constraint by requiring one single source domain only.
However, as Dubey~\etal~\cite{dubey2021adaptive} points out, the optimal model learned from training domains may be far from being optimal for an unseen target domain.

\topic{Test-time adaptation} approaches have been proposed to tackle exactly the same problem.
These methods can be roughly categorized into two groups: 
1) optimizing model parameters at test time with a proxy task~\cite{bartler2021mt3,sun2020test}, prediction pseudo-label~\cite{liang2020we}, or  entropy regularizations~\cite{wang2021tent}, and
2) BatchNorm calibration~\cite{hu2021mixnorm,nado2020evaluating,schneider2020improving}.
These approaches can be applied to update models along with observing each target test data, thus observing the entire test distribution.
Alternatively, they can be used to create an \emph{instance-specific} model for each test example individually.
Despite the flexibility, the optimization-based methods all require time-consuming backprop computation to update model parameters.

\topic{Our work.} In this paper, we present a simple test-time adaptation method for cross-domain segmentation (\figref{teaser}).
Building upon BatchNorm calibration methods~\cite{nado2020evaluating,schneider2020improving}, we propose to learn \emph{instance-specific} calibration rules using strong data augmentations to simulate various pseudo source domains.
Our approach offers several advantages.
First, compared with existing work~\cite{nado2020evaluating,schneider2020improving} with \emph{manually determined} calibration rules that require time-consuming grid searches and may not transfer to different models, our approach is \emph{data-driven} and \emph{instance-specific}.
Second, unlike other test-time adaptation methods~\cite{bartler2021mt3,liang2020we,sun2020test,wang2021tent}, our work does not involve expensive gradient-based optimization for updating model parameters at test time. 
Third, our method learns to calibrate BatchNorm statistics with \emph{one} single instance (i.e., without accessing to a batch of samples).\footnote{As we validated in \tabref{analysis}, even examples come for the same test distribution, batch-wise calibration may be sub-optimal when the batch size is small.}
We validate our proposed methods on cross-domain semantic and panoptic segmentation tasks on several benchmarks. 
Our experiments show a sizable boost over existing adaptation methods.

\topic{Contributions.} In summary, we make the following contributions:
\begin{itemize}
\item We propose a simple instance-specific test-time adaptation method and show its applicability to off-the-shelf segmentation models containing BatchNorm.
\item We conduct a detailed ablation study and analysis to validate our design choices in
semantic segmentation.
Applying the optimal configuration to the more complex panoptic segmentation task leads to promising performance.
\item When combined with the models pre-trained by standard recipes, our method compares favorably with state-of-the-art one-shot UDA methods and 
domain generalizing semantic segmentation methods. 
Our approach can also be combined with existing DG methods to improve the performance further.
\end{itemize}

\section{Related Work}
\label{sec:related}
\secmargin

\topic{Domain adaptation.}
Models trained on one (source) domain often suffers from a severe performance drop when processing samples from unseen (target) domains.
Domain adaptation methods aim to mitigate this issue by adapting a pre-trained model using samples from target domains.
Unsupervised domain adaptation (UDA) methods show promising results by leveraging \emph{unlabeled} target data.
These UDA techniques include 1) domain invariant learning, 2) generative models, and 3) self-training.
Domain invariant learning methods learn invariant features for the source and target domains by 
imposing an adversarial loss~\cite{ganin2016domain,luo2019taking,tzeng2017adversarial,vu2019advent}, 
minimizing the domain distribution distance (e.g., MMD)~\cite{long2015learning,tzeng2014deep} or 
correlation distance~\cite{sun2016deep}.
Applying data augmentation with generative models can also reduce domain gap using image-to-image translation~\cite{bousmalis2017unsupervised,shrivastava2017learning,zhu2017unpaired}, style transfer~\cite{dundar2018domain}, or hybrid methods that integrate with domain invariance learning methods~\cite{chen2019crdoco,hoffman2018cycada}.
Self-training methods~\cite{zou2018unsupervised,zou2019confidence} select confident/reliable target data predictions and convert them into pseudo labels. 
These methods then iterate the fine-tuning and pseudo-labeling procedures until convergence.

While we have observed remarkable progress in UDA, pre-collected target domain data requirement makes it less practical.
Recently, one-shot UDA methods~\cite{benaim2018one,luo2020adversarial} have been proposed to tackle this problem.
Instead of training on \emph{many} unlabeled target data, these approaches require only \emph{one} unlabeled target data.
However, these methods require time-consuming offline training before deploying on the target domain.
In contrast, our proposed method efficiently adapts the model by calibrating BatchNorm on each target example \emph{on the fly}, without offline training on each target domain separately.  
Models trained with our method can be easily applied to many different unseen domains.

\topic{Domain generalization.}
Instead of adapting models to using target domain data, domain generalization~\cite{balaji2018metareg,Li_2019_ICCV,matsuura2020domain,zhao2020domain} aims to train a model on source domains that are \emph{generalizable} to unseen target domains by encouraging the networks to learn domain-invariant representations.
However, these approaches require multiple source domains for training, which poses additional challenges in (labeled) data collection and restricts their feasibility in practical usage.
To mitigate the data collection issues, single domain generalization~\cite{qiao2020learning} trains models on one single source domain only by either exploiting strong data augmentation strategies~\cite{qiao2020learning,volpi2018generalizing,zhao2020maximum} to diversify the source domain training data, or performing feature whitening operations or normalization~\cite{choi2021robustnet,fan2021adversarially,huang2019iterative,pan2019switchable} to remove domain-specific information during training.
Similar to these methods, our method also exploits data augmentation to diversify one single source domain data.
However, instead of enforcing models to learn domain-invariant features, we encourage models to calibrate BatchNorm on \emph{each} unseen target data at test time, by training them on diverse pseudo domains generated with strong data augmentation strategies in training time.
Our proposed method can also complement (single) domain generalization approaches to improve the performance further.

\topic{Test-time adaptation.}
Depending on the use of online training/optimization, test-time adaptation methods can be divided into two groups.
First, \emph{optimization-free} test-time adaptation methods mostly focus on calibrating the running statistics inside BatchNorm layers~\cite{hu2021mixnorm,khurana2021sita,mirza2021norm,nado2020evaluating,schneider2020improving} because these feature statistics carry domain-specific information~\cite{li2016revisiting}.
However, these methods either directly replace the running statistics with current input batch statistics or mix the running statistics and current input batch statistics with a \emph{pre-defined} calibration rule.
In contrast, we propose to learn the instance-specific BatchNorm calibration rule from source domain data.
Second, \emph{test-time optimization} methods adapt the model parameters using a training objective such as entropy minimization~\cite{wang2021tent}, pseudo-labeling~\cite{liang2020we}, or self-supervised proxy tasks~\cite{cohen2020self,sun2020test}.
Our experiments show that integrating our method with
test-time optimization
boosts performance.

\section{Learning Instance-Specific BatchNorm Calibration}
\label{sec:method}
\secmargin
Our method applies to off-the-shelf pre-trained segmentation models containing BatchNorm layers~\cite{ioffe2015batch}, a reasonable assumption in most modern CNN models.
In \secref{sec:pre}, we first review BatchNorm and recent test-time calibration techniques.
We then introduce our method (unconditional and conditional BatchNorm calibration) in \secref{sec:learnable}.

\begin{figure*}[t]

\includegraphics[width=\linewidth]{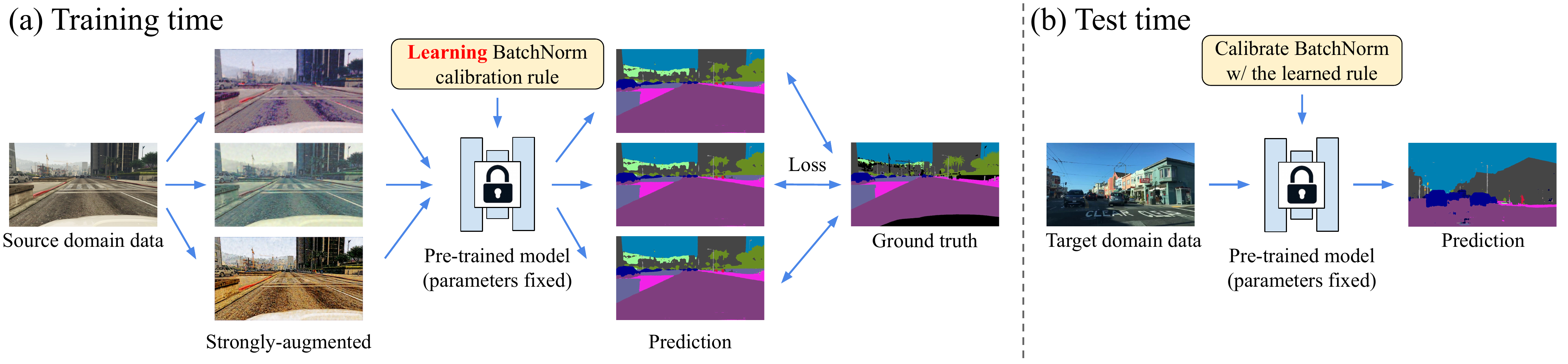}

\figcapmargin
\vspace{-.2cm}
\figcaption{Overview}
{
(a) At training time, we learn the BatchNorm calibration rule (\eqnref{norm}) by training only the newly initialized parameters on the strongly-augmented source domain data;
(b) At test time, we conduct instance-specific BatchNorm calibration using the learned calibration rule.
Note that our method does not perform test-time training or optimization, and thus the model parameters are fixed after training.
}

\label{fig:overview}
\end{figure*}

\subsection{Background}
\label{sec:pre}
\vspace{-2pt}
\topic{A brief review of BatchNorm.}
BatchNorm has been empirically shown to stabilize model training and improve model convergence speed~\cite{wu2021rethinking}, making it an essential component in most modern CNN models.
The inputs to each BatchNorm layer are CNN features $x \in \mathbb{R}^{B \times C \times H \times W}$, where $B$ denotes the batch size, $C$ denotes the number of feature channels, $H \times W$ denotes the spatial size.
BatchNorm conducts normalization followed by an affine transformation on the inputs $x$ to get outputs $y \in \mathbb{R}^{B \times C \times H \times W}$
\begin{equation}
    y = \frac{x-\mu}{\sqrt{\sigma^2+\epsilon}} \times \gamma + \beta,
\end{equation}
where $\epsilon$ is a small constant for numerical stability, $\gamma \in \mathbb{R}^C$ and $\beta \in \mathbb{R}^C$ are learnable parameters. 
We generalize the tensor operations by assuming broadcasting when the dimensions are not exactly the same.
Note that the definition of $\mu$ and $\sigma^2$ differs in training and test time.
In training time, $\mu$ and $\sigma^2$ are set as the  \emph{input batch statistics}.\footnote{Following the ``NamedTensor'' practice~\cite{named_tensor}, this computes the statistics over the B, H, W dimensions and return vectors with dimension C.}
\begin{align}
    \mu_{B} &= \text{mean}(x, \text{axis}=(B, H, W)) \\
    \sigma_{B}^2 &= \text{var}(x, \text{axis}=(B, H, W))
\end{align}

At test time, $\mu$ and $\sigma^2$ are set to \emph{population statistics} $\mu_{pop}, \sigma^2_{pop}$, accumulated in training time using exponential moving averaging
\begin{align}
    \label{eq:mu}
    \mu_{pop, t} &= (1-\alpha)\times\mu_{pop, t-1} + \alpha\times\mu_{B} \\
    \label{eq:sigma}
    \sigma_{pop, t}^2 &= (1-\alpha)\times\sigma^2_{pop, t-1} + \alpha\times\sigma^2_{B}
\end{align}
where $\alpha$ is a scalar called momentum, and the default value is 0.1 (in PyTorch convention).
Note that the population statistics update happens during training in every feed-forward step.

\topic{Manual BatchNorm calibration.}
Despite the empirical success in in-domain testing, models with BatchNorm layers suffer from a significant performance drop when testing on out-of-distribution data.
One potential reason is that the population statistics within the BatchNorm layers carry \emph{domain-specific} information~\cite{li2016revisiting}, and thus these statistics are not suitable for normalizing inputs from a different domain.
Recent studies~\cite{li2016revisiting,schneider2020improving,wang2021tent} show that, by
calibrating the population statistics with input statistics,
the cross-domain performance can be significantly improved:
\begin{equation}
    y = \frac{x-\left((1-m)\times\mu_{pop} + m\times\mu_{ins}\right)}{\sqrt{\left((1-m)\times\sigma^2_{pop} + m\times\sigma^2_{ins}\right)+\epsilon}} \times \gamma + \beta,
\end{equation}
where $m$ indicates \emph{calibration strength}, each method has a different empirically specified value, $\mu_{ins} \in \mathbb{R}^{B \times C}$ and $\sigma^2_{ins} \in \mathbb{R}^{B \times C}$ indicate instance mean and variance.
Note that the above calibration step happens in every BatchNorm layer, and thus the input features in later BatchNorm layers will be increasingly more calibrated.

In our study (\tabref{abl}(a)), we show that simply setting calibration strength $m$ as the default momentum value 0.1 can improve overall cross-domain performance.
However, we find several potential issues.
First, the calibration strength $m$ is specified \emph{empirically}, requiring a grid search to obtain the optimal value. 
Nevertheless, the optimal value for one setting might not well transfer to other settings with different pre-trained models or target domains of interest. 
Second, the calibration strength $m$ is a scalar. 
However, different feature channels may encode different semantic information~\cite{yosinski2015understanding}. 
Therefore, we may use different calibration strengths for different feature channels.
Third, calibrating the mean and variance with the same strength leads to sub-optimal results.

\subsection{The proposed method}
\label{sec:learnable}
\topic{Learning to calibrate BatchNorm (InstCal-U).}
To address the aforementioned issues, we propose to learn the calibration strengths \emph{during training} instead of manually specifying them at test time.
For simplicity, we define the following function
\begin{equation}
    f_c(a, b, \mathbf{m}) = (\vec{1}-\mathbf{m}) \times a + \mathbf{m} \times b
\end{equation}
The proposed calibration and normalization process can thus be written as
\begin{equation}
\label{eq:norm}
    y = \frac{x-f_c\left( \mu_{pop}, \mu_{ins}, \mathbf{m}_{\mu} \right)}{\sqrt{f_c\left( \sigma^2_{pop}, \sigma^2_{ins}, \mathbf{m}_{\sigma} \right)+\epsilon}} \times \gamma + \beta,
\end{equation}
where $\mathbf{m}_{\mu} \in \mathbb{R}^C$ and $\mathbf{m}_{\sigma} \in \mathbb{R}^C$ are two learnable parameters.
More specifically, we initialize two learnable parameters $\mathbf{m}_{\mu}$ and $\mathbf{m}_{\sigma}$ with the default momentum value 0.1.
We provide a detailed PyTorch implementation of this module in the supplementary material.

Given an off-the-shelf model, we convert all BatchNorm layers into the instance-specific calibrated format in \eqnref{norm}.
Note that we only train the newly initialized calibration parameters $\mathbf{m}_{\mu}$ and $\mathbf{m}_{\sigma}$, and we keep the other learnable parameters (including $\gamma$ and $\beta$) fixed.

Using training data in the source domain, we train parameters $\mathbf{m}_{\mu}$ and $\mathbf{m}_{\sigma}$ on a diverse set of domains.
Our intuition is that, by exposing the model to diverse (simulated) domains, we implicitly constrain the learnable calibration parameters $\mathbf{m}_{\mu}$ and $\mathbf{m}_{\sigma}$ to be robust and invariant to unseen target domains.
However, since we only have one single source domain, we need to generate multiple pseudo domains based on the source domain.
Instead of adopting complex generative models to generate pseudo domains, we find that applying appropriate strong data augmentation during training leads to promising results.
We explore three different augmentation strategies: RandAugment~\cite{cubuk2020randaugment}, AugMix~\cite{hendrycks2020augmix}, and DeepAugment~\cite{hendrycks2021many}, and empirically find that DeepAugment performs the best (\tabref{abl}(b)).
The details of augmentations are in supplementary materials.

\topic{Learning to conditionally calibrate BatchNorm (InstCal-C).}
While the goal is to learn the BatchNorm calibration parameters so that the models can adapt to unseen domains at test time, the learnable parameters $\mathbf{m}_{\mu}$ and $\mathbf{m}_{\sigma}$ are \emph{fixed} after training.
We propose an optional module to enable \emph{conditional} calibration to increase the flexibility.

Instead of directly learning the parameters $\mathbf{m}_{\mu}$ and $\mathbf{m}_{\sigma}$, we propose to learn a set of parameters $\mathbf{m}_{\mu, i}$ and $\mathbf{m}_{\sigma, i}$ for mean and variance, respectively.
These parameters can be viewed as the \emph{basis} of calibration rules. 
We will use two lightweight MLPs (one for each statistic) to predict the coefficients to combine the basis to get the actual calibration strength for each test example, given the concatenation of instance and population statistics.
Take $\mathbf{m}_\mu$ as an example, the computation step can be written as follows
\begin{align}
\label{eq:K}
    \{c_{\mu,i}\}_1^K &= \textbf{Softmax} \left( g_{\mu}\left( \textbf{Concat}( \mu_{pop}, \mu_{ins}) \right) \right) \\
    \mathbf{m}_{\mu} &= \sum_i^K c_{\mu,i}  \mathbf{m}_{\mu,i}
\end{align}
where $c_{\mu,i}$ is a scalar, $\textbf{Concat}(\cdot)$ is the channel-wise concatenation operation, and $g_\mu(\cdot)$ is a small 2-layer MLP.
The computation of $\mathbf{m}_\sigma$ is similar.
We provide a detailed PyTorch implementation of this module in the supplementary material.

Thanks to the redesign, we further increase the learnable instance-specific BatchNorm calibration flexibility by setting the calibration rule to be conditional on the input features. 
As a result, the calibration rule is now \emph{dynamically changing} according to different test target samples, while the inference process is still done within one forward pass.
As shown in \secref{sec:results}, in general, the performance of instance-specific calibration (InstCal-C) improves upon the unconditional calibration (InstCal-U) on synthetic-to-real settings where significant domain shifts exist.

\section{Experimental Results}
\label{sec:results}
\secmargin

We mainly validate and analyze our method using the semantic segmentation tasks.
In~\secref{sec:panoptic}, we also apply the proposed method to panoptic segmentation and observe promising results.

\subsection{Experimental setup}
We conduct experiments on the public semantic segmentation benchmarks: GTA5~\cite{richter2016playing}, 
SYNTHIA~\cite{ros2016synthia},
Cityscapes~\cite{cordts2016cityscapes},
BDD100k~\cite{yu2020bdd100k}, Mapillary~\cite{neuhold2017mapillary}, and WildDash2~\cite{Zendel_2018_ECCV} datasets.
The GTA5 and SYNTHIA datasets are synthetic, while the others are real-world datasets. For both synthetic datasets, we split the data following Chen~\etal~\cite{Chen_2019_ICCV}.
For the WildDash2 dataset, we only evaluate the 19 classes overlapping with Cityscapes and ignore the remaining classes.
We evaluate model performance using the standard mean intersection-over-union (mIoU) metric.
We provide the implementation details in the supplementary material.
We will release our source code and pre-trained models for reproducibility.

\subsection{Ablation study}
We use GTA5 as the source domain for ablation experiments and Cityscapes as the unseen target domain.
We use the DeepLabv2 model with a ResNet-101 backbone.

\begin{table}[htbp]
\centering
\caption{\tb{Ablation study.}
We show results from a DeepLabv2 model with a ResNet-101 backbone.
We train models on the GTA5 dataset and treat the Cityscapes dataset as the unseen target domain for evaluation.
}

\mpage{0.48}{(a) Calibration parameters to learn}\hfill
\mpage{0.48}{(b) Different augmentations}\hfill
\\
\mpage{0.48}{
\resizebox{0.7\linewidth}{!}{
\begin{tabular}{lc}
\toprule
Strategy & mIoU (\%) \\
\midrule
Pre-trained & 35.7 \\
$m=0.1$, fixed & 40.1 \\
\midrule
$\mathbf{m} \in \mathbb{R}$ & 39.8 \\
$\mathbf{m} \in \mathbb{R}^C$ & 41.1 \\
$\mathbf{m}_\mu \in \mathbb{R}^C$, $\mathbf{m}_{\sigma} \in \mathbb{R}^C$ & \textbf{41.5} \\
\bottomrule
\end{tabular}
}
}
\hfill
\mpage{0.48}{
\resizebox{0.9\linewidth}{!}{
\begin{tabular}{lcc}
\toprule
Augmentation & InstCal-U & InstCal-C %
\\
\midrule
Default & 39.7 & 40.9
\\
AugMix~\cite{hendrycks2020augmix} & 40.6 & 41.3 %
\\
RandAugment~\cite{cubuk2020randaugment} & 41.1 & 40.0 %
\\
DeepAugment~\cite{hendrycks2021many} & \textbf{41.5} & \textbf{42.2} %
\\
\bottomrule
\end{tabular}

}
}
\hfill
\\
\vspace{2mm}

\mpage{0.54}
{(c) Not enough to pre-train with strong aug.}
\hfill
\mpage{0.42}
{(d) Number of basis for InstCal-C}
\\
\mpage{0.48}{
\resizebox{0.6\linewidth}{!}{
\begin{tabular}{lc}
\toprule
Augmentation & mIoU (\%)
\\
\midrule
Default & 35.7 
\\
AugMix~\cite{hendrycks2020augmix} & 35.9 
\\
RandAugment~\cite{cubuk2020randaugment} & \textbf{37.9} 
\\
DeepAugment~\cite{hendrycks2021many} & 31.7 
\\
\bottomrule
\end{tabular}
}
}
\mpage{0.48}{
\resizebox{0.4\linewidth}{!}{
\begin{tabular}{lc}
\toprule
\#basis & mIoU (\%) \\
\midrule
2 & 40.9 \\
4 & 41.6 \\
8 & \textbf{42.2} \\
16 & 40.7 \\
\bottomrule
\end{tabular}
}
}

\label{tab:abl}
\end{table}

\topic{What calibration parameters should we learn?}
We first conduct experiments to study what calibration parameters should be learned.
As shown in~\tabref{abl}(a), suppose we directly learn a scalar parameter shared by the mean and variance. 
The performance is worse than using a default value of calibration strength (0.1) to calibrate BatchNorm.
Learning a \emph{vector} parameter works much better than a single \emph{scalar} and outperforms the baseline calibration.
Separating the learned vector for mean and variance leads to further improved performance.

\topic{Which data augmentation strategy should we use?}
As we mentioned in~\secref{sec:learnable}, since we only require one source domain for model training, we need to use strong data augmentation to simulate a diverse set of training domains.
In this experiment, we study the impact of data augmentation methods.

\tabref{abl}(b) shows that using the default weak augmentation, e.g., random scaling, cropping, the performance is even worse than the default baseline.
While RandAugment~\cite{cubuk2020randaugment} and AugMix~\cite{hendrycks2020augmix} work well for InstCal-U or InstCal-C separately, these two augmentation strategies do not work well in both variants.
Our results show that DeepAugment~\cite{hendrycks2021many} achieves the best overall performance.
We thus adopt DeepAugment as our default strong data augmentation strategy.

\topic{Pre-training with strong data augmentation is not sufficient.}
In the previous study, we show that the selection of strong data augmentation is critical.
One may wonder if pre-training with strong data augmentation \emph{without the proposed adaptation (InstCal-U and InstCal-C)} is sufficient for performance improvement.
\tabref{abl}(c) shows that pre-training models using strong data augmentations do not achieve the models trained with our proposed adaptation methods.
AugMix~\cite{hendrycks2020augmix} and RandAugment~\cite{cubuk2020randaugment} can improve the performance over the baseline with standard weak augmentation, but not as significant as using them in InstCal-U or InstCal-C.
If we directly use DeepAugment~\cite{hendrycks2021many} for model pre-training, the performance even drops significantly.
The results suggest that it is necessary to apply strong data augmentation, but we need to use them in the InstCal-U/InstCal-C training stage instead of simply using them during pre-training.

\topic{Number of basis for conditional calibration.}
In~\tabref{abl}(d), we study the impact of number of basis ($K$ in \eqnref{K}) for InstCal-C.
Using eight basis leads to the best results among several options.

\subsection{Comparison with other test-time adaptation methods}
We conduct experiments using the DeepLabv2 model with a ResNet-101 backbone.
We first construct a baseline using a default value ($m=0.1$) to calibrate BatchNorm statistics. 
We compare with one optimization-based approach (TENT~\cite{wang2021tent}) and two BatchNorm calibration based methods (AdaptiveBN~\cite{schneider2020improving} and PT-BN~\cite{nado2020evaluating}).
We use the same protocols to separately conduct test-time adaptation on multiple unseen target domains.
(\ie setting test batch size to 1 and adapting to each test example individually).

Note that AdaptiveBN~\cite{schneider2020improving}, PT-BN~\cite{nado2020evaluating}, and our baseline share the same formulation (\eqnref{norm}) but using different $m$ values.
As shown in~\tabref{v2}, while both the simple baseline and AdaptiveBN~\cite{schneider2020improving} show improved results, PT-BN~\cite{nado2020evaluating} even hurts the pre-trained performance in many cases.
TENT~\cite{wang2021tent} also shows strong results in some of the test settings, but with the price of significantly increased computation time.
In contrast, the proposed InstCal-U and InstCal-C outperform these test-time adaptation methods in most settings.
We also note that InstCal-U performs better in real-world cross-domain settings, while InstCal-C achieves more promising results in synthetic-to-real settings.

\begin{figure*}[t]

\hfill
\mpage{0.22}
{\includegraphics[width=\linewidth]{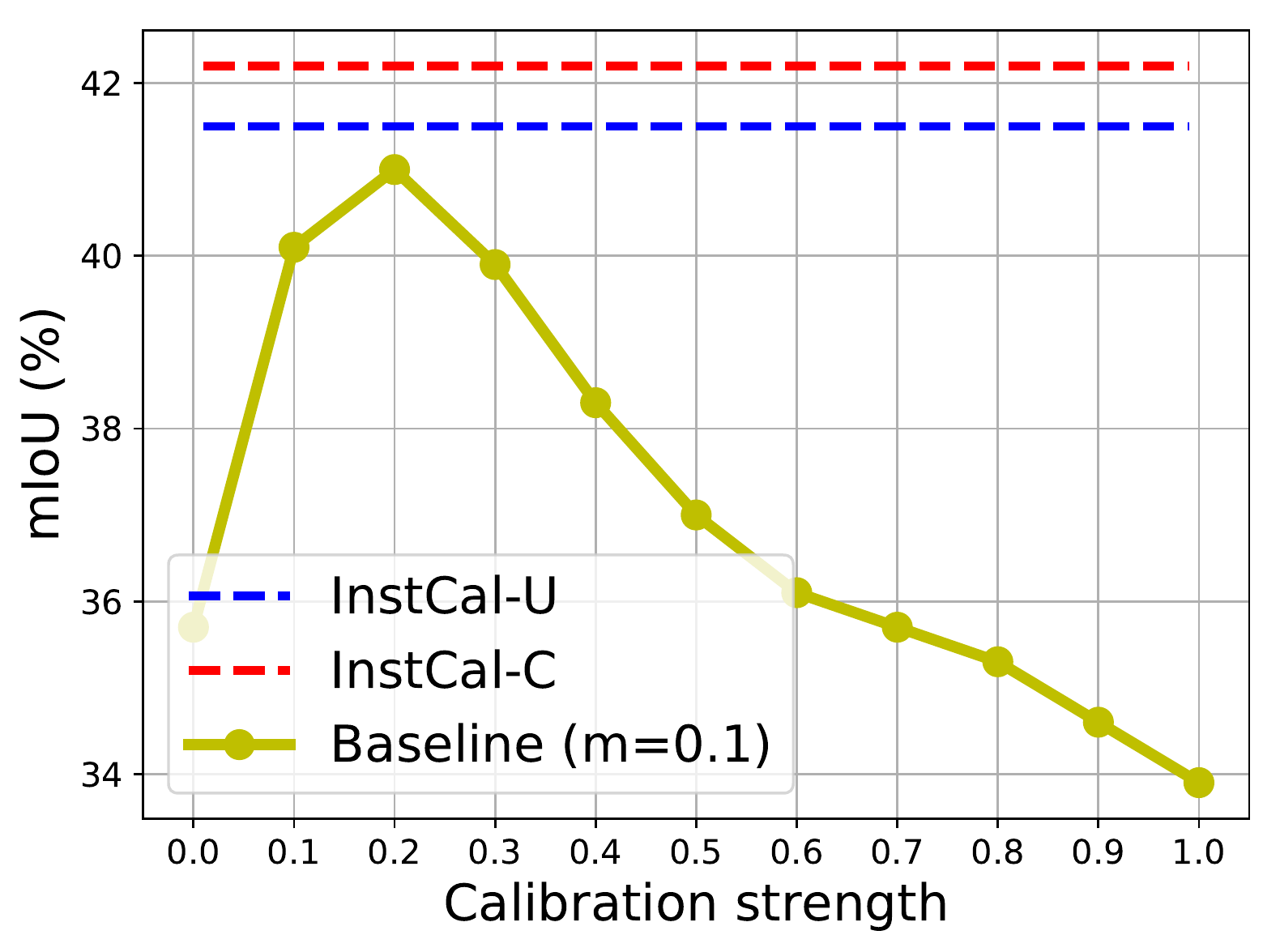}}
\hfill
\mpage{0.22}
{\includegraphics[width=\linewidth]{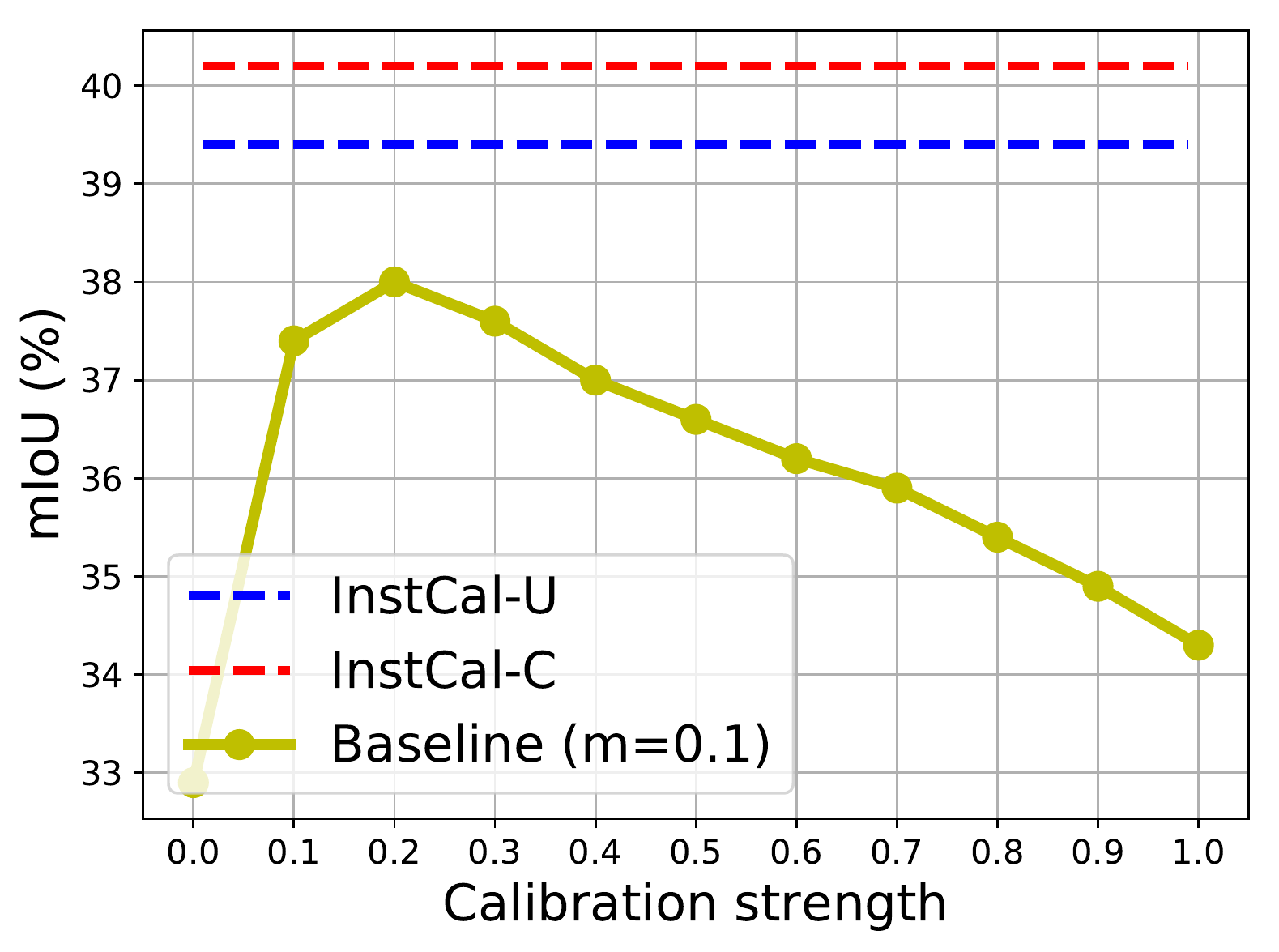}}
\hfill
\mpage{0.22}
{\includegraphics[width=\linewidth]{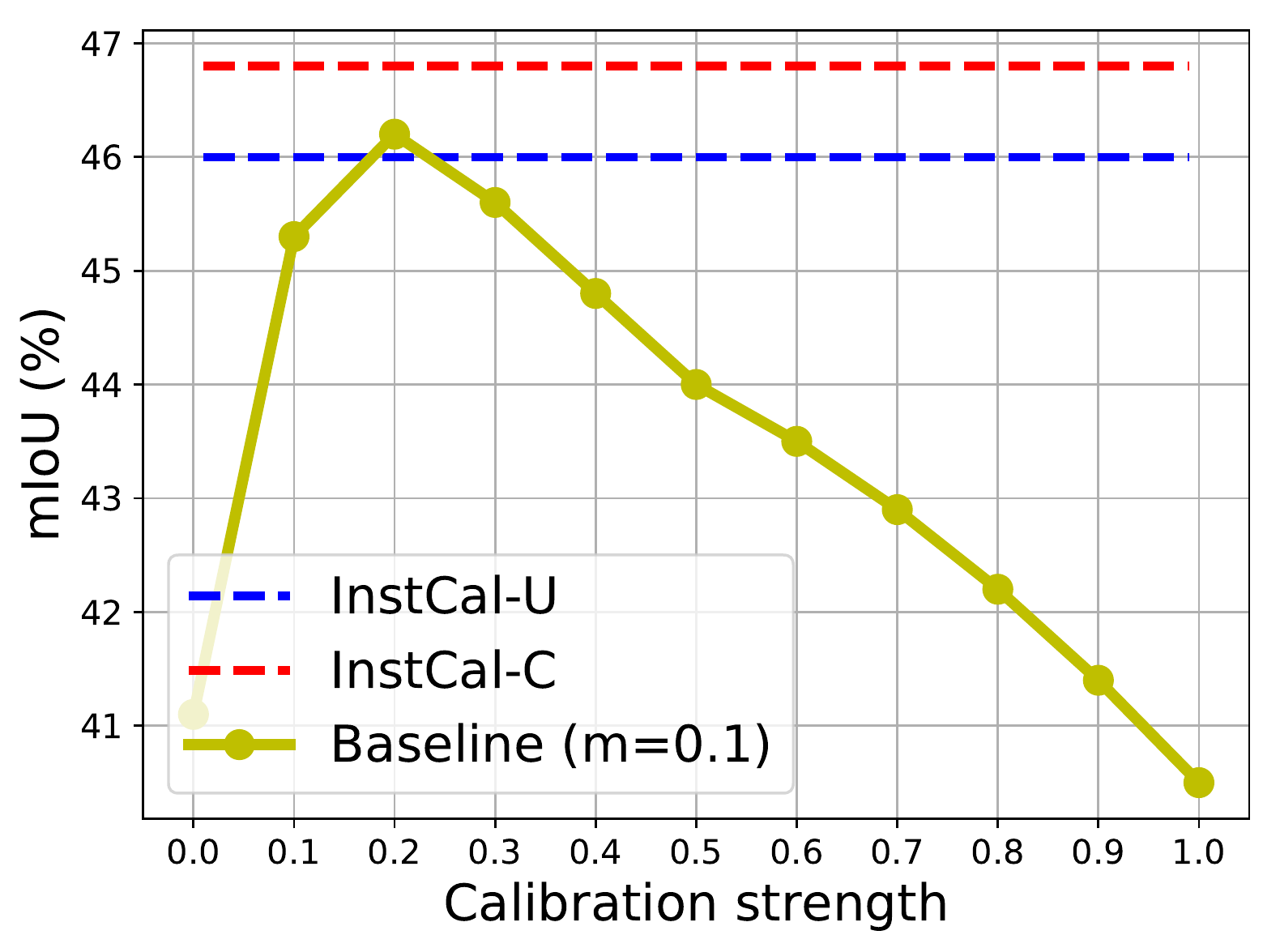}}
\hfill
\mpage{0.22}
{\includegraphics[width=\linewidth]{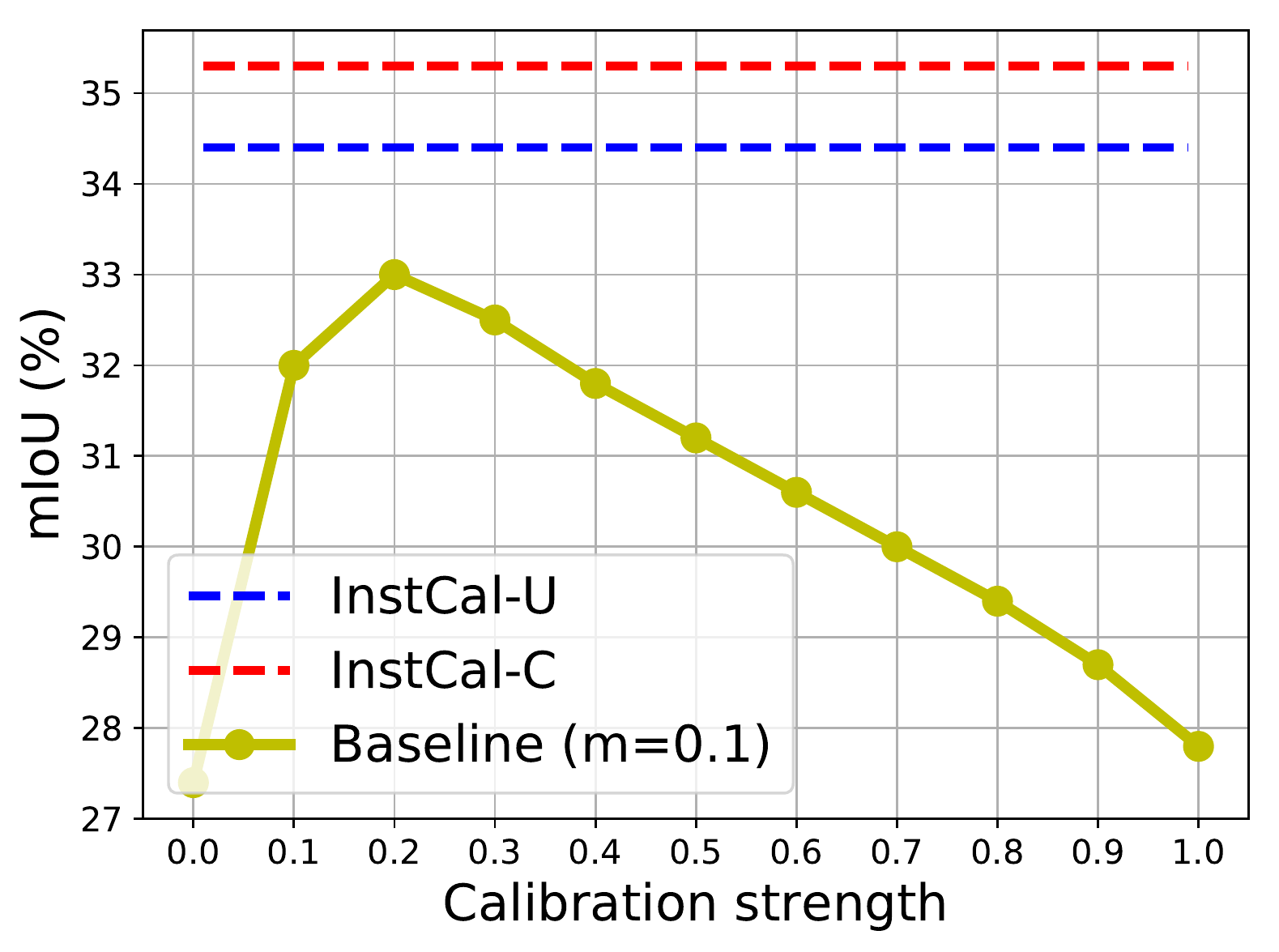}}
\hfill
\\
\hfill
\mpage{0.23}{(a) Cityscape}
\hfill
\mpage{0.23}{(b) BDD100k}
\hfill
\mpage{0.23}{(c) Mapillary}
\hfill
\mpage{0.23}{(d) WildDash2}
\hfill
\\

\figcapmargin
\vspace{-.2cm}
\figcaption{Manually set calibration strength $m$}
{We show results from a DeepLabv2 model with a ResNet-101 backbone. The source domain is the GTA5 dataset.}
\label{fig:momentum}
\end{figure*}

In addition to setting the calibration strength to the default value (0.1), we also experiment with different values.
We try from 0.0 to 1.0 with a step size of 0.1, and visualize the results in~\figref{momentum}.
As we can see, using scalar as strength to calibrate BatchNorm is highly sensitive to selecting the values.
On the contrary, the proposed InstCal-U and InstCal-C consistently perform well across unseen target domains.

\begin{table}[htbp]
\centering
\caption{\tb{Generalizing across multiple domains.}
We show results from DeepLabv2 models with a ResNet-101 backbone.
The baseline uses calibration strength $m=0.1$.
``C'' indicates Cityscapes, ``B'' indicates BDD100k, ``M'' indicates Mapillary, and ``W'' indicates WildDash2.
The best performance is in \best{bold} and the second best is \second{underlined}.
}

\resizebox{0.7\linewidth}{!}{
\begin{tabular}{lcccccccccc}
\toprule
& \multicolumn{5}{c}{Source: GTA5}
&& \multicolumn{4}{c}{Source: Cityscapes}
\\
\cmidrule{2-6}
\cmidrule{8-11}
Method & C & B & M & W & Avg. 
&& B & M & W & Avg. 
\\
\midrule
Pre-trained
& 35.7 & 32.9 & 41.1 & 27.4 & 34.3
&& 41.2 & 49.5 & 33.9 & 41.5
\\
\midrule
Baseline ($m=0.1$)
& 40.1 & {37.4} & 45.3 & 32.0 & 38.7
&& 42.8 & {51.9} & 37.6 & 44.1
\\

AdaptiveBN~\cite{schneider2020improving}
& 39.0 & 36.3 & 44.3 & 30.8 & 37.6
&& 43.0 & \best{52.4} & 37.2 & 44.2
\\

PT-BN~\cite{nado2020evaluating}
& 33.9 & 34.3 & 40.5 & 27.8 & 34.1
&& 34.4 & 39.1 & 28.8 & 34.1
\\

TENT~\cite{wang2021tent}
& 38.1 & 37.8 & 44.7 & 32.5 & 38.3
&& 44.2 & \second{52.3} & 36.8 & 44.4
\\

InstCal-U (Ours)
& \second{41.5} & \second{39.4} & \second{46.0} & \second{34.4} & \second{40.3}
&& \best{45.1} & 52.2 & \best{40.3} & \best{45.9}
\\
InstCal-C (Ours)
& \best{42.2} & \best{40.2} & \best{46.8} & \best{35.3} & \best{41.1}
&& \second{44.3} & 51.5 & \second{39.3} & \second{45.0}
\\
\bottomrule
\end{tabular}
}

\label{tab:v2}
\end{table}

\subsection{Analysis}
For the following studies, we use DeepLabv2 models with a ResNet-101 backbone.

\topic{Improvement on in-domain performance.}
We test if our models can improve the performance for \emph{source domain}.
We do so by evaluating the trained model on the \emph{test split} of the source data.
We report in \tabref{analysis}(a) that our learned BatchNorm calibration (both unconditional and conditional) still acheive sizable performance gain.

\begin{table}[htbp]
\centering
\caption{\tb{Analysis.}
Results are from DeepLabv2 models with a ResNet-101 backbone.
}

\mpage{0.48}{(a) In-domain performance}\hfill
\mpage{0.48}{(b) Input batch statistics}\hfill
\\
\mpage{0.48}{
\resizebox{0.6\linewidth}{!}{
\begin{tabular}{lcc}
\toprule
Method & GTA5 & Cityscapes \\
\midrule
Pre-trained & 69.1 & 66.1 \\
InstCal-U & \second{70.3} & \second{66.6} \\
InstCal-C & \best{70.5} & \best{66.8} \\
\bottomrule
\end{tabular}
}
}
\hfill
\mpage{0.48}{
\resizebox{0.8\linewidth}{!}{
\begin{tabular}{lccccc}
\toprule
Batch size & 1 & 2 & 4 & 8 & 16 \\
\midrule
Baseline ($m=1$)  & \textbf{40.1} & 39.8 & 39.7 & 39.7 & 39.6 \\
InstCal-U & \textbf{41.5} & 41.2 & 40.9 & 40.8 & 40.7 \\
InstCal-C & \textbf{42.2} & 41.8 & 41.5 & 41.5 & 41.4 \\
\bottomrule
\end{tabular}
}
}
\\
\mpage{0.48}{(c) Model calibration}\hfill
\mpage{0.48}{(d) Test-time optimization}\hfill
\\
\mpage{0.48}{
\includegraphics[width=0.8\linewidth]{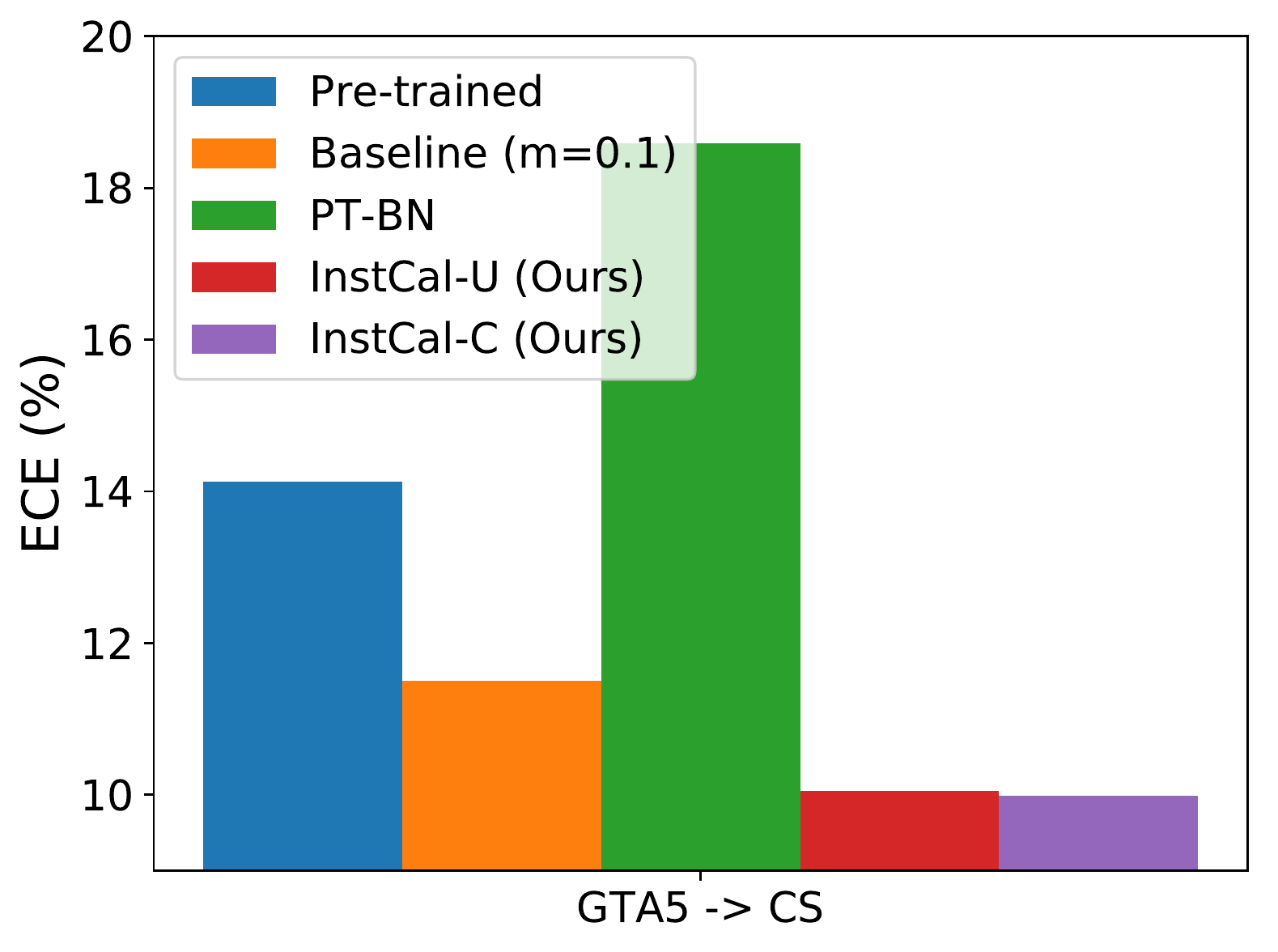}
}
\hfill
\mpage{0.48}{
\resizebox{0.75\linewidth}{!}{
\begin{tabular}{lc}
\toprule
Method & mIoU (\%) \\
\midrule
Pre-trained & 35.7 \\
TENT~\cite{wang2021tent} & 38.1 \\
\midrule
InstCal-U & 41.5 \\
InstCal-U + entropy min.~\cite{wang2021tent} & \second{44.1} \\
\midrule
InstCal-C & 42.2 \\
InstCal-C + entropy min.~\cite{wang2021tent} & \best{44.2} \\
\bottomrule
\end{tabular}
}
}
\hfill
\\

\label{tab:analysis}
\end{table}

\topic{Input batch statistics v.s. instance statistics}
As mentioned in~\secref{sec:learnable}, we compute input instance statistics for \emph{each} test example, instead of computing the batch statistics for mixing statistics across different examples within a mini-batch.
We validate this design choice by replacing instance statistics with batch statics using different batch sizes during test time.
\tabref{analysis}(b) shows that the performance drops as we increase the batch size, even though the test examples come from the same target distribution.
We conjecture the performance will worsen if the mini-batch contains test examples from multiple target domains.
Thus, we stick to using the input instance statistics.

\topic{Improvement on model calibration.}
We compute the expected calibration error (ECE)~\cite{guo2017calibration} for the pre-trained model, baseline update ($m=0.1$), PT-BN~\cite{nado2020evaluating}, and the proposed InstCal-U/InstCal-C.
As shown in \tabref{analysis}(c), calibrating BatchNorm statistics indeed reduces model calibration error.

\topic{Compatible with test-time optimization.}
We incorporate the optimization-based method, TENT~\cite{wang2021tent}, into our methods, by optimizing the prediction entropy at test-time.
Following TENT~\cite{wang2021tent}, we only optimize the weight $\gamma$ and bias $\beta$ in BatchNorm layers. 
Moreover, we conduct instance-specific adaptation.
\tabref{analysis}(d) shows the complementary nature of these two strategies.

\topic{Running time.}
We test the inference speed on Cityscapes on a single V100 GPU.
The pre-trained model takes 39 ms to process each testing sample (1024$\times$512 resolution).
The BatchNorm calibration method~\cite{nado2020evaluating} induce a 60 ms overhead. 
Our method increases the inference time by 58 ms (for InstCal-U) and 149 ms (InstCal-C).

\subsection{Comparison with one-shot unsupervised domain adaptation}
This section compares the proposed method with recent state-of-the-art one-shot UDA methods.
One-shot UDA methods adapt source domain pre-trained models on one single unlabeled target example offline. In contrast, InstCal-U and InstCal-C adapt pre-trained models on the fly on each test example individually.
Conceptually, one-shot UDA methods and InstCal-U/InstCal-C use the same amount of data for adaptation.
However, one-shot UDA methods usually require time-consuming offline training, and thus it is impossible to adapt models on each target example separately.
So these methods only adapt the models using one single unlabeled (training) example and then deploy the adapted model at test-time without adaptation.
As shown in~\tabref{uda_all}, simply augmenting pre-trained models with InstCal-U/InstCal-C, compares favorably with recent one-shot UDA methods and even outperforms the state of the arts by a large margin in Synthia$\rightarrow$Cityscapes setting.

\begin{table}[htbp]
\centering
\caption{\tb{Comparison with one-shot unsupervised domain adaptation.}
{All results are from modified DeepLabv2 models (specific for domain adaptation). 
The best performance is in \best{bold} and the second best is \second{underlined}.
}
}
\tabcapmargin
\resizebox{0.8\textwidth}{!}{

\begin{tabular}{lcccccc}
\toprule
& GTA5$\rightarrow$Cityscapes &&& \multicolumn{2}{c}{Synthia$\rightarrow$Cityscapes} \\
\cmidrule{2-3}
\cmidrule{5-6}
Method & mIoU &&& mIoU (13-class) & mIoU (16-class) \\
\midrule

CLAN~\cite{luo2019taking} & 37.7 &&& 40.4 & - \\

AdvEnt~\cite{vu2019advent} & 36.1 &&& 39.9 & - \\

CBST~\cite{zou2018unsupervised} & 37.1 &&& 38.5 & - \\

OST~\cite{benaim2018one} & 42.3 &&& 42.8 & -
\\

ASM~\cite{luo2020adversarial} & \best{43.2} &&& 40.7 & 34.6
\\

\midrule

Source-only pre-trained & 36.2 &&& 36.2 & 31.6 \\

+ InstCal-U (Ours)
& \second{42.4} &&& \second{43.5} & \second{37.7}
\\

+ InstCal-C (Ours)
& 42.2 &&& \best{44.1} & \best{38.1}    %
\\

\bottomrule
\end{tabular}

}

\label{tab:uda_all}
\end{table}

\subsection{Comparison with domain generalizing segmentation}
This section compares our InstCal-U/InstCal-C with recent domain generalizing (DG) semantic segmentation approaches.
We use the DeepLabv3+ model with a ResNet-50 backbone.
As shown in~\tabref{v3}, upgrading non-DG pre-trained \emph{weak} models with InstCal-U/InstCal-C compares favorably with these \emph{strong} domain generalizing segmentation methods across different testing settings.
Our method even outperforms all the methods except ISW~\cite{choi2021robustnet} by a large margin.

\begin{figure*}[t]
\mpage{0.31}
{\includegraphics[width=\linewidth]{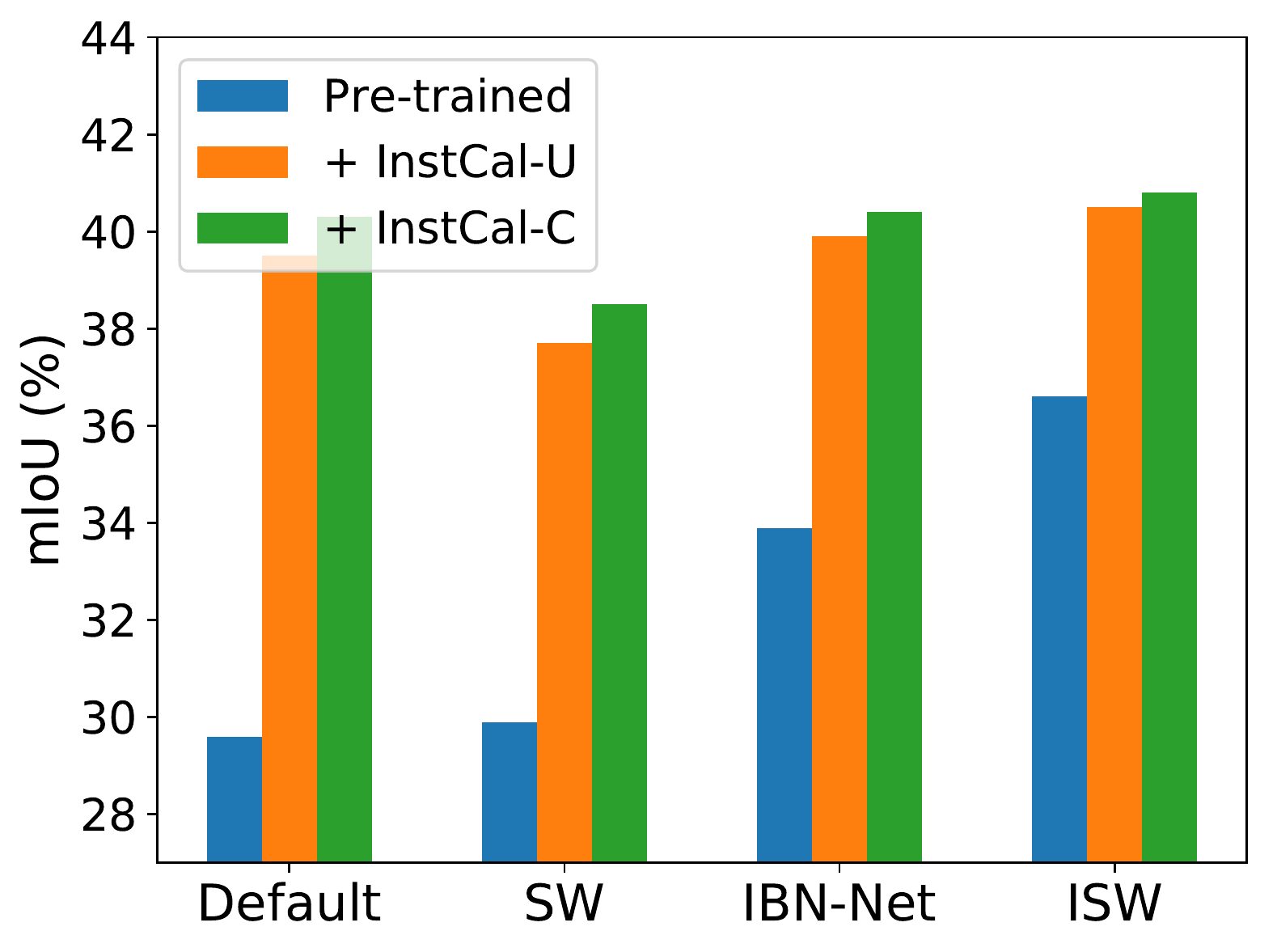}}
\hfill
\mpage{0.31}
{\includegraphics[width=\linewidth]{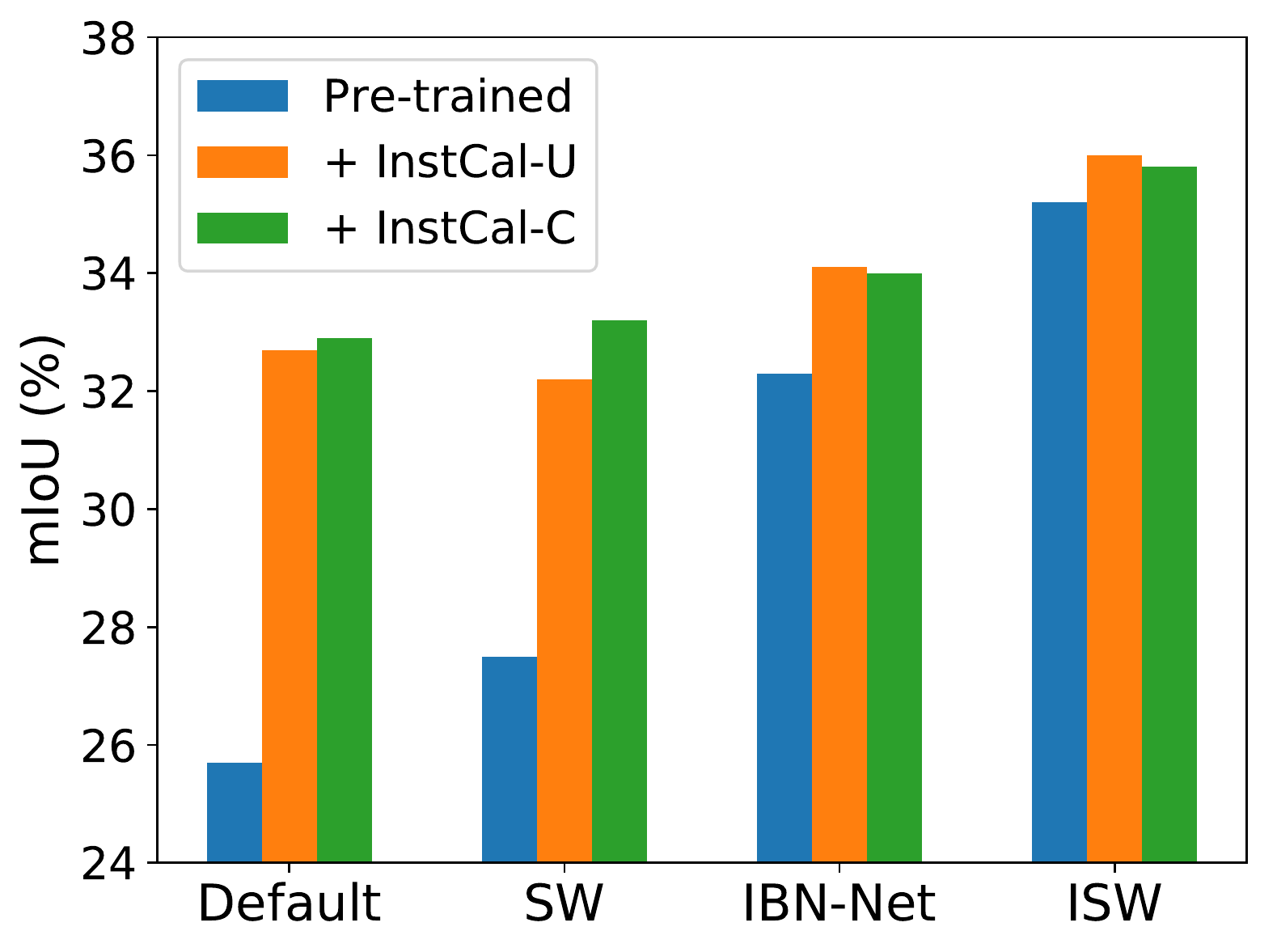}}
\hfill
\mpage{0.31}
{\includegraphics[width=\linewidth]{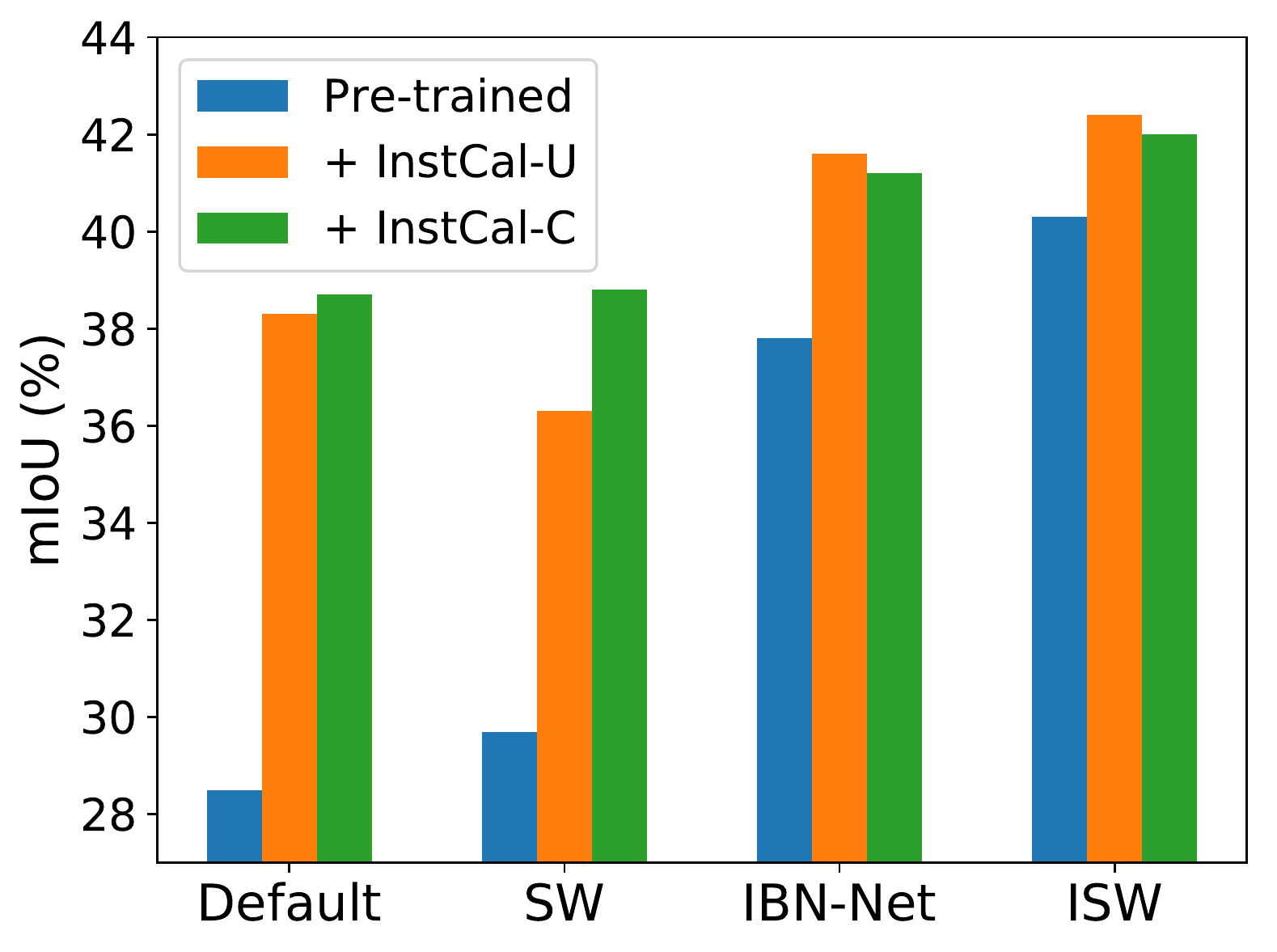}}
\hfill
\\

\mpage{0.31}{(a) GTA5$\rightarrow$Cityscapes}\hfill
\mpage{0.31}{(b) GTA5$\rightarrow$BDD100k}\hfill
\mpage{0.31}{(c) GTA5$\rightarrow$Mapillary}\hfill

\figcapmargin
\vspace{-.2cm}
\figcaption{Combining domain generalization with our method}
{
In addition to the model pre-trained with the standard recipe (non-DG, labeled as ``Default''), we choose three DG methods: SW~\cite{pan2019switchable}, IBN-Net~\cite{pan2018two}, and ISW~\cite{choi2021robustnet}.
All methods use DeepLabv3+ models with a ResNet-50 backbone, trained on the GTA5 dataset.
}
\label{fig:dg}
\end{figure*}

Note that our method and these domain generalizing methods complement each other.
Thus, we can also incorporate our methods on top of these domain generalizing segmentation methods.
As shown in~\figref{dg}, the proposed method consistently improves the performance of these methods.
Our method can even improve the strong ISW~\cite{choi2021robustnet} approach and achieve a new state of the art.

\begin{table}[htbp]
\centering
\caption{\tb{Comparison with state-of-the-art domain generalizing semantic segmentation methods.}
We show results from a DeepLabv3+ model with a ResNet-50 backbone. 
``C'' indicates Cityscapes, ``B'' indicates BDD100k, and ``M'' indicates Mapillary.
The best performance is in \best{bold} and the second best is \second{underlined}.
}

\resizebox{0.6\linewidth}{!}{
\begin{tabular}{lcccccccc}
\toprule
& \multicolumn{4}{c}{Source: GTA5}
&& \multicolumn{3}{c}{Source: Cityscapes}
\\
\cmidrule{2-5}
\cmidrule{7-9}
Method & C & B & M & Avg. 
&& B & M & Avg. 
\\
\midrule
SW~\cite{pan2019switchable}
& 29.9 & 27.5 & 29.7 & 29.0
&& 48.5 & 55.8 & 52.2
\\
IBN-Net~\cite{pan2018two}
& 33.9 & 32.3 & 37.8 & 34.7
&& 48.6 & 57.0 & 52.8
\\
IterNorm~\cite{huang2019iterative}
& 31.8 & 32.7 & 33.9 & 32.8
&& 49.2 & 56.3 & 52.8
\\
ISW~\cite{choi2021robustnet}
& 36.6 & \best{35.2} & \best{40.3} & \best{37.4}
&& \second{50.7} & \best{58.6} & \second{54.7}
\\
\midrule
Non-DG pre-trained
& 29.6 & 25.7 & 28.5 & 27.9
&& 46.1 & 52.5 & 49.3
\\
+ InstCal-U (Ours)
& \second{39.8} & \second{32.9} & 38.6 & 37.1
&& \best{51.1} & \second{58.5} & \best{54.8}
\\
+ InstCal-C (Ours)
& \best{40.3} & \second{32.9} & \second{38.7} & \second{37.3}
&& 50.5 & 57.7 & 54.1
\\
\bottomrule
\end{tabular}
}

\label{tab:v3}
\end{table}

\subsection{Backbone network agnostic}
In previous sections, we have shown the proposed method can improve pre-trained model performance on multiple unseen target domains.
However, we only conduct experiments using the ResNet backbones.
In this section, we use the DeepLabv3+ model with ShuffleNetV2~\cite{ma2018shufflenet} and MobileNetV2~\cite{sandler2018mobilenetv2} as backbones, to demonstrate our methods is network-agnostic.
As shown in~\tabref{network_v3}, the proposed methods can also improve pre-trained models with these backbones by a large margin, outperforming the recent state of the arts.

\begin{table}[htbp]
\centering
\caption{\tb{The proposed module is backbone network agnostic.}
We show results from a DeepLabv3+ model with ShuffleNetV2 and MobileNetV2 as backbones.
These models are trained on the GTA5 dataset.
``C'' indicates Cityscapes, ``B'' indicates BDD100k, and ``M'' indicates Mapillary.
The best performance is in \best{bold} and the second best is \second{underlined}.
}

\resizebox{0.6\linewidth}{!}{
\begin{tabular}{lccccccccc}
\toprule
& \multicolumn{4}{c}{ShuffleNetV2}
&& \multicolumn{4}{c}{MobileNetV2}
\\
\cmidrule{2-5}
\cmidrule{7-10}
Method & C & B & M & Avg. 
&& C & B & M & Avg. 
\\
\midrule
IBN-Net~\cite{pan2018two}
& 27.1 & 31.8 & 34.9 & 31.3
&& 30.1 & 27.7 & 27.1 & 28.3
\\
ISW~\cite{choi2021robustnet}
& 31.0 & \best{32.1} & 35.3 & 32.8
&& 30.9 & \second{30.1} & 30.7 & 30.6
\\
\midrule
Non-DG pre-trained
& 25.7 & 22.1 & 28.3 & 25.4
&& 27.1 & 27.5 & 27.3 & 27.3
\\
+ InstCal-U (Ours)
& \second{35.8} & \second{31.1} & \best{36.4} & \best{34.4}
&& \second{37.2} & \best{31.2} & \best{34.5} & \best{34.3}
\\
+ InstCal-C (Ours)
& \best{35.9} & 30.8 & \second{35.4} & \second{34.0}
&& \best{37.8} & 30.0 & \second{33.9} & \second{33.9}
\\
\bottomrule
\end{tabular}
}

\label{tab:network_v3}
\end{table}

\subsection{Panoptic segmentation results}
\label{sec:panoptic}
In this section, we directly apply InstCal-U/InstCal-C to an even more challenging task, panoptic segmentation~\cite{kirillov2019panoptic}.
We start with the off-the-shelf models from Panoptic-DeepLab~\cite{cheng2020panoptic} and train these models on the Cityscapes dataset.
We test the models on Foggy Cityscapes~\cite{SDV18}, which inserts synthetic fog into the original Cityscapes clear images with three strength levels (0.005, 0.01, and 0.02).
We adopt panoptic quality (PQ), mean intersection-over-union (mIoU), and mean average precision (mAP) as the evaluation metrics.
As shown in~\tabref{panoptic}, InstCal-U/InstCal-C greatly improves off-the-shelf Panoptic-DeepLab performance on out-of-distribution foggy scenes by a large margin, validating the proposed method is universally applicable to different image segmentation tasks without further tuning.
We also provide visual results in \figref{teaser} and supplementary material.

\begin{table}[htbp]
\centering
\caption{\tb{Panoptic segmentation results.}
{We show results on two Panoptic-DeepLab model variants (w/ and w/o depthwise separable convolution).
The best performance is in \best{bold} and the second best is \second{underlined}.
}
}
\tabcapmargin
\resizebox{0.75\textwidth}{!}{

\begin{tabular}{lccccccccccccc}
\toprule
&&& \multicolumn{11}{c}{Synthetic fog strength} \\
&&& \multicolumn{3}{c}{0.005} && \multicolumn{3}{c}{0.01} && \multicolumn{3}{c}{0.02} \\
\cmidrule{4-6}
\cmidrule{8-10}
\cmidrule{12-14}
Method & w/ DSConv && PQ & mIoU & mAP && PQ & mIoU & mAP && PQ & mIoU & mAP  \\
\midrule
Pre-trained & $\times$
&& 53.3 & 72.2 & 25.3
&& 45.0 & 64.9 & 18.8
&& 32.6 & 52.8 & 11.6
\\

+ InstCal-U & $\times$
&& \best{56.6} & \best{75.7} & \best{28.9}
&& \second{51.1} & \best{71.9} & \second{24.3}
&& \best{42.8} & \second{64.5} & \best{18.5}
\\

+ InstCal-C & $\times$
&& \best{56.6} & \best{75.7} & \second{28.5}
&& \best{51.2} & \best{71.9} & \best{24.4}
&& \second{42.4} & \best{64.8} & \second{18.0}
\\

\midrule

Pre-trained & \checkmark
&& 53.0 & 73.2 & 24.5
&& 45.3 & 66.5 & 18.3
&& 33.1 & 54.8 & 11.5
\\

+ InstCal-U & \checkmark
&& \second{55.5} & \second{76.0} & \second{27.2}
&& \best{49.1} & \second{71.3} & \second{21.8}
&& \second{40.4} & \second{63.2} & \second{16.2}
\\

+ InstCal-C & \checkmark
&& \best{55.6} & \best{76.3} & \best{27.6}
&& \second{48.9} & \best{71.6} & \best{22.4}
&& \best{40.9} & \best{64.1} & \best{16.8}
\\

\bottomrule
\end{tabular}

}

\label{tab:panoptic}
\end{table}

\section{Discussions}
\vspace{-.2cm}
\label{sec:conclusions}
\secmargin

This paper proposes a simple learning-based test-time adaptation method for cross-domain segmentation. 
The proposed method is learned to perform \emph{instance-specific} BatchNorm calibration during training, without time-consuming test-time parameter optimization. 
As a result, our method is efficient and effective, demonstrating competitive performance across multiple cross-domain image segmentation settings.

\topic{Limitations.}
Currently, we conduct calibration for every BatchNorm layers. It will be interesting to study which layer is more important and thus only calibrate specific layers to increase inference speed.
And it will be interesting to extend our method to other normalization layers (\eg LayerNorm for Vision Transformers~\cite{dosovitskiy2020image}) and other challenging tasks.
We leave these as future work.

\section*{Acknowledgement}
We thank Sayna Ebrahimi, Shih-Yang Su, and Yun-Chun Chen for their valuable comments.
Y. Zou and J.-B. Huang were supported in part by NSF under Grant No. (\#1755785).

{\small
\bibliographystyle{splncs04}
\bibliography{main}
}
\addcontentsline{toc}{section}{Supplementary Material}
\renewcommand{\thesubsection}{\Alph{subsection}}
\section*{Supplementary Matarial}

In this supplementary document, we provide additional experimental results and details to complement the main manuscript.
First, we provide the implementation details of our experiments.
Second, we describe the three data augmentation strategies we explore.
Lastly, we show qualitative results of different cross-domain segmentation settings.

\subsection{Implementation details}
We conduct all the experiments using the PyTorch framework with one single V100 GPU.

\topic{Semantic segmentation.}
We start with a DeepLabv2 model~\cite{chen2017deeplab} with a ResNet-101 backbone pre-trained on ImageNet~\cite{russakovsky2015imagenet}.
We follow Chen~\etal~\cite{Chen_2019_ICCV} to pre-train the \emph{source-only} model.
In our proposed learning stage, we freeze the model parameters except for the proposed module and train the model with the SGD optimizer with momentum 0.9 and weight decay 5$\times 10^{-4}$.
We use a learning rate of 2.5$\times 10^{-3}$ for InstCal-U and a learning rate of 2.5$\times 10^{-2}$ for InstCal-C.
We adopt the polynomial learning rate decay schedule as in Chen~\etal~\cite{Chen_2019_ICCV} and set the total number of training iterations to 80,000.
We set the batch size to one.

To compare with recent domain generalizing semantic segmentation methods, we adopt the implementation and off-the-shelf models from RobustNet~\cite{choi2021robustnet}.
We do not conduct pre-training and directly train these off-the-shelf models using the proposed method.
For these DeepLabv3+ models, we set the learning rate as 2.5$\times 10^{-3}$ and reduce the batch size for single GPU training (\ie 8 for ResNet-50, 4 for ShuffleNetV2 and MobileNetV2). The other training hyper-parameters are kept the same. 
Please refer to Choi~\etal~\cite{choi2021robustnet} for more details.

\topic{Panoptic segmentation.}
We adopt off-the-shelf models from Panoptic-DeepLab~\cite{cheng2020panoptic}, implemented in the PyTorch Detectron2 codebase.
We use a learning rate of 2.5$\times 10^{-4}$. We set the batch size to 4. 
The other training hyper-parameters are kept the same. Please refer to the PyTorch implementation for more details.\footnote{\url{https://github.com/facebookresearch/detectron2/tree/main/projects/Panoptic-DeepLab}}

\subsection{Strong data augmentation strategies}
We provide details for the three strong data augmentation strategies we adopted.

\topic{RandAugment~\cite{cubuk2020randaugment}.}
RandAugment samples $m$ operations from a pre-defined list of image augmentation operations and composes them to form the final augmentation for each input data.
In this paper, we set $m=2$, and we only adopt the color-related augmentation operations in the pre-defined list: Identity, AutoContrast, Invert, Equalize, Solarize, Posterize, Color, Brightness, Sharpness.

\topic{AugMix~\cite{hendrycks2020augmix}.}
AugMix is proposed to improve model robustness and uncertainty estimation.
AugMix constructs three augmentation paths with cascaded one, two, and three augmentation operations for each input data. 
The three augmented data are then linearly combined, where the combination weights are sampled from a Dirichlet distribution. 
Finally, the original and augmented images are linearly combined to construct the final augmented image, where the combination weight is sampled from a Beta distribution.
We adopt all the augmentation operations, including geometric transforms, and we use the original annotation as the ground truth for the final augmented image.
We only use the augmentation but not the Jensen-Shannon divergence consistency loss described in Hendrycks~\etal~\cite{hendrycks2020augmix}.

\topic{DeepAugment~\cite{hendrycks2021many}.}
Instead of composing basic image augmentation operators to construct a strong data augmentation, DeepAugment transforms the input image by feeding it into an image-to-image translation network and randomly perturbing the intermediate feature representations.
There are three options provided in the official implementation. 
We adopt the CAE~\cite{theis2017lossy} variant, which is initially used for the image compression task.

\subsection{Qualitative results}
\begin{figure*}[htbp]
\mpage{0.17}{\includegraphics[width=1.0\linewidth]{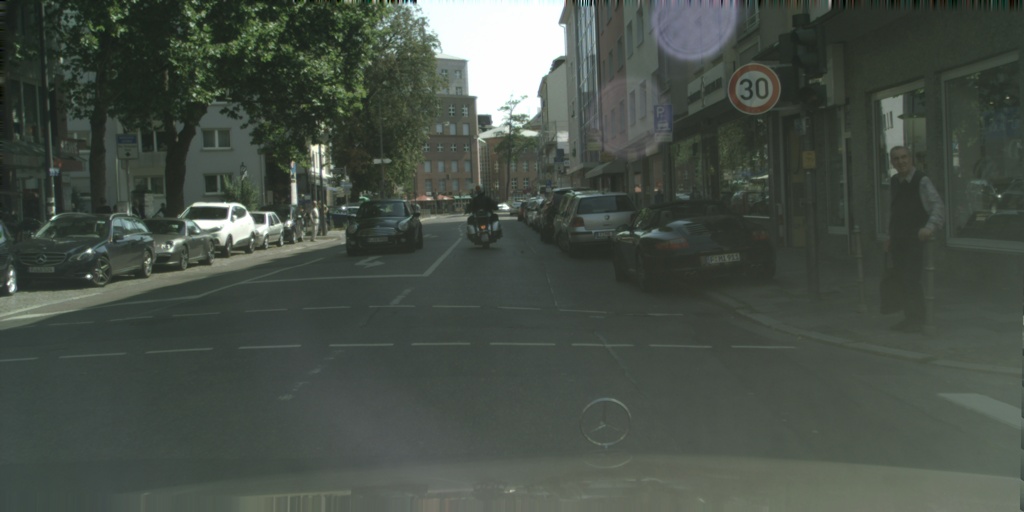}}
\hfill
\mpage{0.17}{\includegraphics[width=1.0\linewidth]{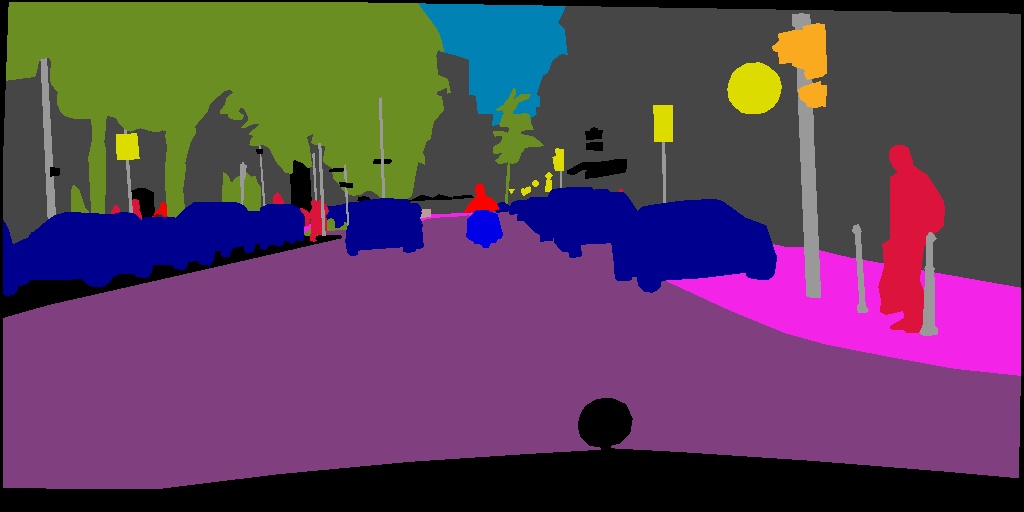}}
\hfill
\mpage{0.17}{\includegraphics[width=1.0\linewidth]{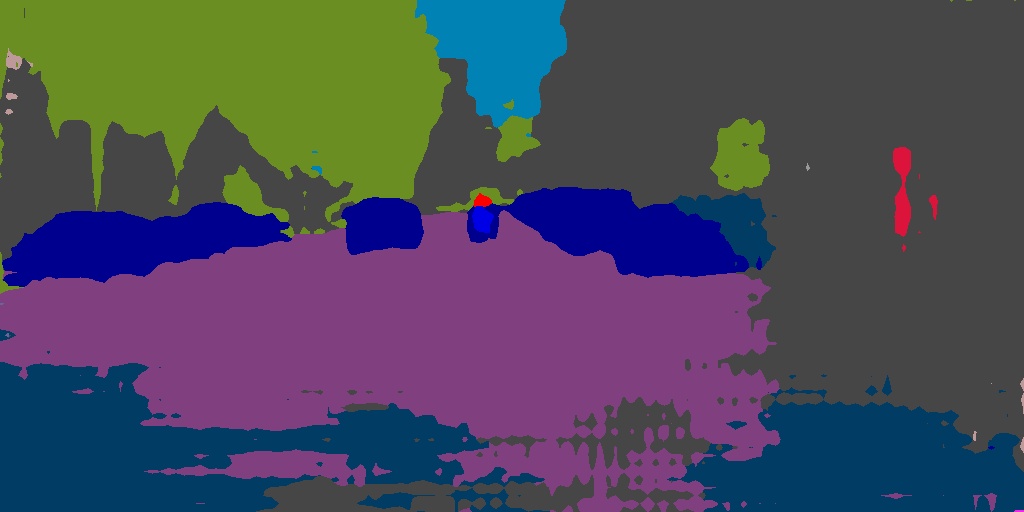}}
\hfill
\mpage{0.17}{\includegraphics[width=1.0\linewidth]{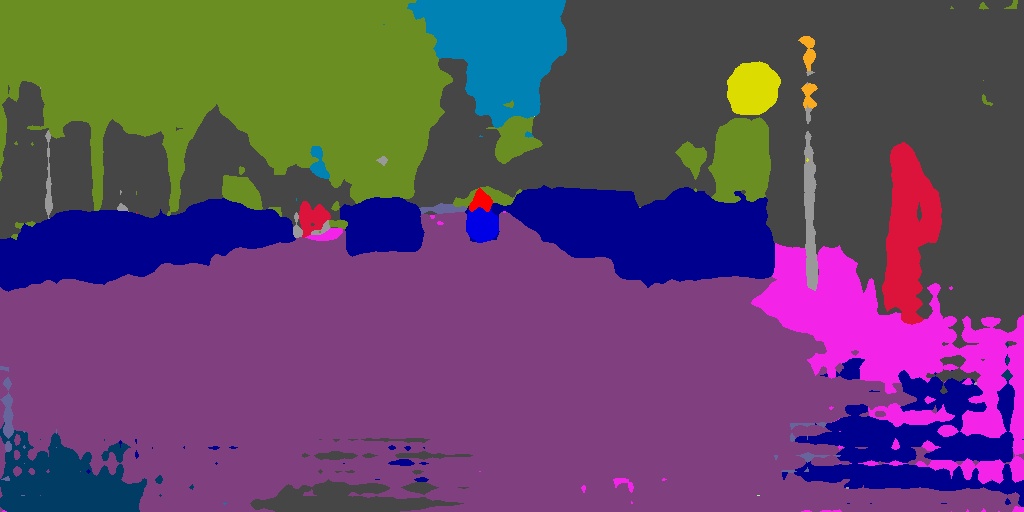}}
\hfill
\mpage{0.17}{\includegraphics[width=1.0\linewidth]{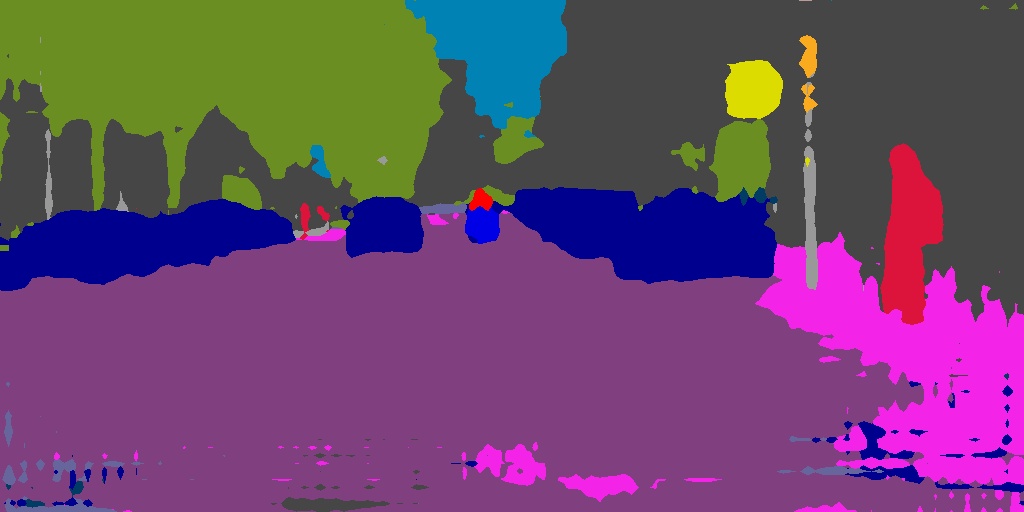}}
\hfill
\\
\mpage{0.17}{\includegraphics[width=1.0\linewidth]{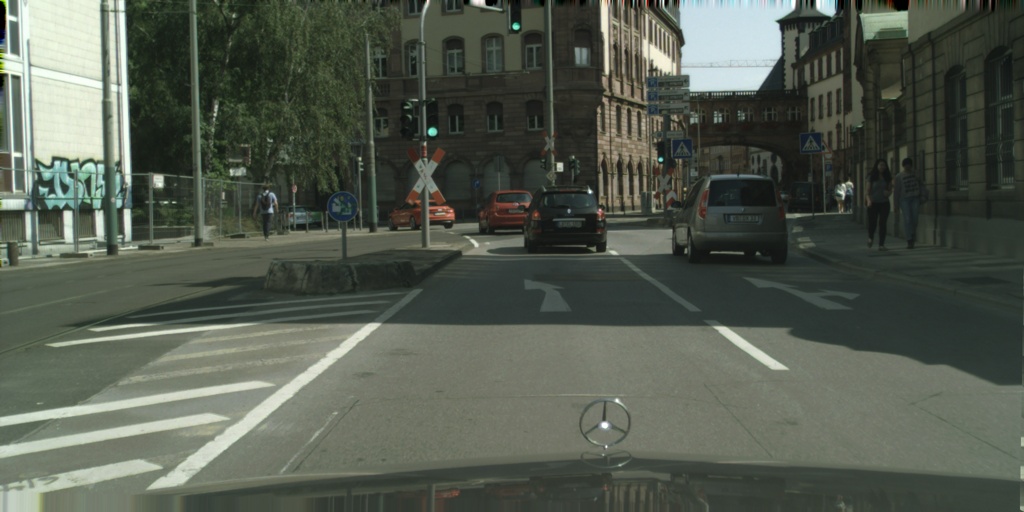}}
\hfill
\mpage{0.17}{\includegraphics[width=1.0\linewidth]{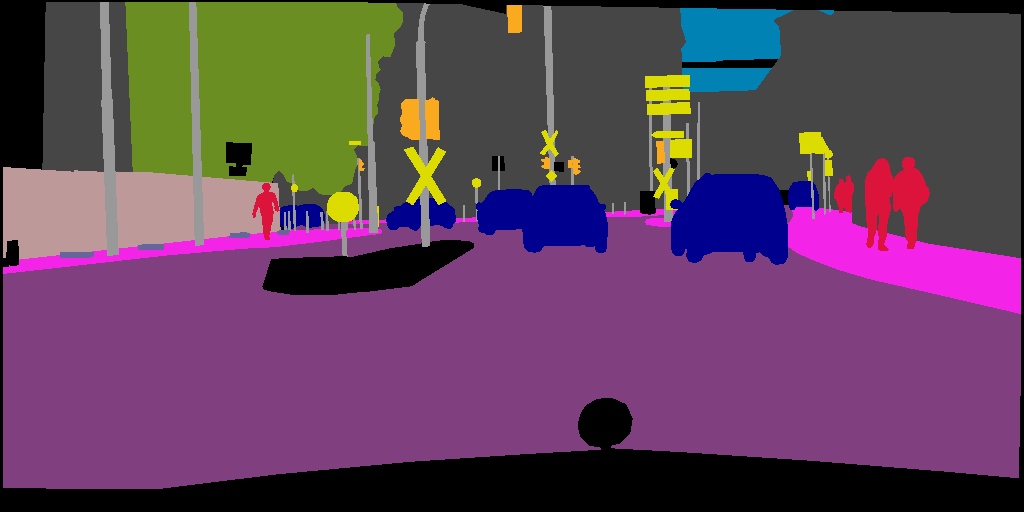}}
\hfill
\mpage{0.17}{\includegraphics[width=1.0\linewidth]{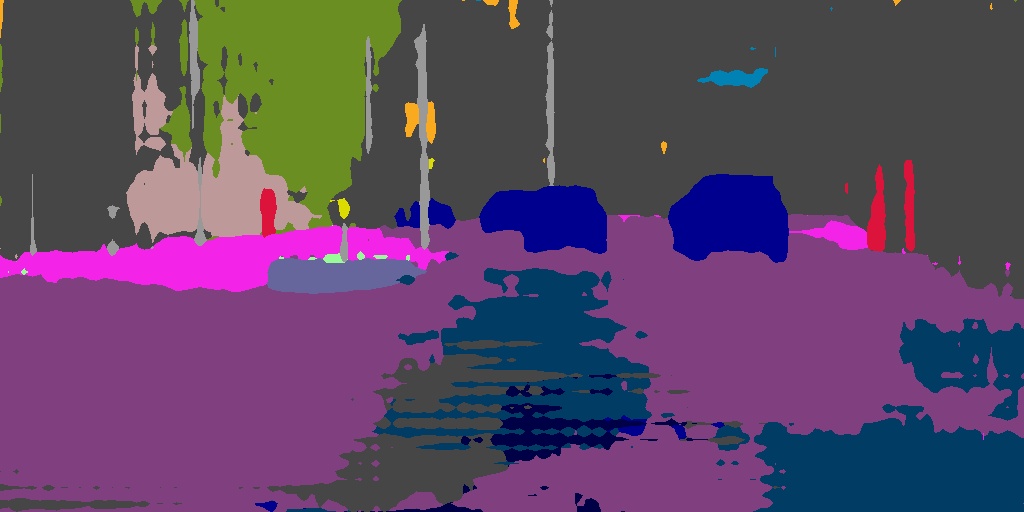}}
\hfill
\mpage{0.17}{\includegraphics[width=1.0\linewidth]{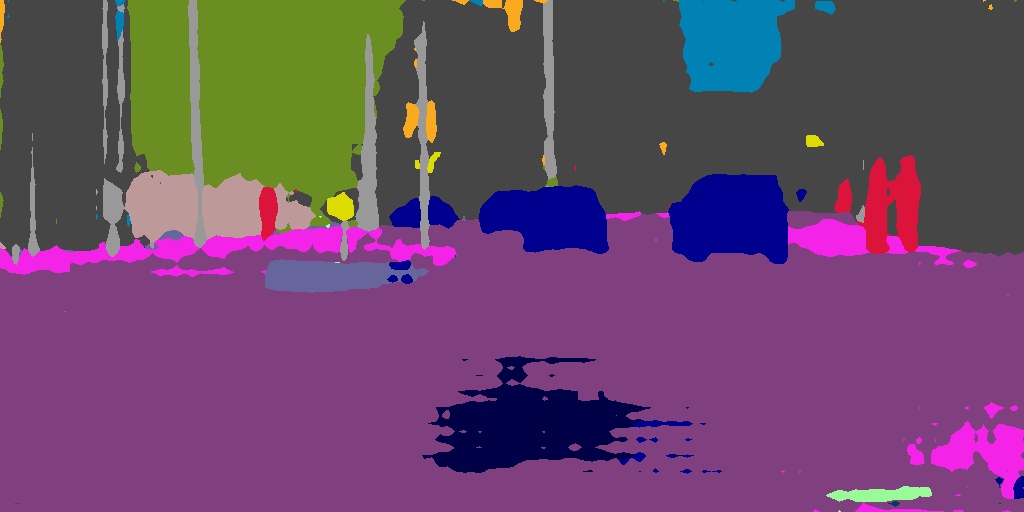}}
\hfill
\mpage{0.17}{\includegraphics[width=1.0\linewidth]{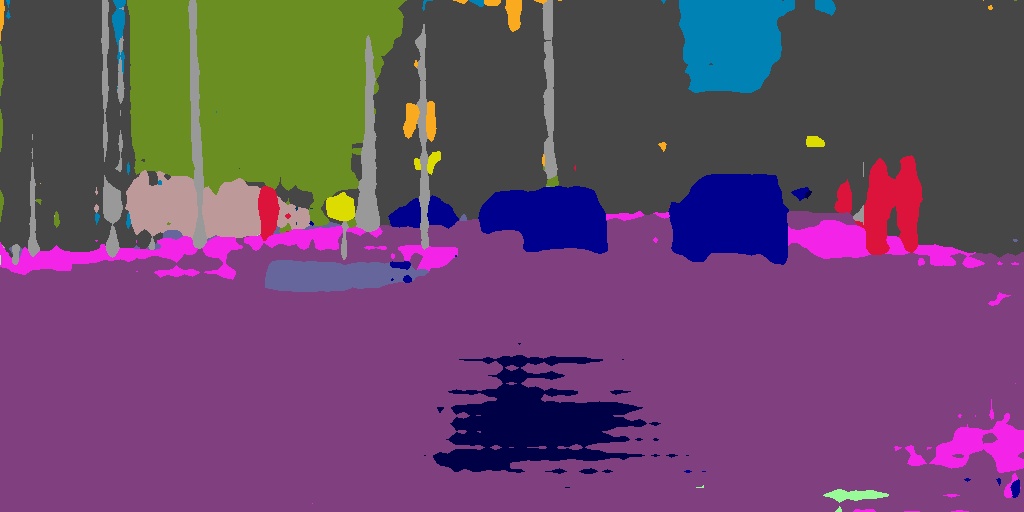}}
\hfill
\\
\mpage{0.17}{\includegraphics[width=1.0\linewidth]{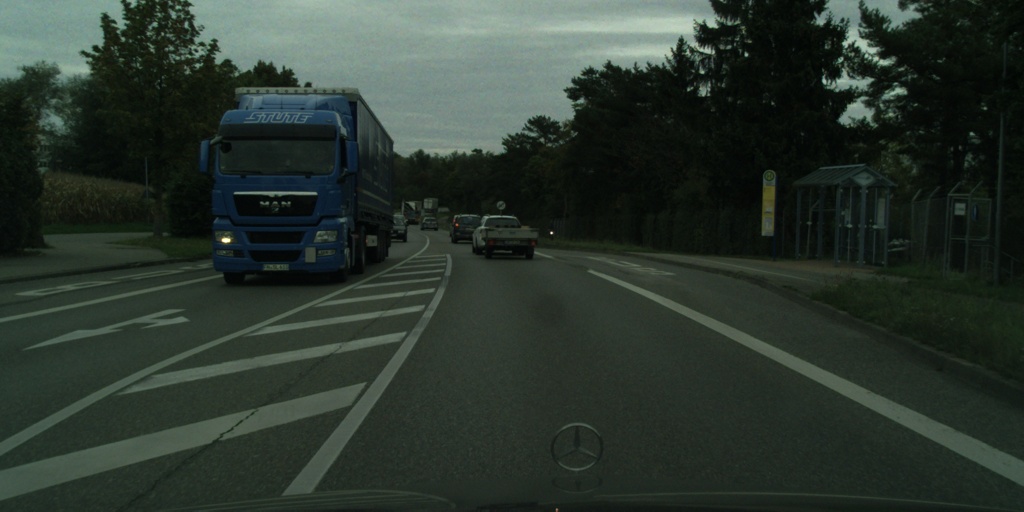}}
\hfill
\mpage{0.17}{\includegraphics[width=1.0\linewidth]{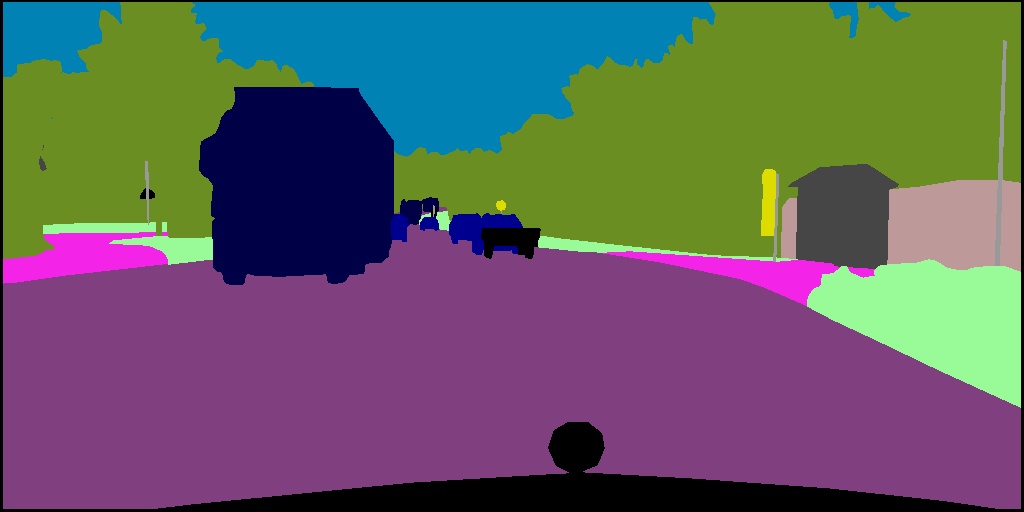}}
\hfill
\mpage{0.17}{\includegraphics[width=1.0\linewidth]{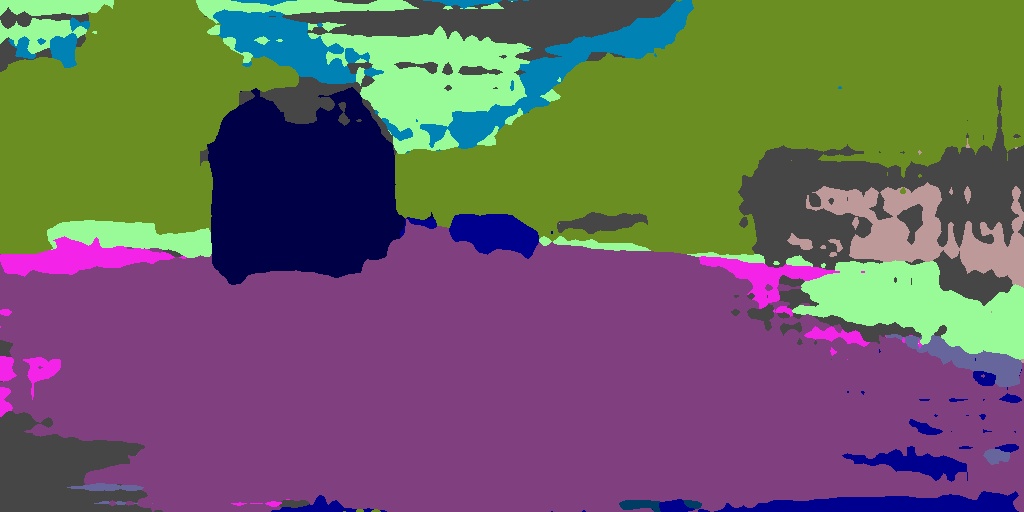}}
\hfill
\mpage{0.17}{\includegraphics[width=1.0\linewidth]{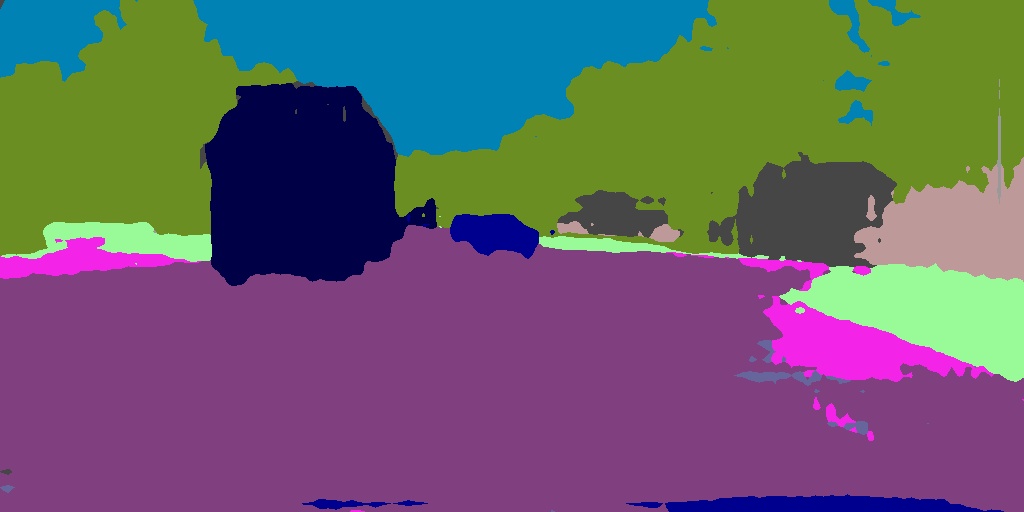}}
\hfill
\mpage{0.17}{\includegraphics[width=1.0\linewidth]{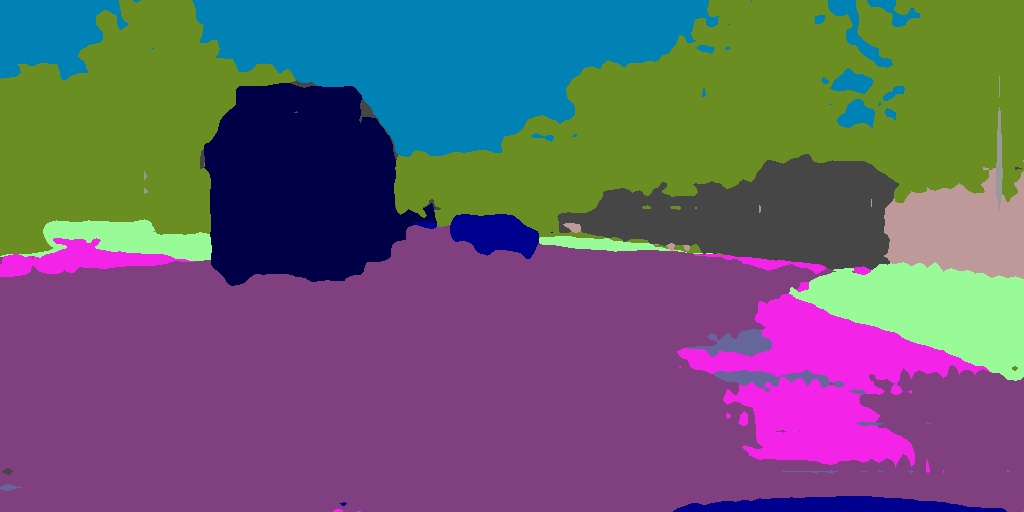}}
\hfill
\\
\mpage{0.17}{\includegraphics[width=1.0\linewidth]{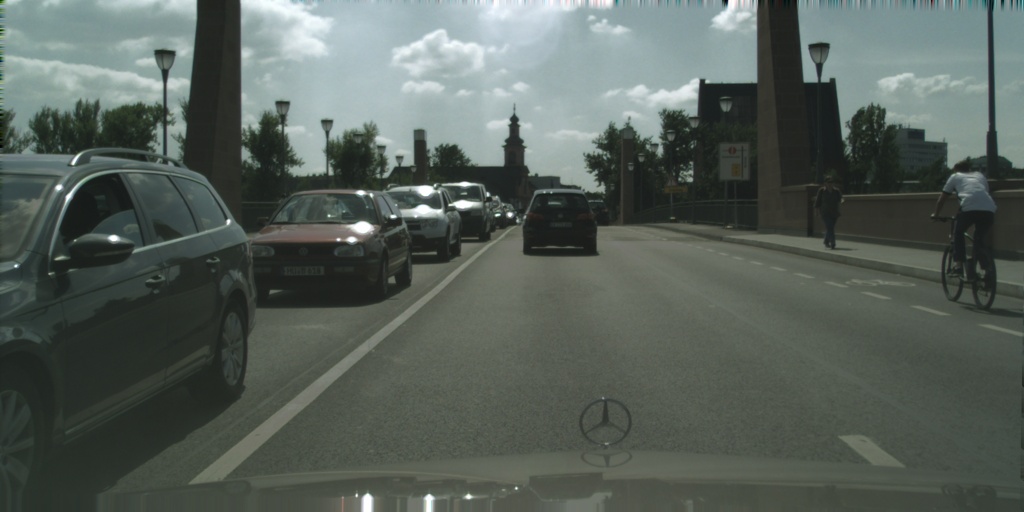}}
\hfill
\mpage{0.17}{\includegraphics[width=1.0\linewidth]{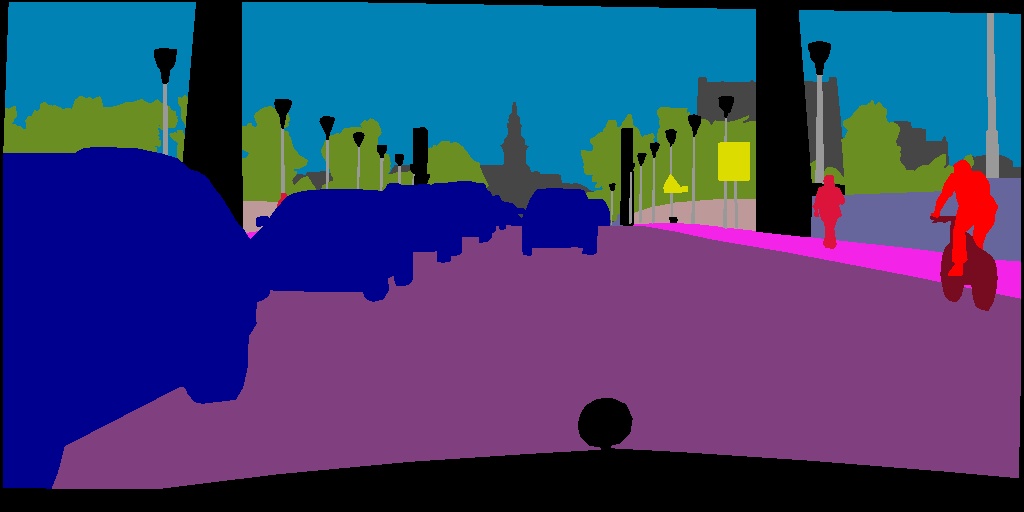}}
\hfill
\mpage{0.17}{\includegraphics[width=1.0\linewidth]{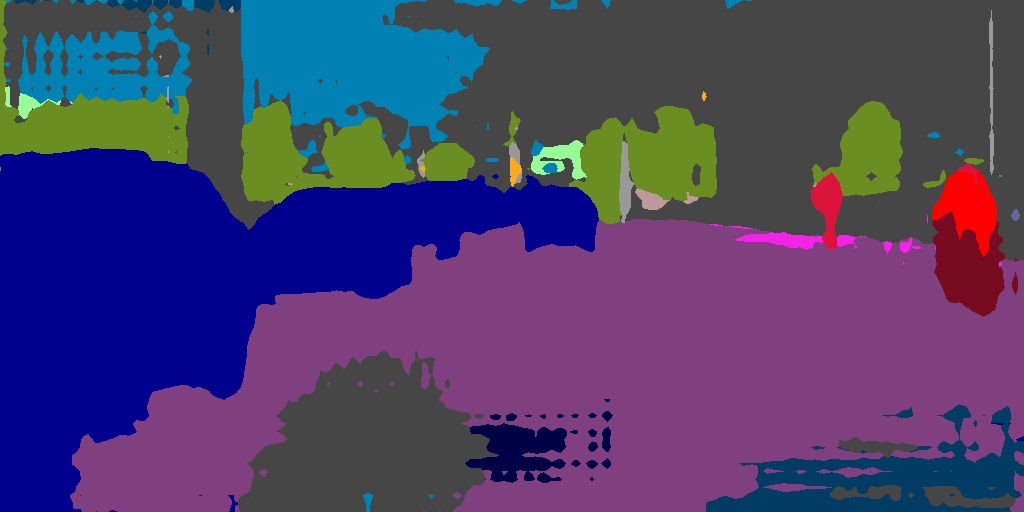}}
\hfill
\mpage{0.17}{\includegraphics[width=1.0\linewidth]{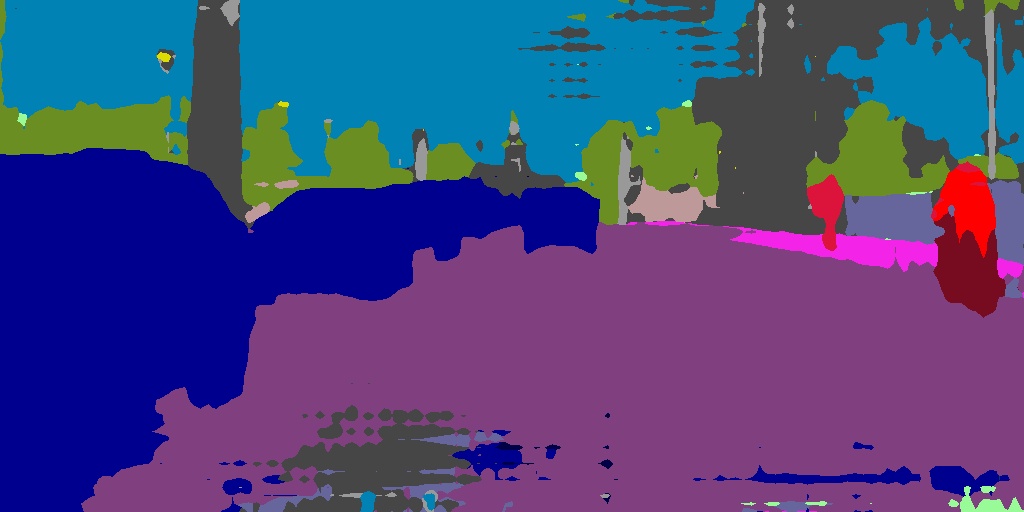}}
\hfill
\mpage{0.17}{\includegraphics[width=1.0\linewidth]{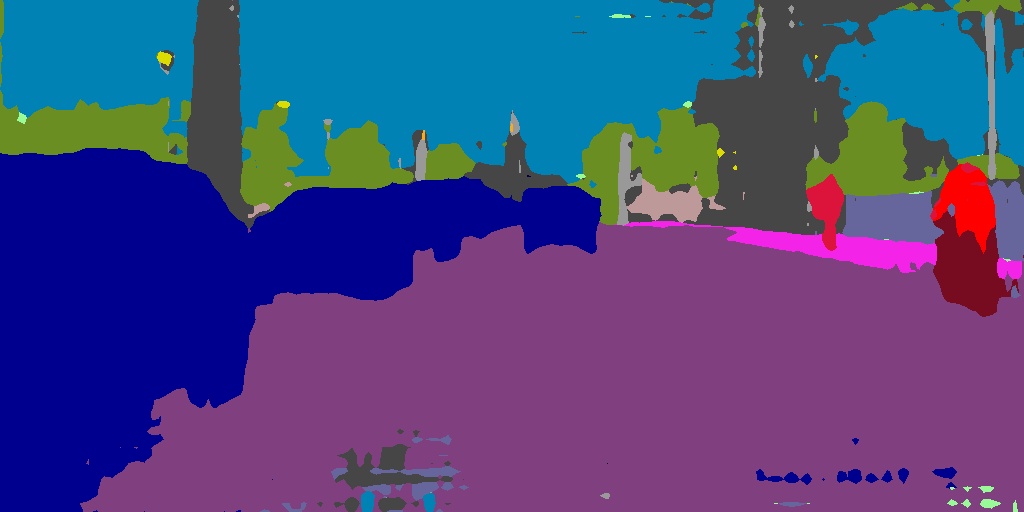}}
\hfill
\\
\mpage{0.17}{\includegraphics[width=1.0\linewidth]{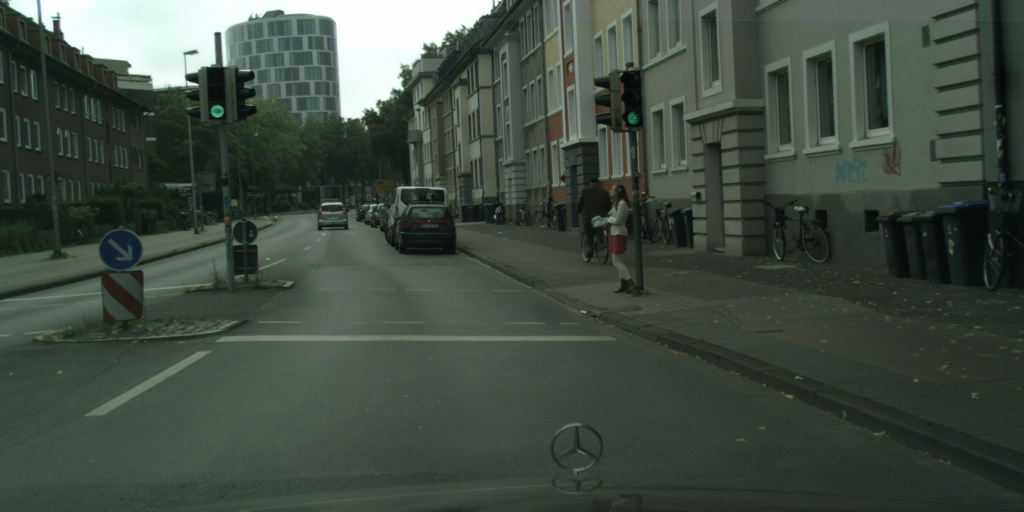}}
\hfill
\mpage{0.17}{\includegraphics[width=1.0\linewidth]{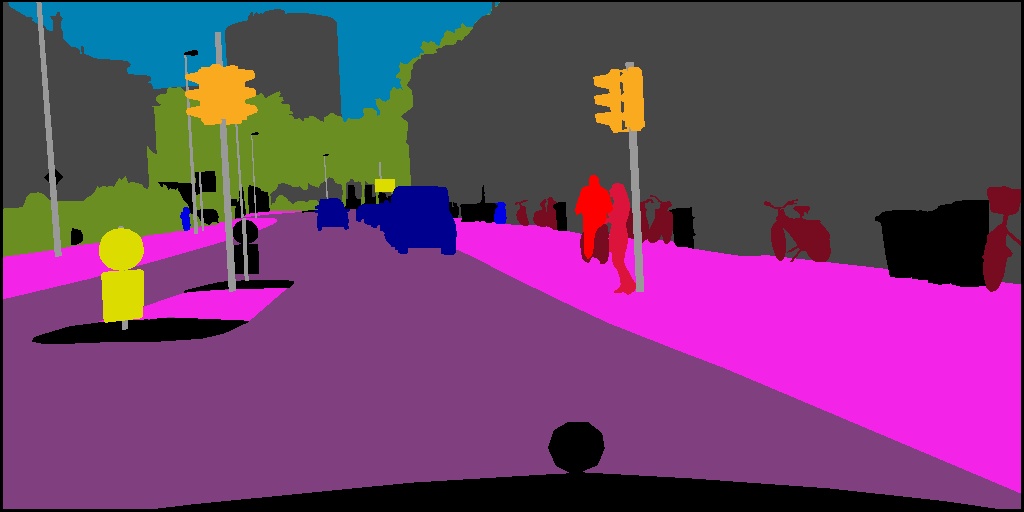}}
\hfill
\mpage{0.17}{\includegraphics[width=1.0\linewidth]{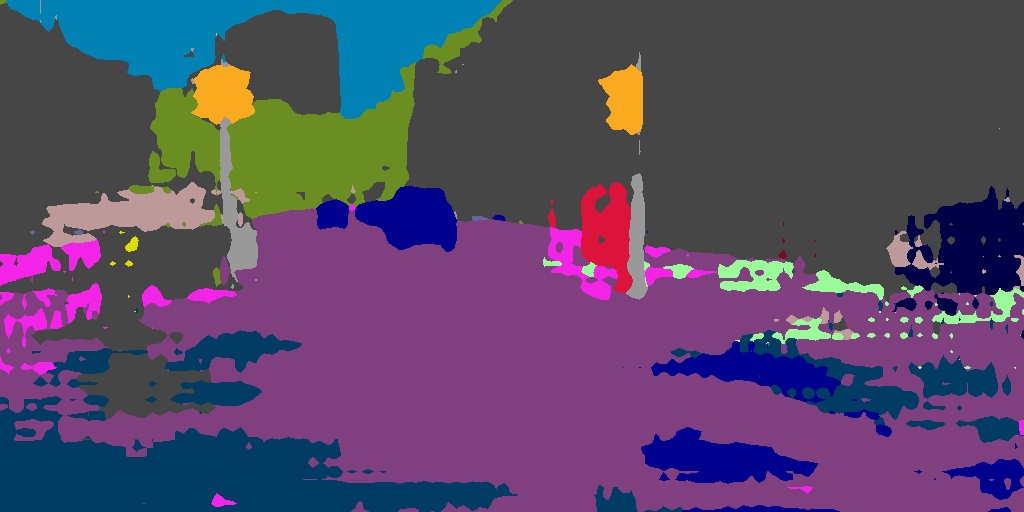}}
\hfill
\mpage{0.17}{\includegraphics[width=1.0\linewidth]{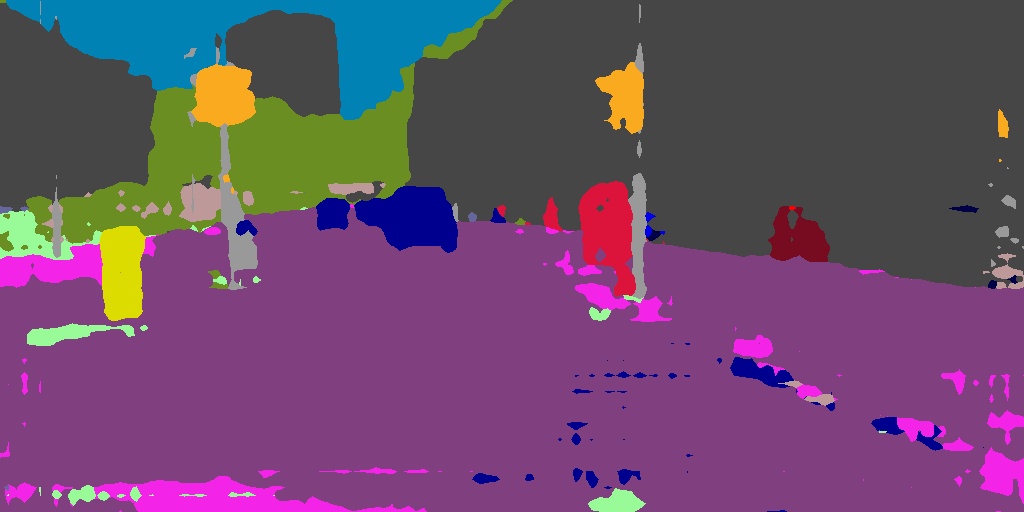}}
\hfill
\mpage{0.17}{\includegraphics[width=1.0\linewidth]{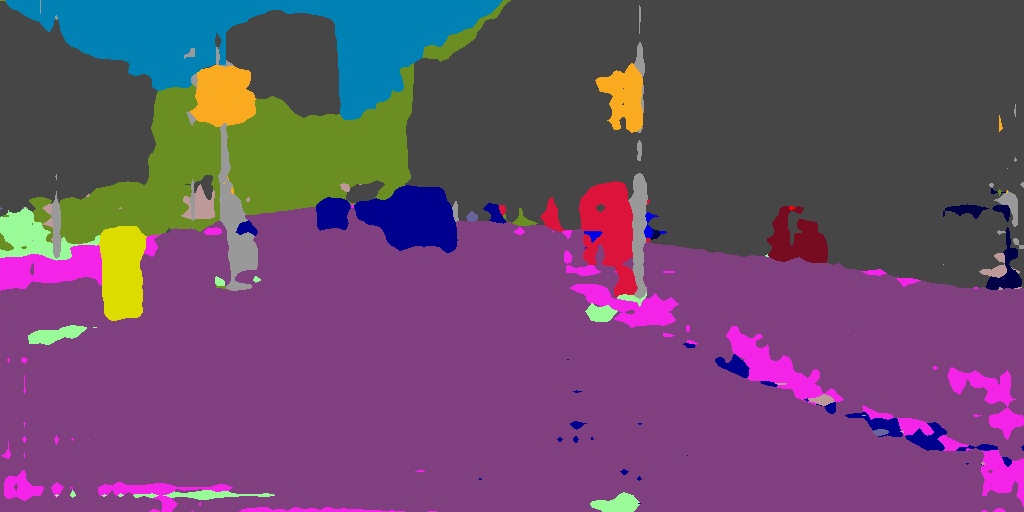}}
\hfill
\\
\mpage{0.17}{\includegraphics[width=1.0\linewidth]{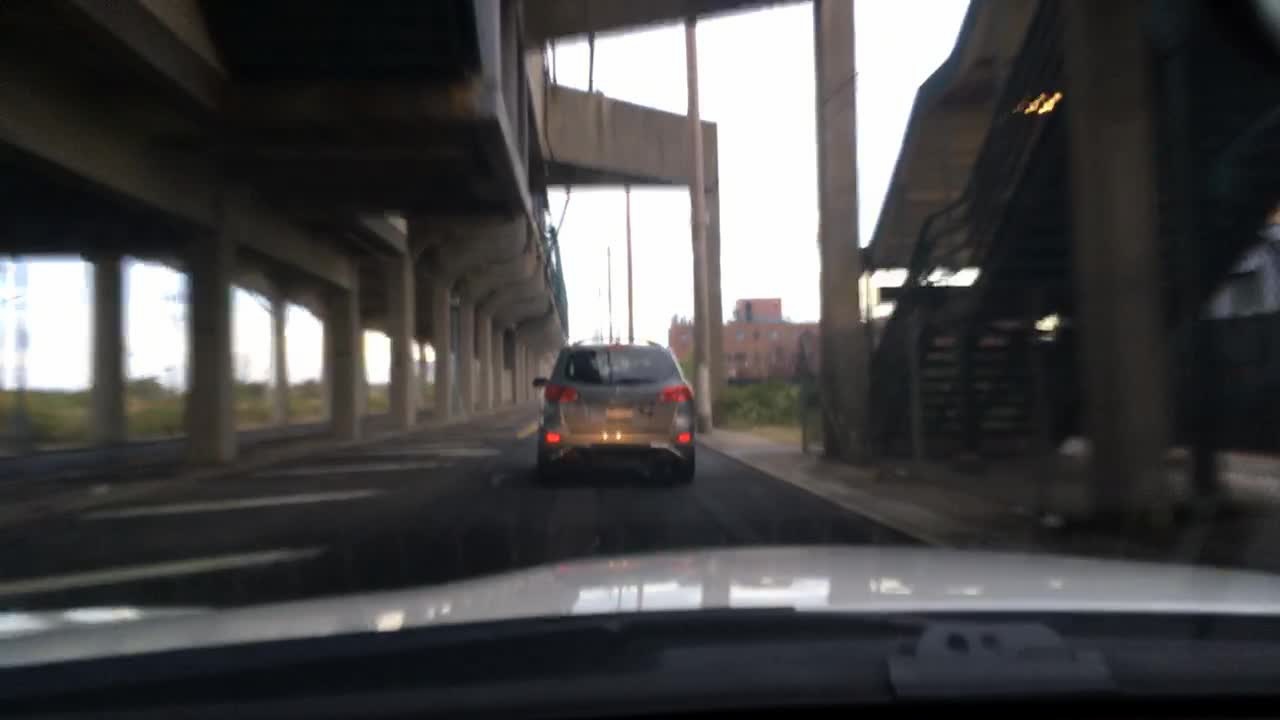}}
\hfill
\mpage{0.17}{\includegraphics[width=1.0\linewidth]{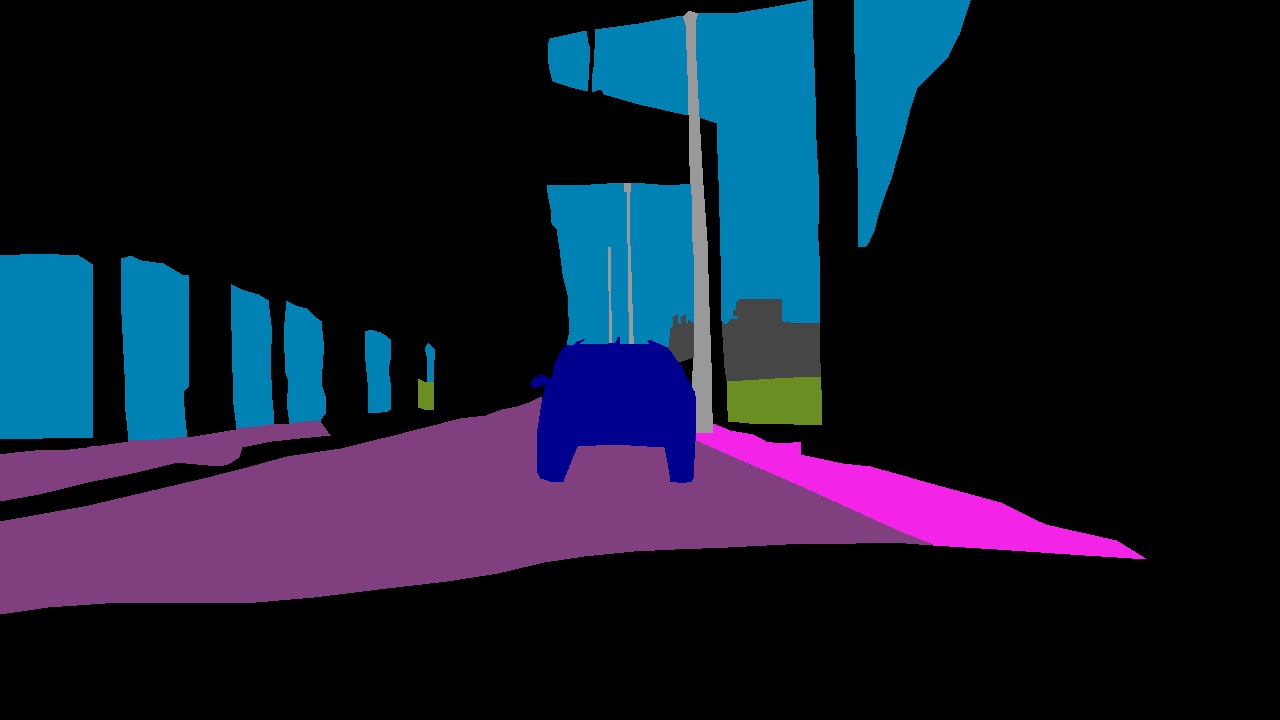}}
\hfill
\mpage{0.17}{\includegraphics[width=1.0\linewidth]{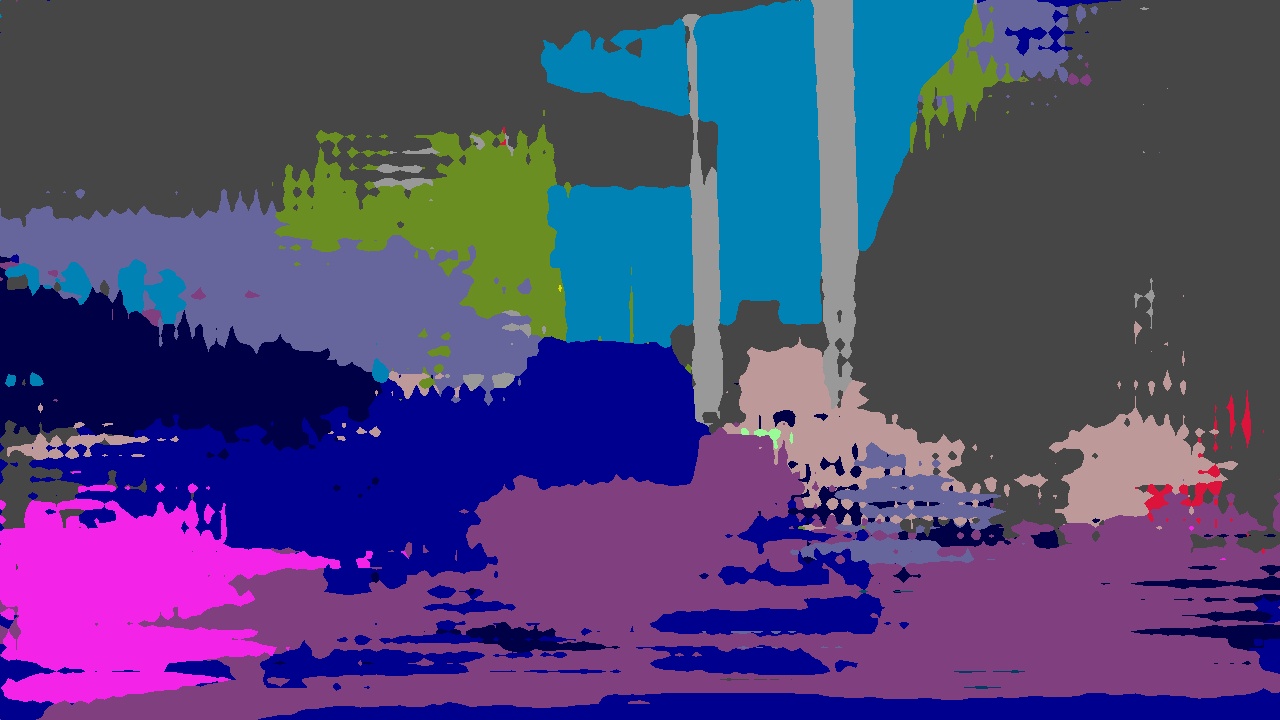}}
\hfill
\mpage{0.17}{\includegraphics[width=1.0\linewidth]{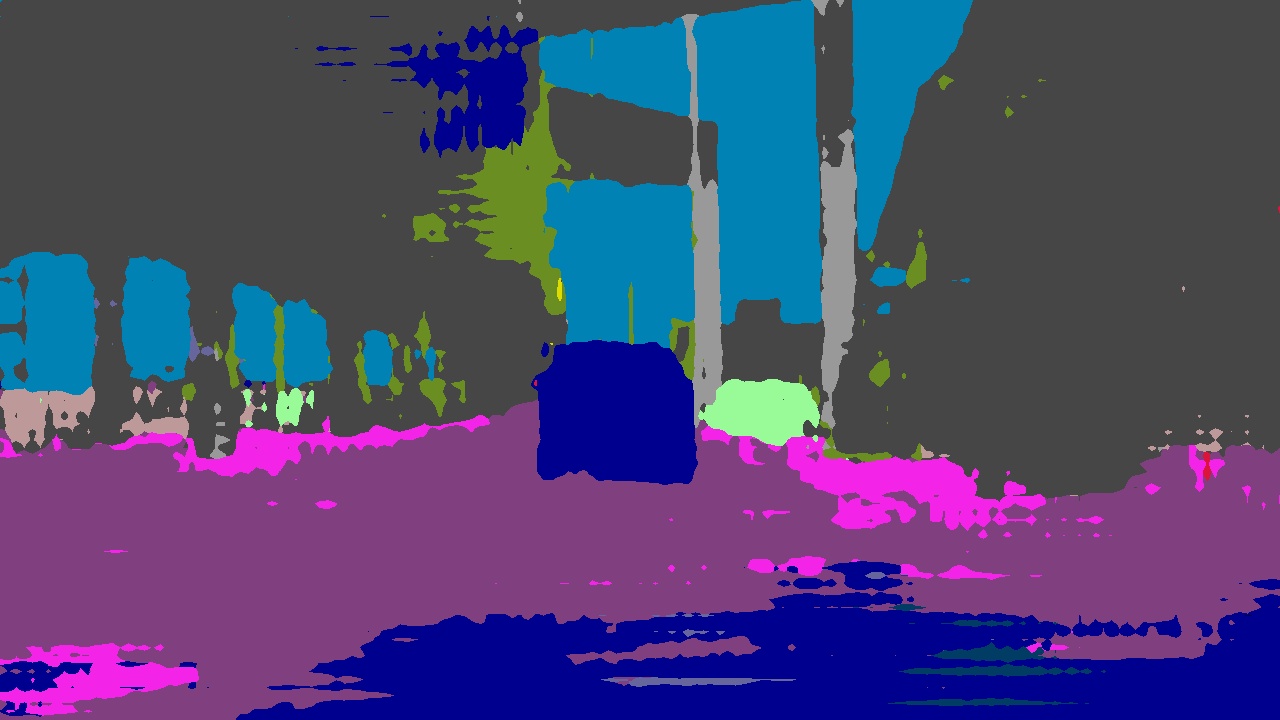}}
\hfill
\mpage{0.17}{\includegraphics[width=1.0\linewidth]{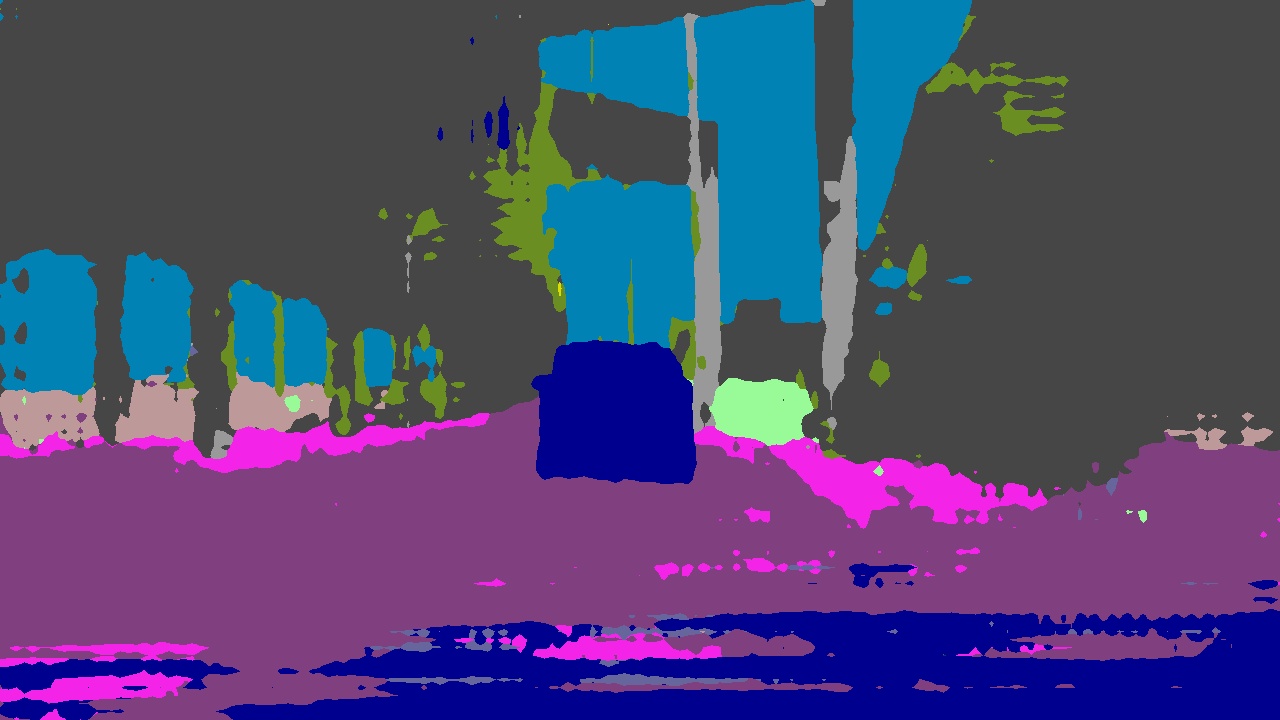}}
\hfill
\\
\mpage{0.17}{\includegraphics[width=1.0\linewidth]{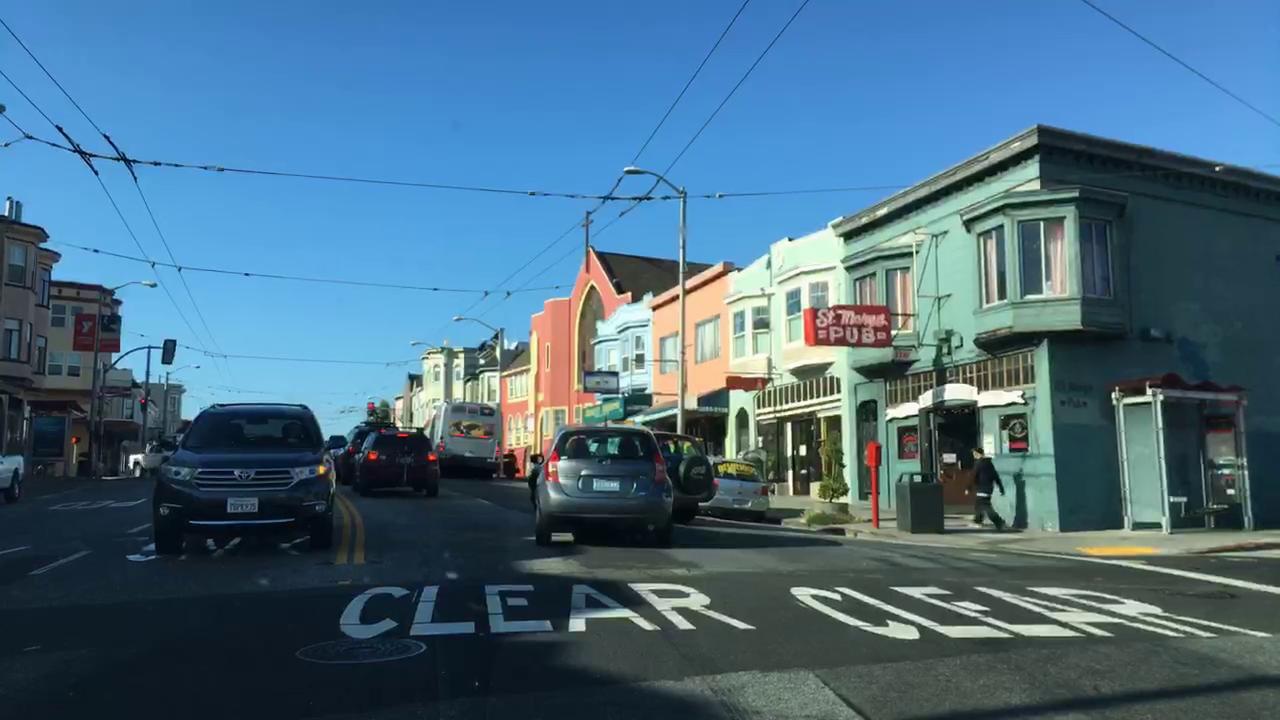}}
\hfill
\mpage{0.17}{\includegraphics[width=1.0\linewidth]{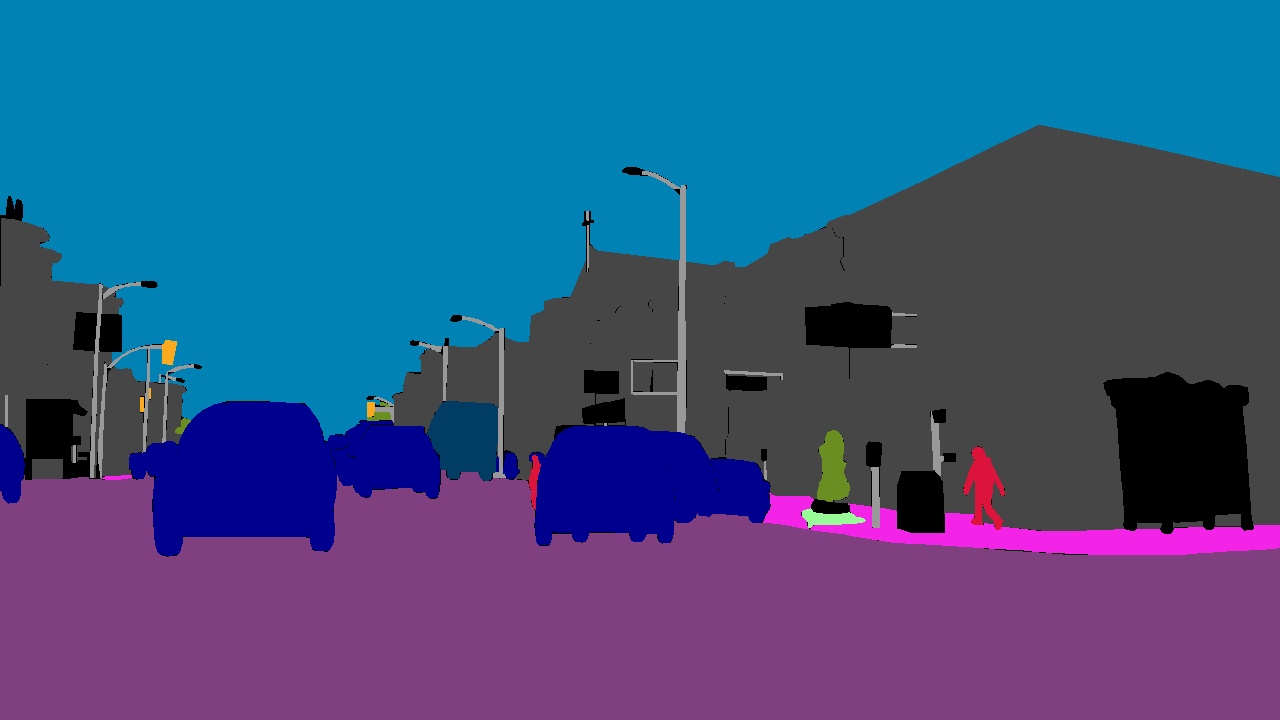}}
\hfill
\mpage{0.17}{\includegraphics[width=1.0\linewidth]{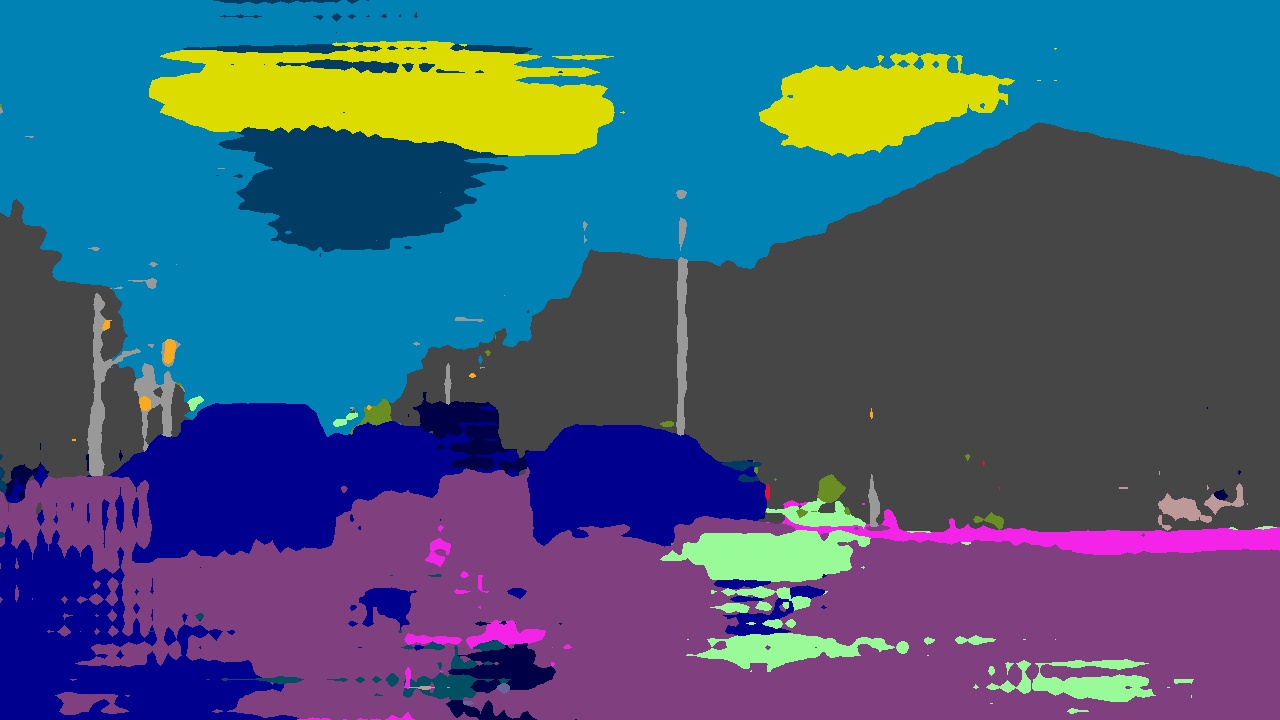}}
\hfill
\mpage{0.17}{\includegraphics[width=1.0\linewidth]{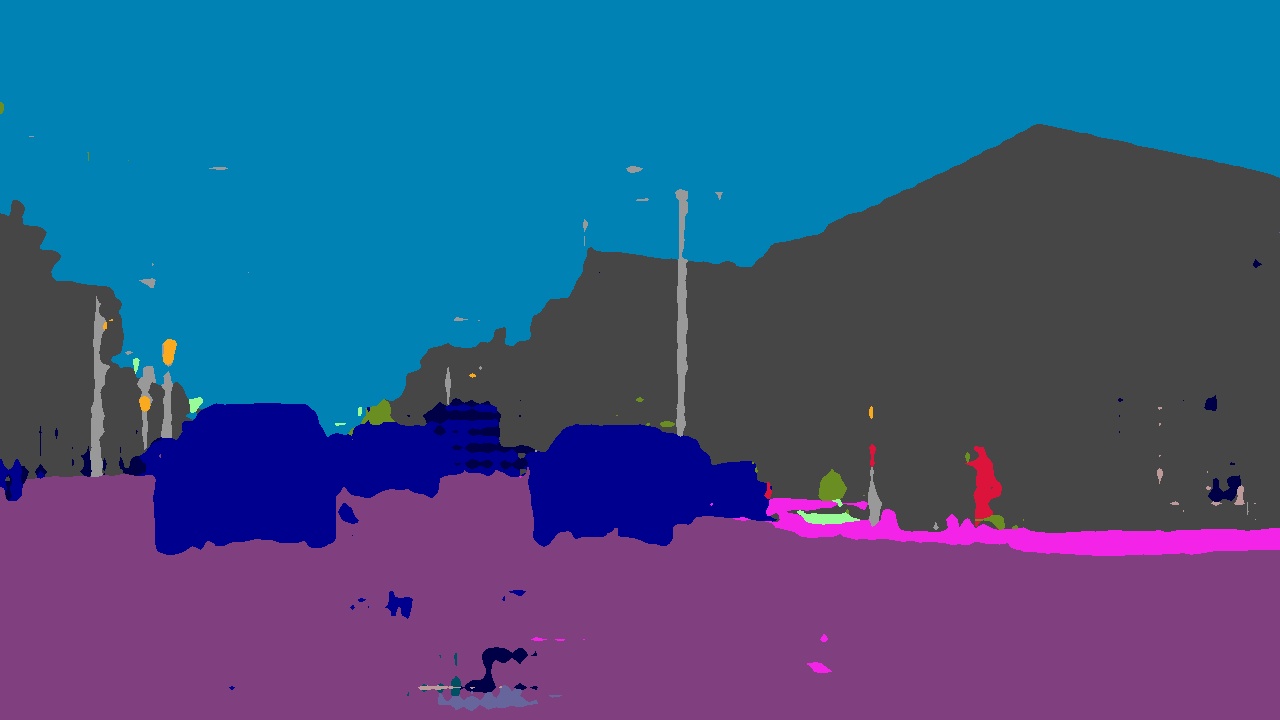}}
\hfill
\mpage{0.17}{\includegraphics[width=1.0\linewidth]{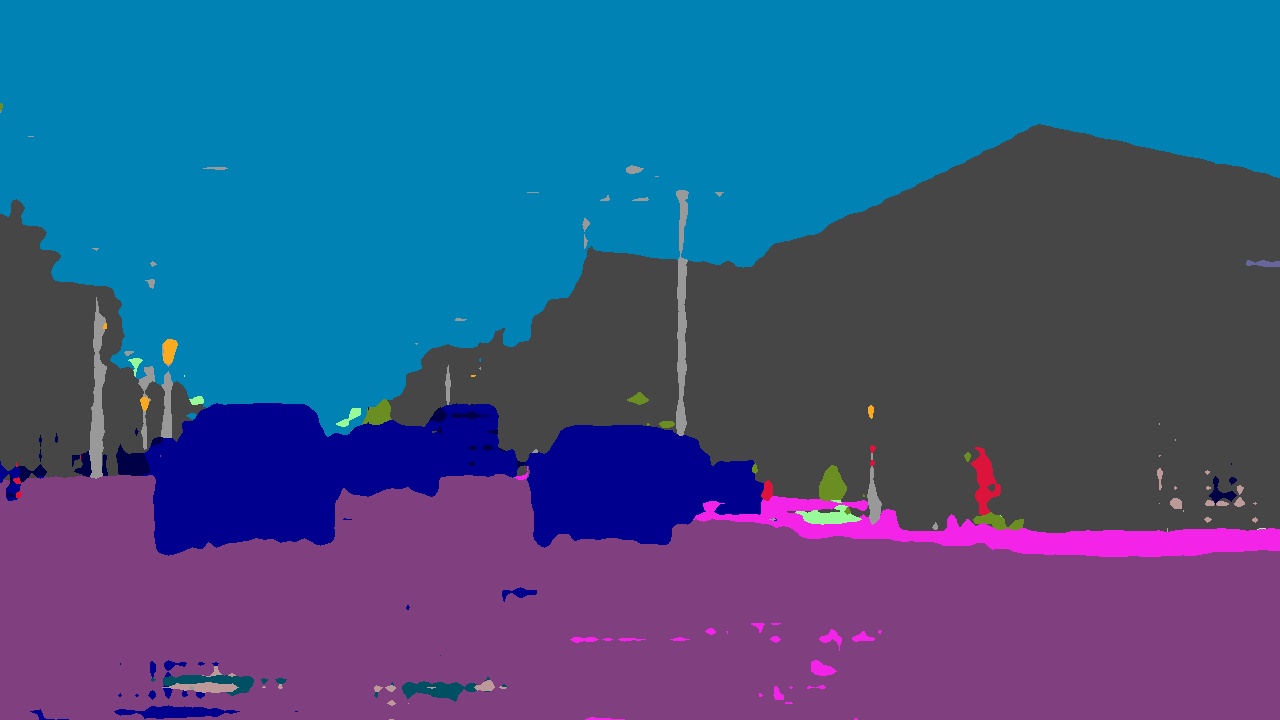}}
\hfill
\\
\mpage{0.17}{\includegraphics[width=1.0\linewidth]{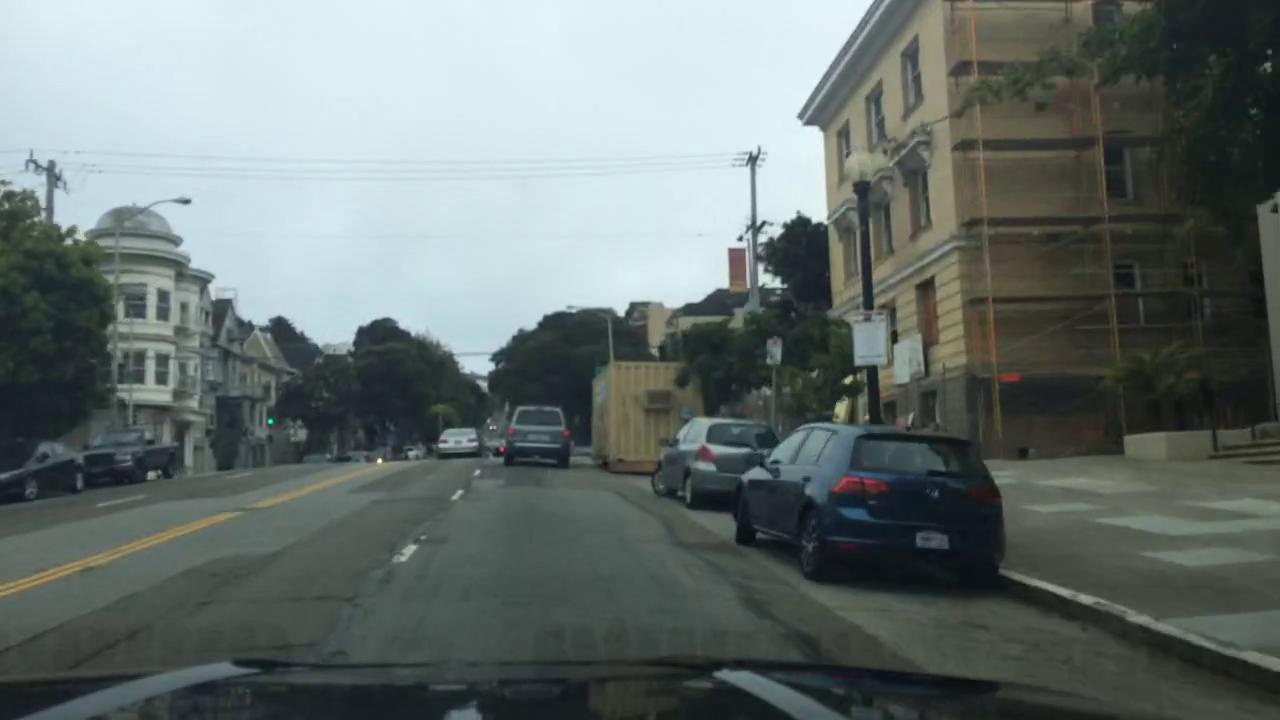}}
\hfill
\mpage{0.17}{\includegraphics[width=1.0\linewidth]{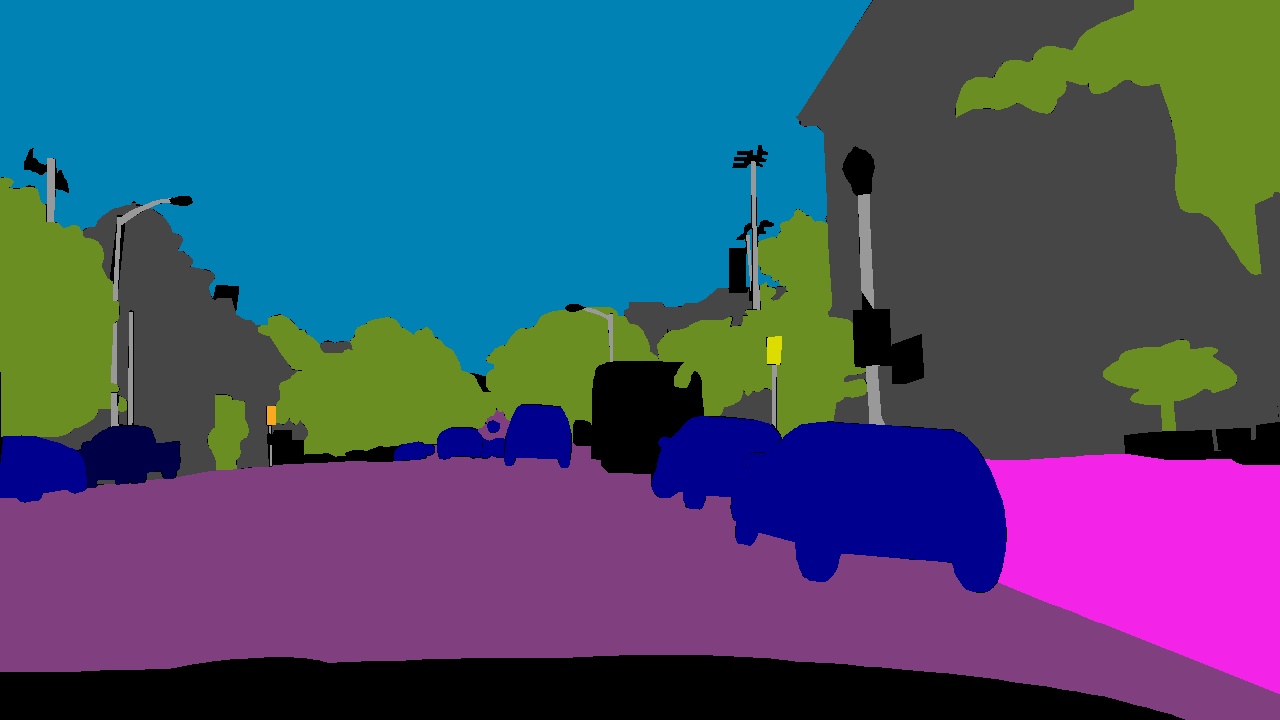}}
\hfill
\mpage{0.17}{\includegraphics[width=1.0\linewidth]{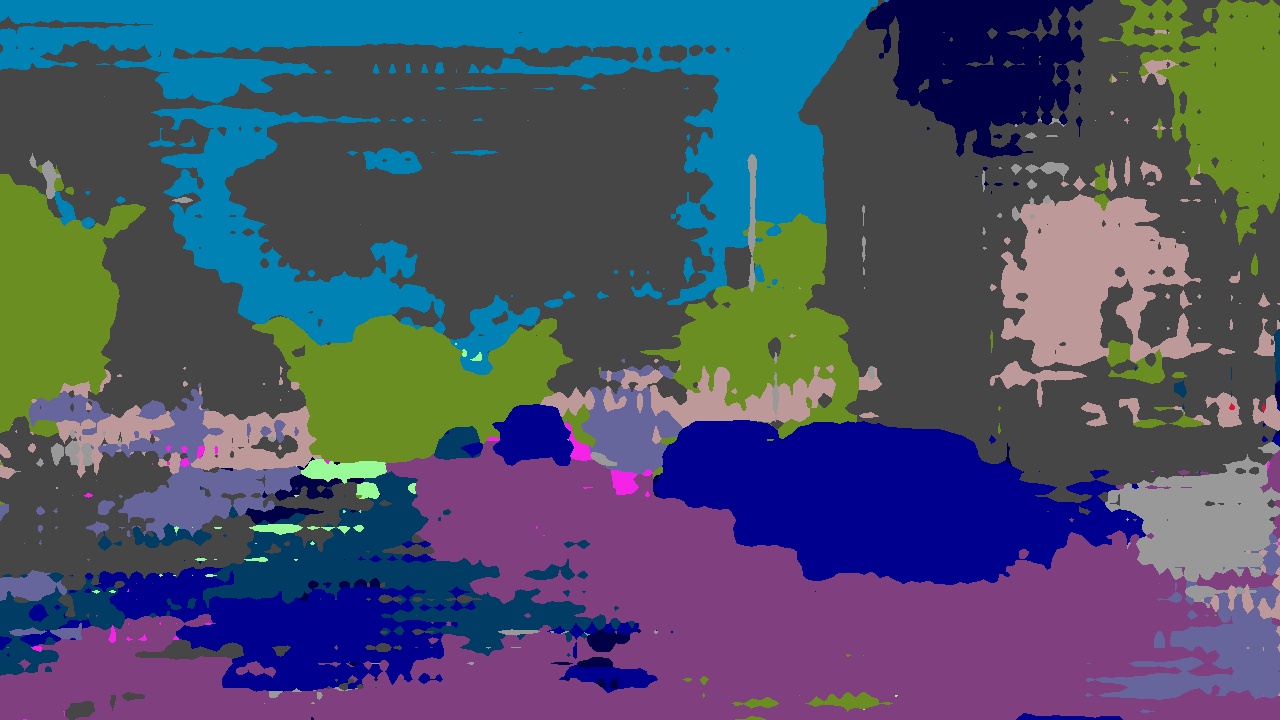}}
\hfill
\mpage{0.17}{\includegraphics[width=1.0\linewidth]{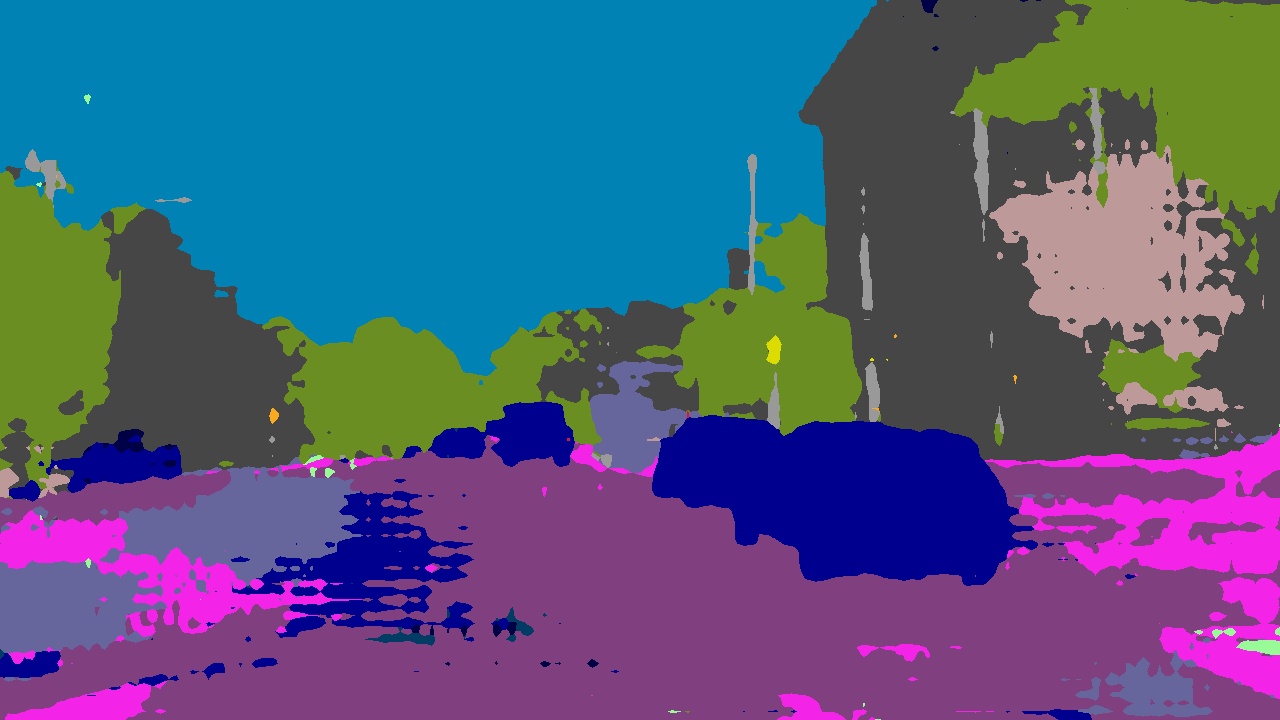}}
\hfill
\mpage{0.17}{\includegraphics[width=1.0\linewidth]{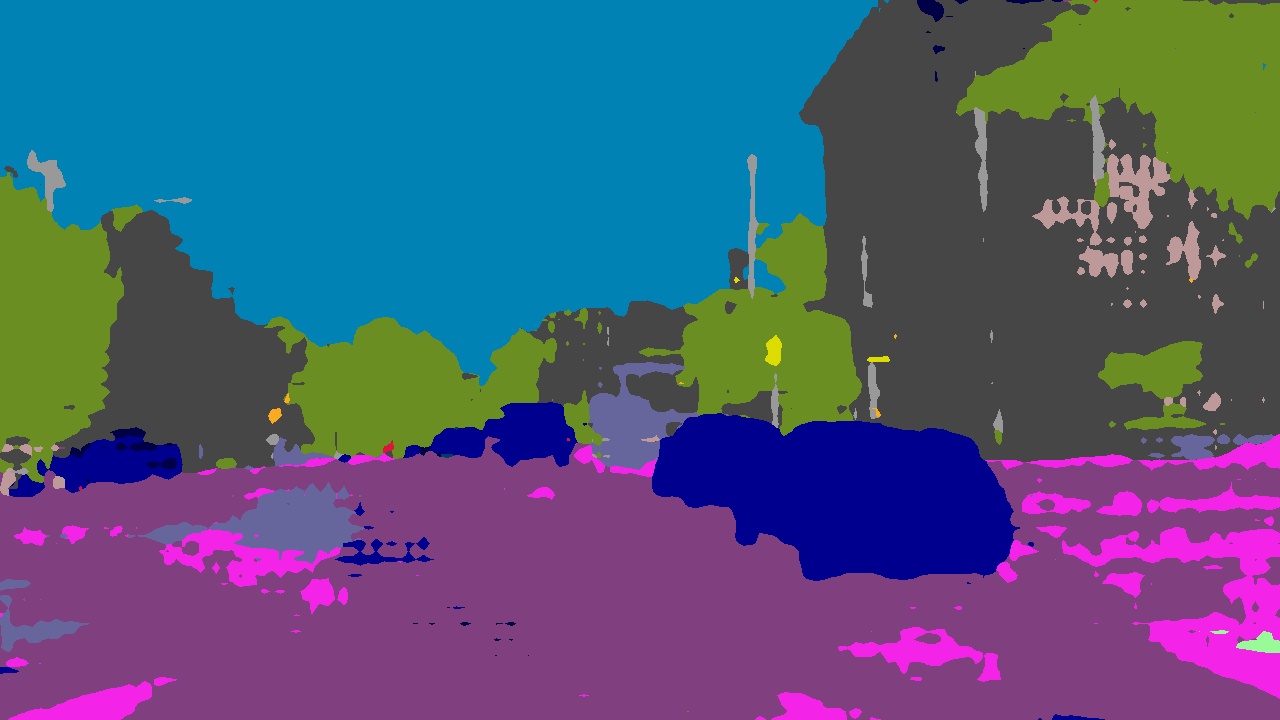}}
\hfill
\\
\mpage{0.17}{\includegraphics[width=1.0\linewidth]{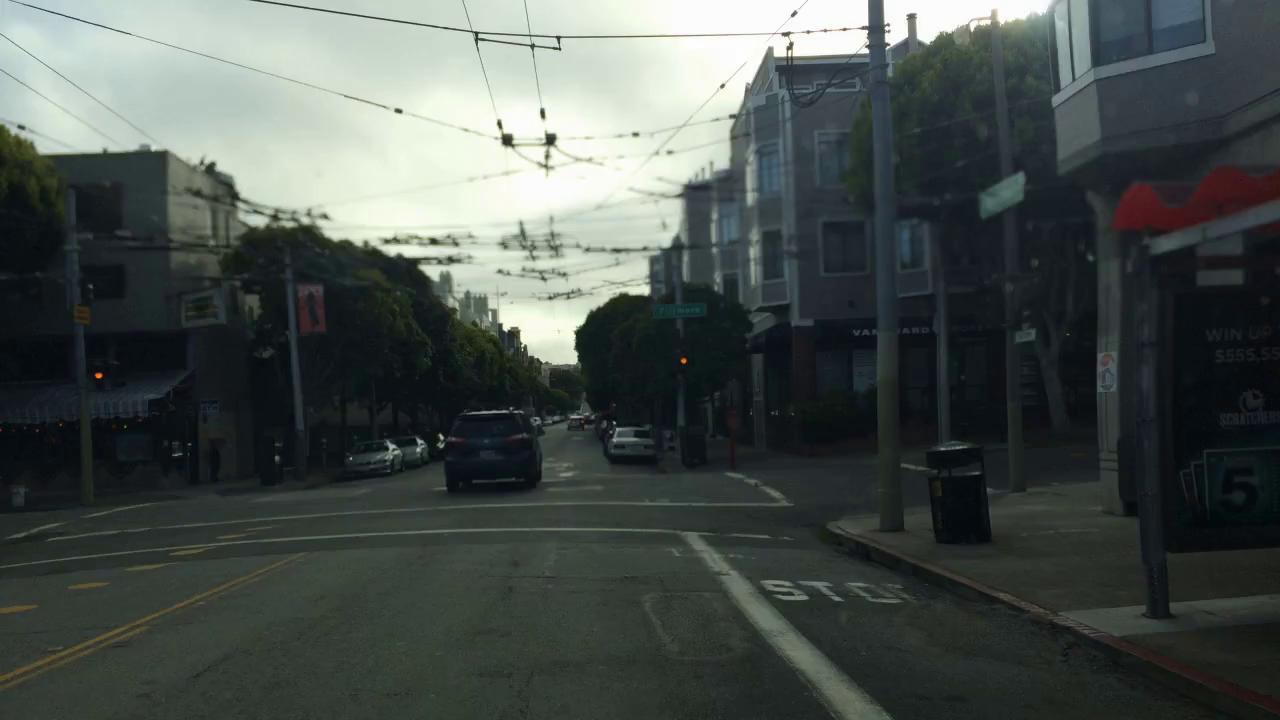}}
\hfill
\mpage{0.17}{\includegraphics[width=1.0\linewidth]{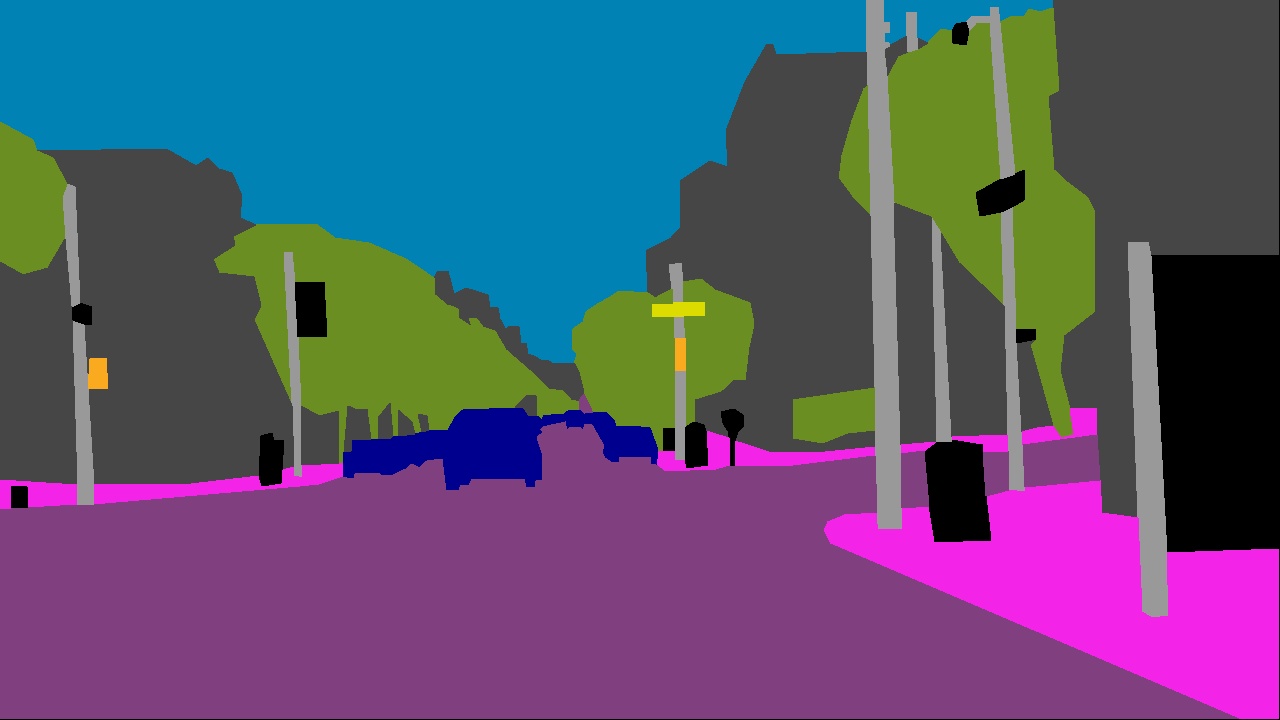}}
\hfill
\mpage{0.17}{\includegraphics[width=1.0\linewidth]{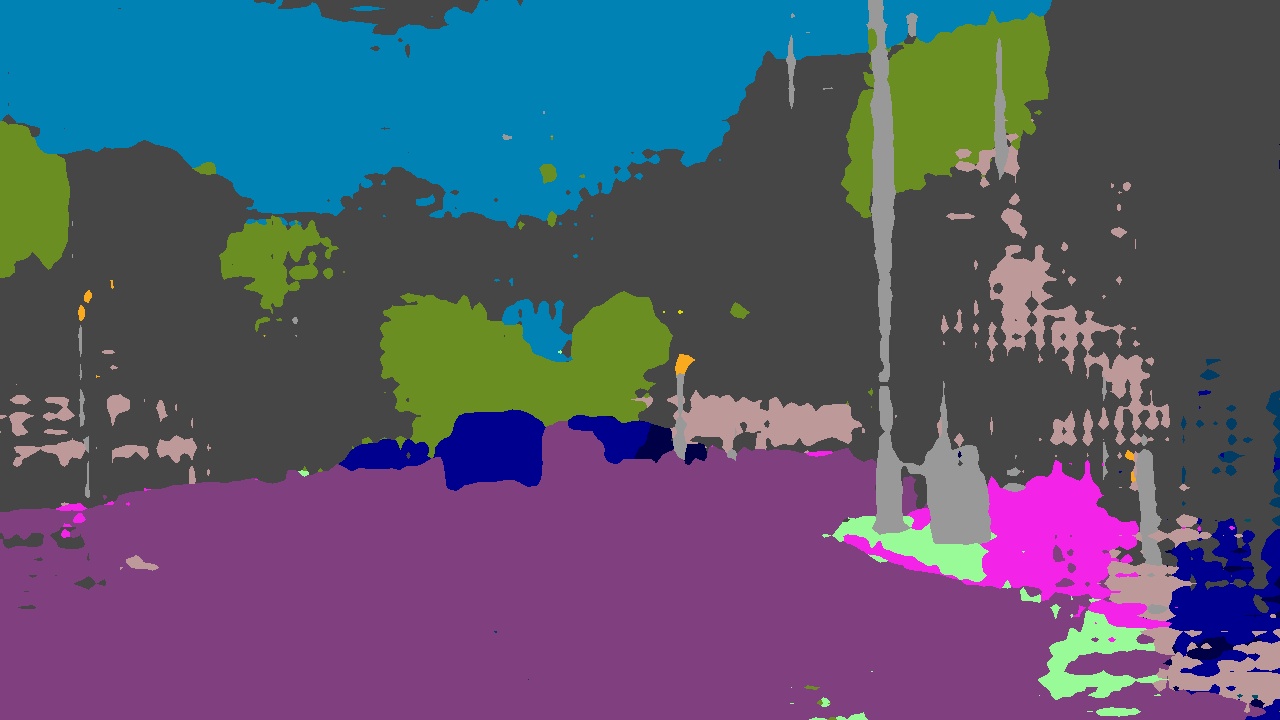}}
\hfill
\mpage{0.17}{\includegraphics[width=1.0\linewidth]{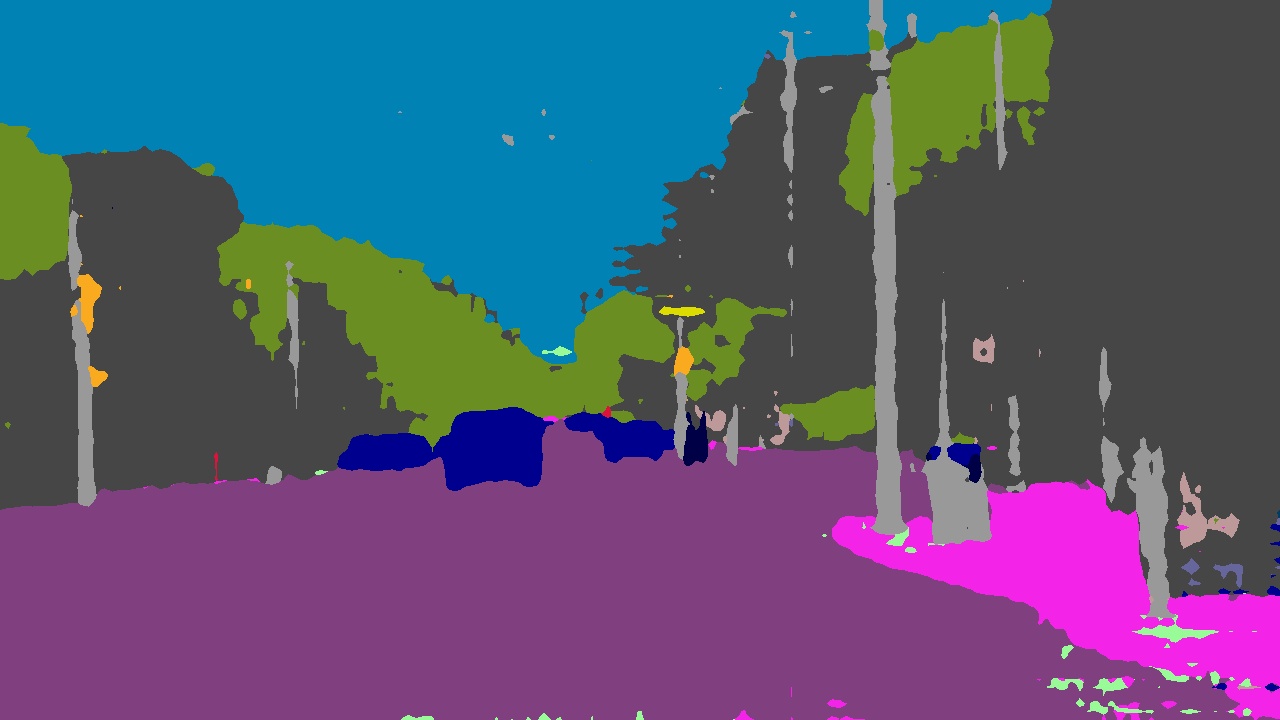}}
\hfill
\mpage{0.17}{\includegraphics[width=1.0\linewidth]{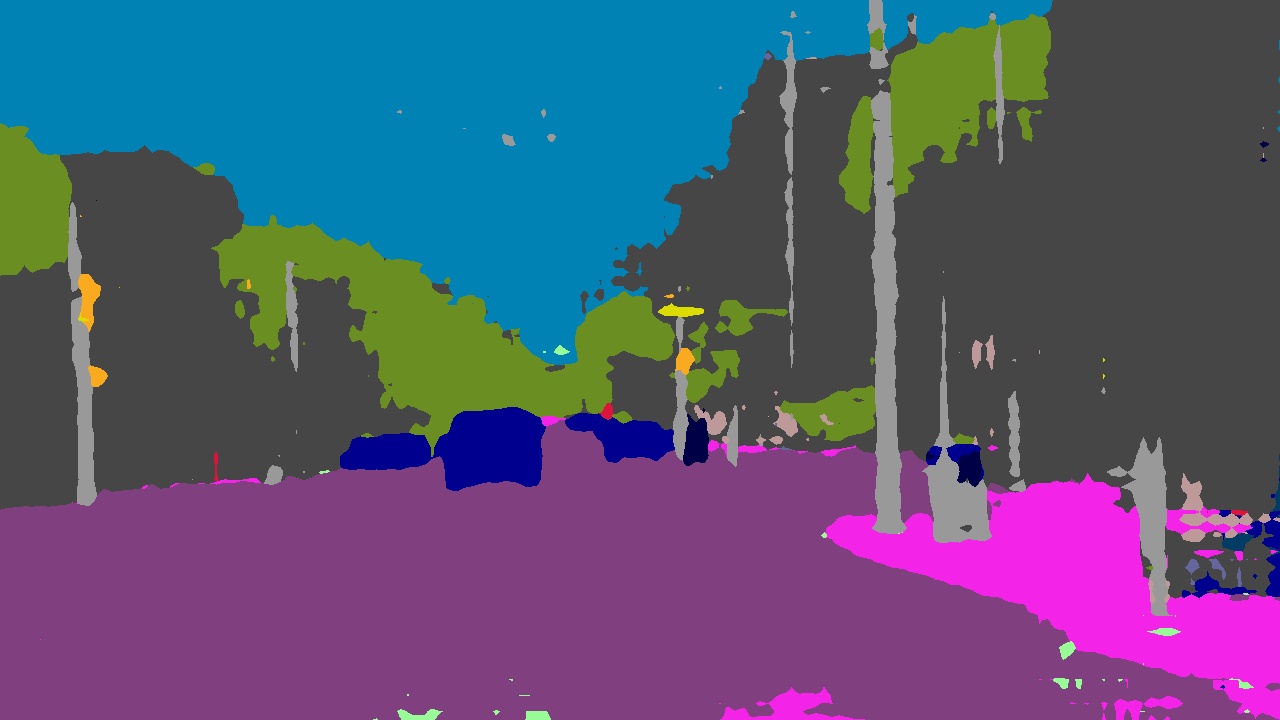}}
\hfill
\\
\mpage{0.17}{\includegraphics[width=1.0\linewidth]{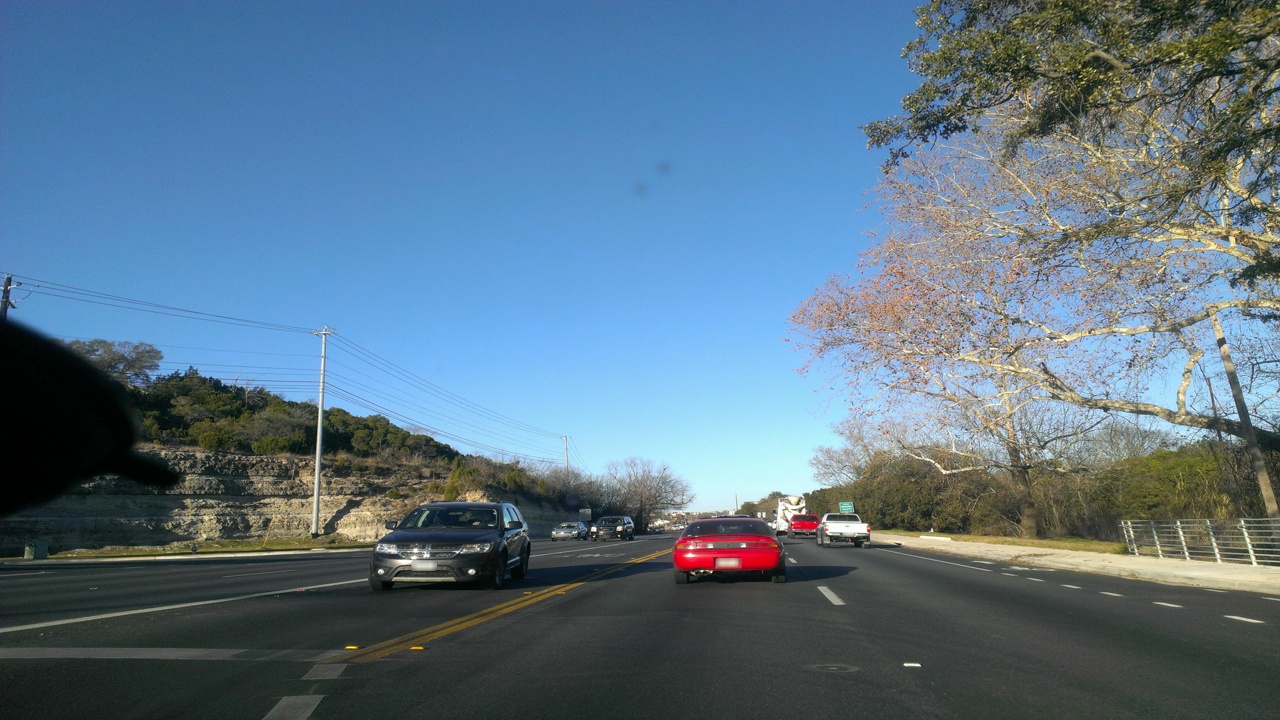}}
\hfill
\mpage{0.17}{\includegraphics[width=1.0\linewidth]{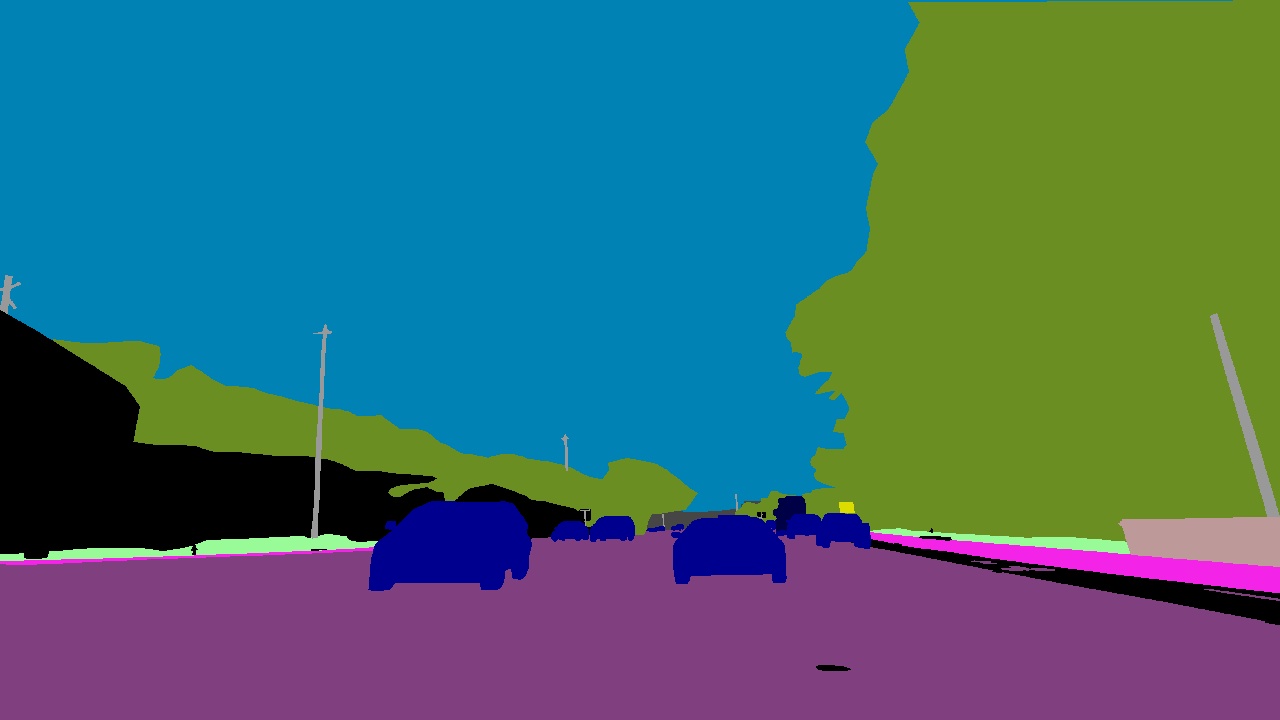}}
\hfill
\mpage{0.17}{\includegraphics[width=1.0\linewidth]{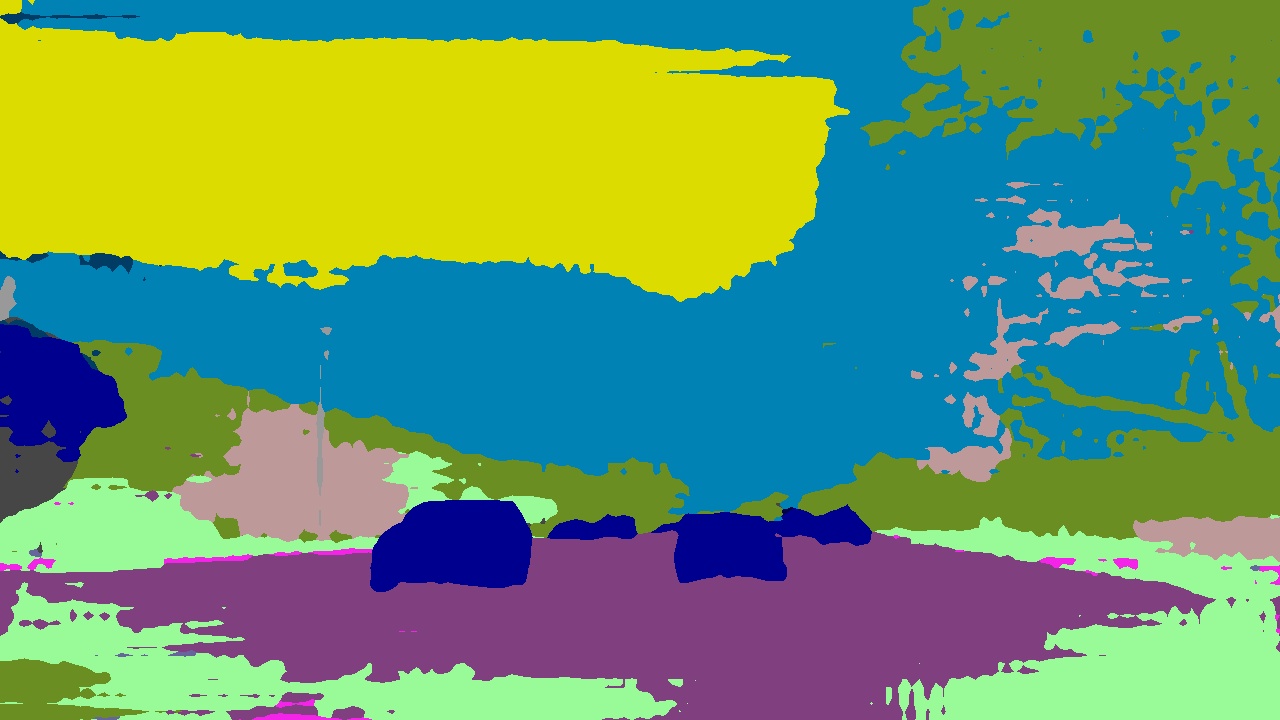}}
\hfill
\mpage{0.17}{\includegraphics[width=1.0\linewidth]{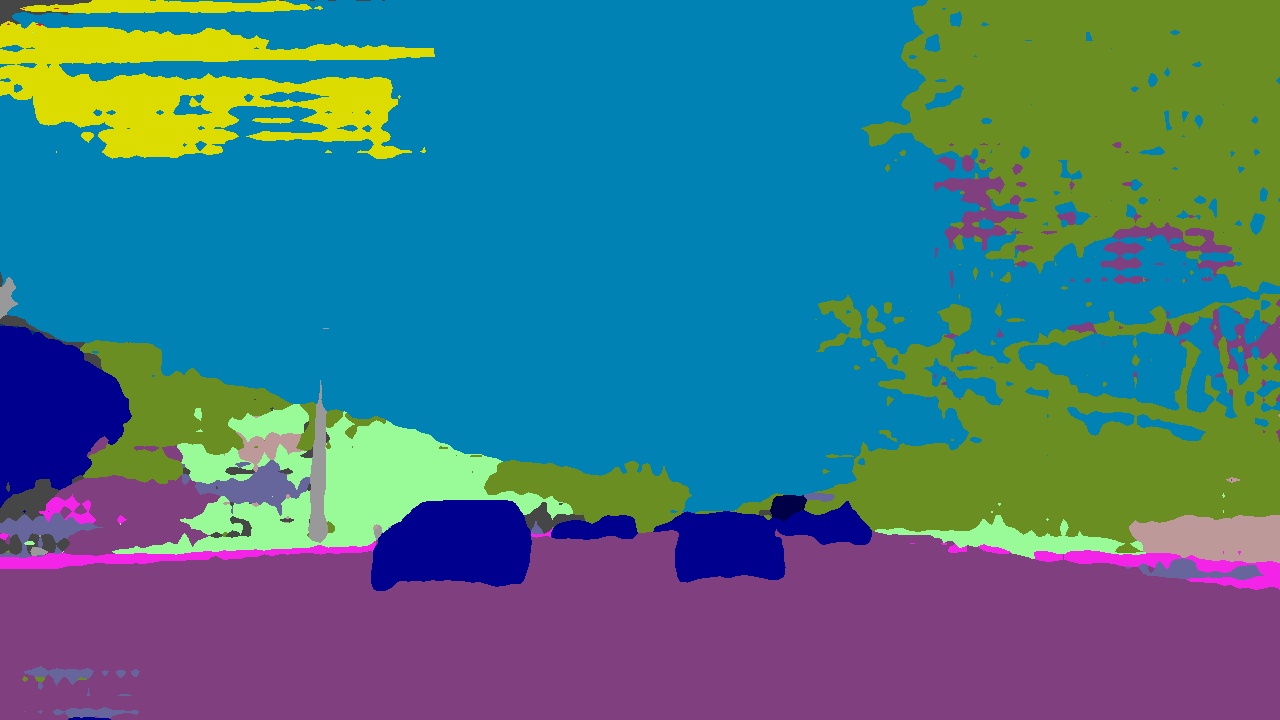}}
\hfill
\mpage{0.17}{\includegraphics[width=1.0\linewidth]{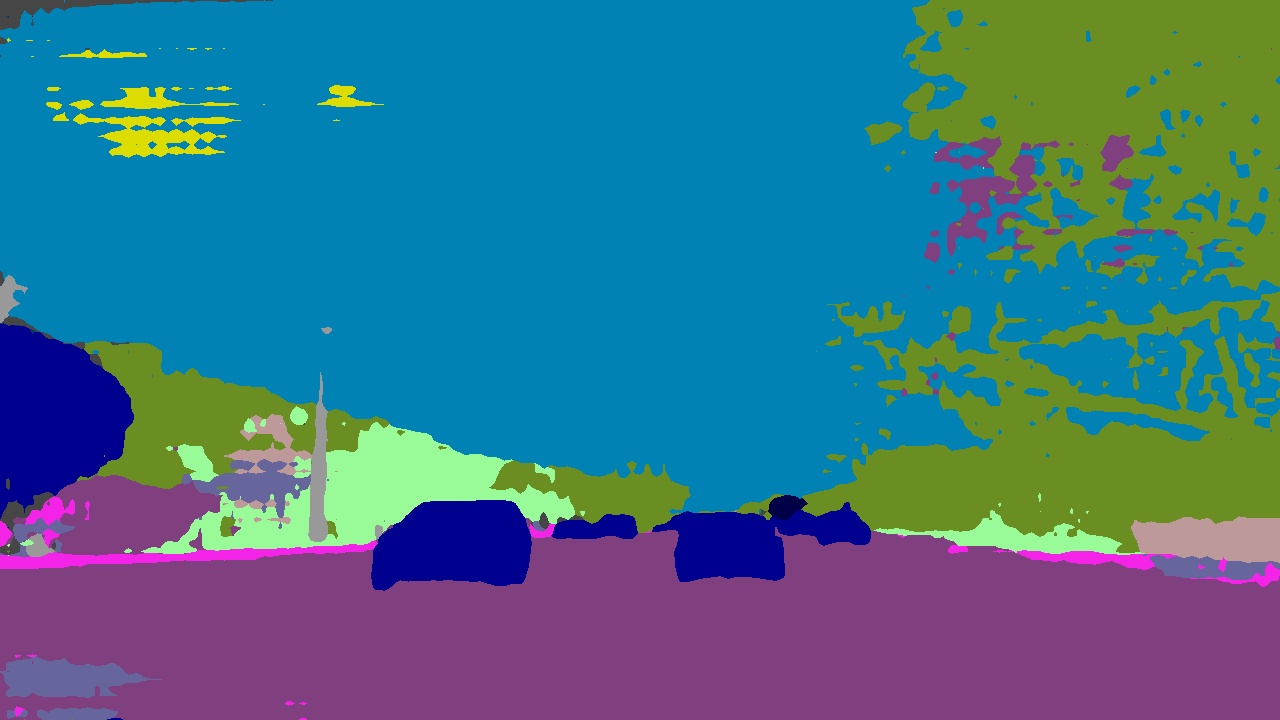}}
\hfill
\\
\mpage{0.17}{\includegraphics[width=1.0\linewidth]{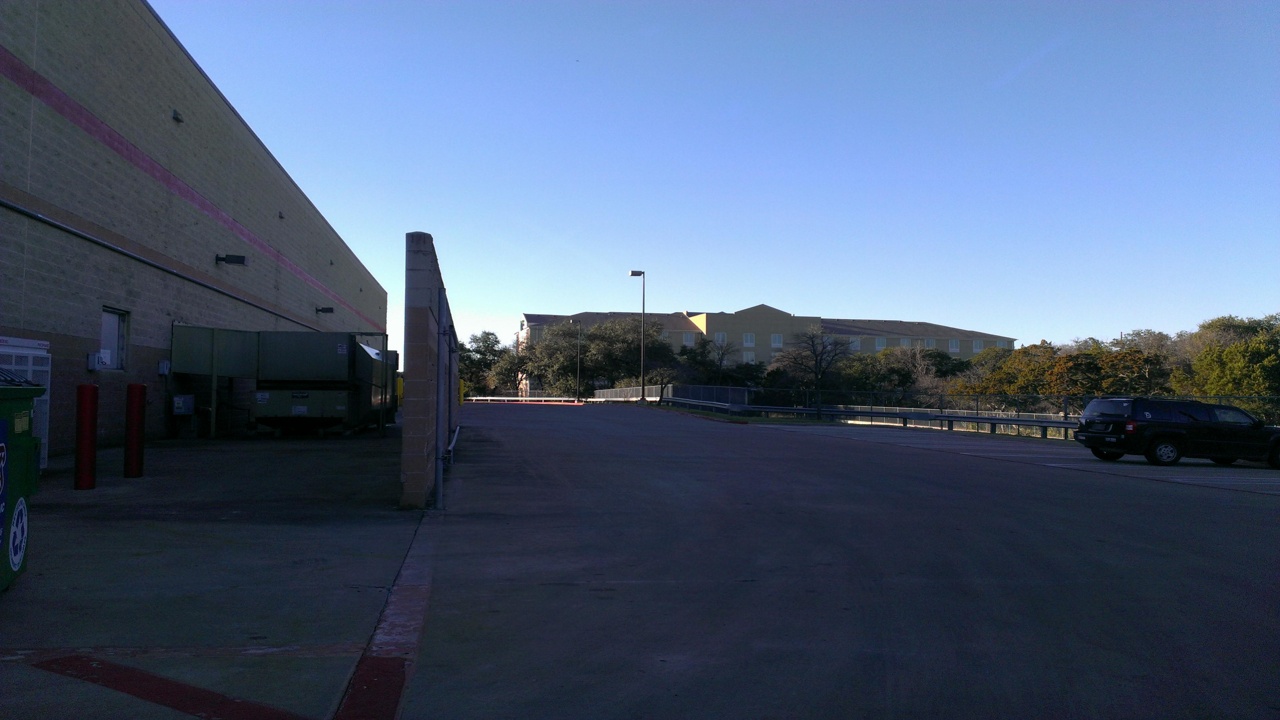}}
\hfill
\mpage{0.17}{\includegraphics[width=1.0\linewidth]{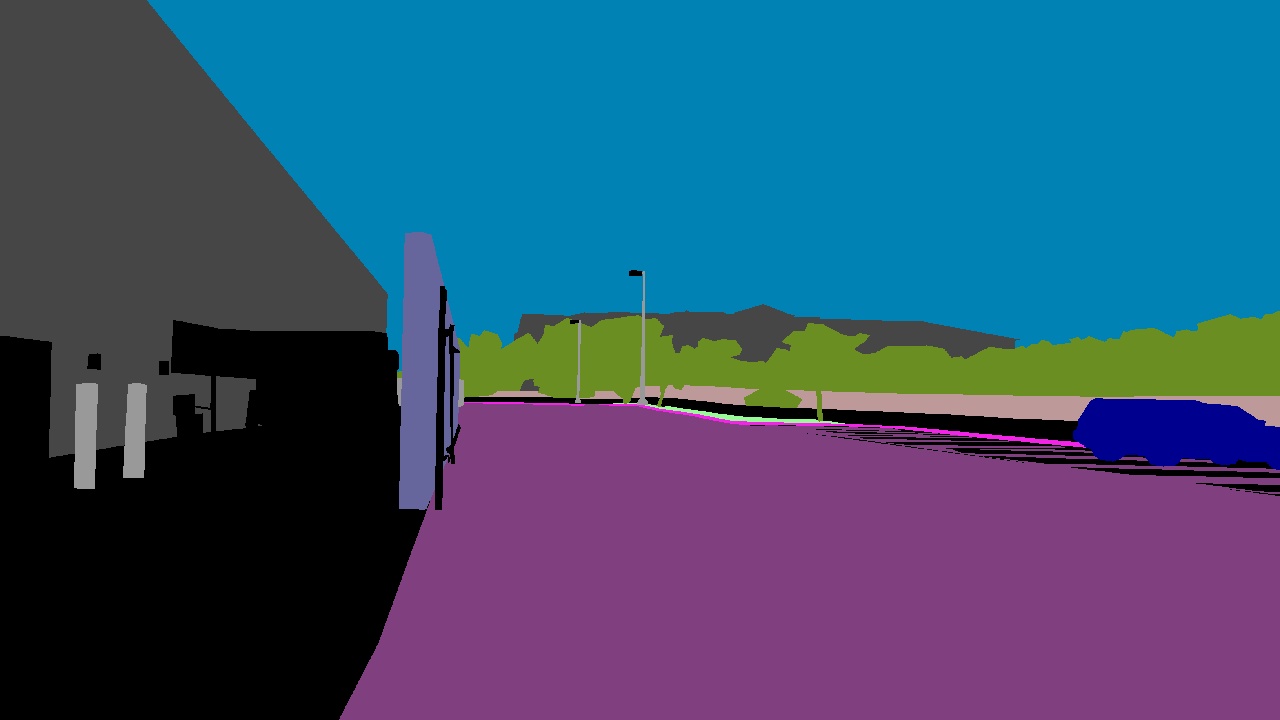}}
\hfill
\mpage{0.17}{\includegraphics[width=1.0\linewidth]{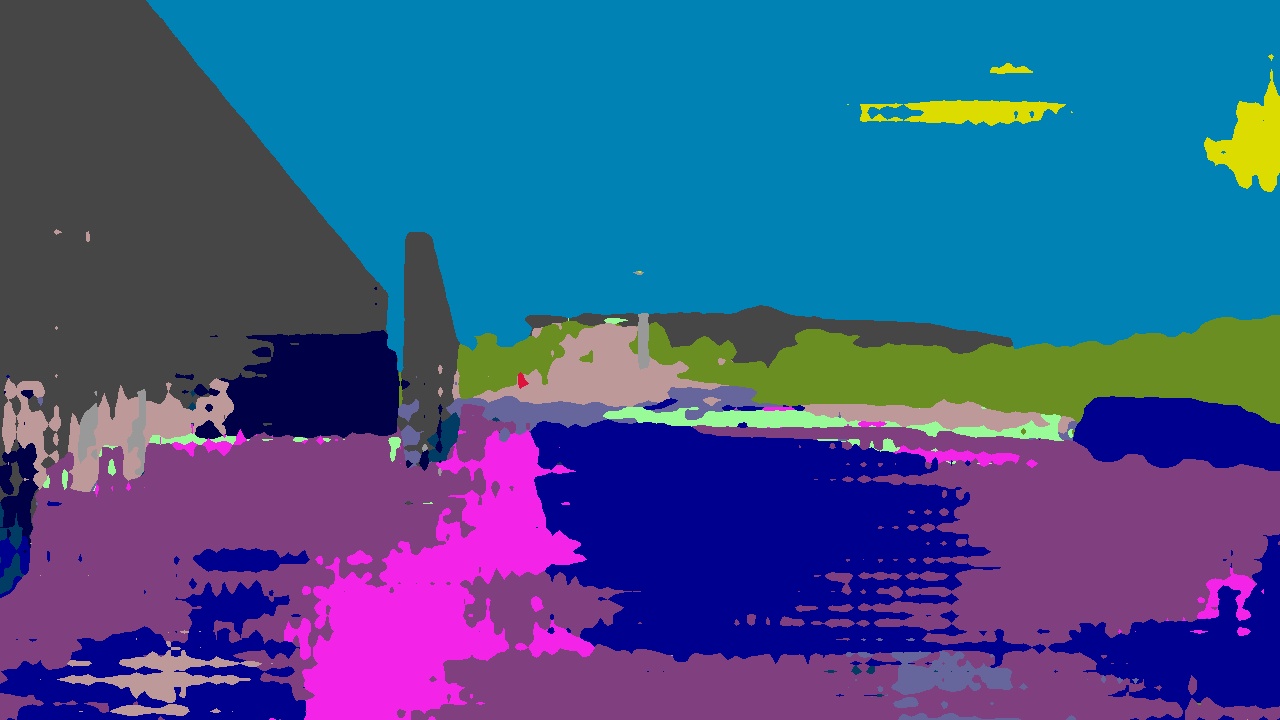}}
\hfill
\mpage{0.17}{\includegraphics[width=1.0\linewidth]{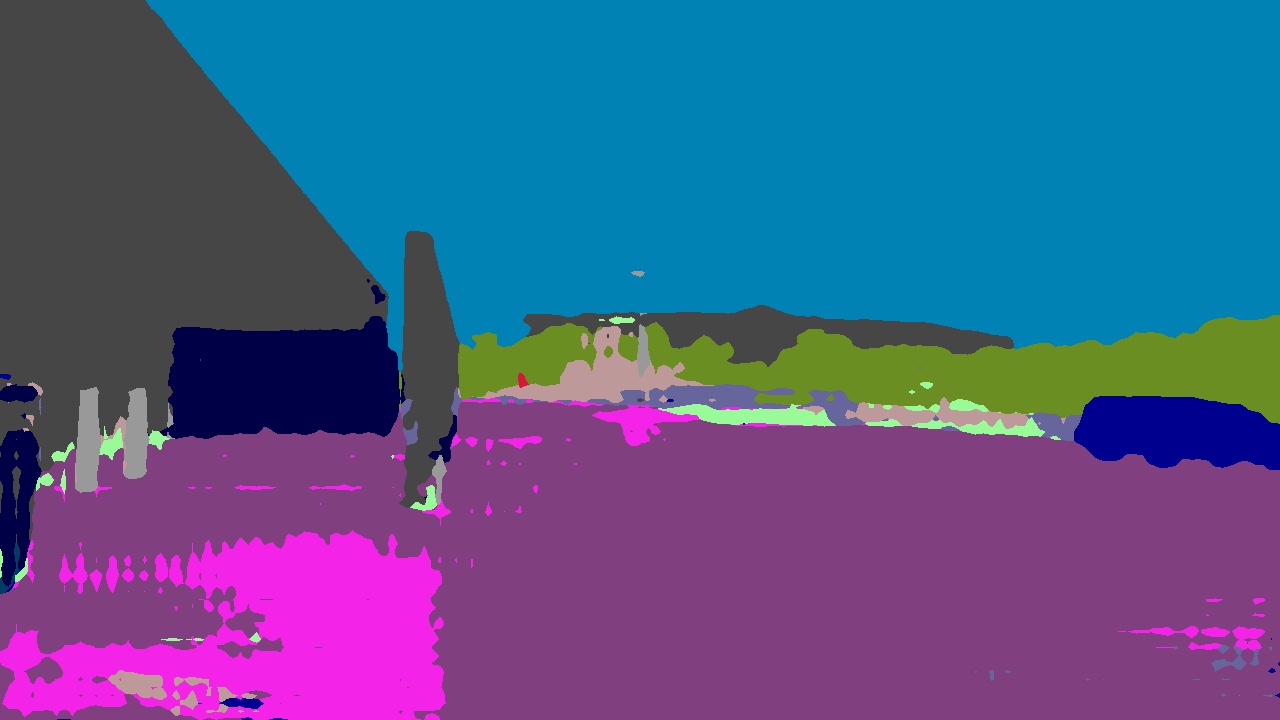}}
\hfill
\mpage{0.17}{\includegraphics[width=1.0\linewidth]{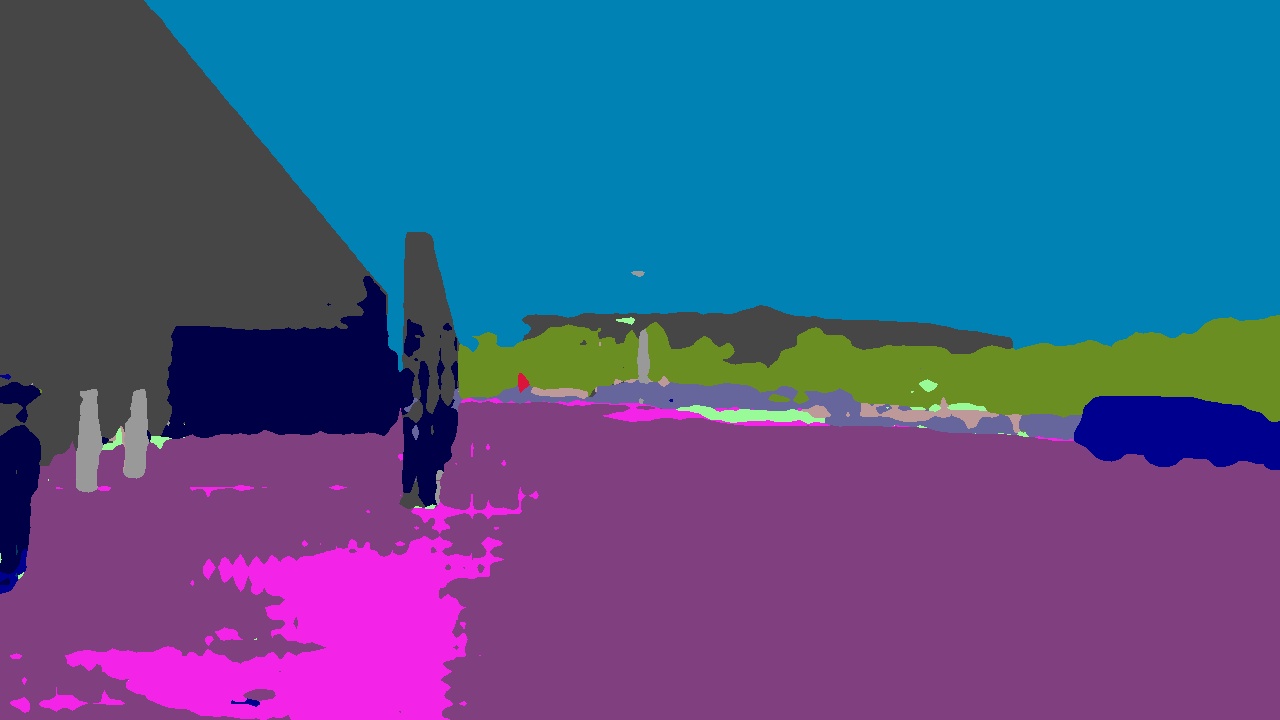}}
\hfill
\\
\mpage{0.17}{\includegraphics[width=1.0\linewidth]{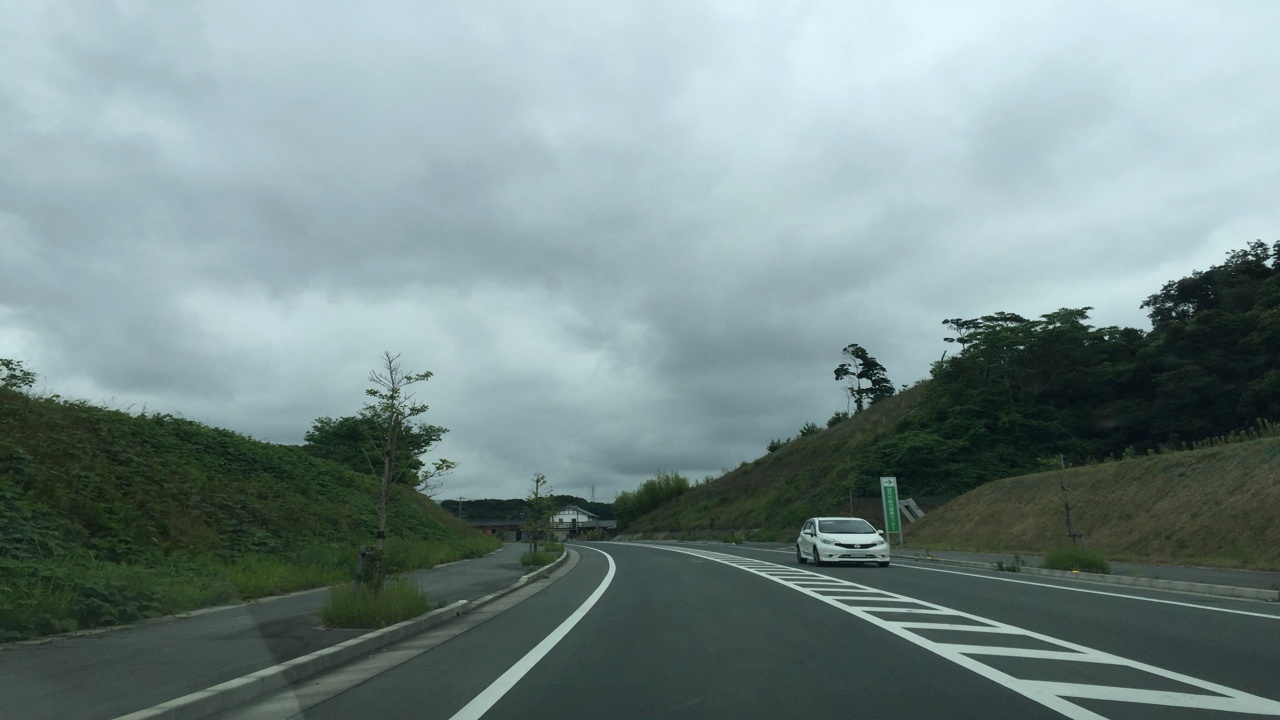}}
\hfill
\mpage{0.17}{\includegraphics[width=1.0\linewidth]{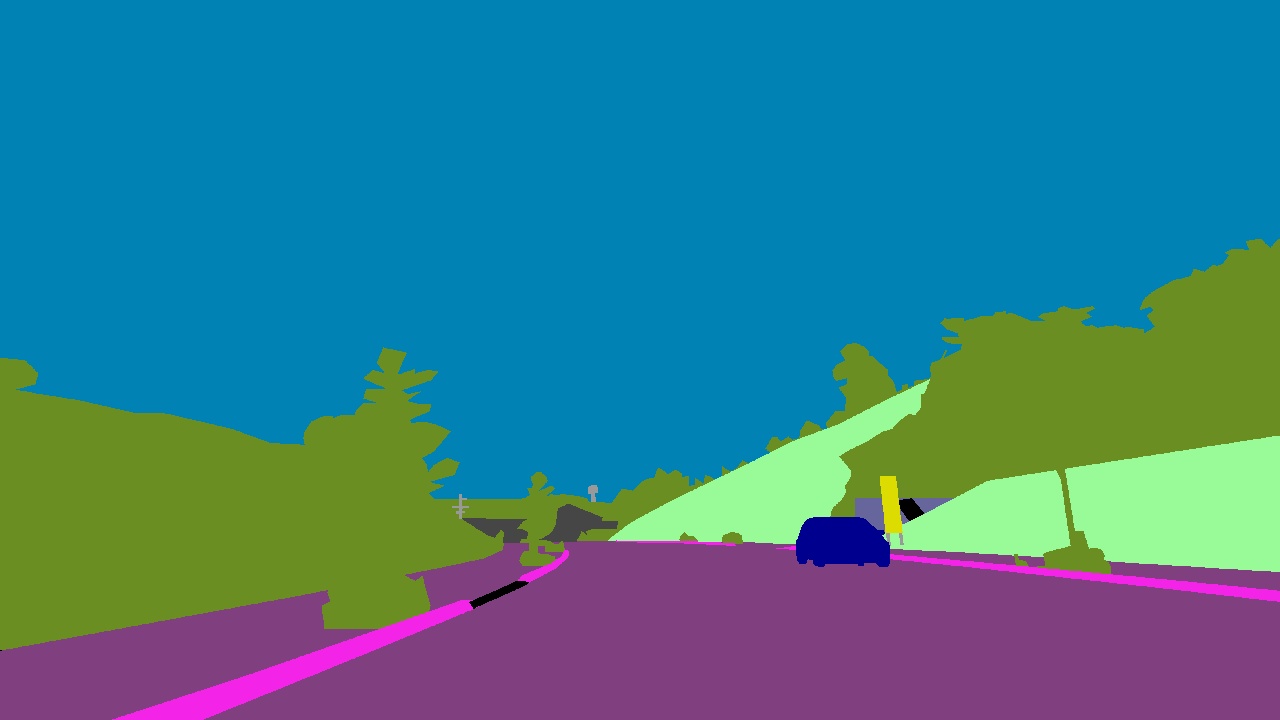}}
\hfill
\mpage{0.17}{\includegraphics[width=1.0\linewidth]{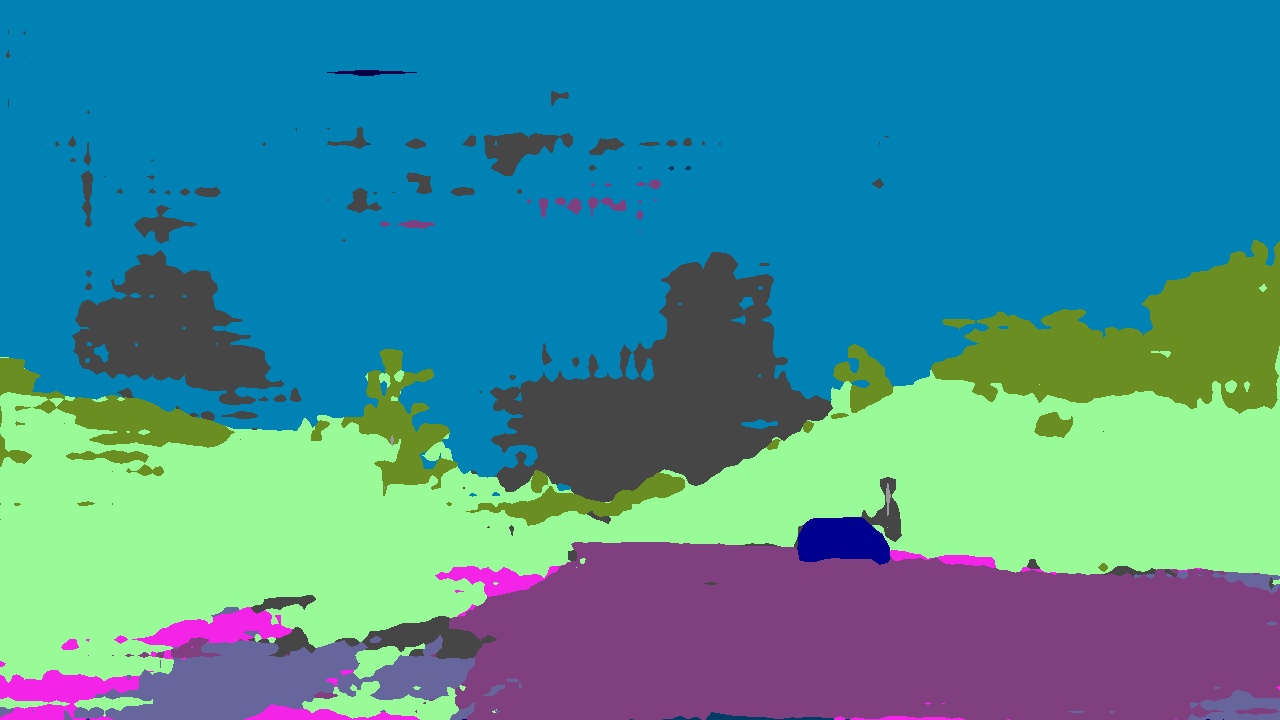}}
\hfill
\mpage{0.17}{\includegraphics[width=1.0\linewidth]{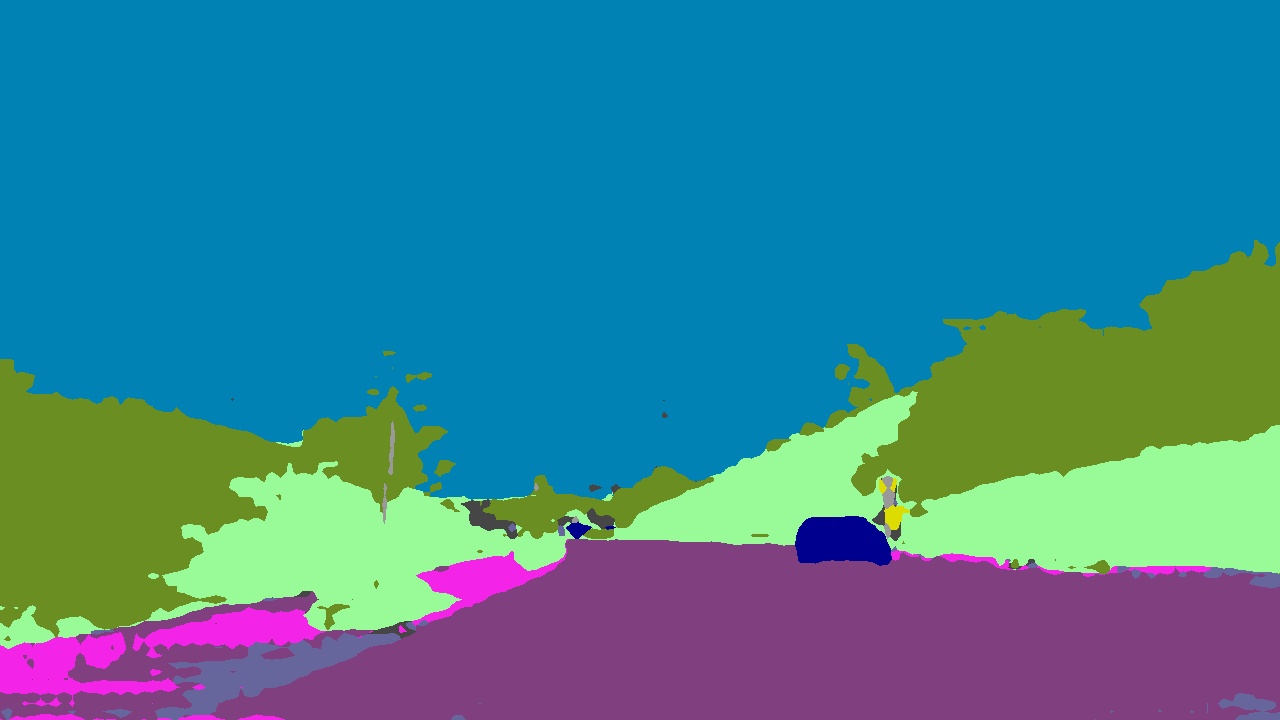}}
\hfill
\mpage{0.17}{\includegraphics[width=1.0\linewidth]{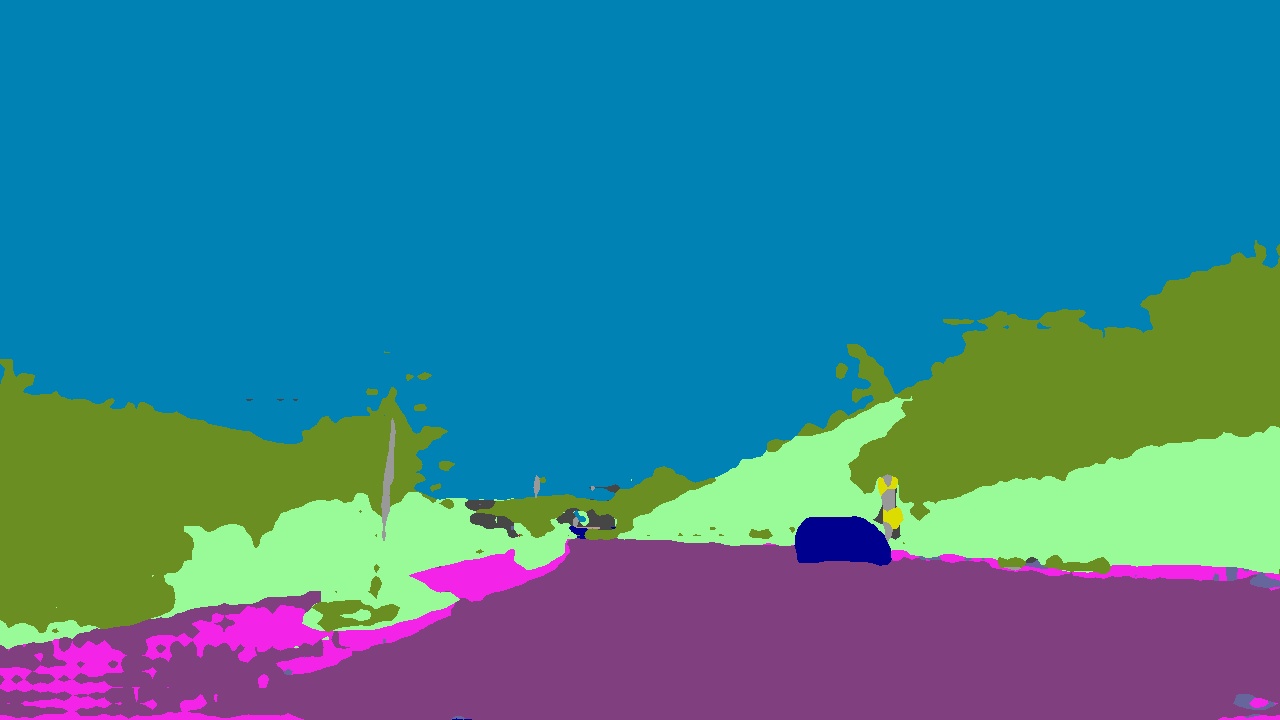}}
\hfill
\\
\mpage{0.17}{\includegraphics[width=1.0\linewidth]{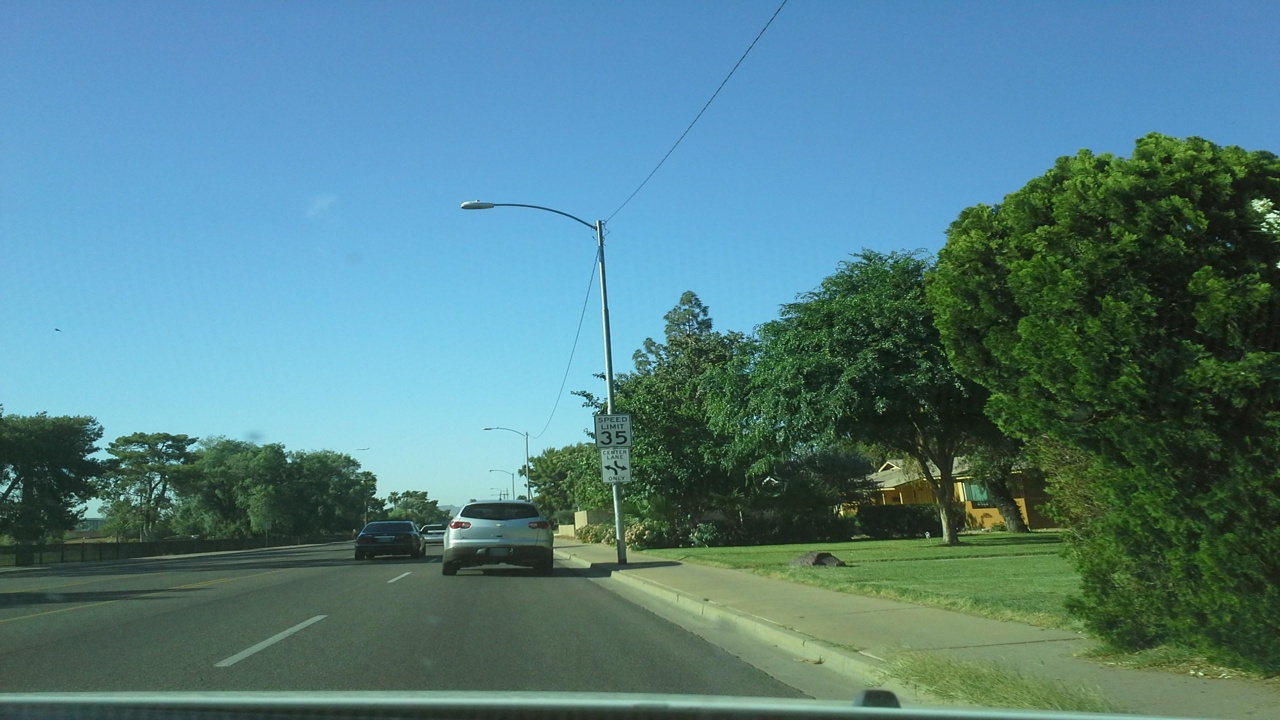}}
\hfill
\mpage{0.17}{\includegraphics[width=1.0\linewidth]{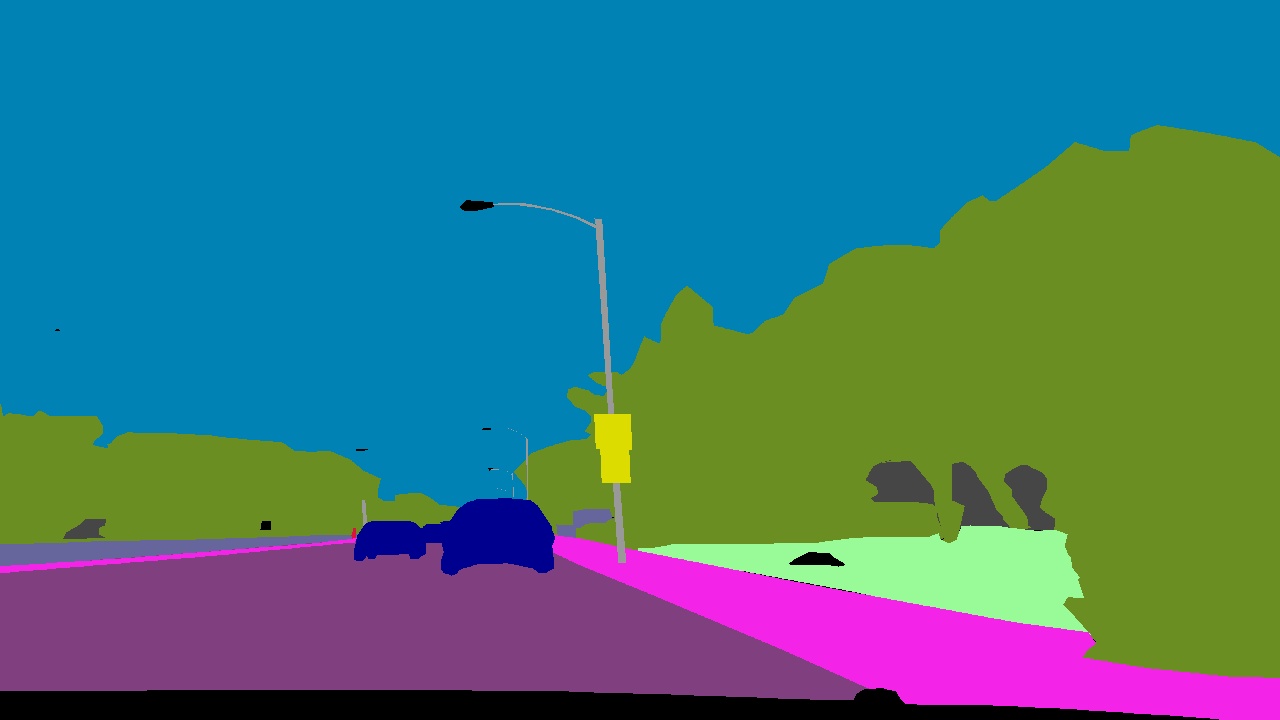}}
\hfill
\mpage{0.17}{\includegraphics[width=1.0\linewidth]{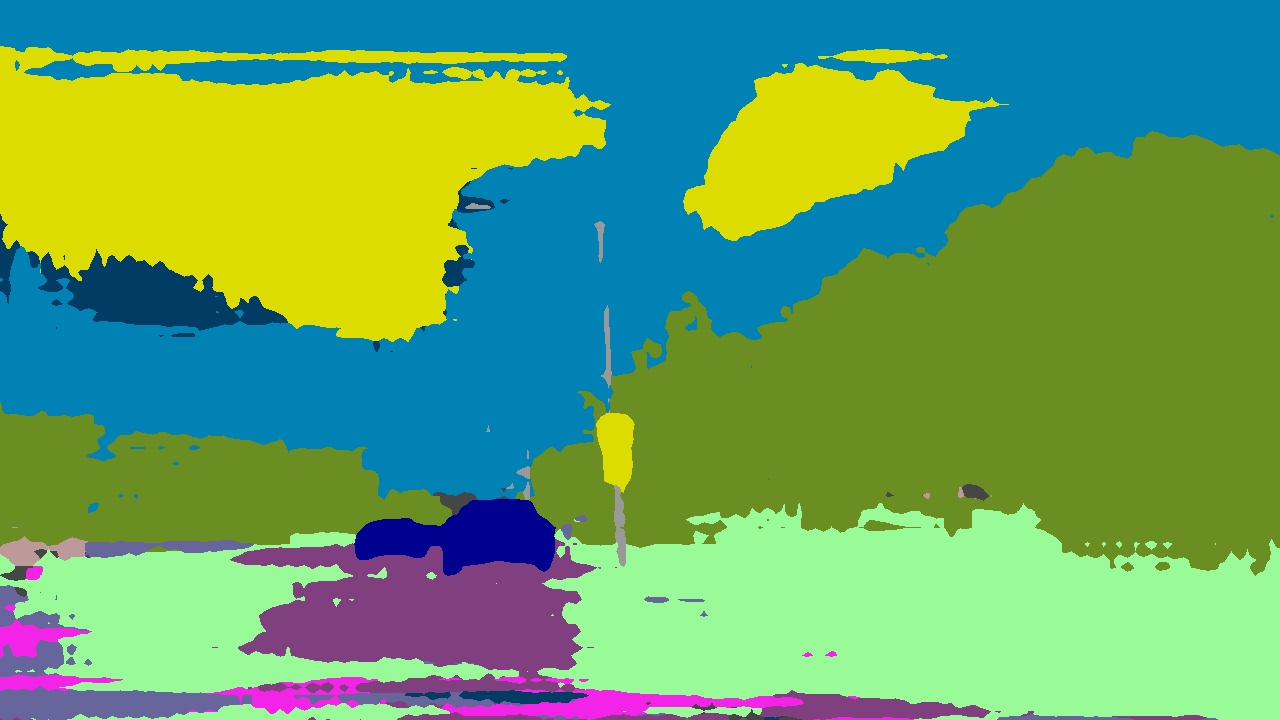}}
\hfill
\mpage{0.17}{\includegraphics[width=1.0\linewidth]{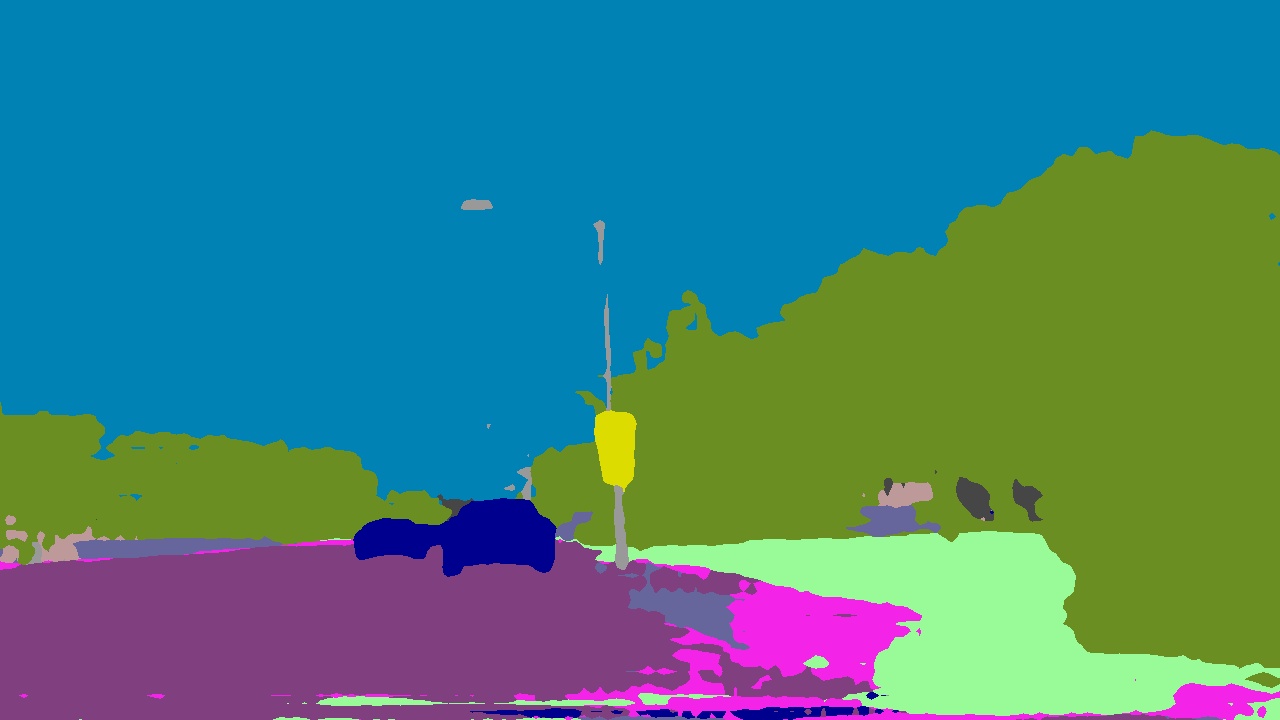}}
\hfill
\mpage{0.17}{\includegraphics[width=1.0\linewidth]{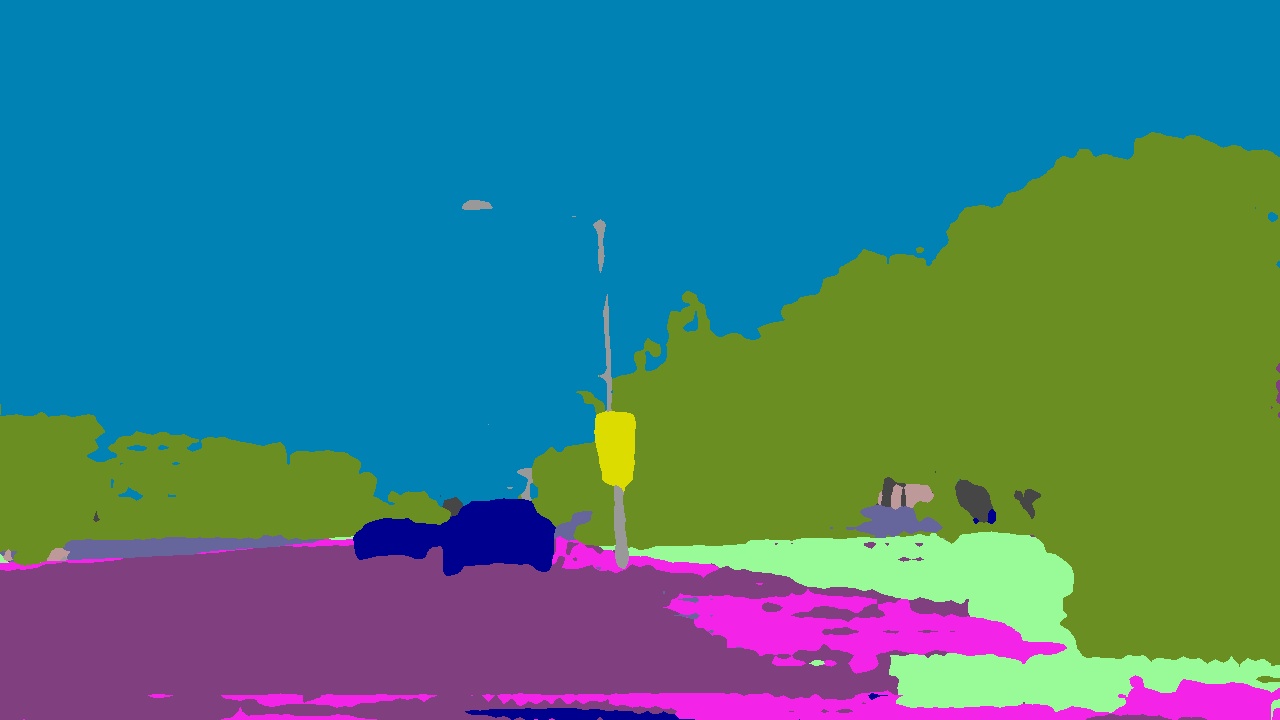}}
\hfill
\\
\mpage{0.17}{\includegraphics[width=1.0\linewidth]{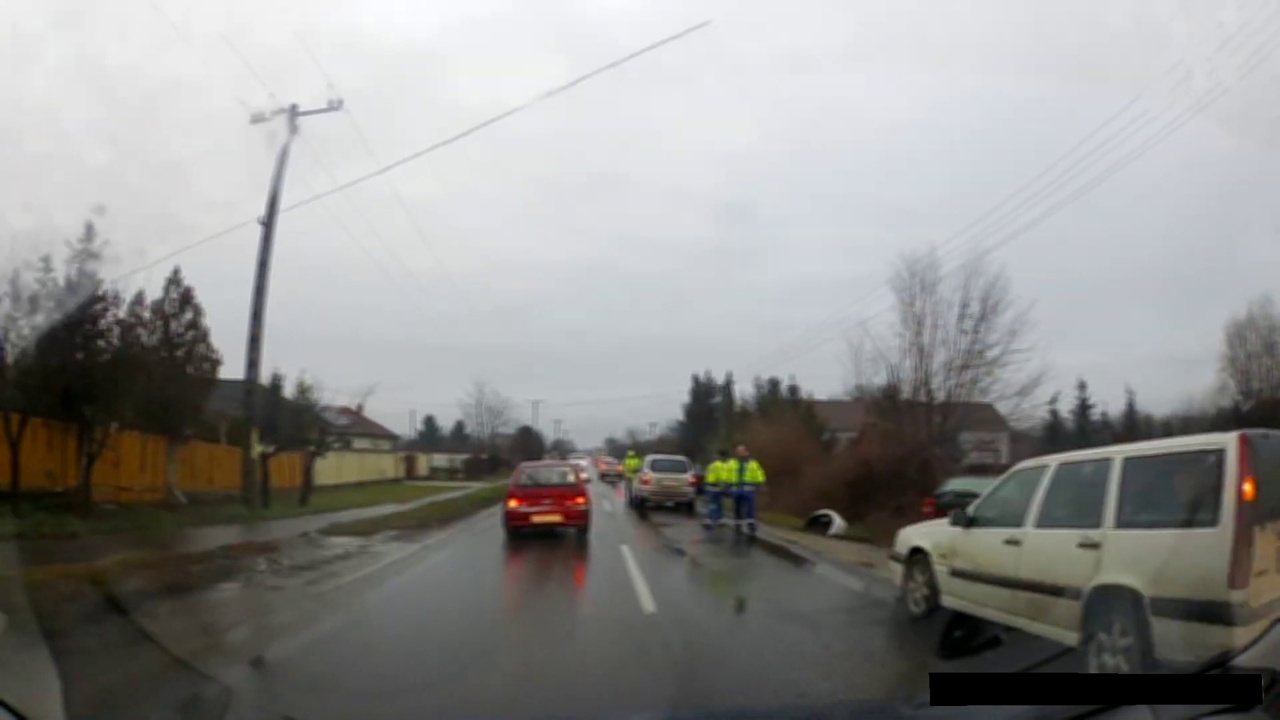}}
\hfill
\mpage{0.17}{\includegraphics[width=1.0\linewidth]{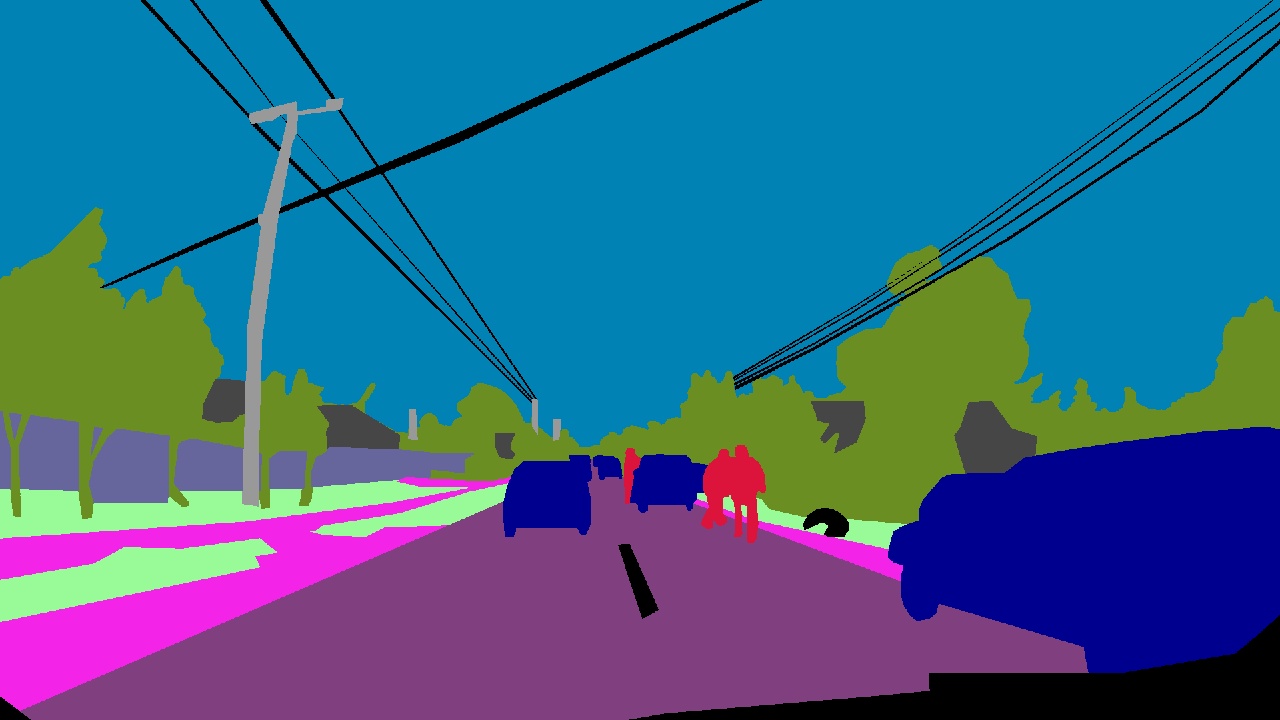}}
\hfill
\mpage{0.17}{\includegraphics[width=1.0\linewidth]{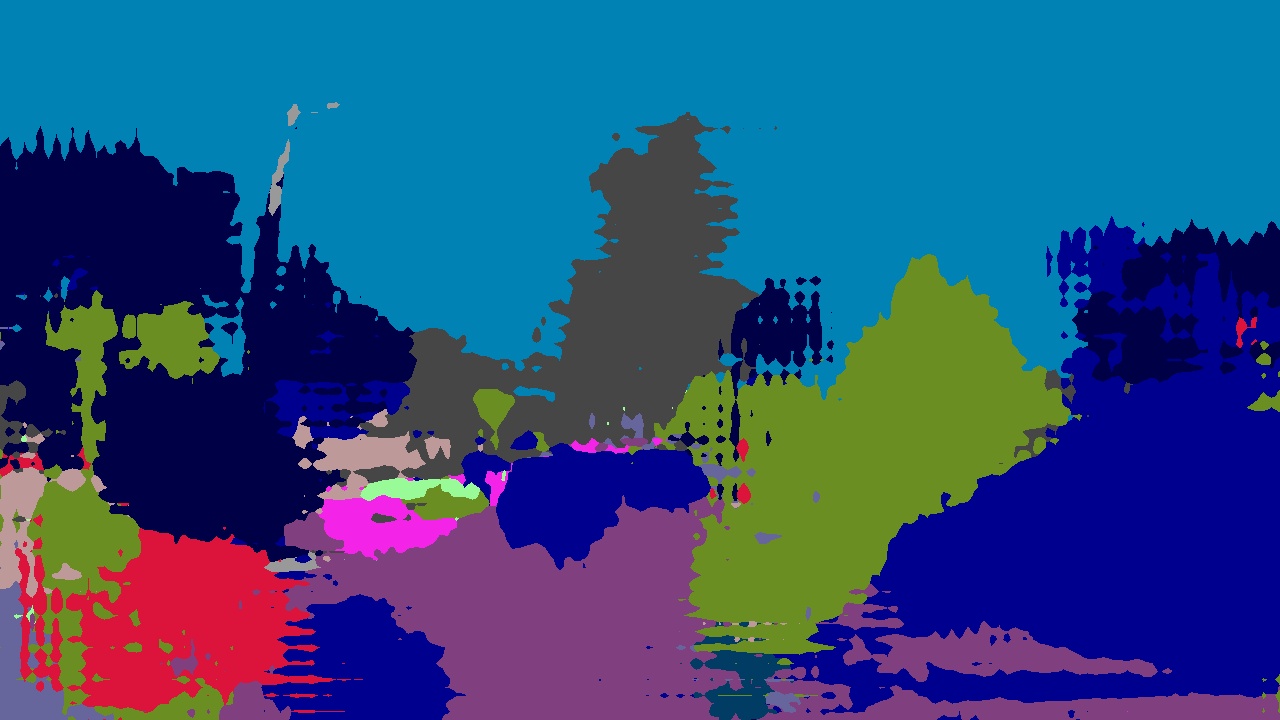}}
\hfill
\mpage{0.17}{\includegraphics[width=1.0\linewidth]{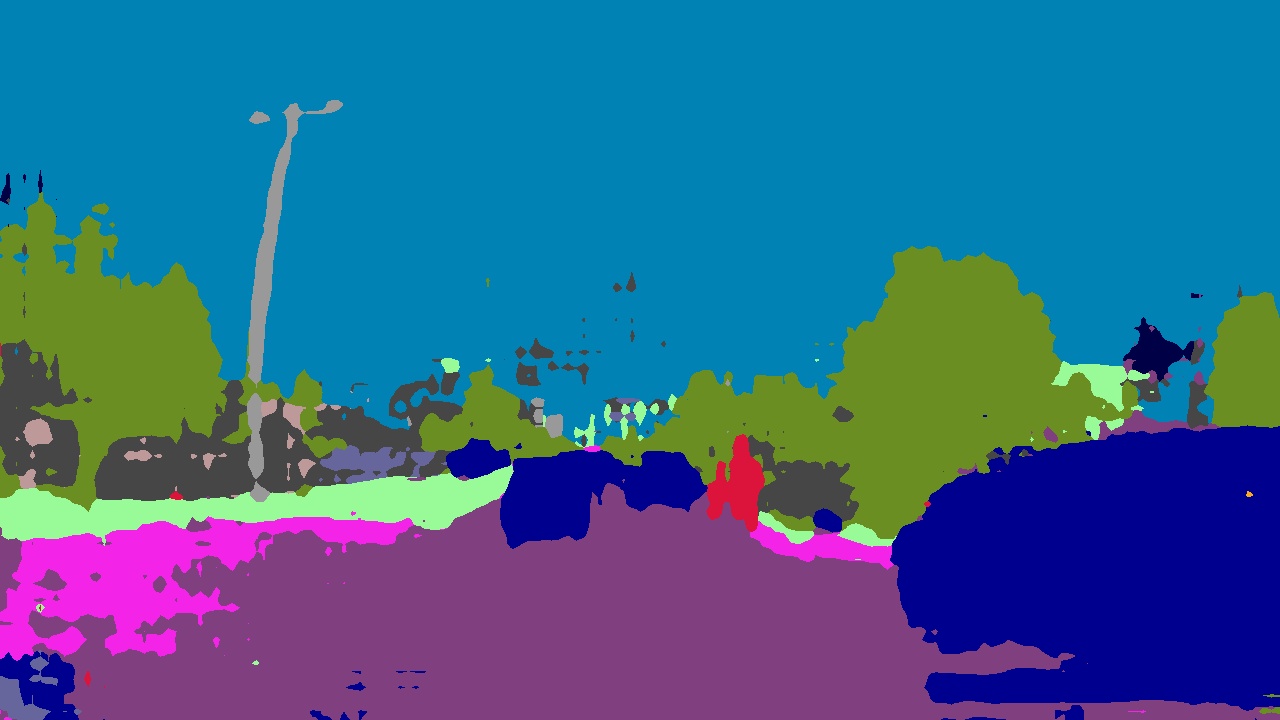}}
\hfill
\mpage{0.17}{\includegraphics[width=1.0\linewidth]{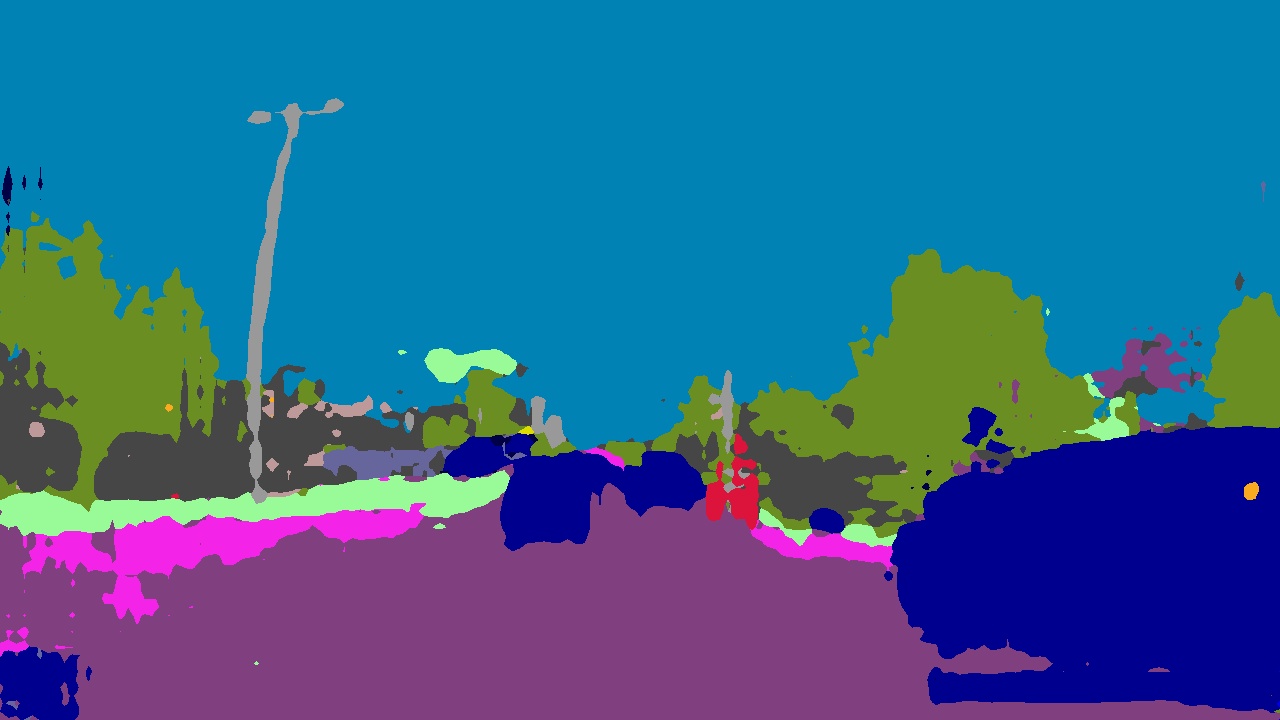}}
\hfill
\\
\mpage{0.17}{\includegraphics[width=1.0\linewidth]{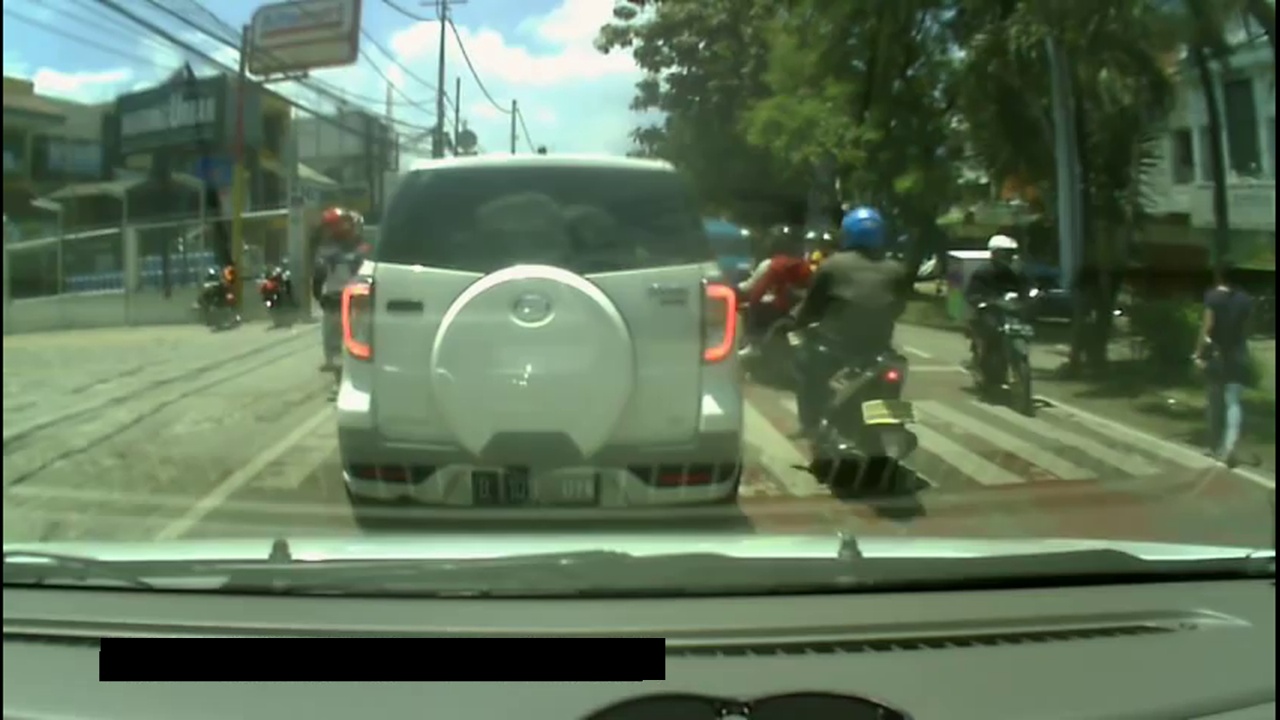}}
\hfill
\mpage{0.17}{\includegraphics[width=1.0\linewidth]{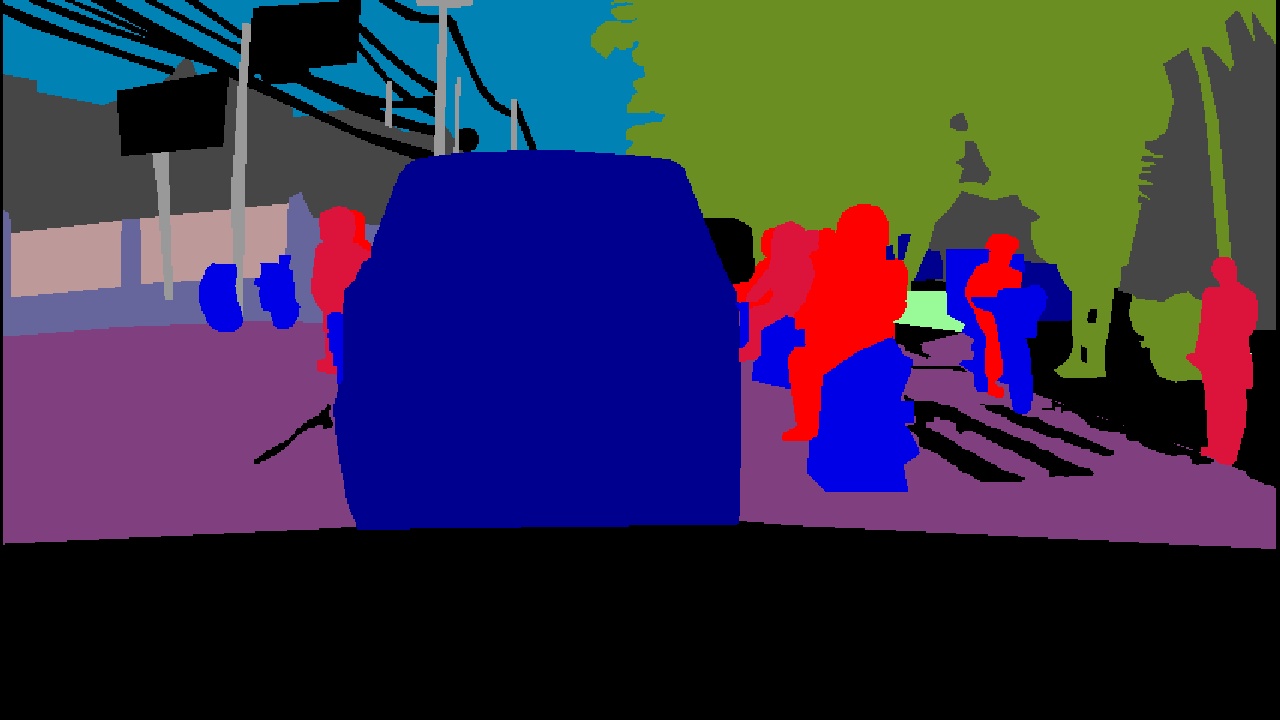}}
\hfill
\mpage{0.17}{\includegraphics[width=1.0\linewidth]{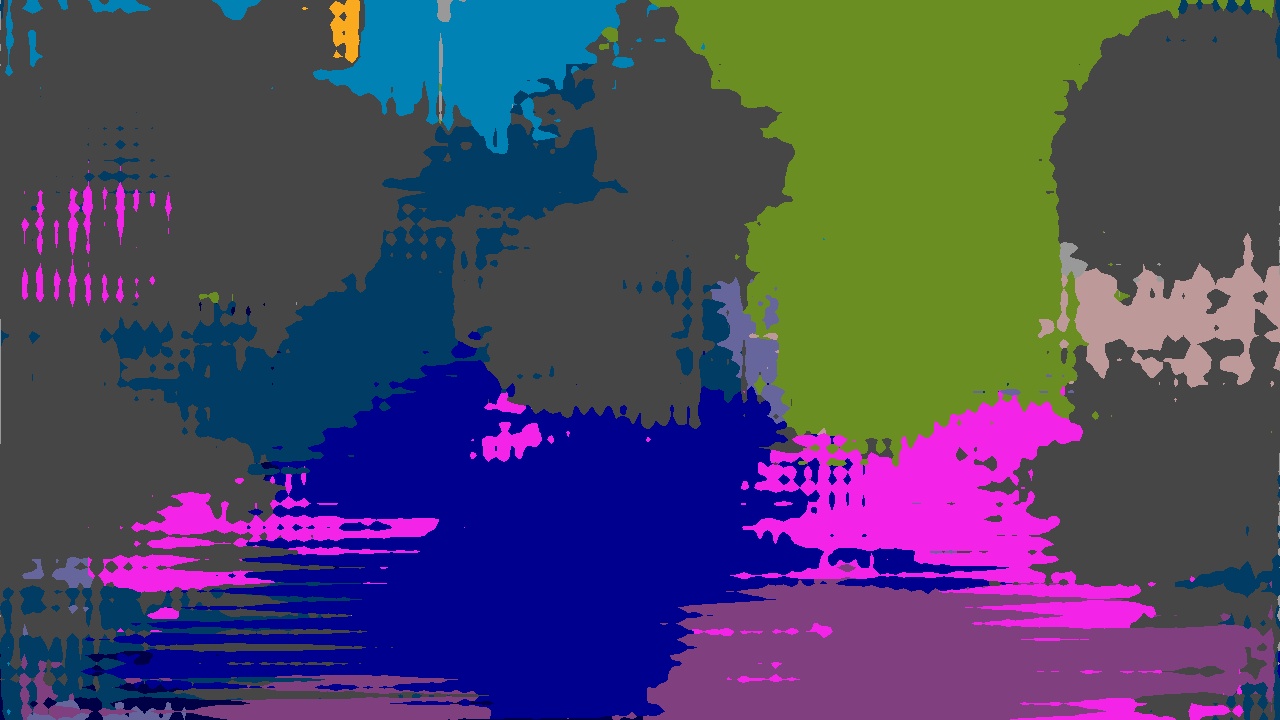}}
\hfill
\mpage{0.17}{\includegraphics[width=1.0\linewidth]{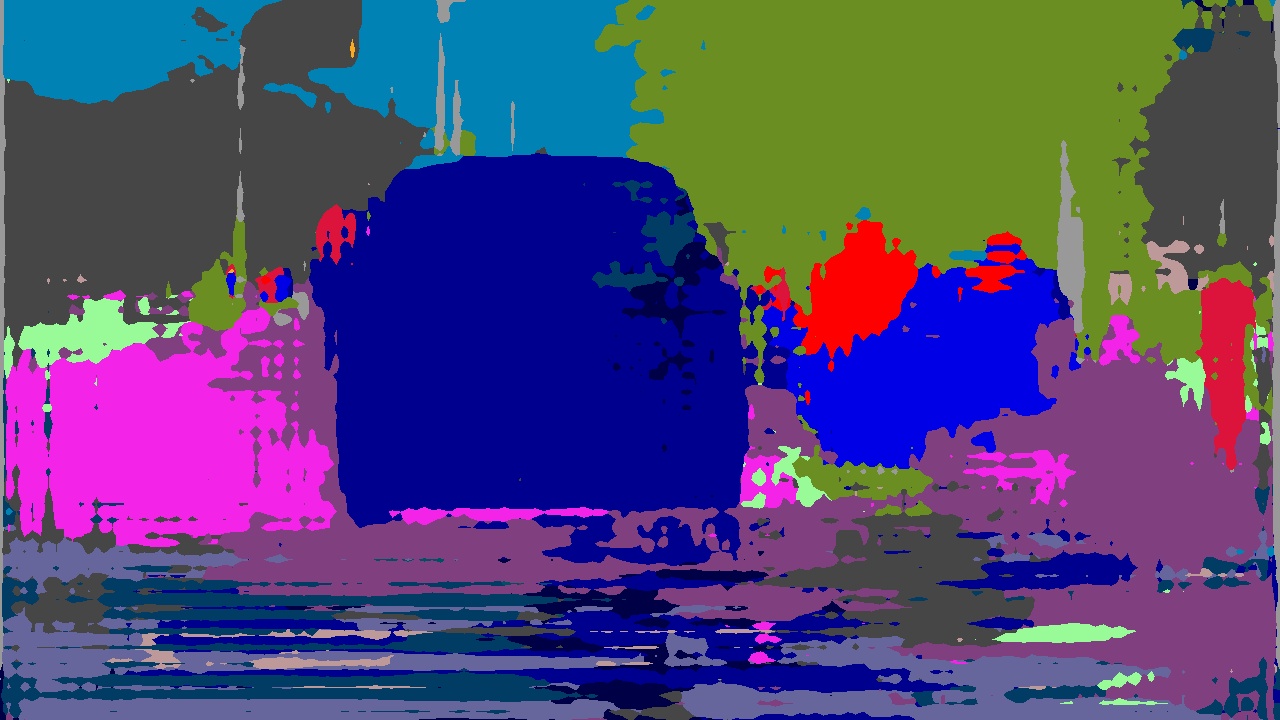}}
\hfill
\mpage{0.17}{\includegraphics[width=1.0\linewidth]{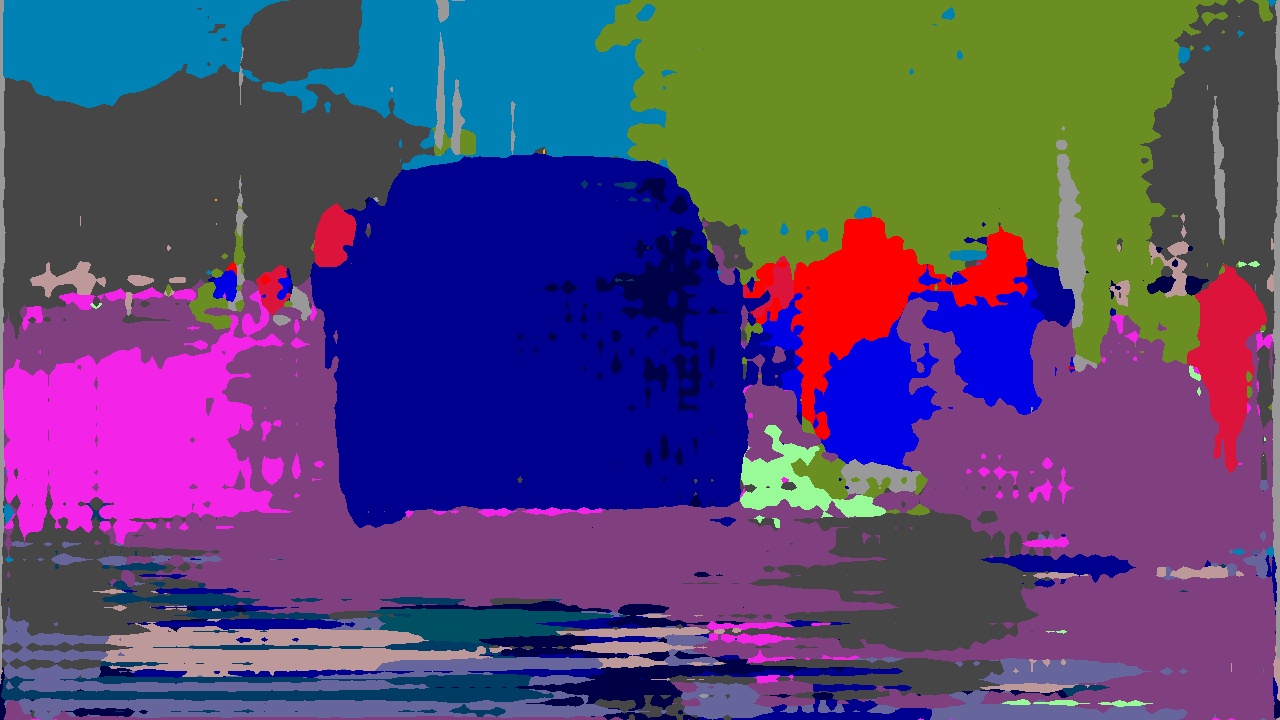}}
\hfill
\\
\mpage{0.17}{\includegraphics[width=1.0\linewidth]{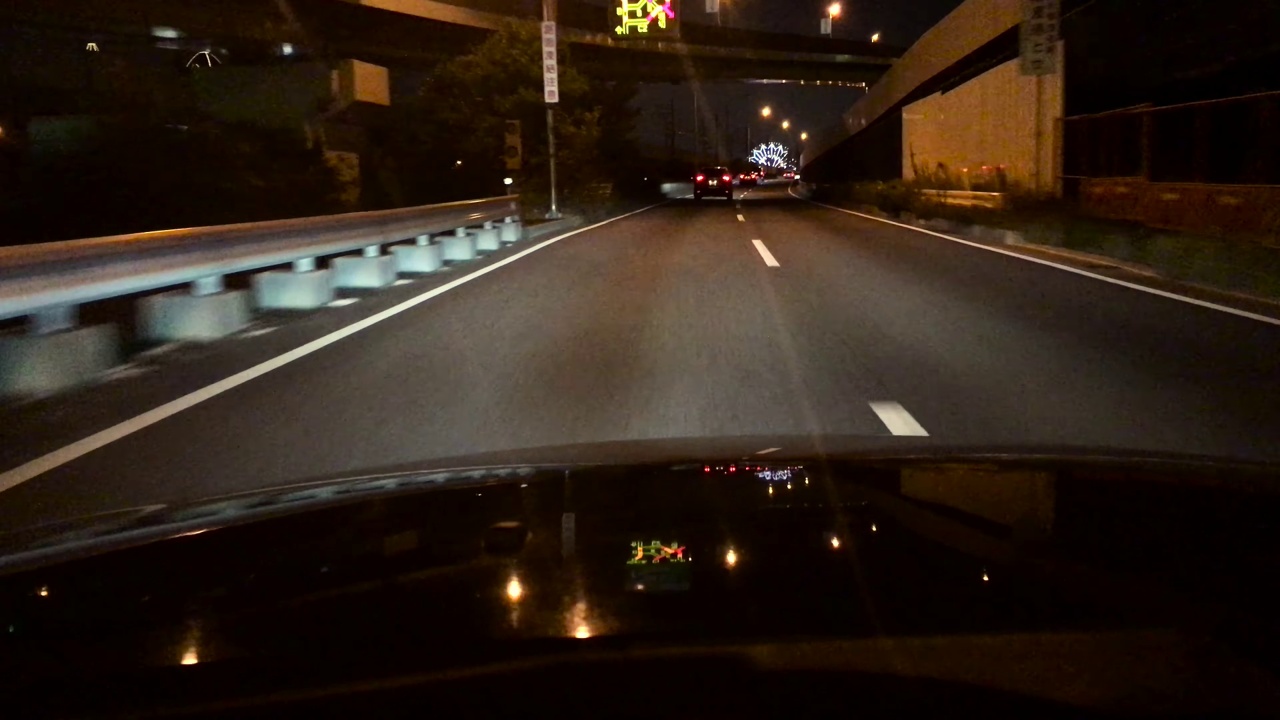}}
\hfill
\mpage{0.17}{\includegraphics[width=1.0\linewidth]{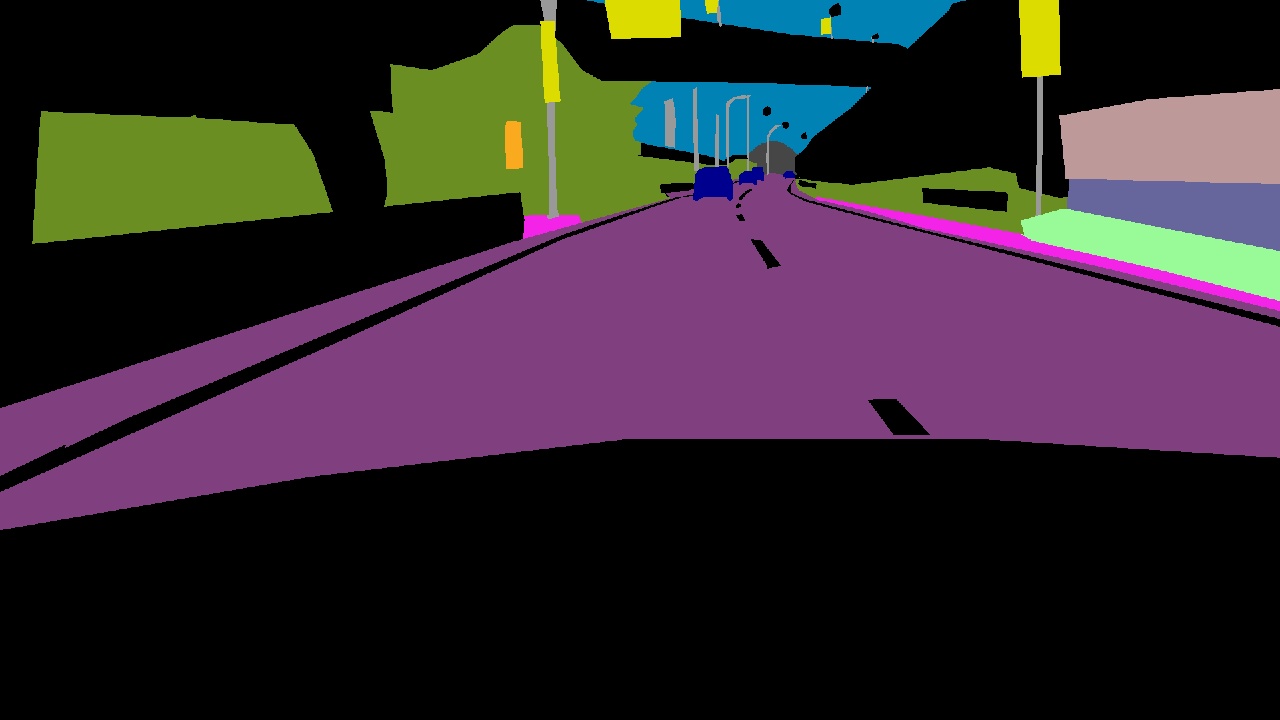}}
\hfill
\mpage{0.17}{\includegraphics[width=1.0\linewidth]{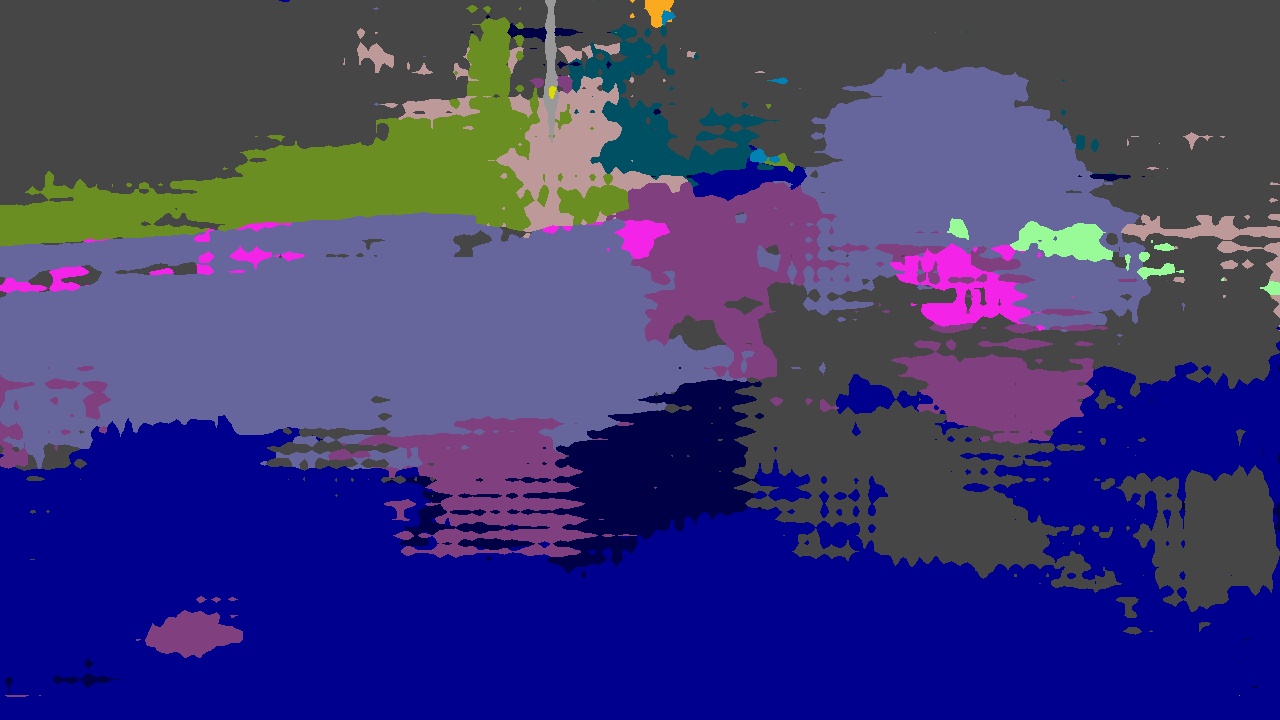}}
\hfill
\mpage{0.17}{\includegraphics[width=1.0\linewidth]{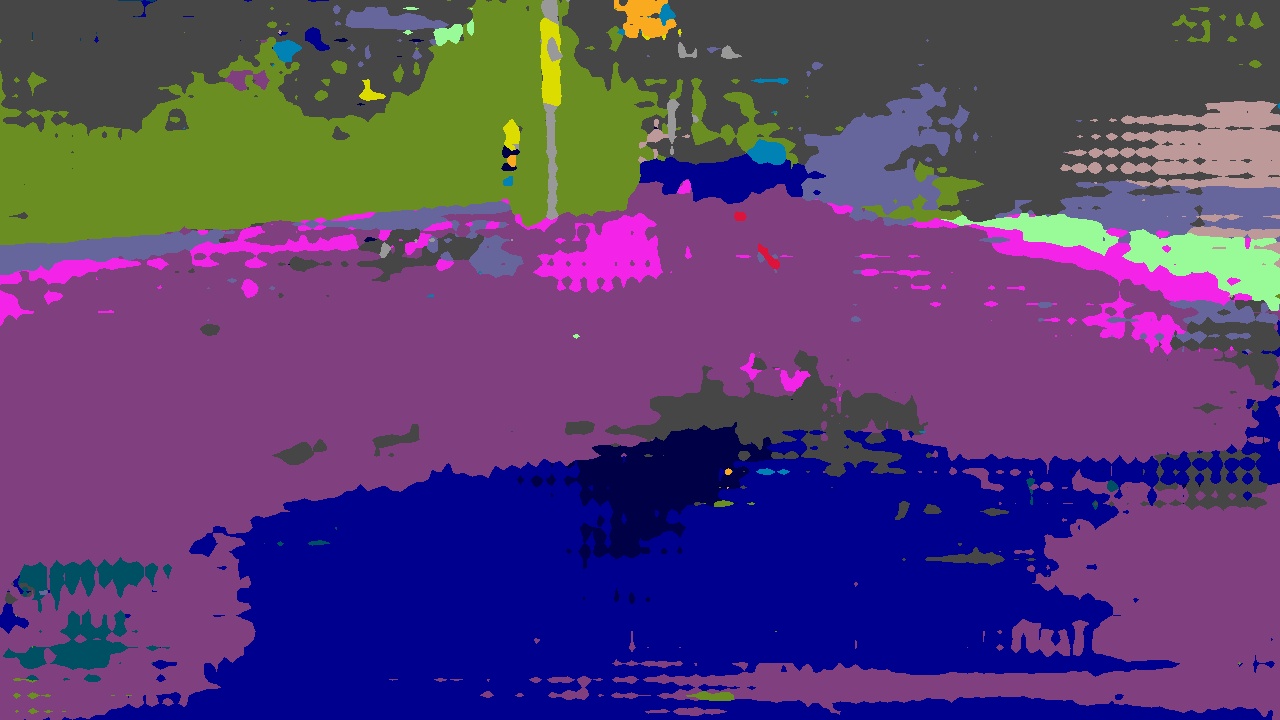}}
\hfill
\mpage{0.17}{\includegraphics[width=1.0\linewidth]{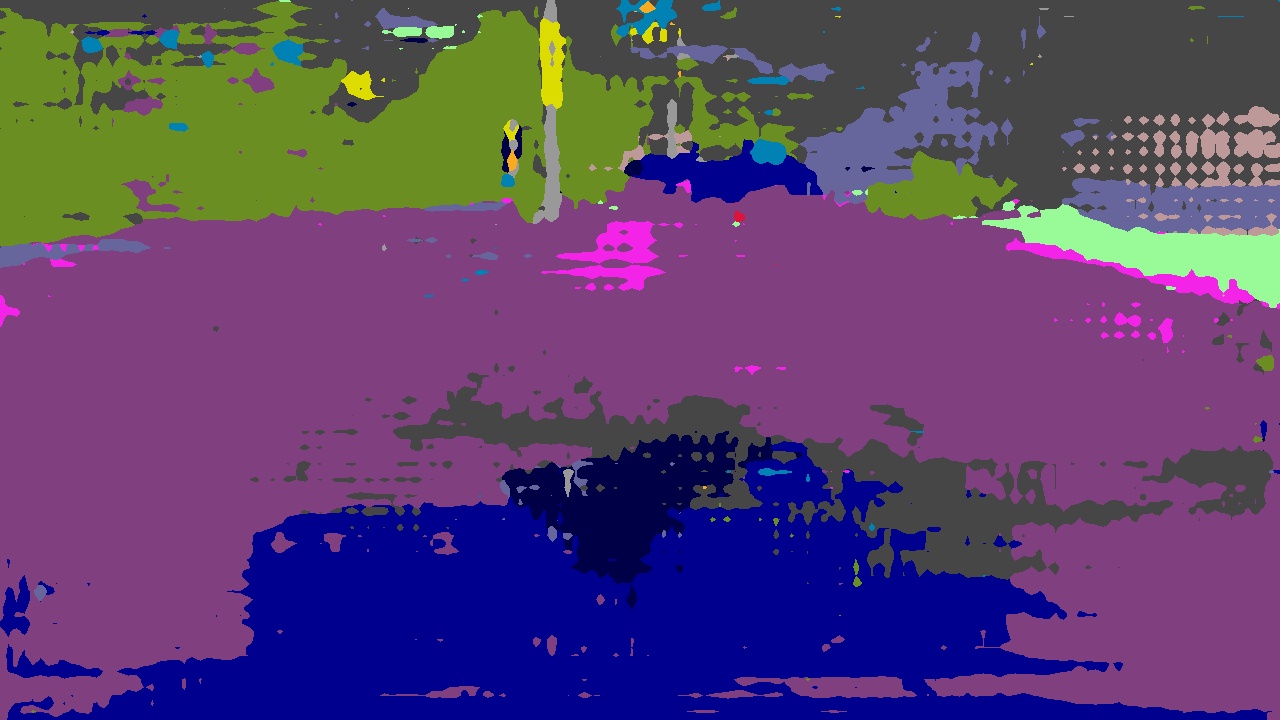}}
\hfill
\\
\mpage{0.17}{Input}
\hfill
\mpage{0.17}{Ground truth}
\hfill
\mpage{0.17}{Pre-trained}
\hfill
\mpage{0.17}{InstCal-U}
\hfill
\mpage{0.17}{InstCal-C}
\hfill

\figcapmargin
\figcaption{Qualitative results of semantic segmentation}
{Models are trained on the GTA5 dataset. We treat the Cityscapes, BDD100k, Mapillary, and WildDash2 datasets as the unseen target domains.
}
\label{fig:supp_semantic}
\end{figure*}

\begin{figure*}[htbp]
\mpage{0.22}{\includegraphics[width=1.0\linewidth]{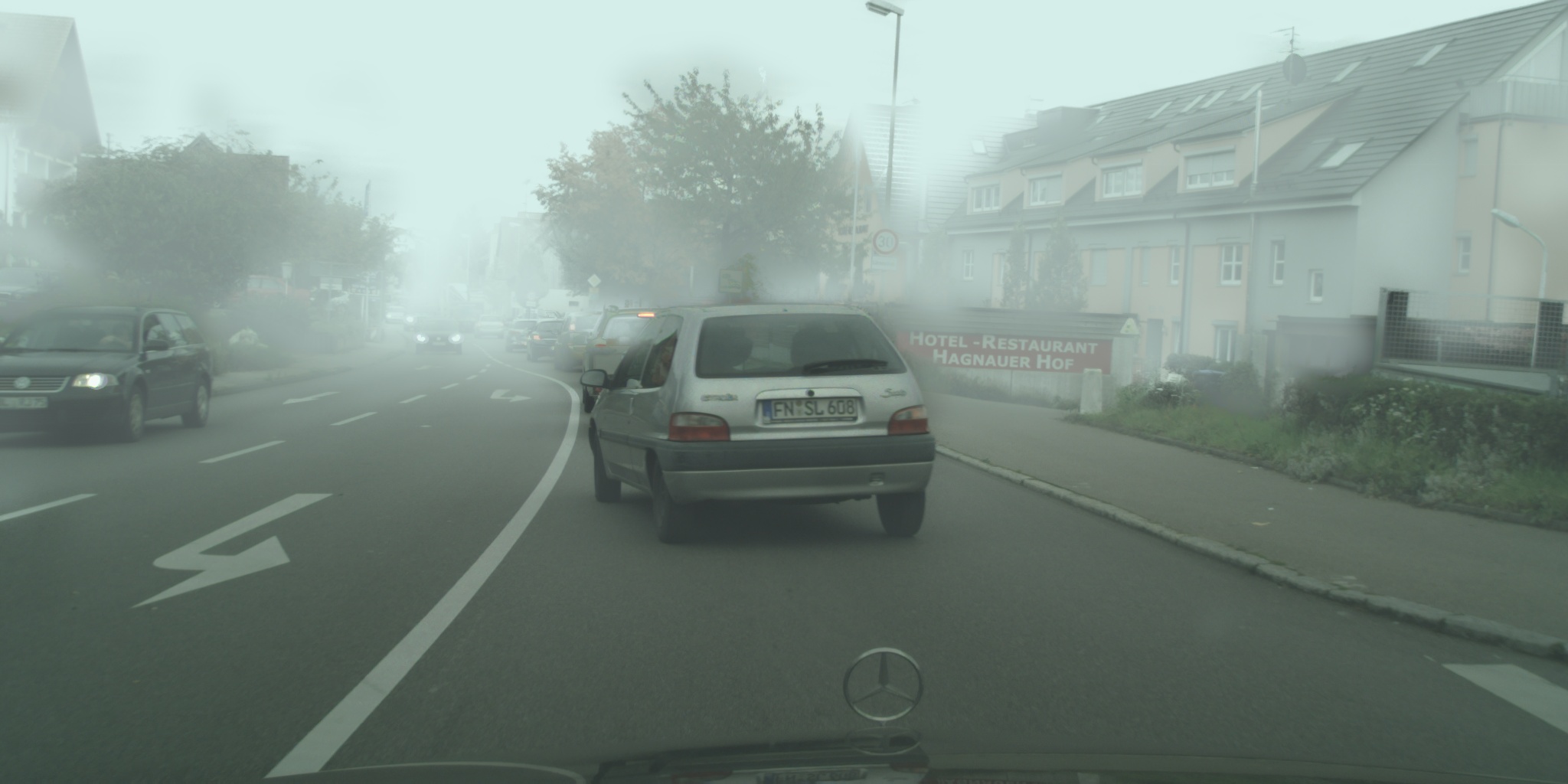}}
\hfill
\mpage{0.22}{\includegraphics[width=1.0\linewidth]{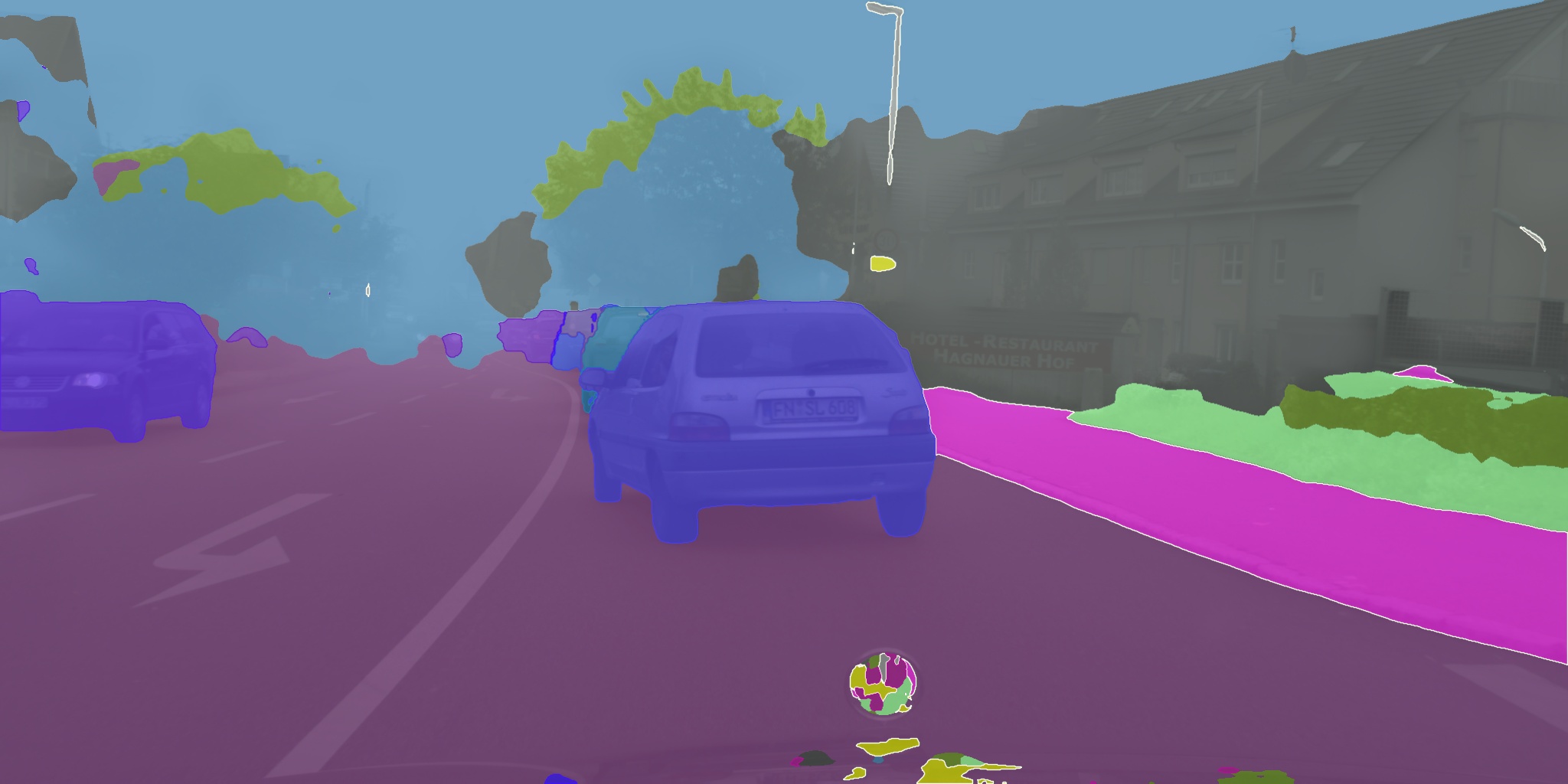}}
\hfill
\mpage{0.22}{\includegraphics[width=1.0\linewidth]{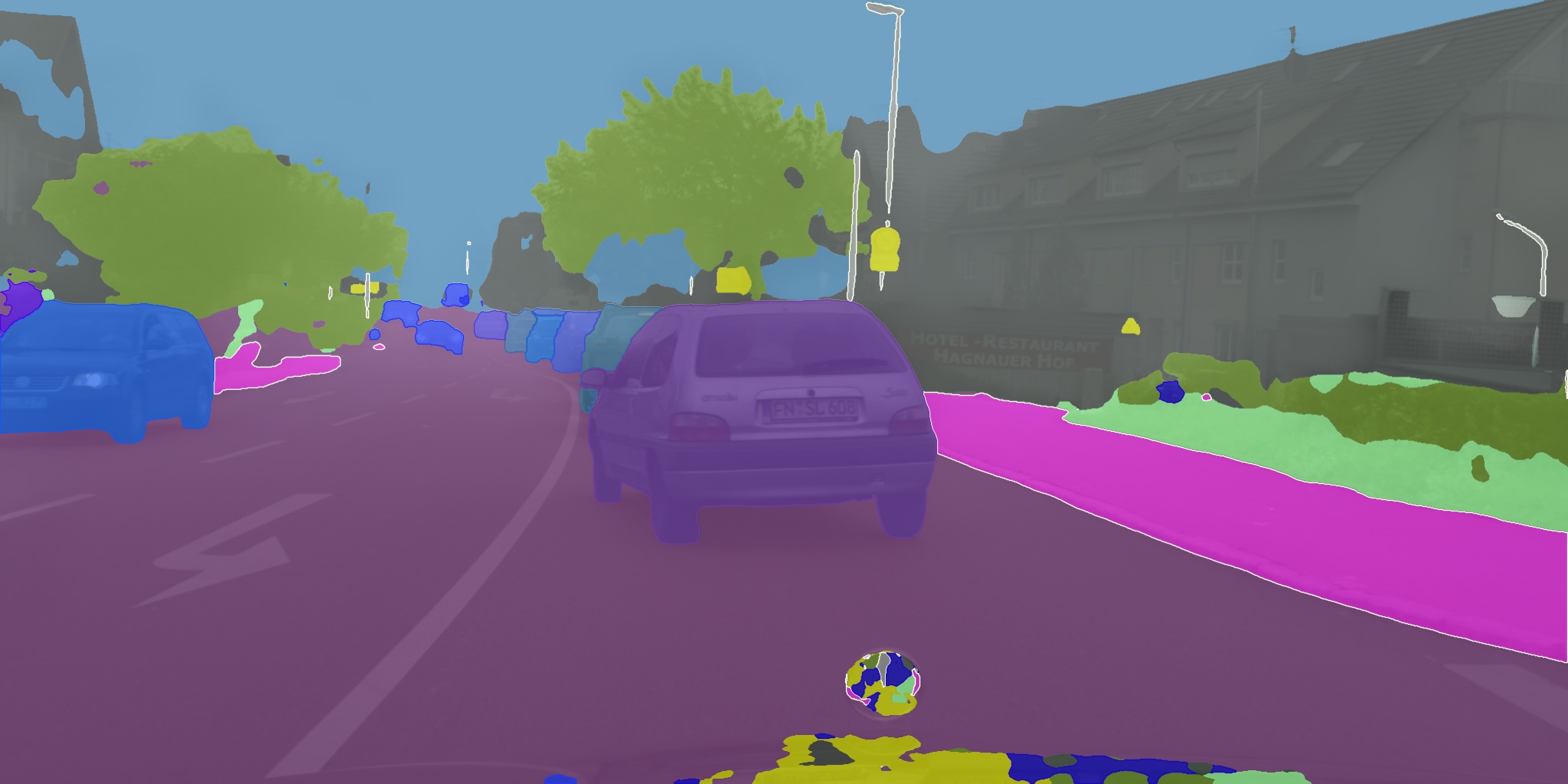}}
\hfill
\mpage{0.22}{\includegraphics[width=1.0\linewidth]{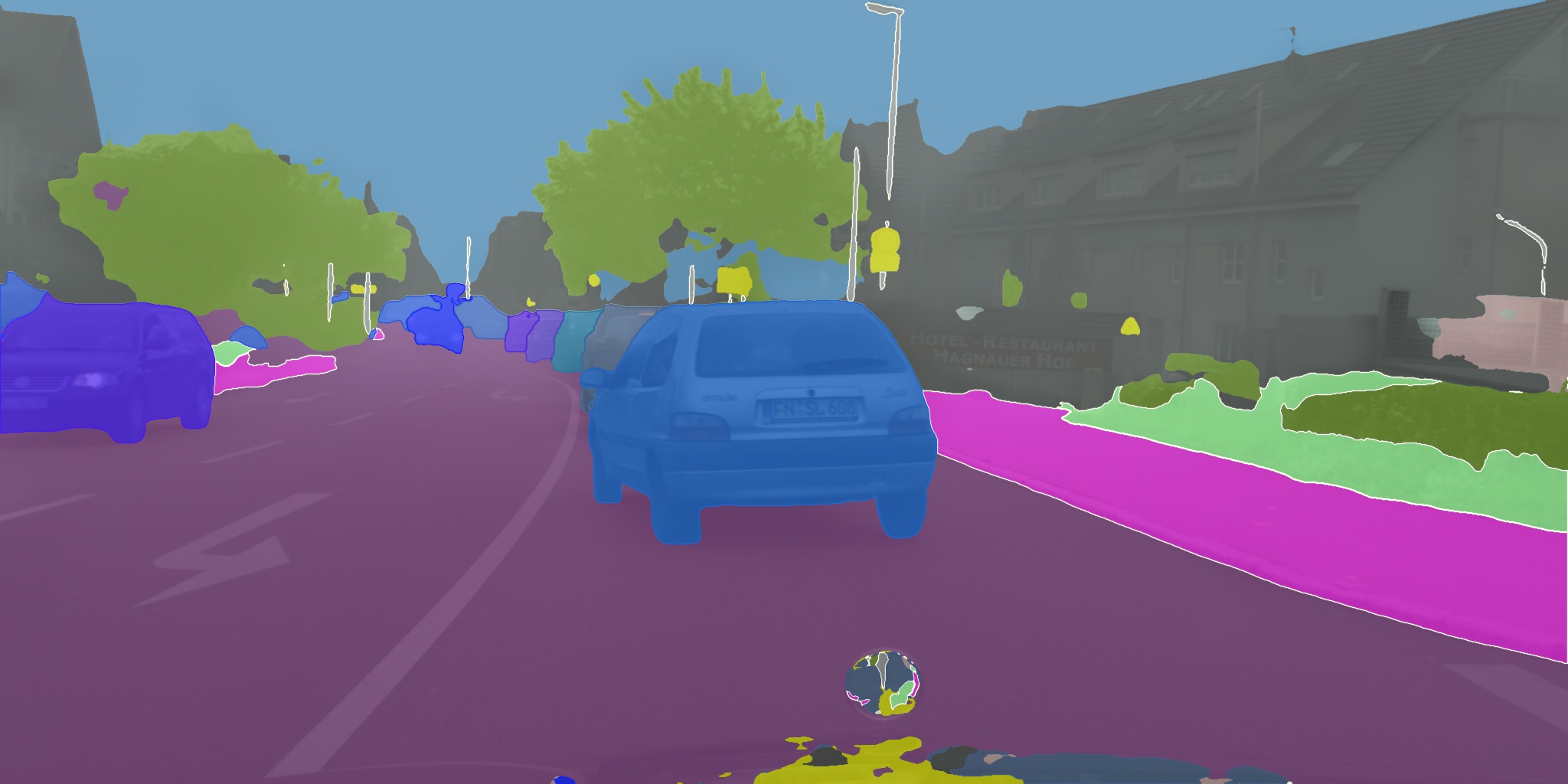}}
\hfill
\\
\mpage{0.22}{\includegraphics[width=1.0\linewidth]{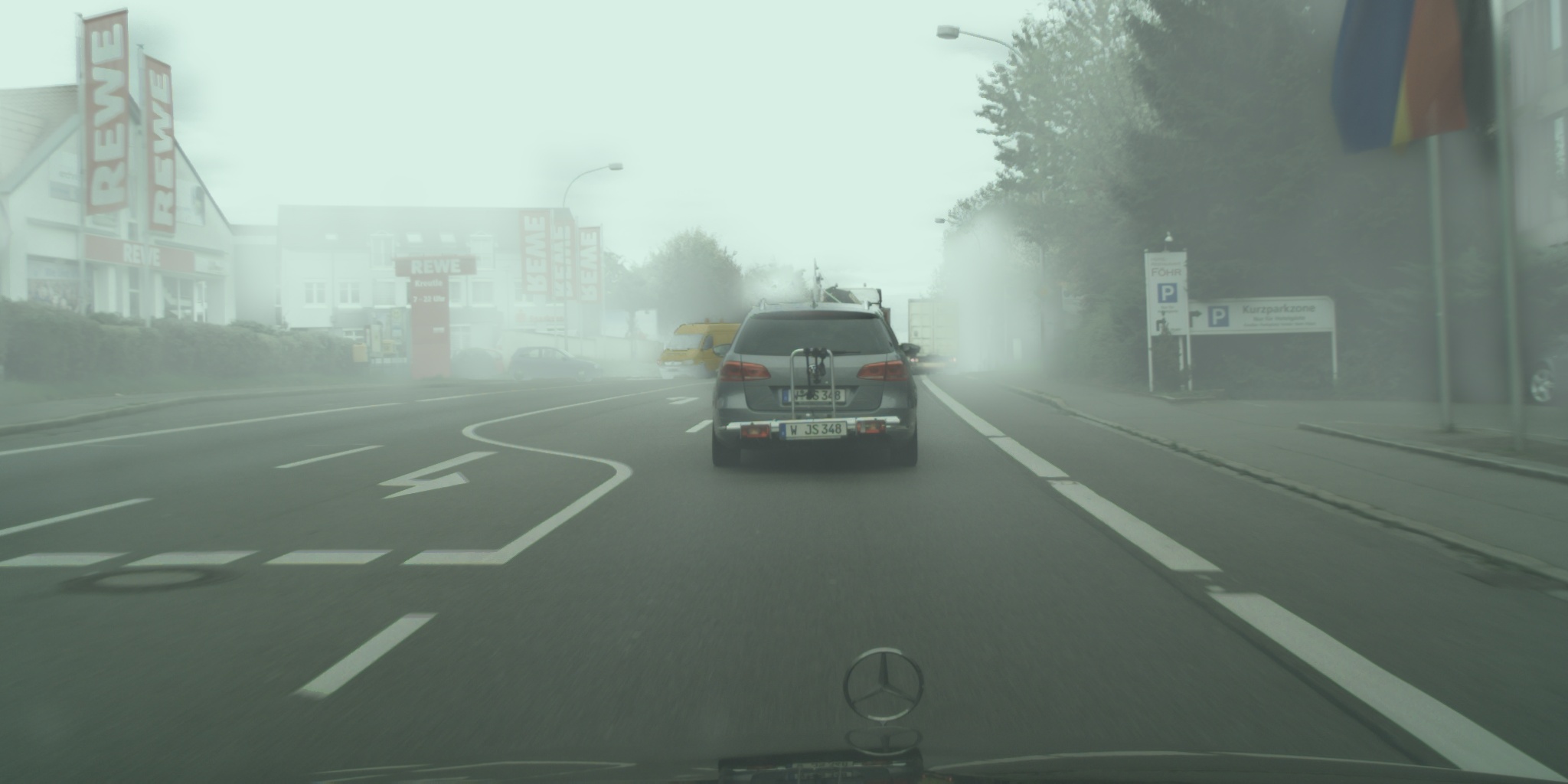}}
\hfill
\mpage{0.22}{\includegraphics[width=1.0\linewidth]{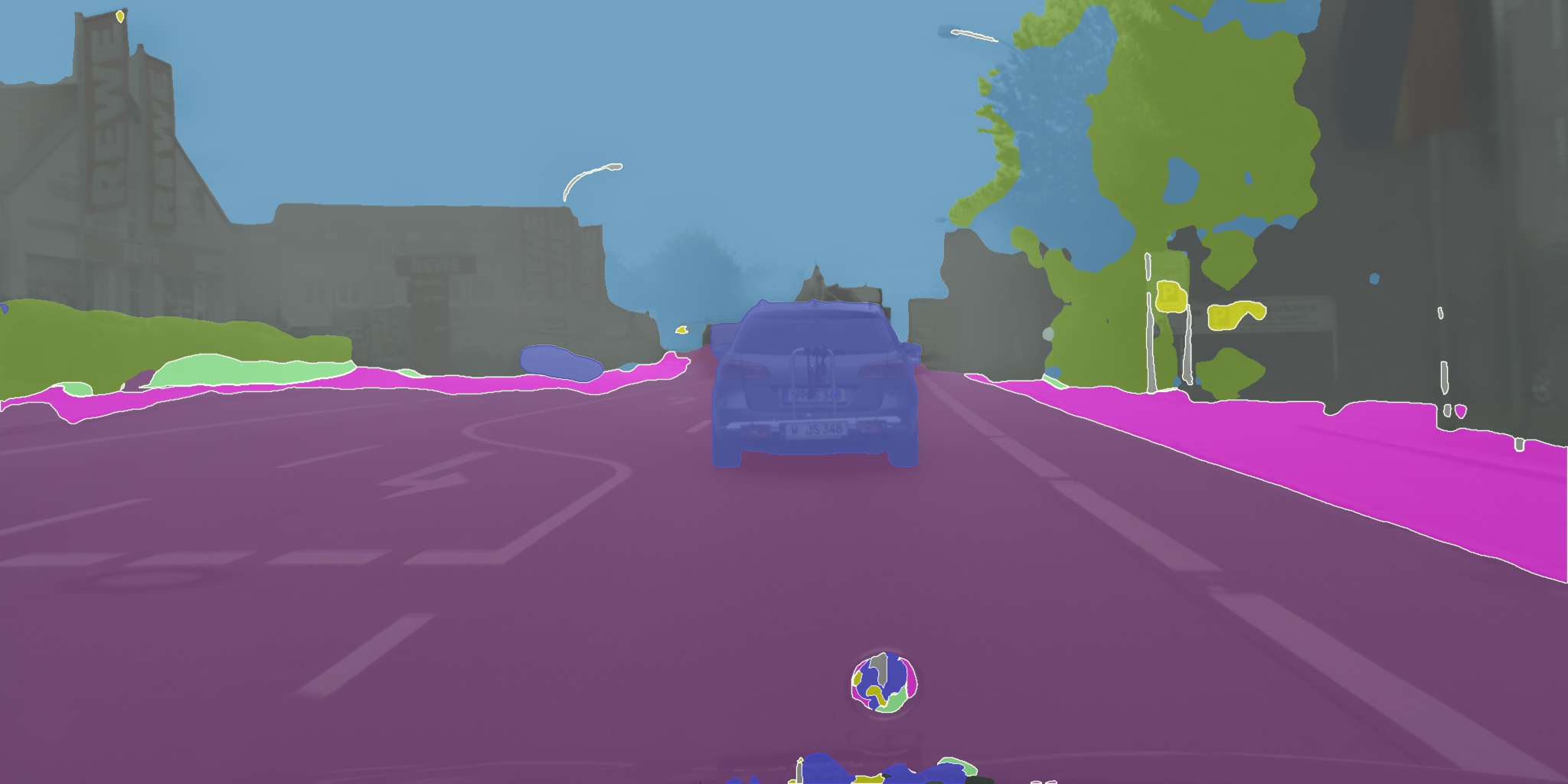}}
\hfill
\mpage{0.22}{\includegraphics[width=1.0\linewidth]{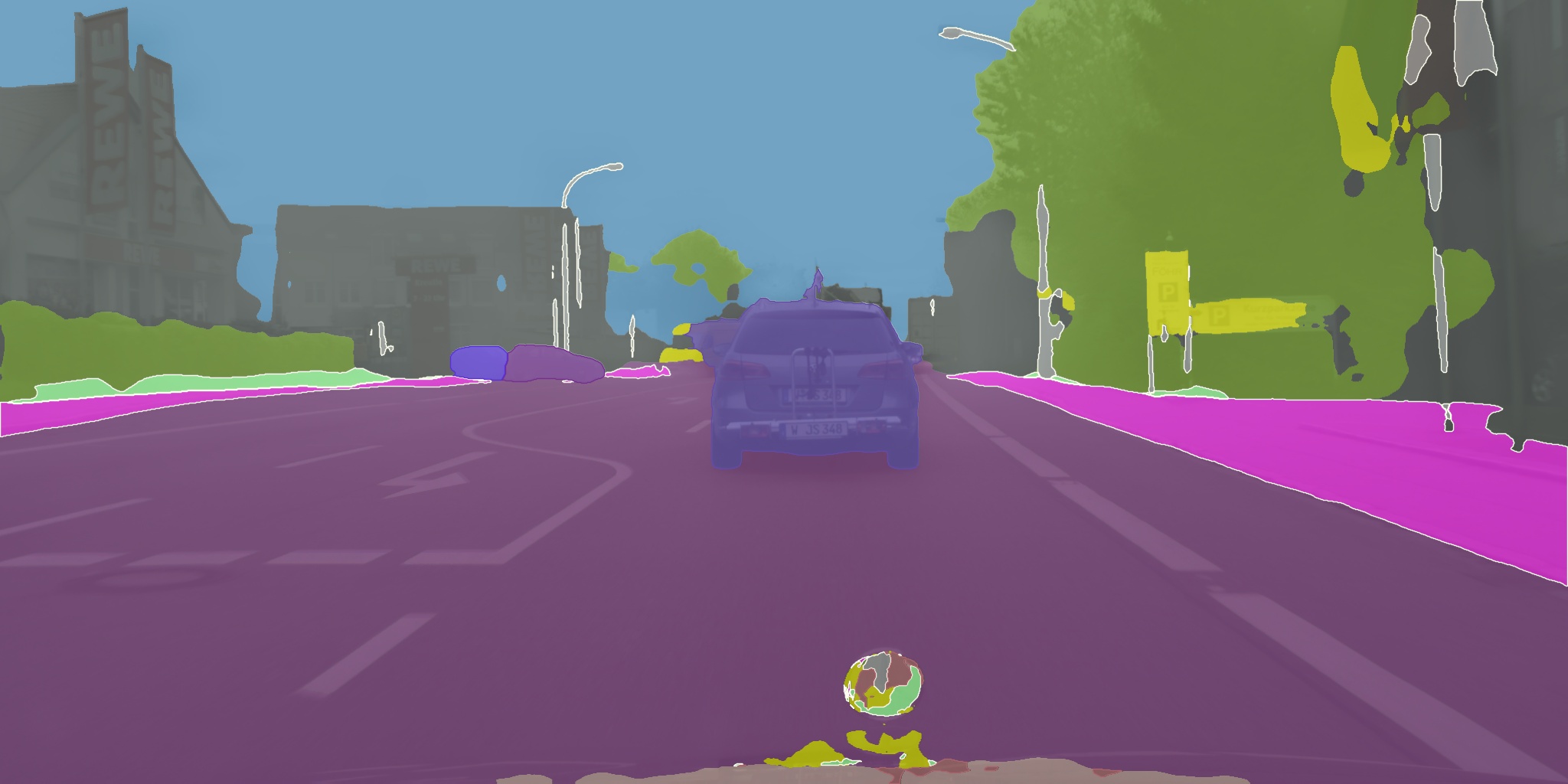}}
\hfill
\mpage{0.22}{\includegraphics[width=1.0\linewidth]{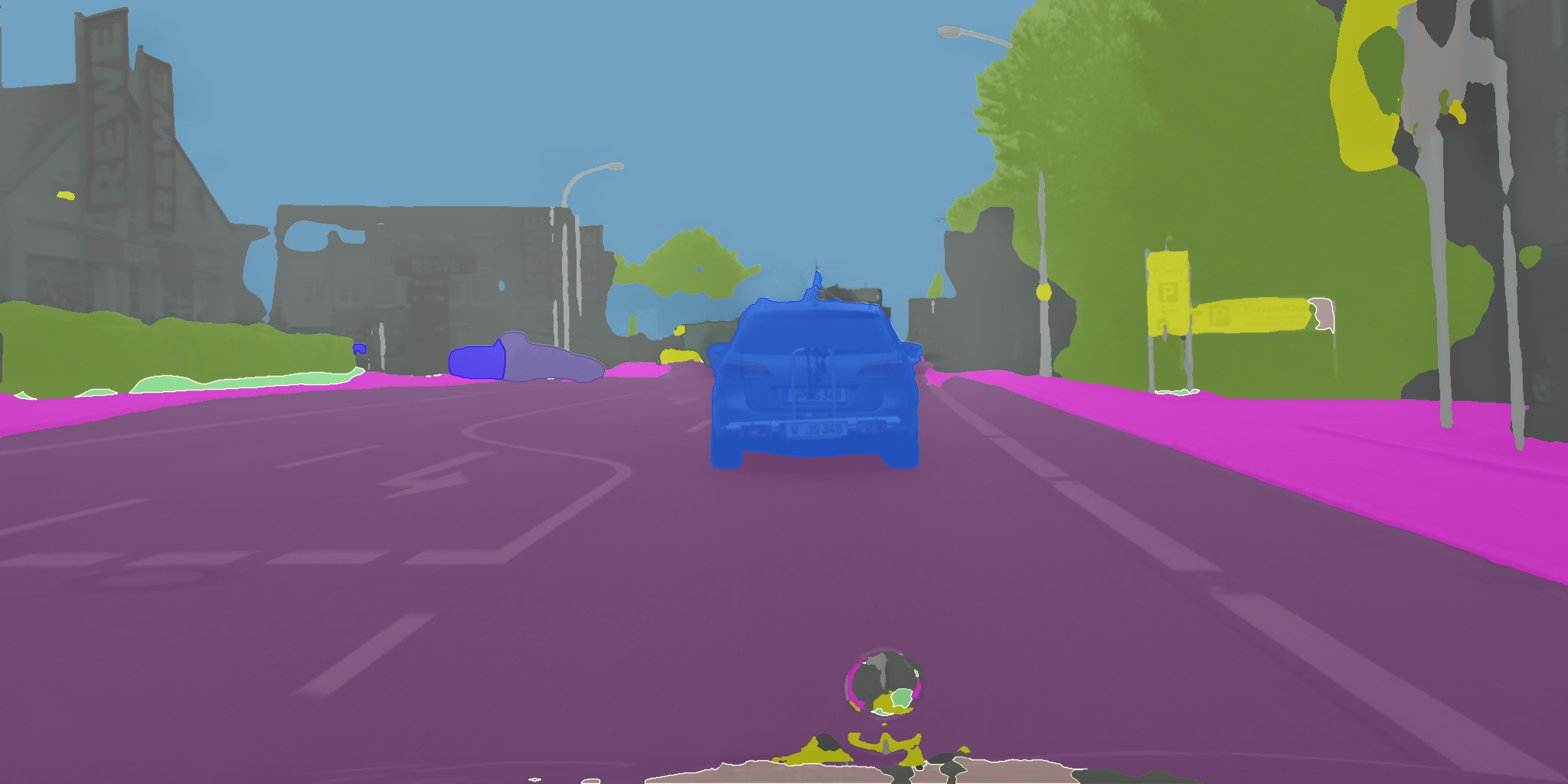}}
\hfill
\\
\mpage{0.22}{\includegraphics[width=1.0\linewidth]{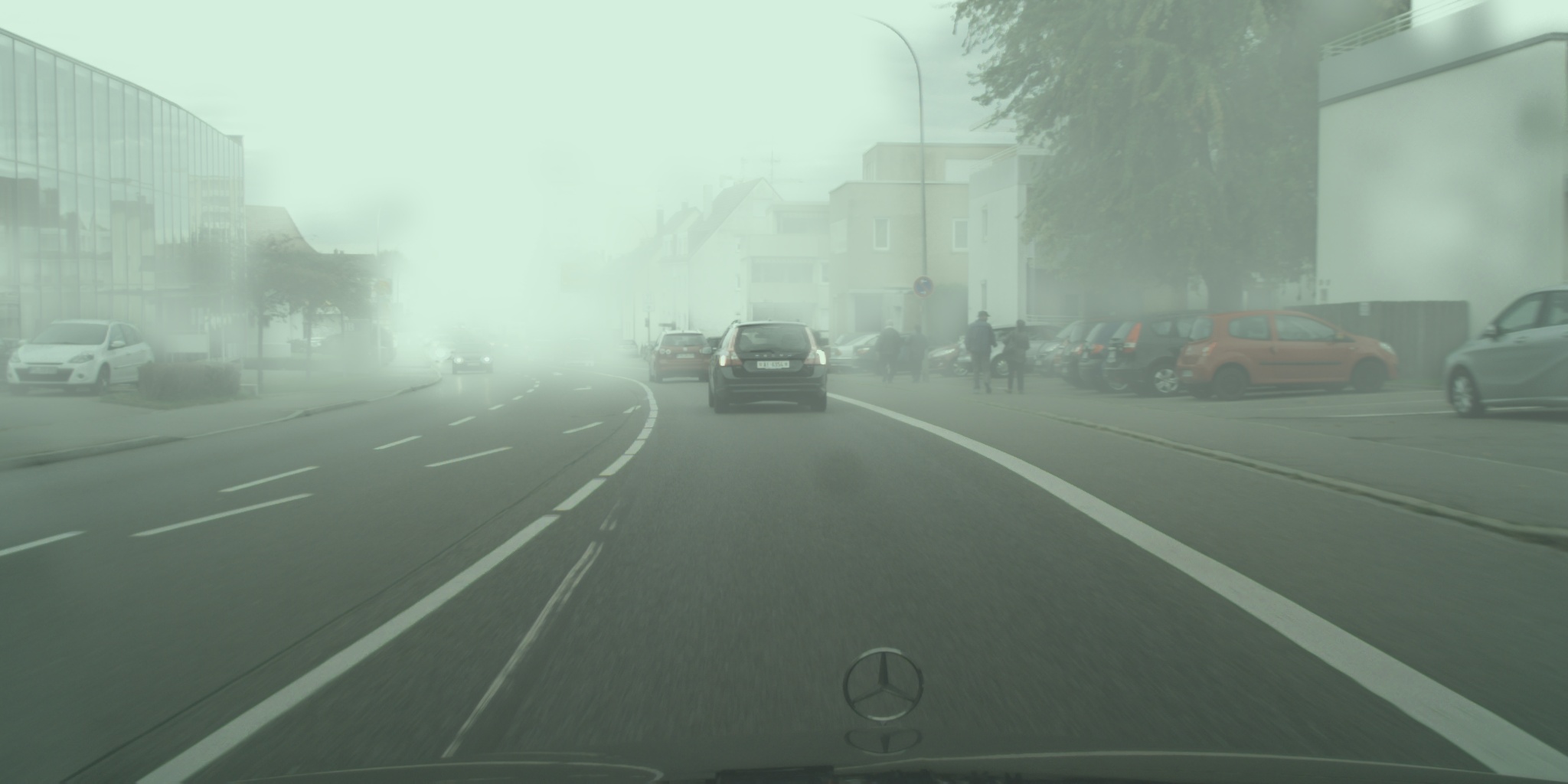}}
\hfill
\mpage{0.22}{\includegraphics[width=1.0\linewidth]{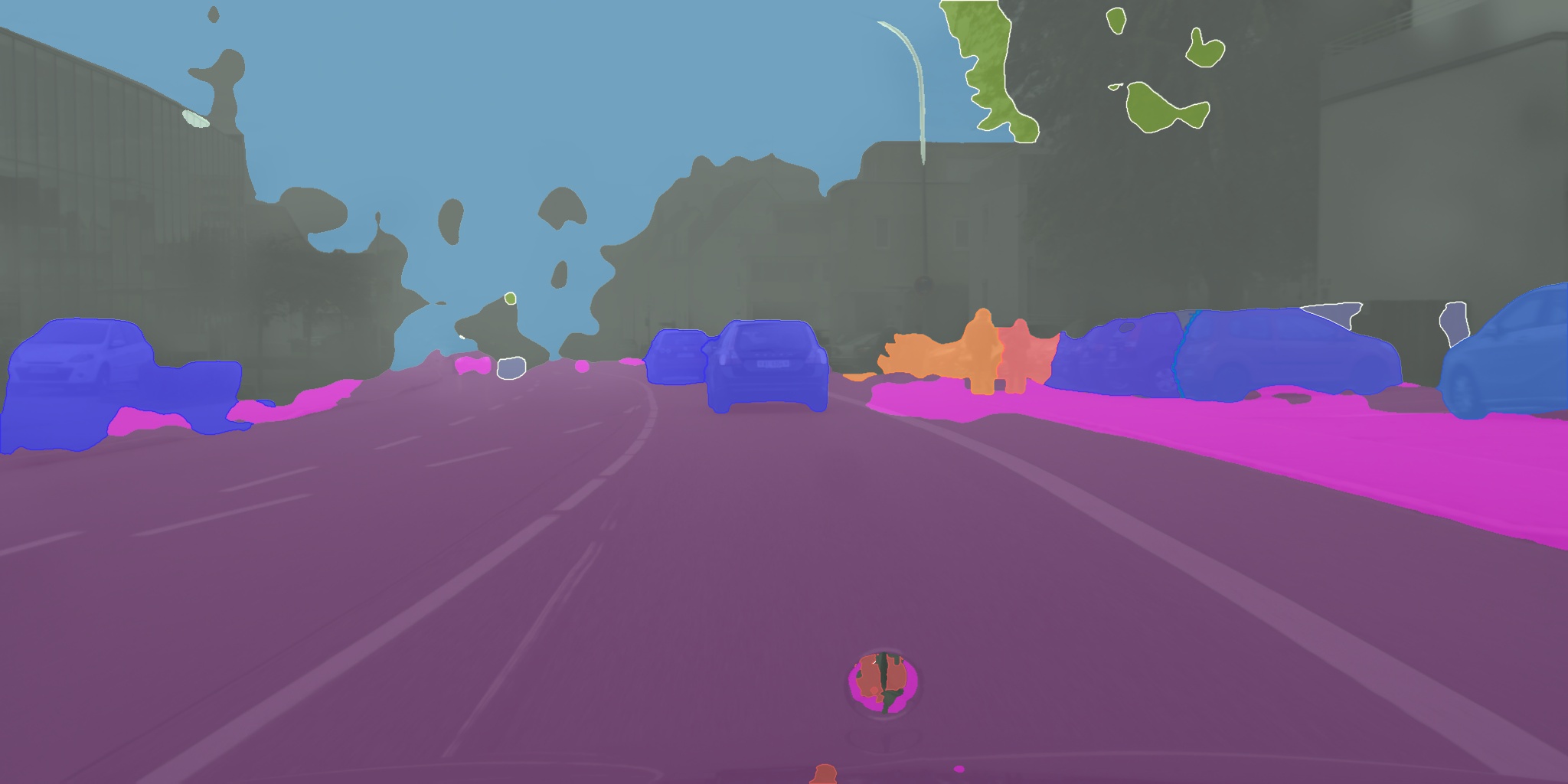}}
\hfill
\mpage{0.22}{\includegraphics[width=1.0\linewidth]{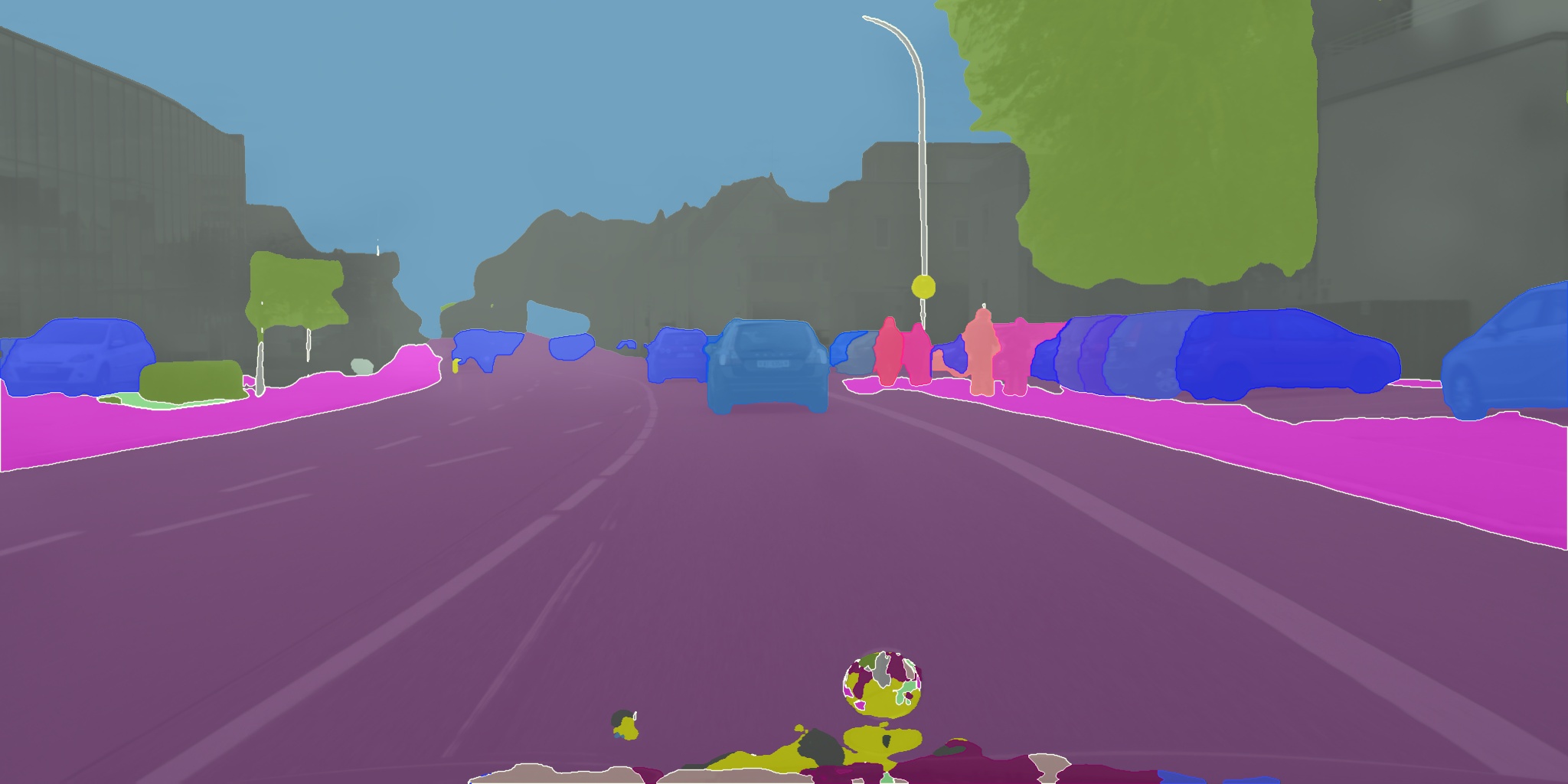}}
\hfill
\mpage{0.22}{\includegraphics[width=1.0\linewidth]{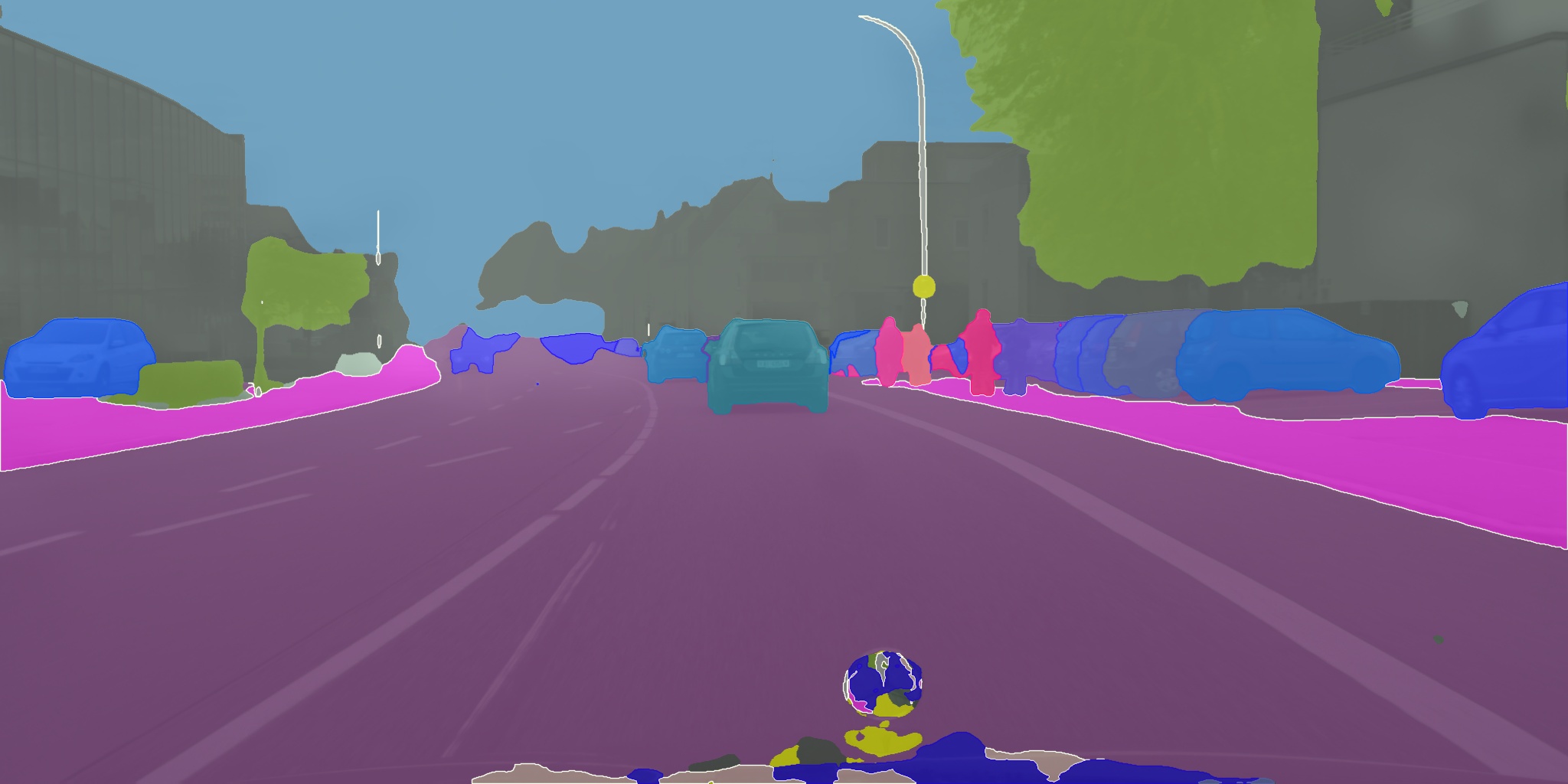}}
\hfill
\\
\mpage{0.22}{\includegraphics[width=1.0\linewidth]{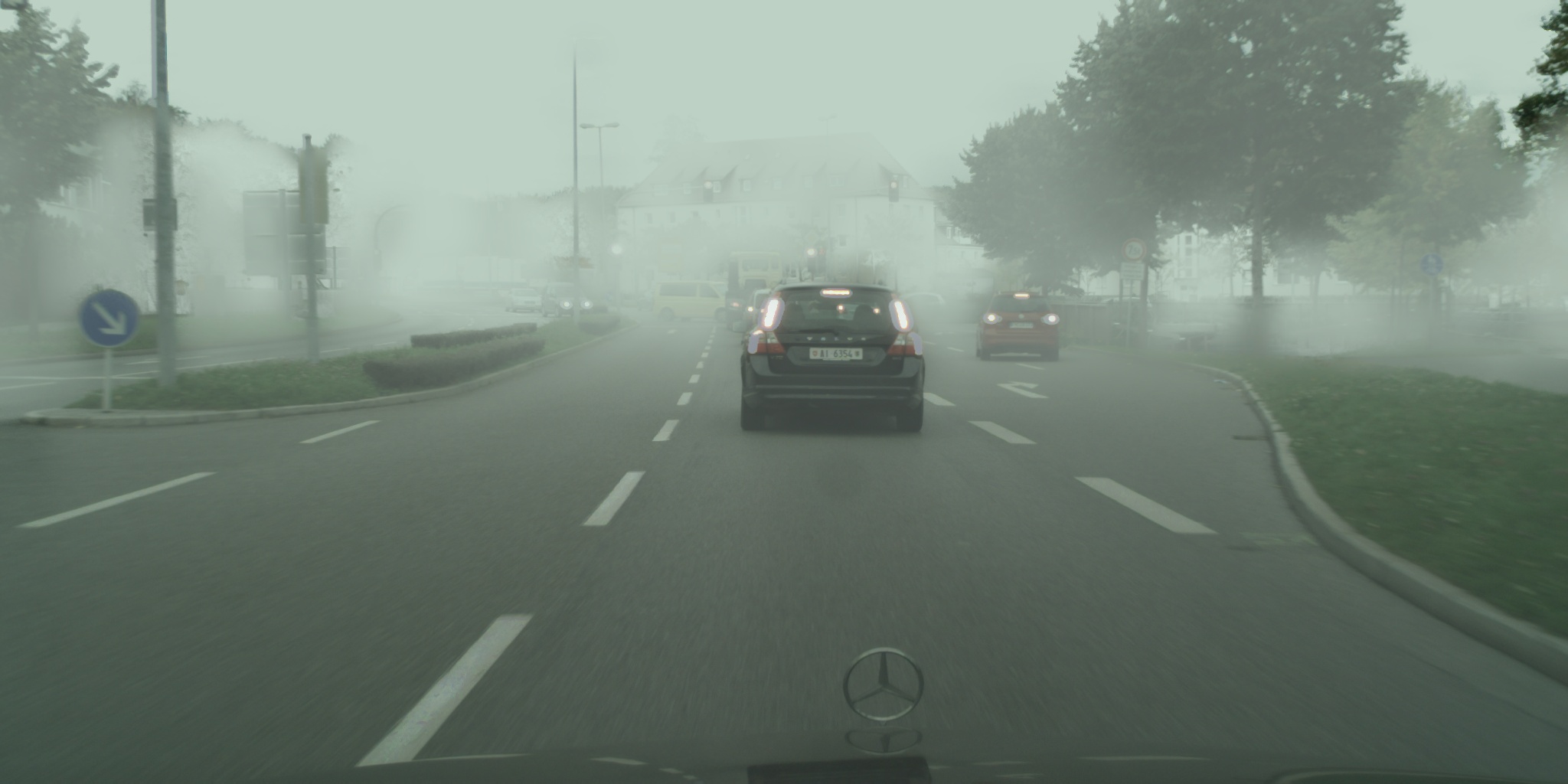}}
\hfill
\mpage{0.22}{\includegraphics[width=1.0\linewidth]{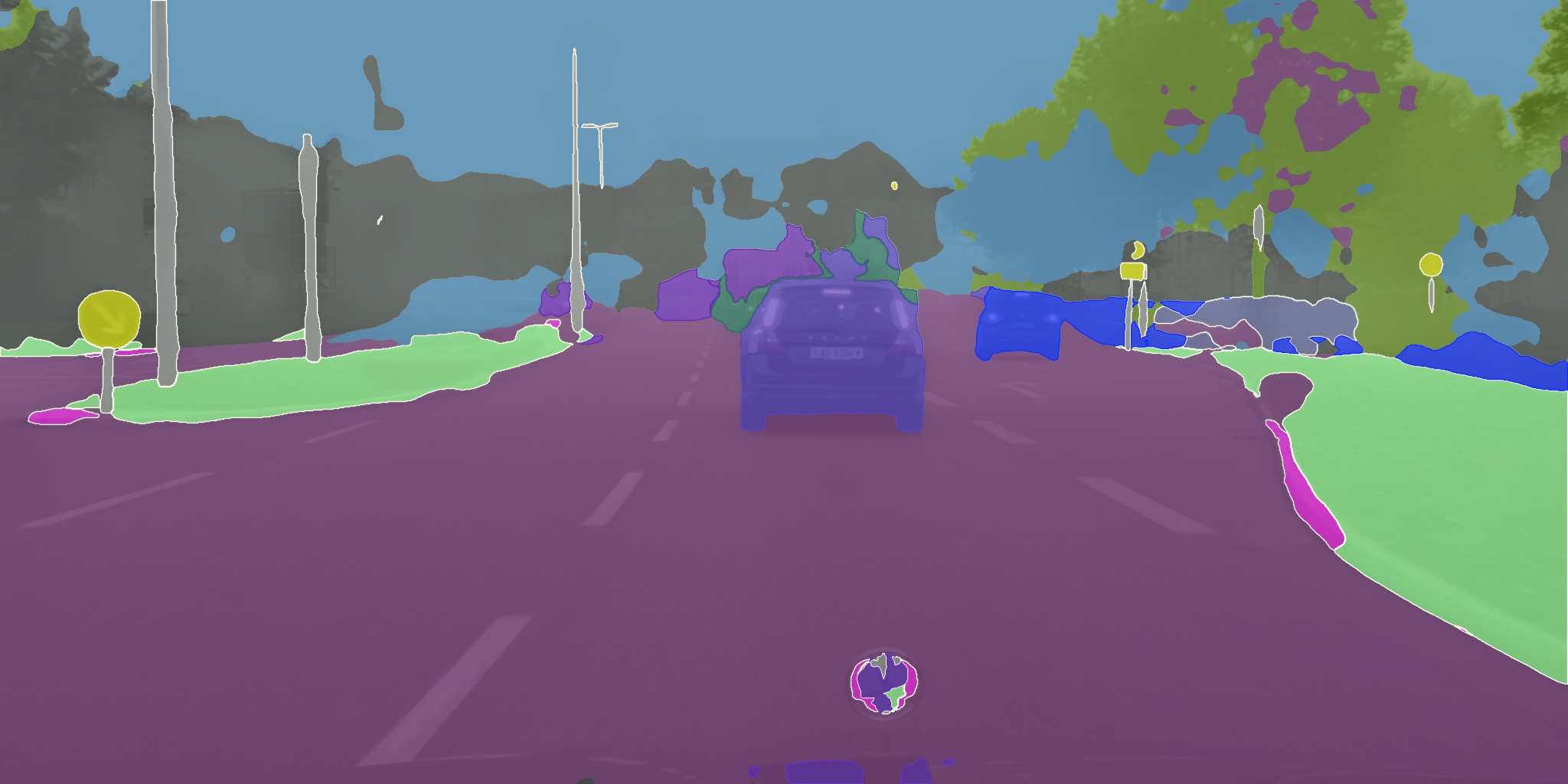}}
\hfill
\mpage{0.22}{\includegraphics[width=1.0\linewidth]{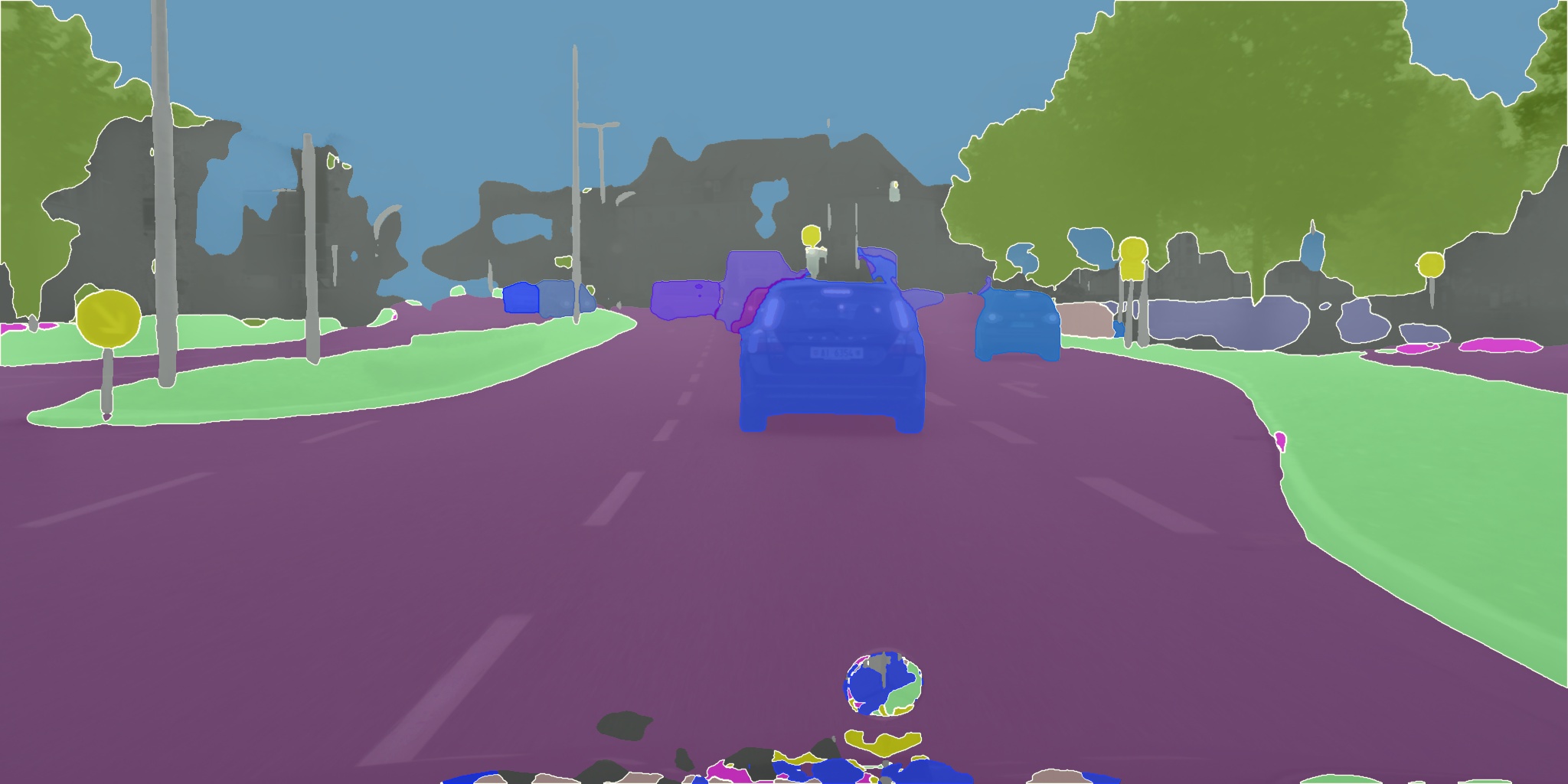}}
\hfill
\mpage{0.22}{\includegraphics[width=1.0\linewidth]{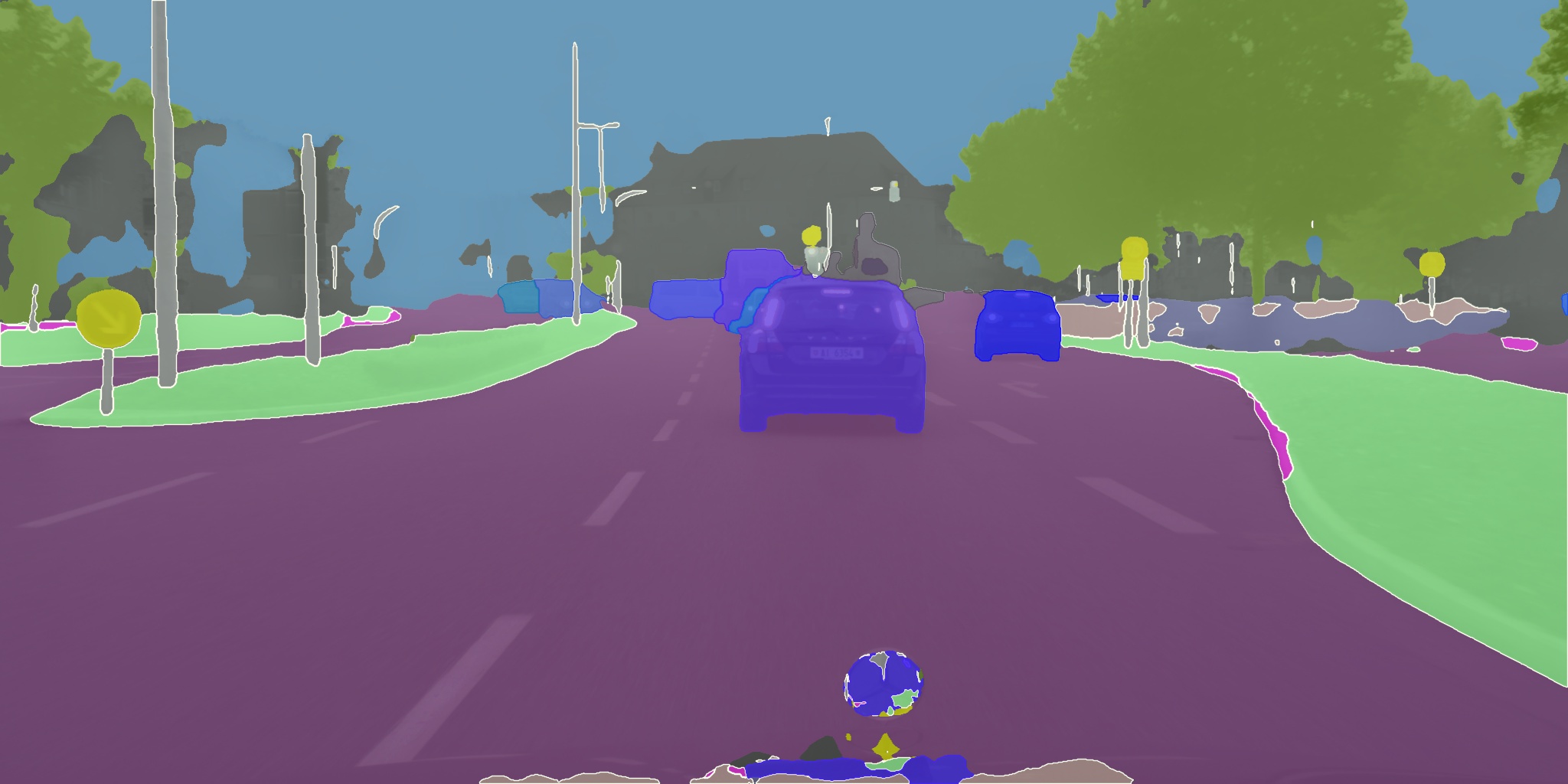}}
\hfill
\\
\mpage{0.22}{\includegraphics[width=1.0\linewidth]{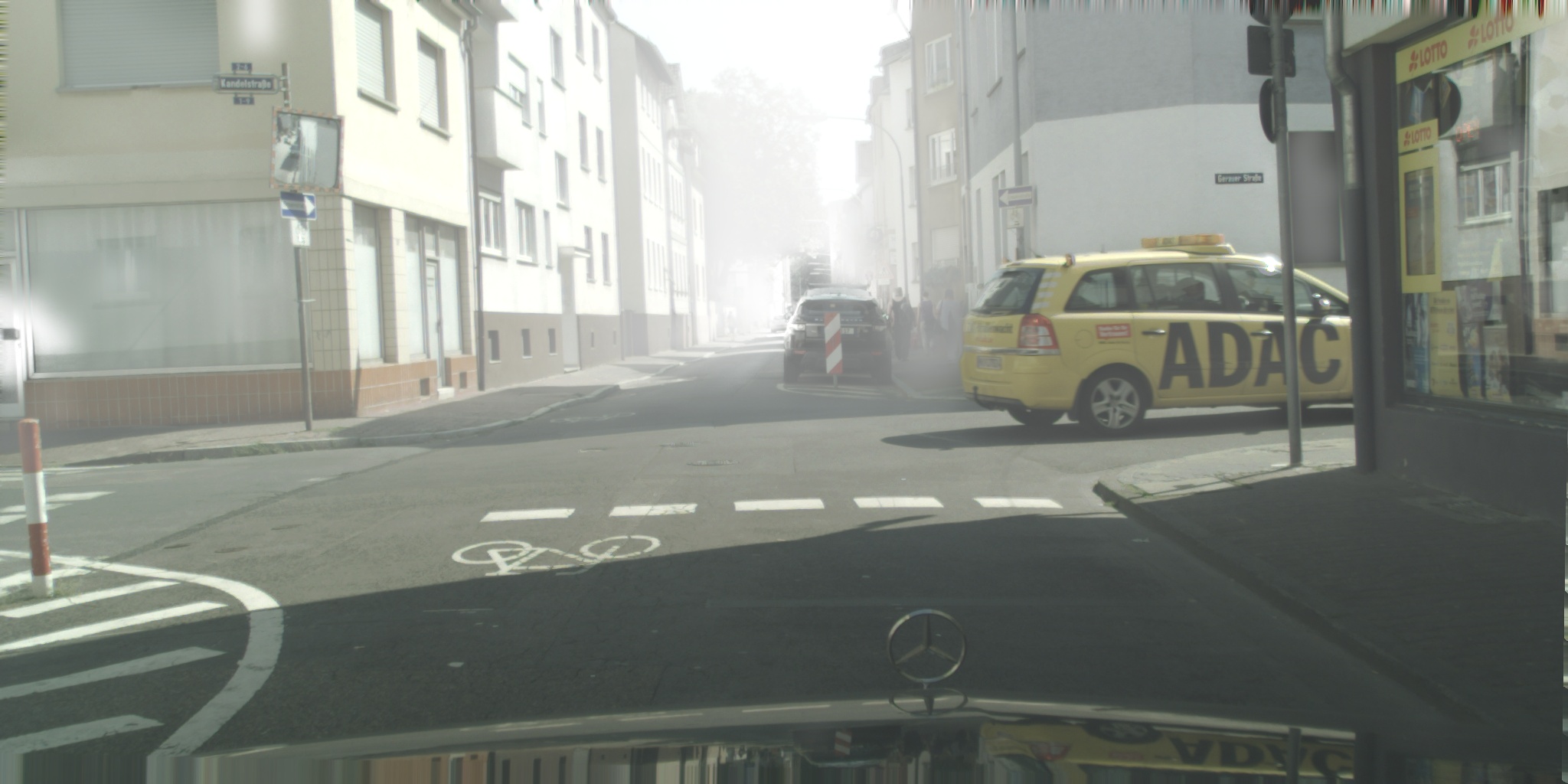}}
\hfill
\mpage{0.22}{\includegraphics[width=1.0\linewidth]{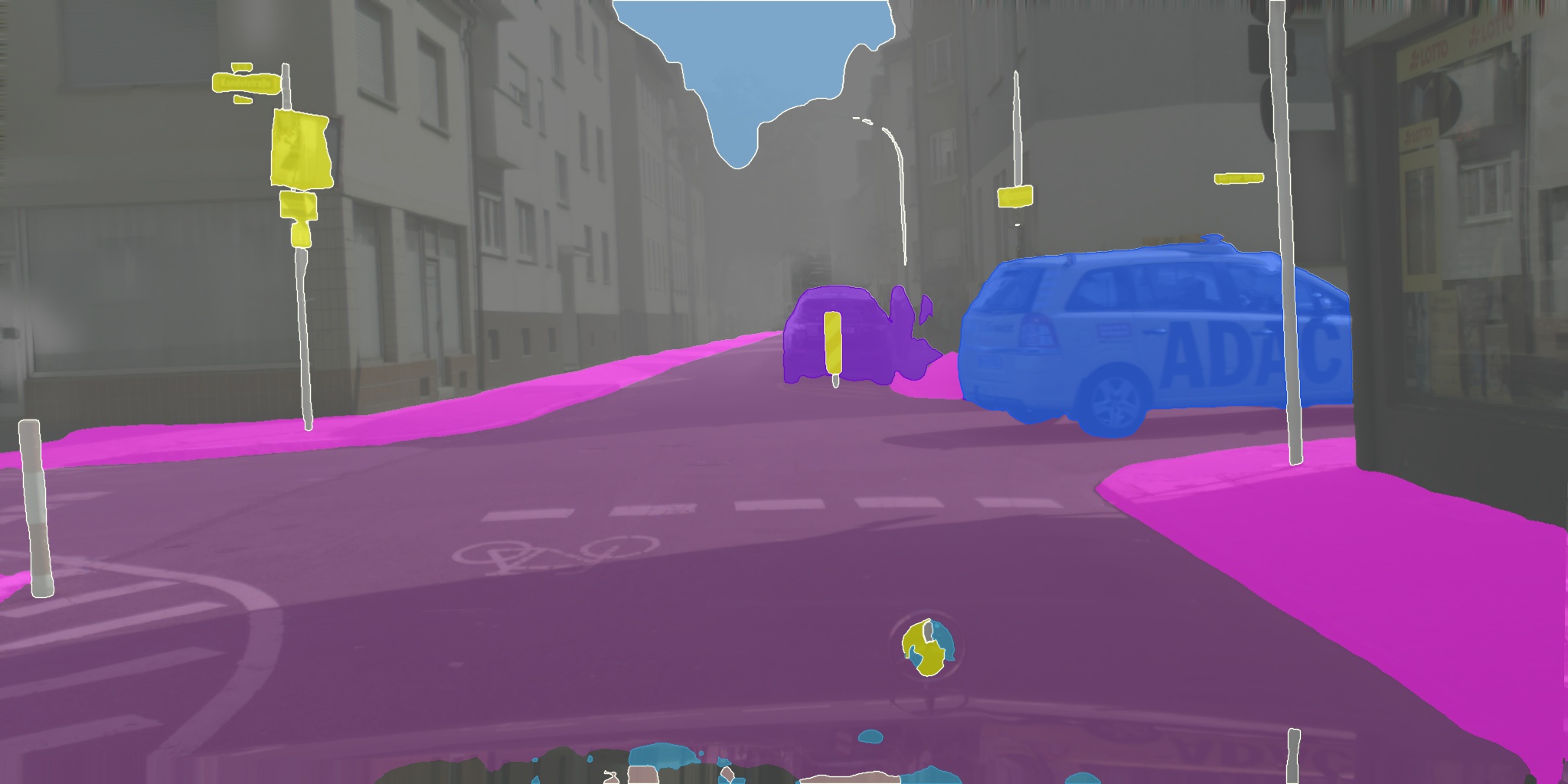}}
\hfill
\mpage{0.22}{\includegraphics[width=1.0\linewidth]{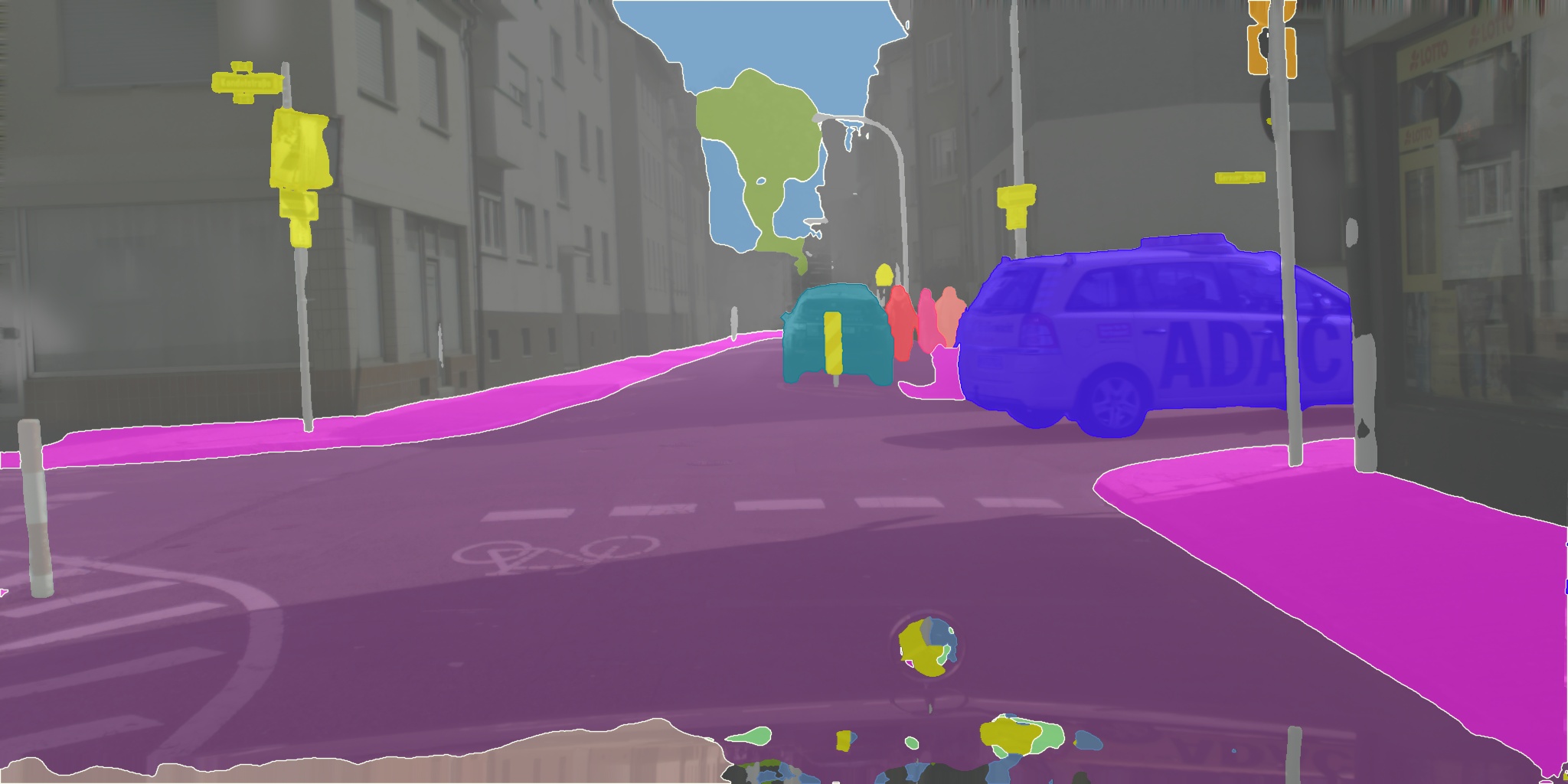}}
\hfill
\mpage{0.22}{\includegraphics[width=1.0\linewidth]{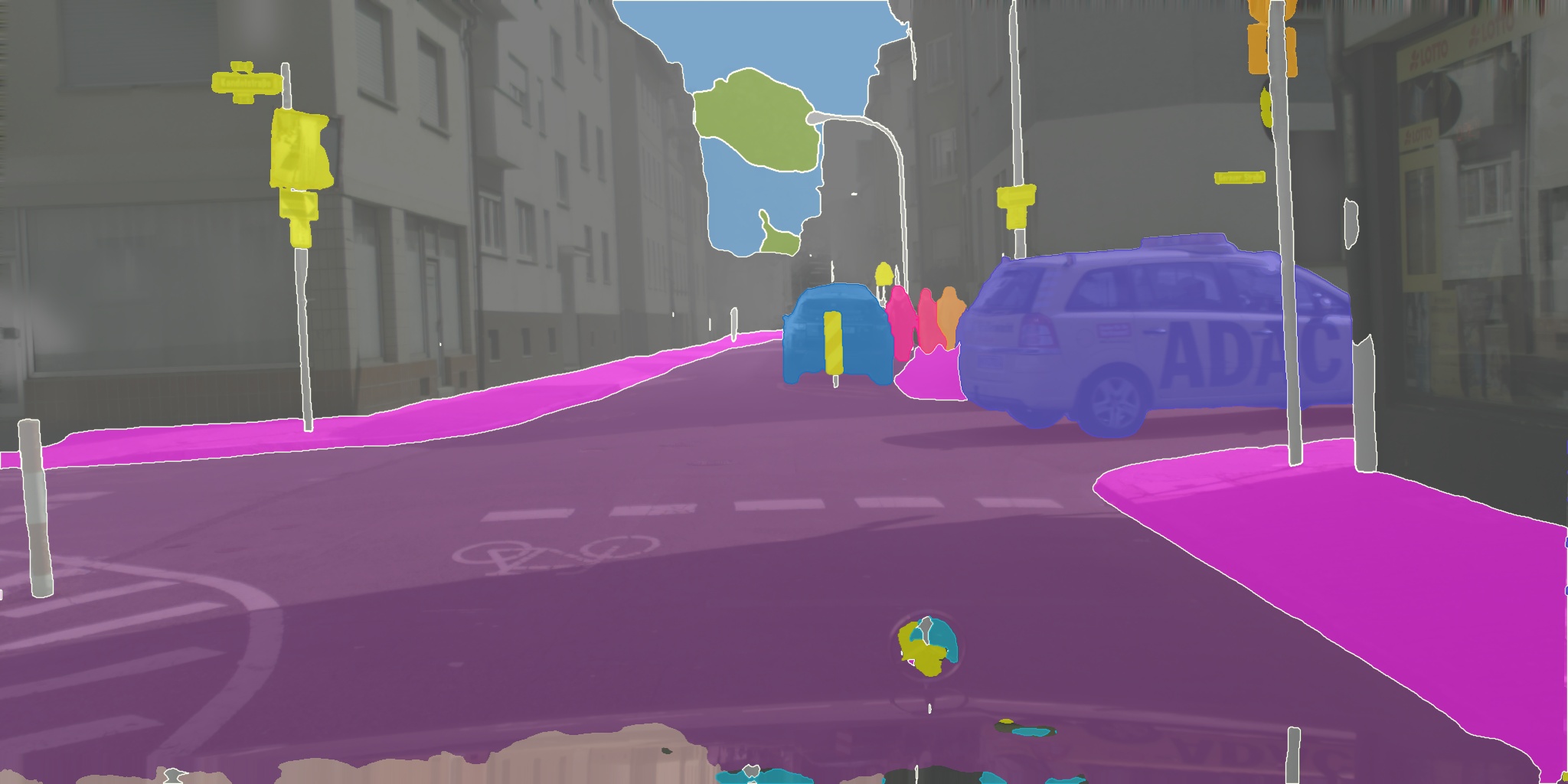}}
\hfill
\\
\mpage{0.22}{\includegraphics[width=1.0\linewidth]{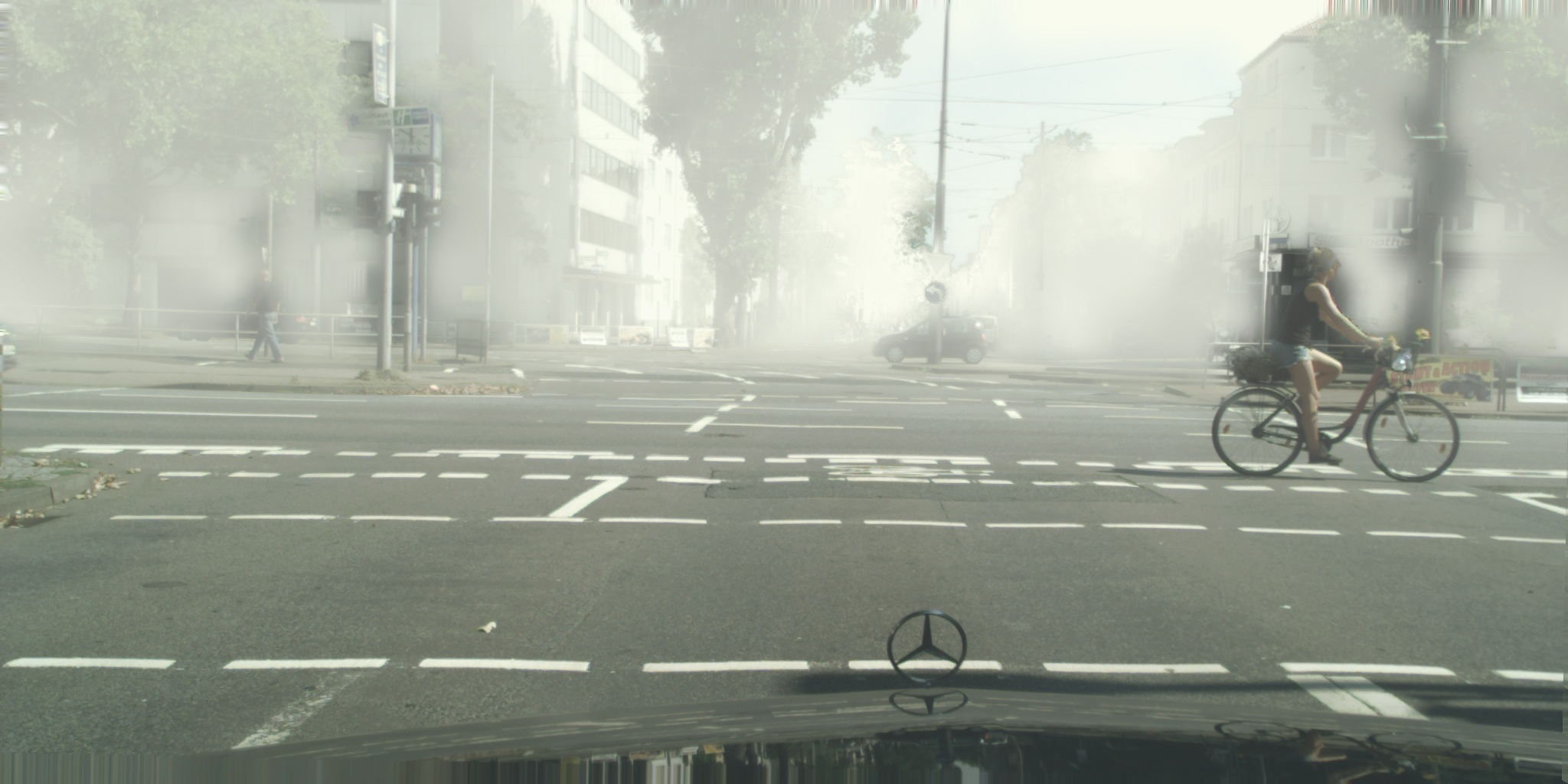}}
\hfill
\mpage{0.22}{\includegraphics[width=1.0\linewidth]{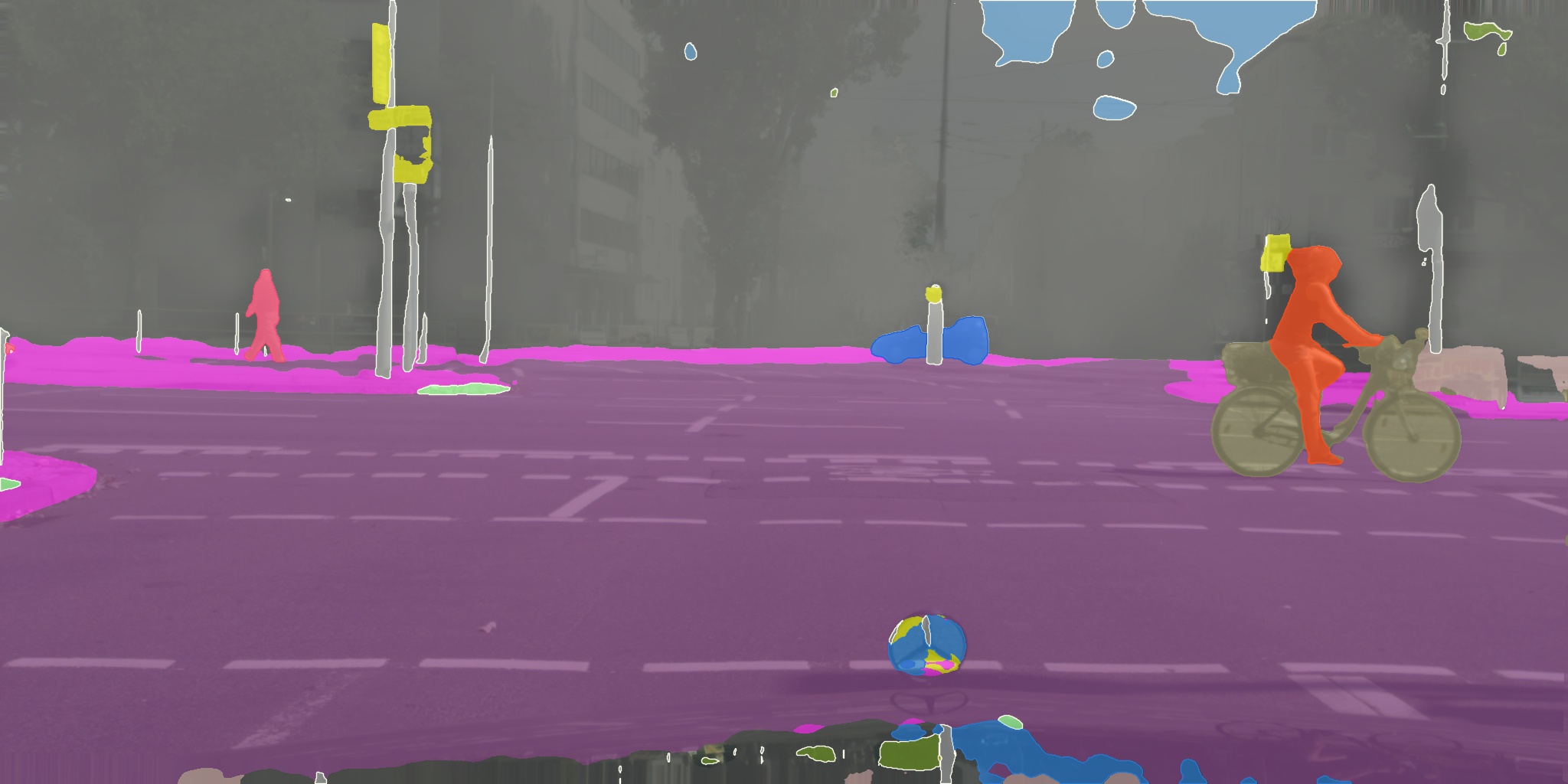}}
\hfill
\mpage{0.22}{\includegraphics[width=1.0\linewidth]{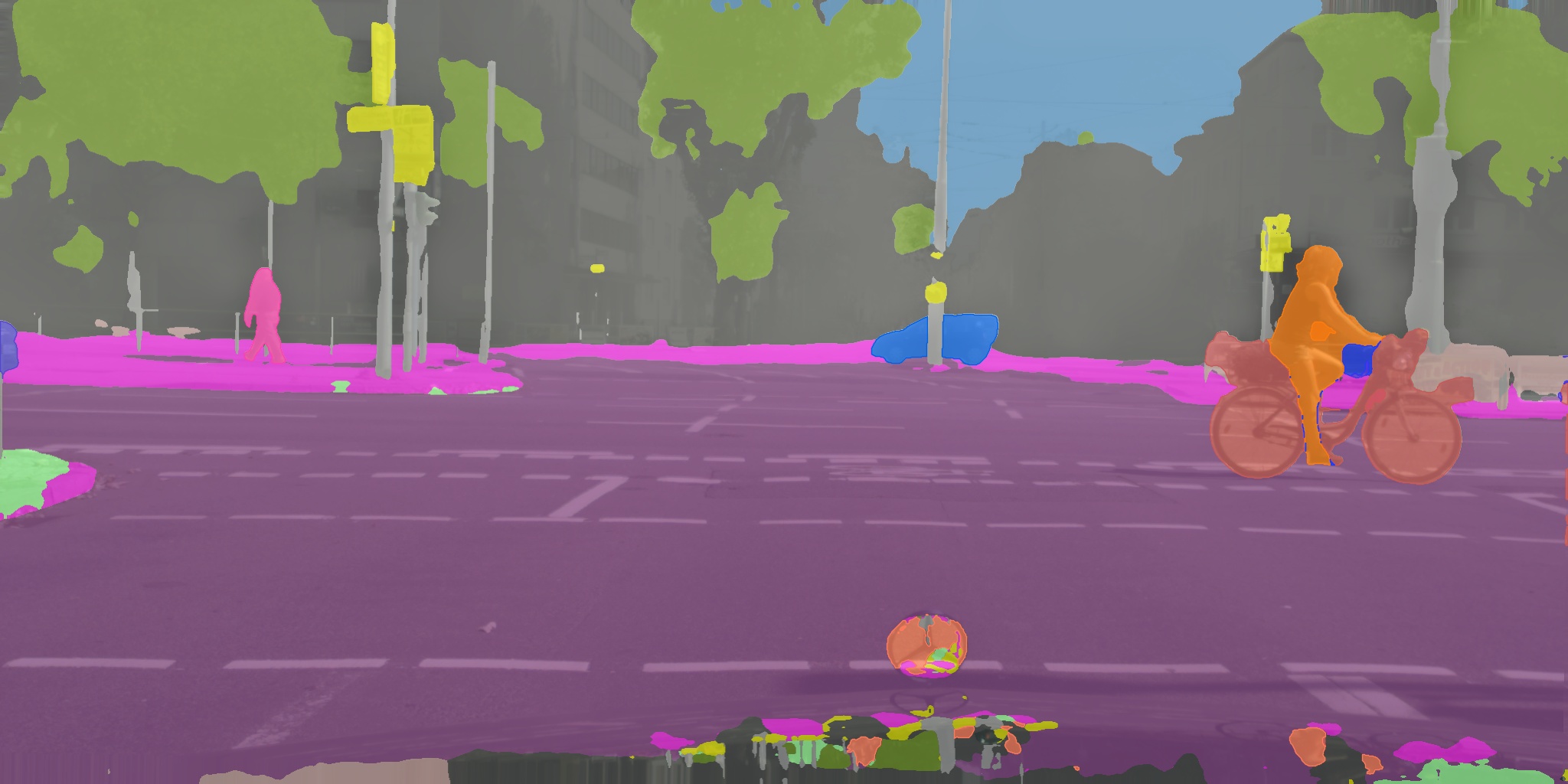}}
\hfill
\mpage{0.22}{\includegraphics[width=1.0\linewidth]{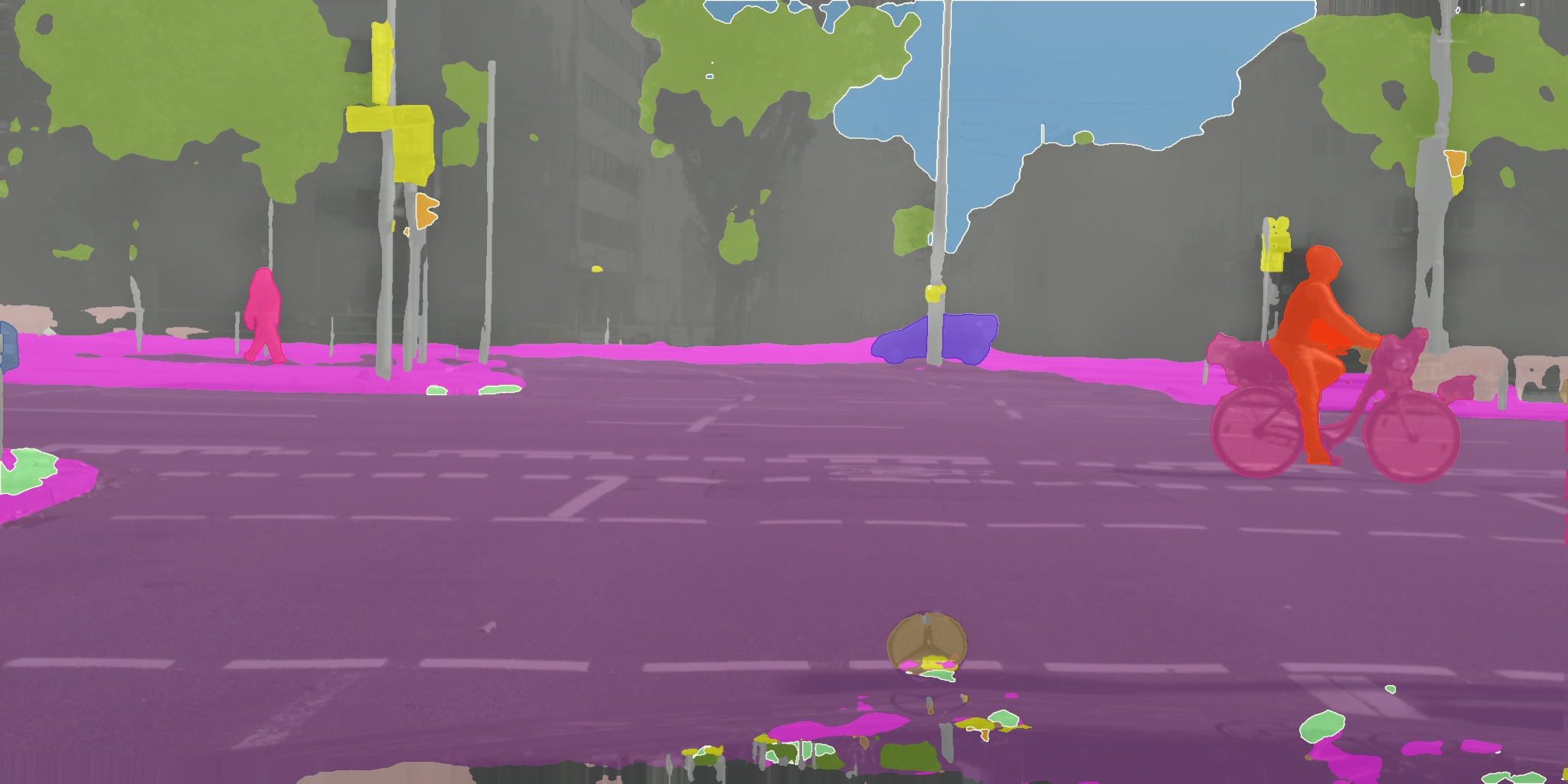}}
\hfill
\\
\mpage{0.22}{\includegraphics[width=1.0\linewidth]{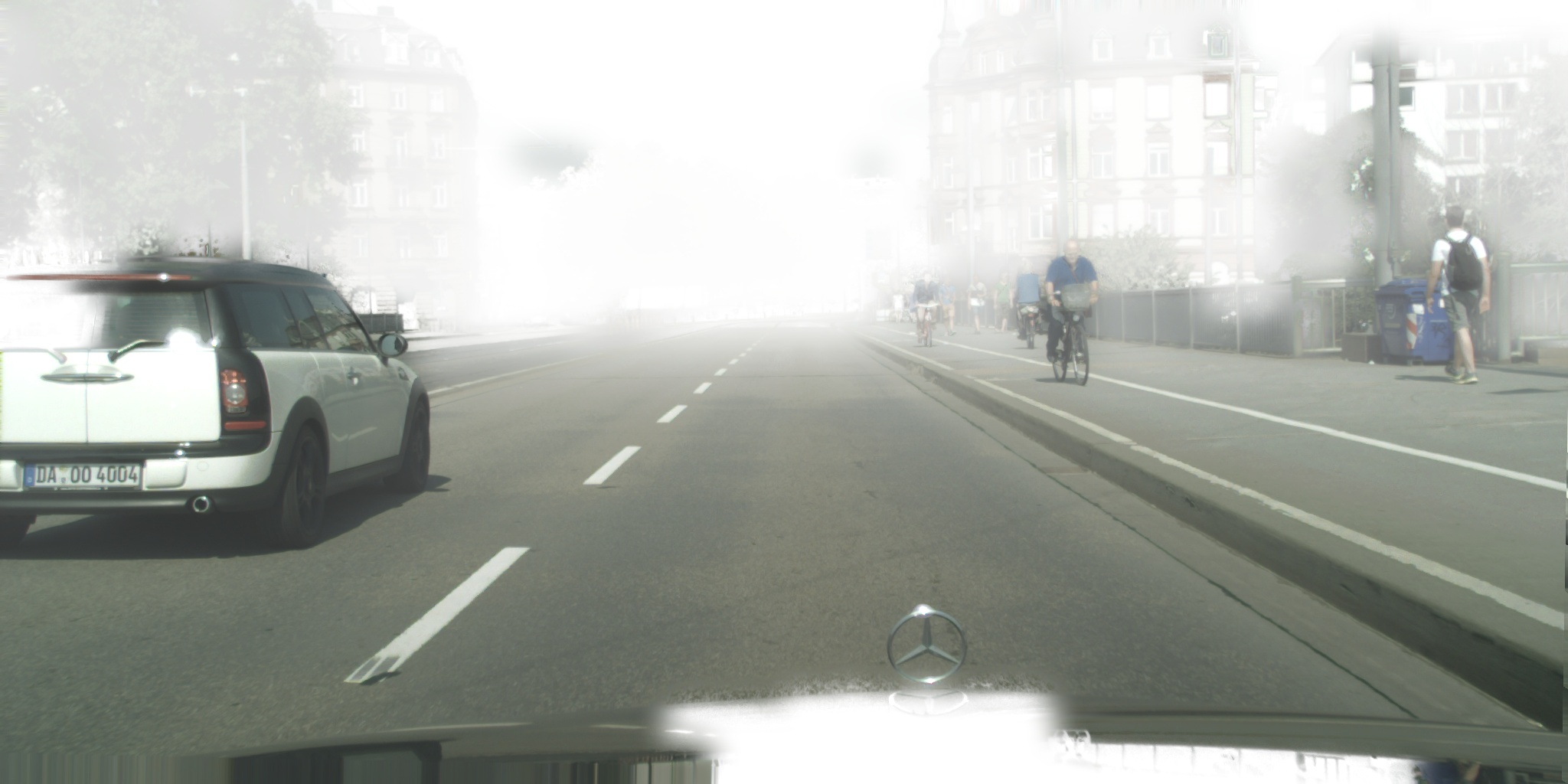}}
\hfill
\mpage{0.22}{\includegraphics[width=1.0\linewidth]{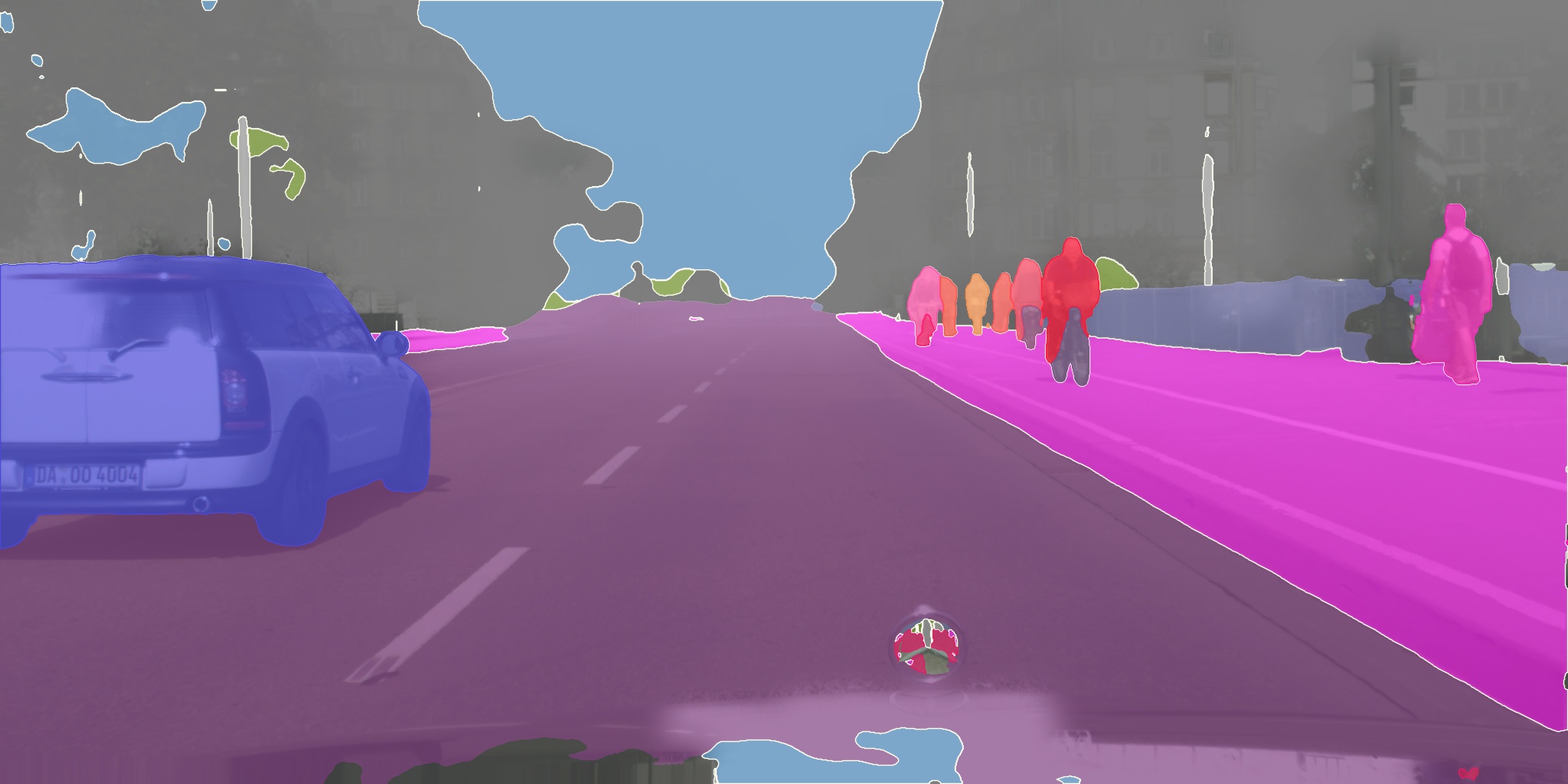}}
\hfill
\mpage{0.22}{\includegraphics[width=1.0\linewidth]{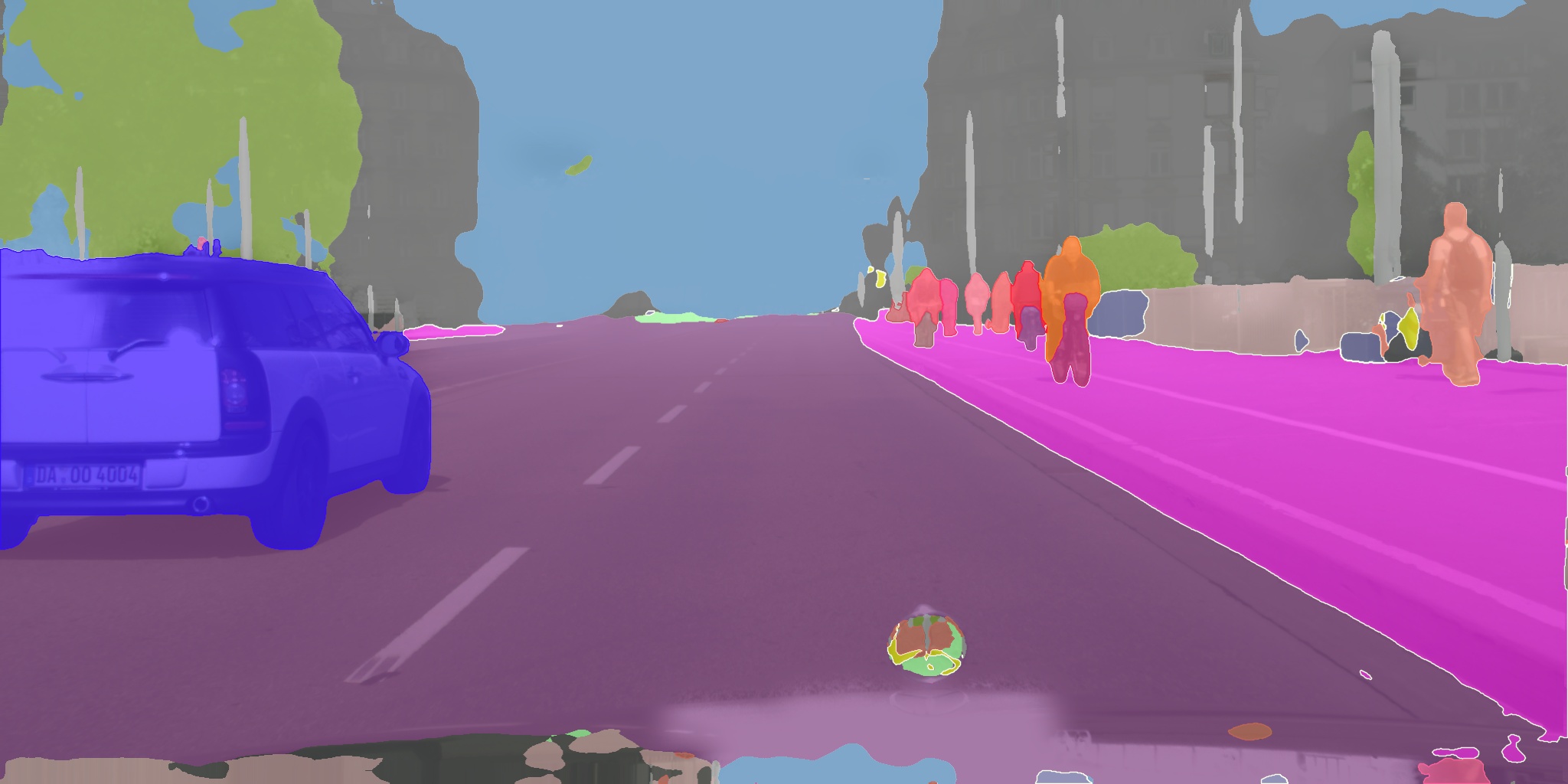}}
\hfill
\mpage{0.22}{\includegraphics[width=1.0\linewidth]{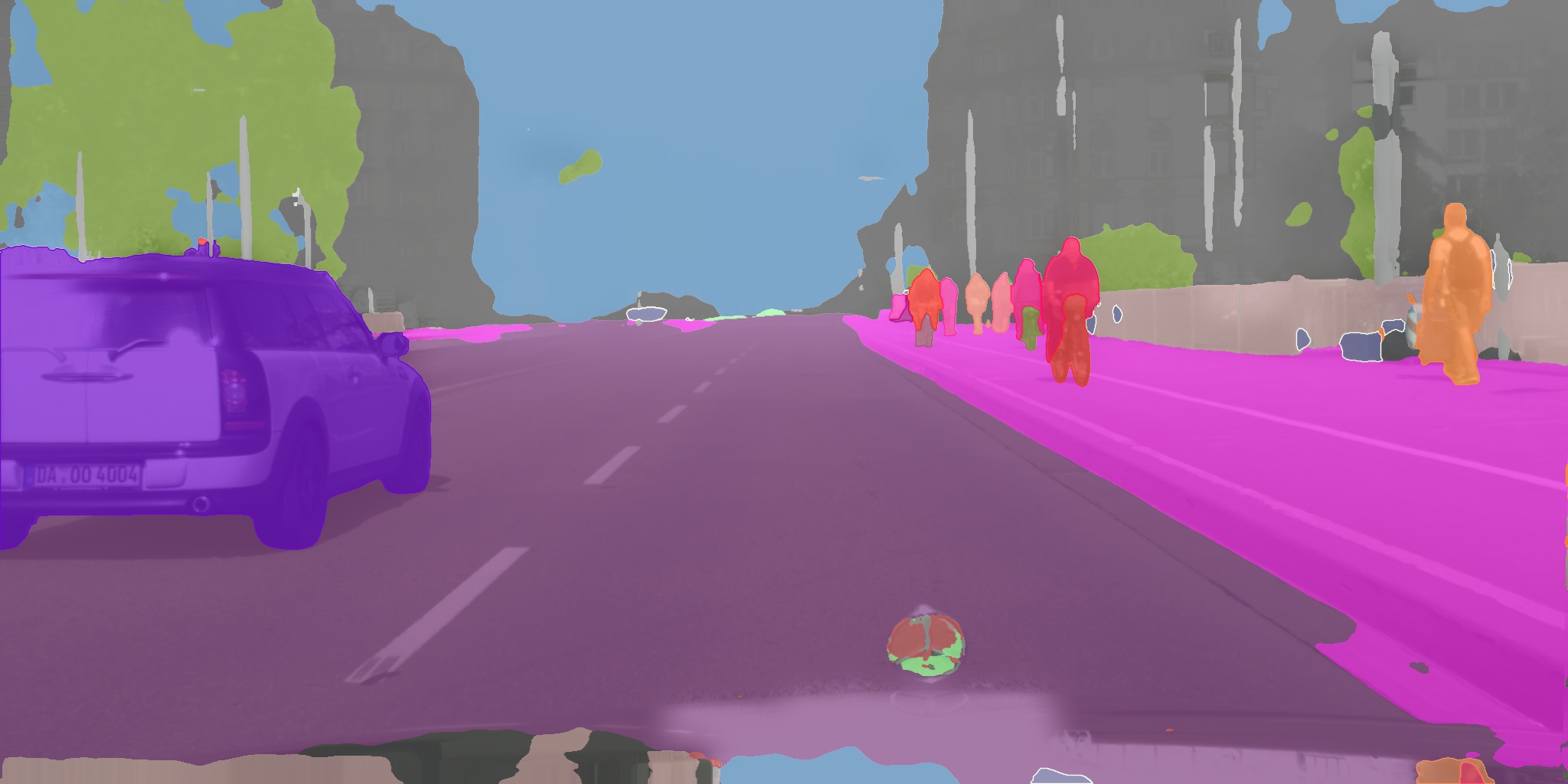}}
\hfill
\\
\mpage{0.22}{\includegraphics[width=1.0\linewidth]{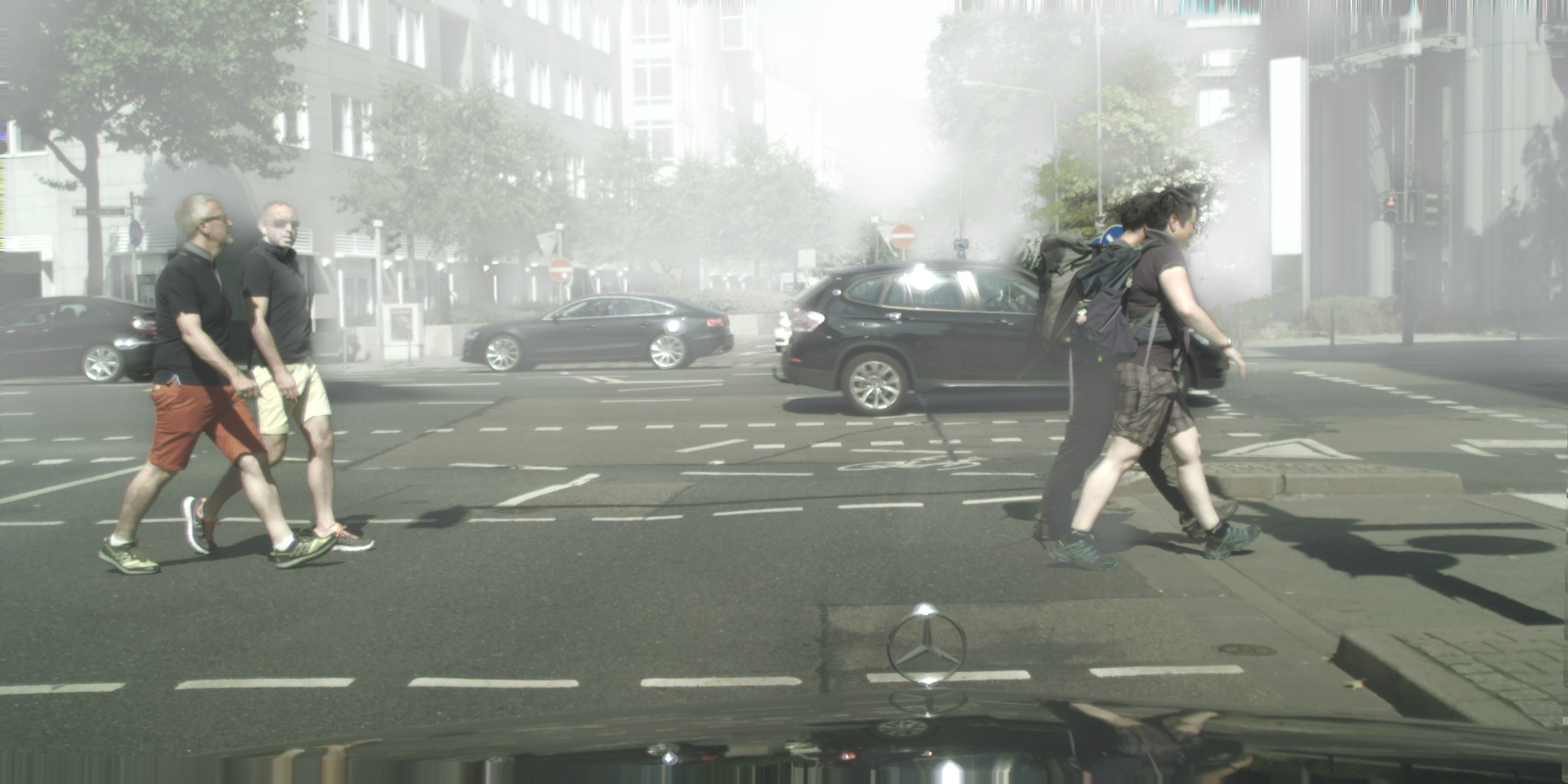}}
\hfill
\mpage{0.22}{\includegraphics[width=1.0\linewidth]{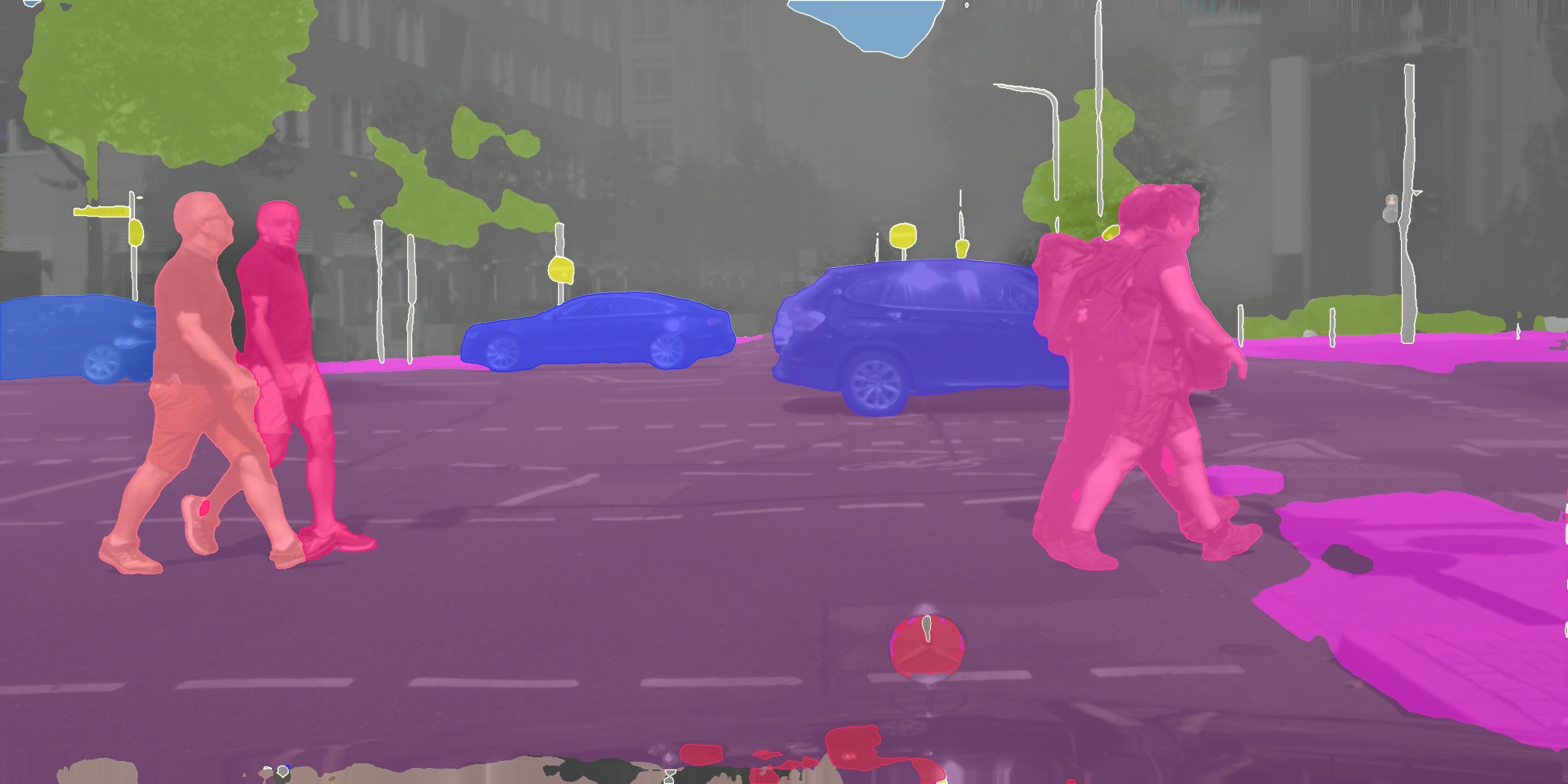}}
\hfill
\mpage{0.22}{\includegraphics[width=1.0\linewidth]{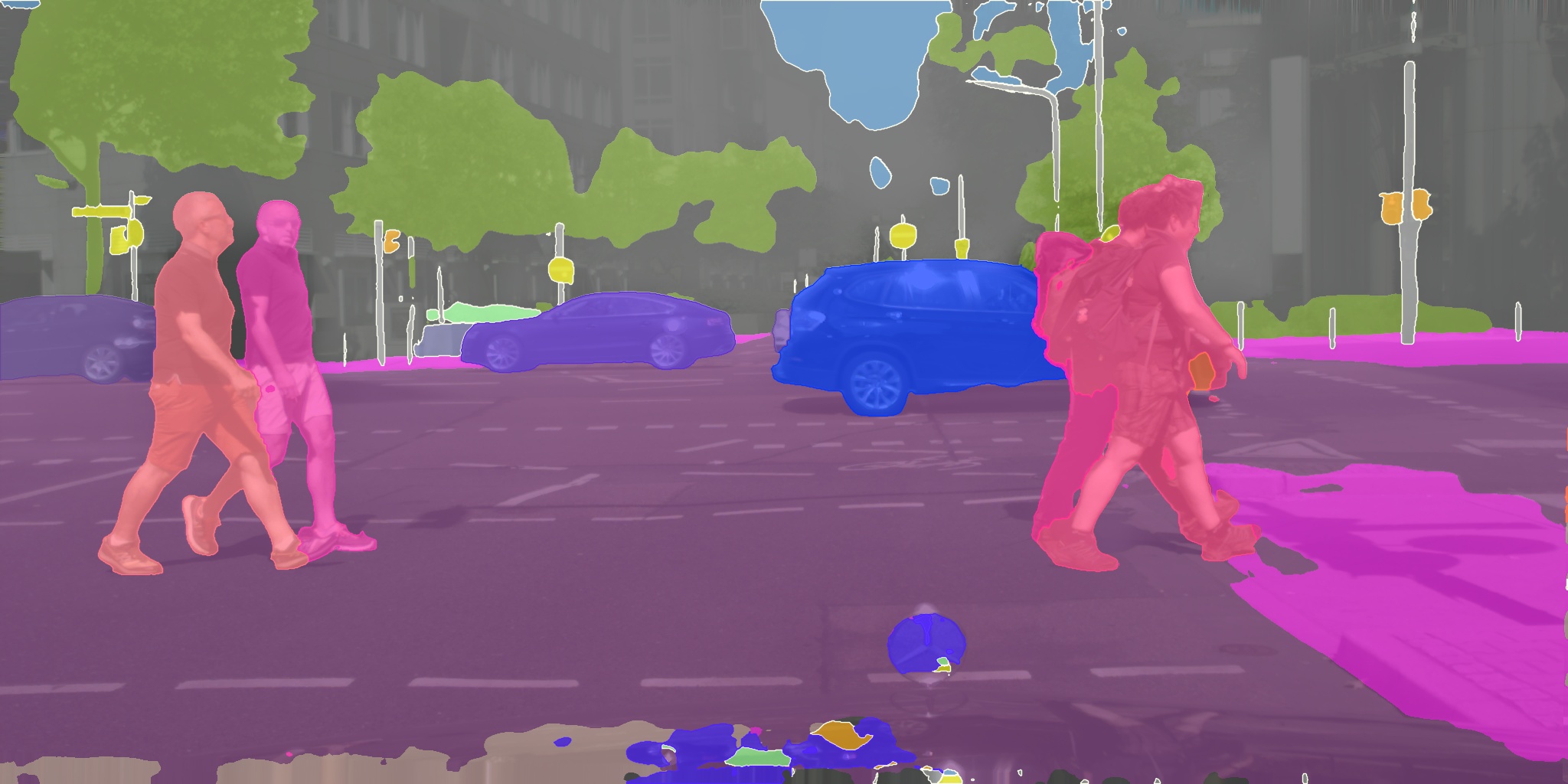}}
\hfill
\mpage{0.22}{\includegraphics[width=1.0\linewidth]{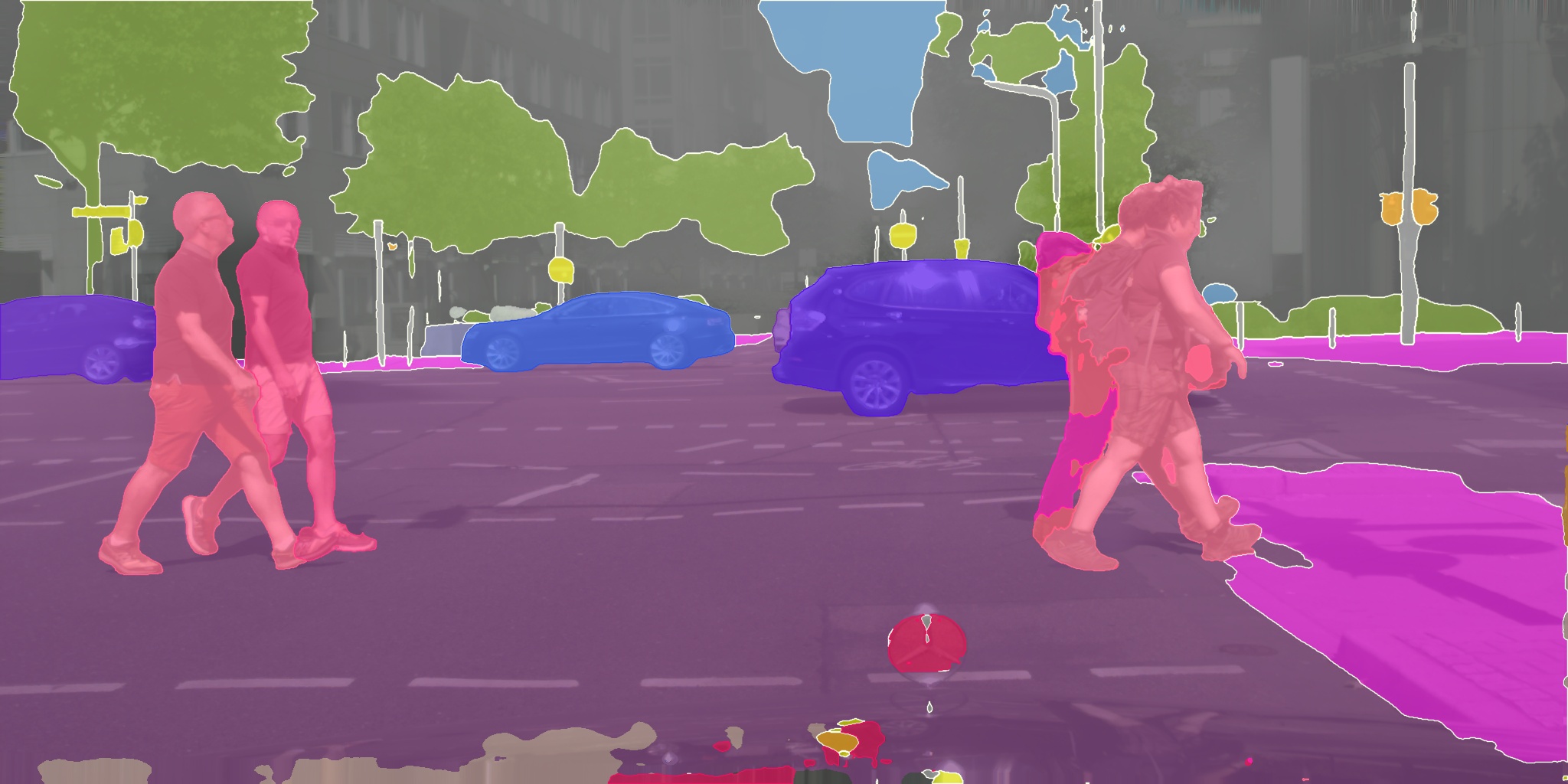}}
\hfill
\\
\mpage{0.22}{\includegraphics[width=1.0\linewidth]{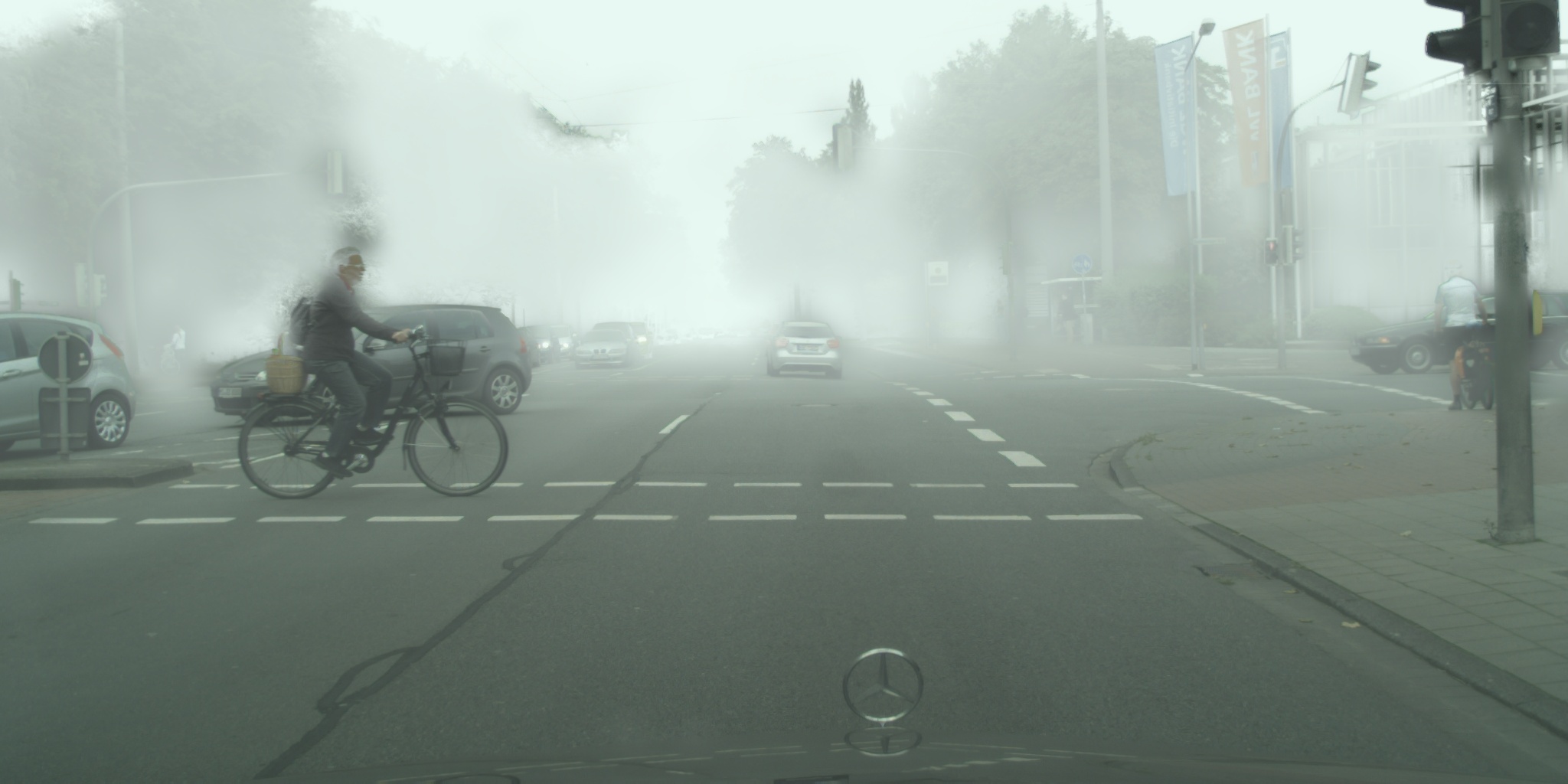}}
\hfill
\mpage{0.22}{\includegraphics[width=1.0\linewidth]{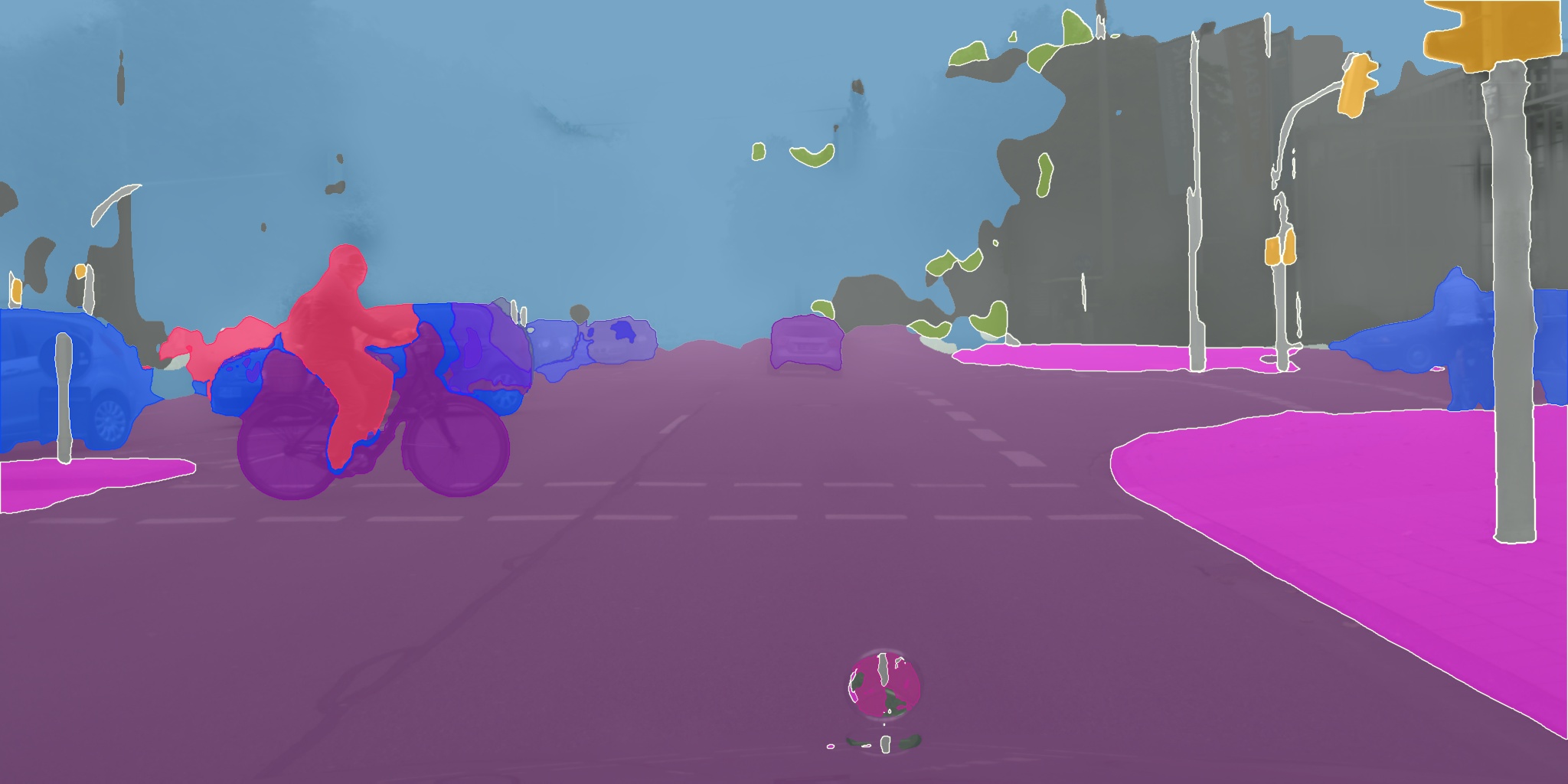}}
\hfill
\mpage{0.22}{\includegraphics[width=1.0\linewidth]{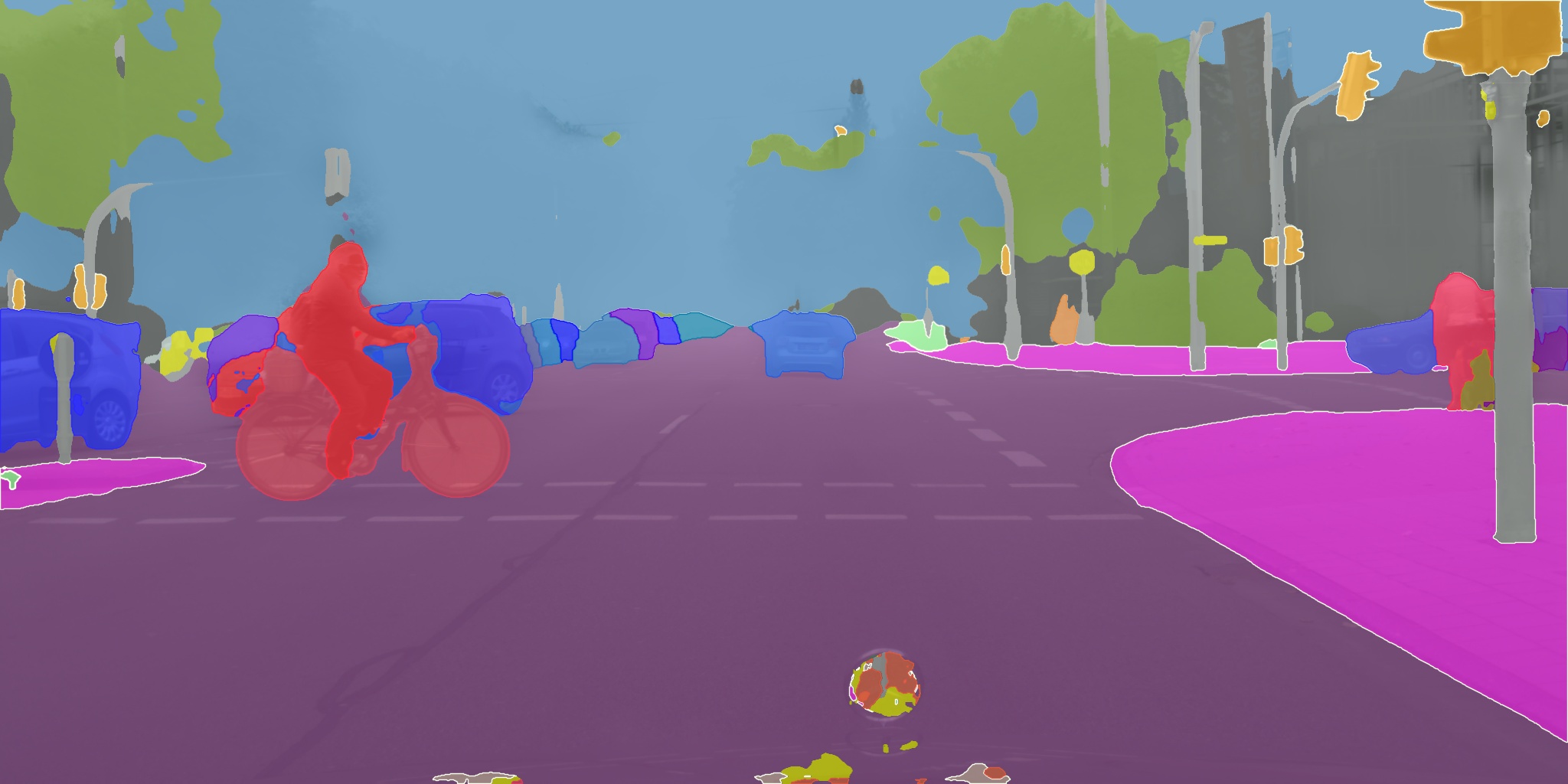}}
\hfill
\mpage{0.22}{\includegraphics[width=1.0\linewidth]{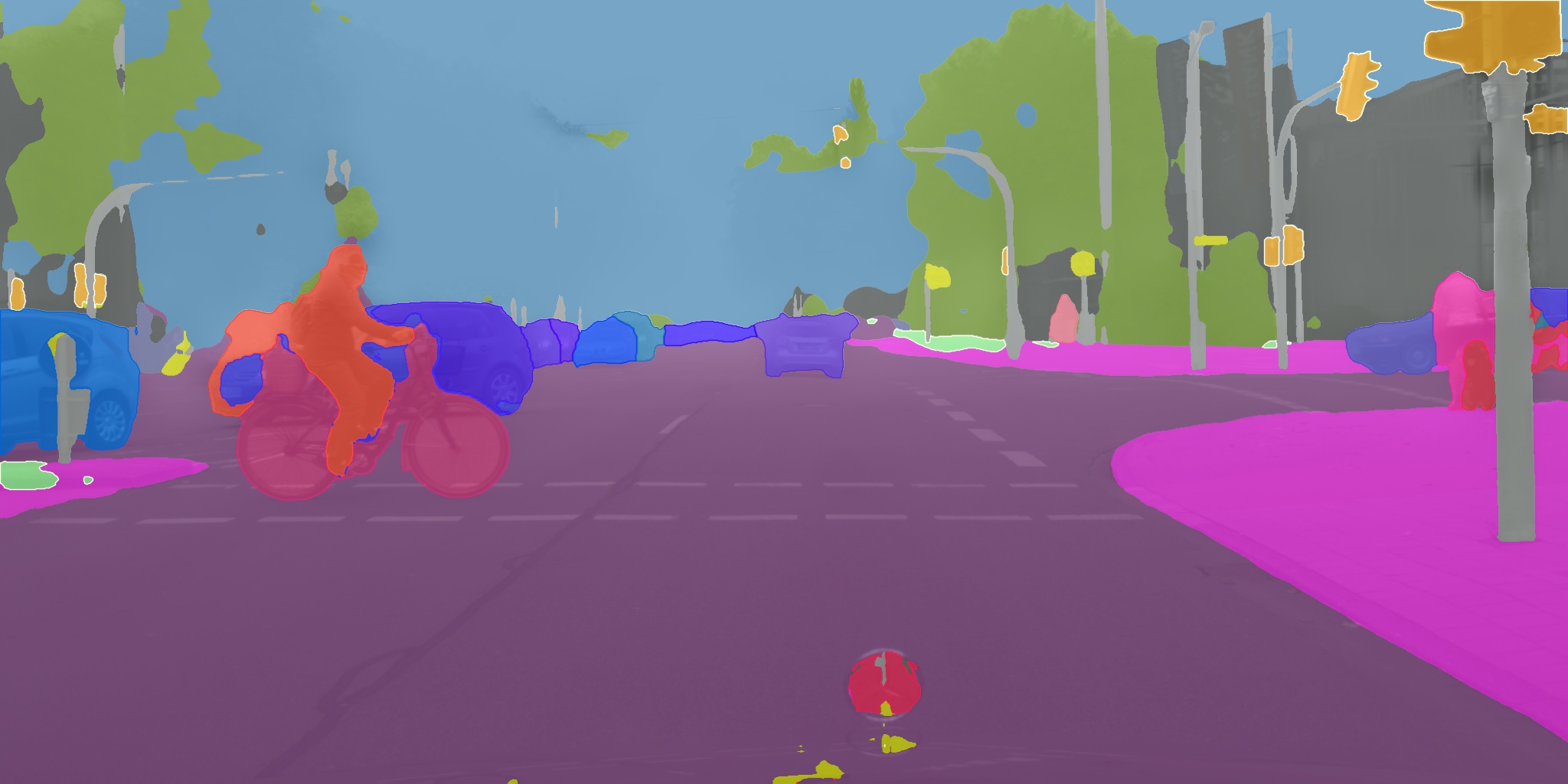}}
\hfill
\\
\mpage{0.22}{\includegraphics[width=1.0\linewidth]{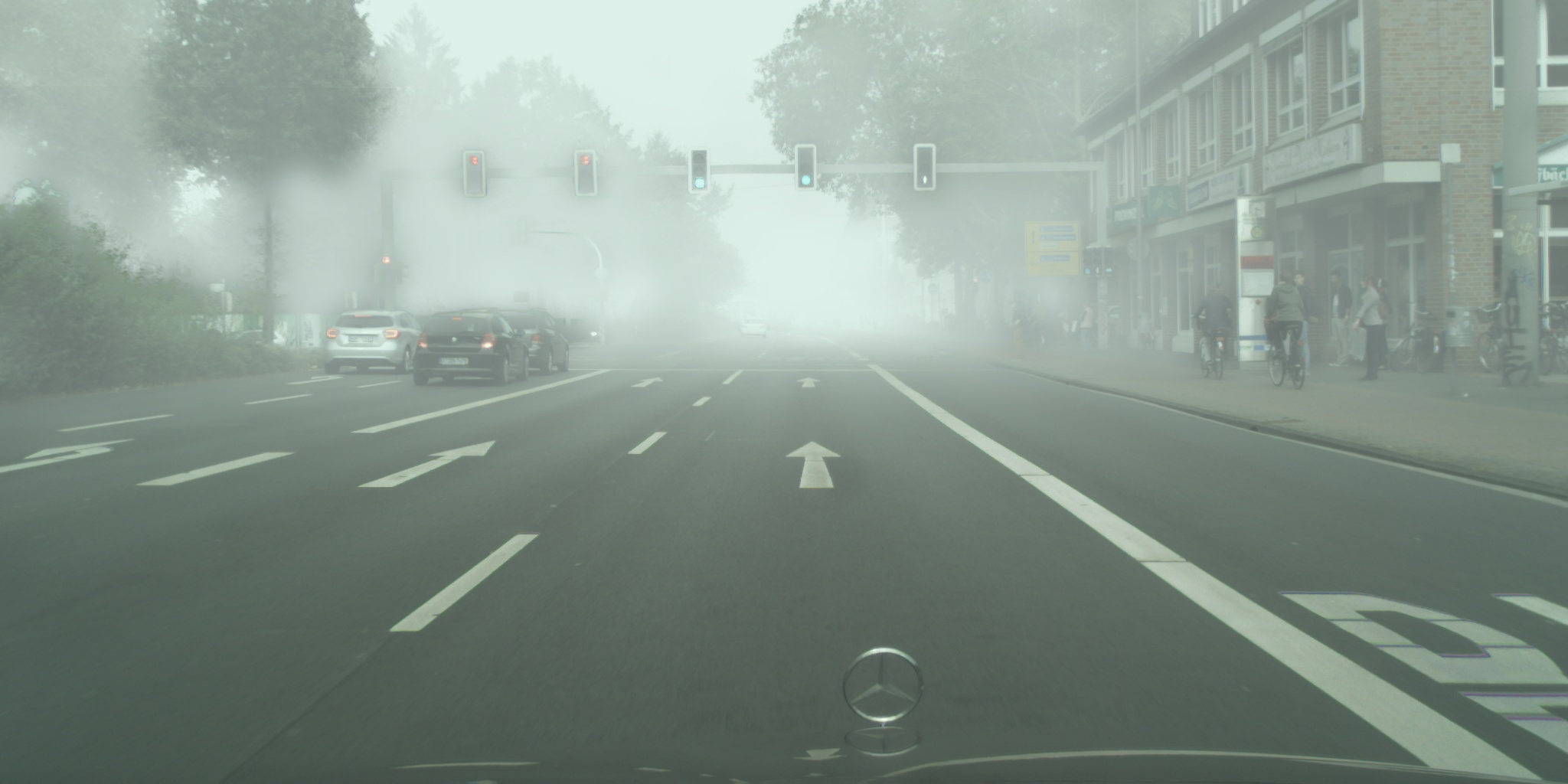}}
\hfill
\mpage{0.22}{\includegraphics[width=1.0\linewidth]{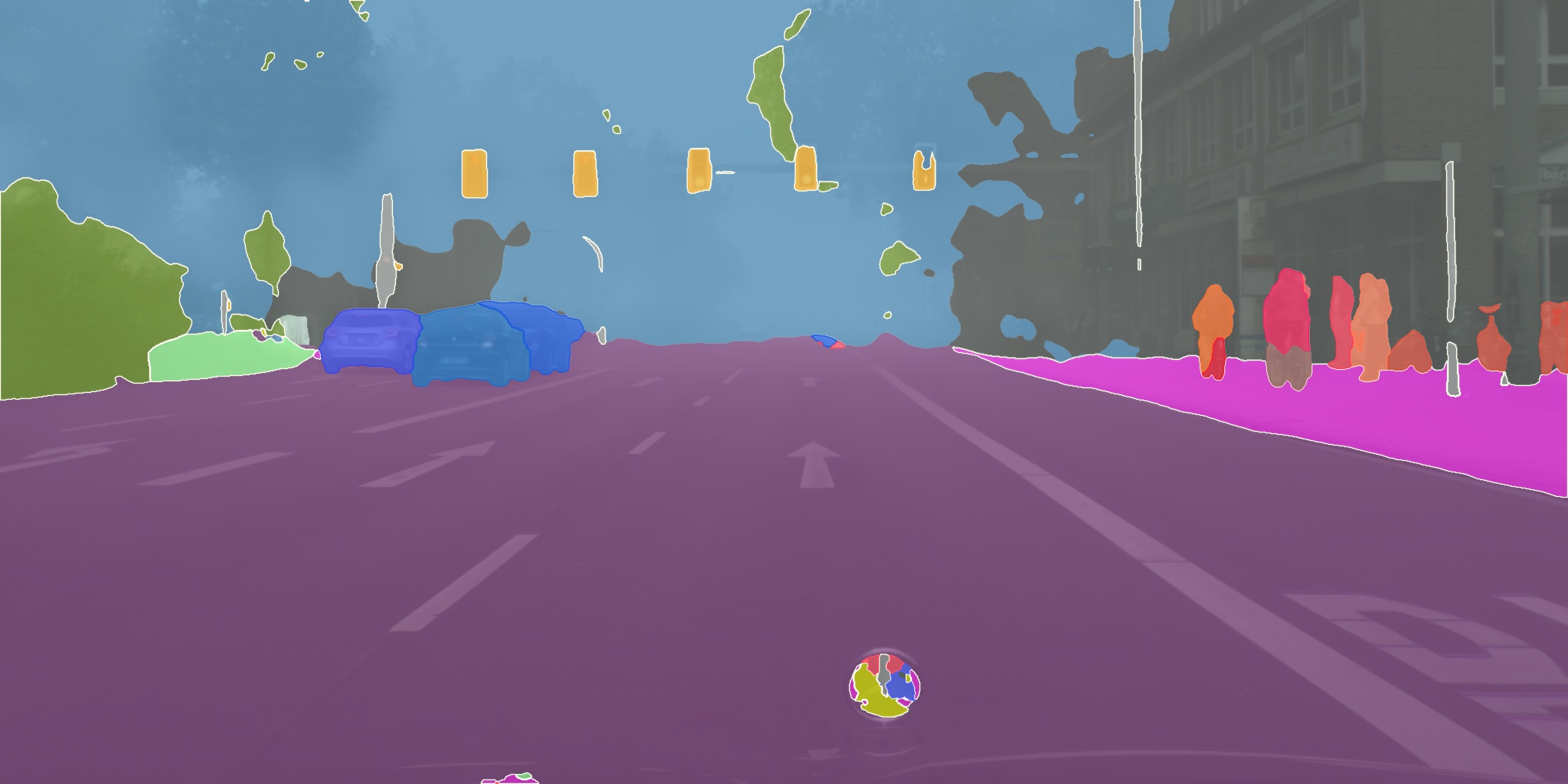}}
\hfill
\mpage{0.22}{\includegraphics[width=1.0\linewidth]{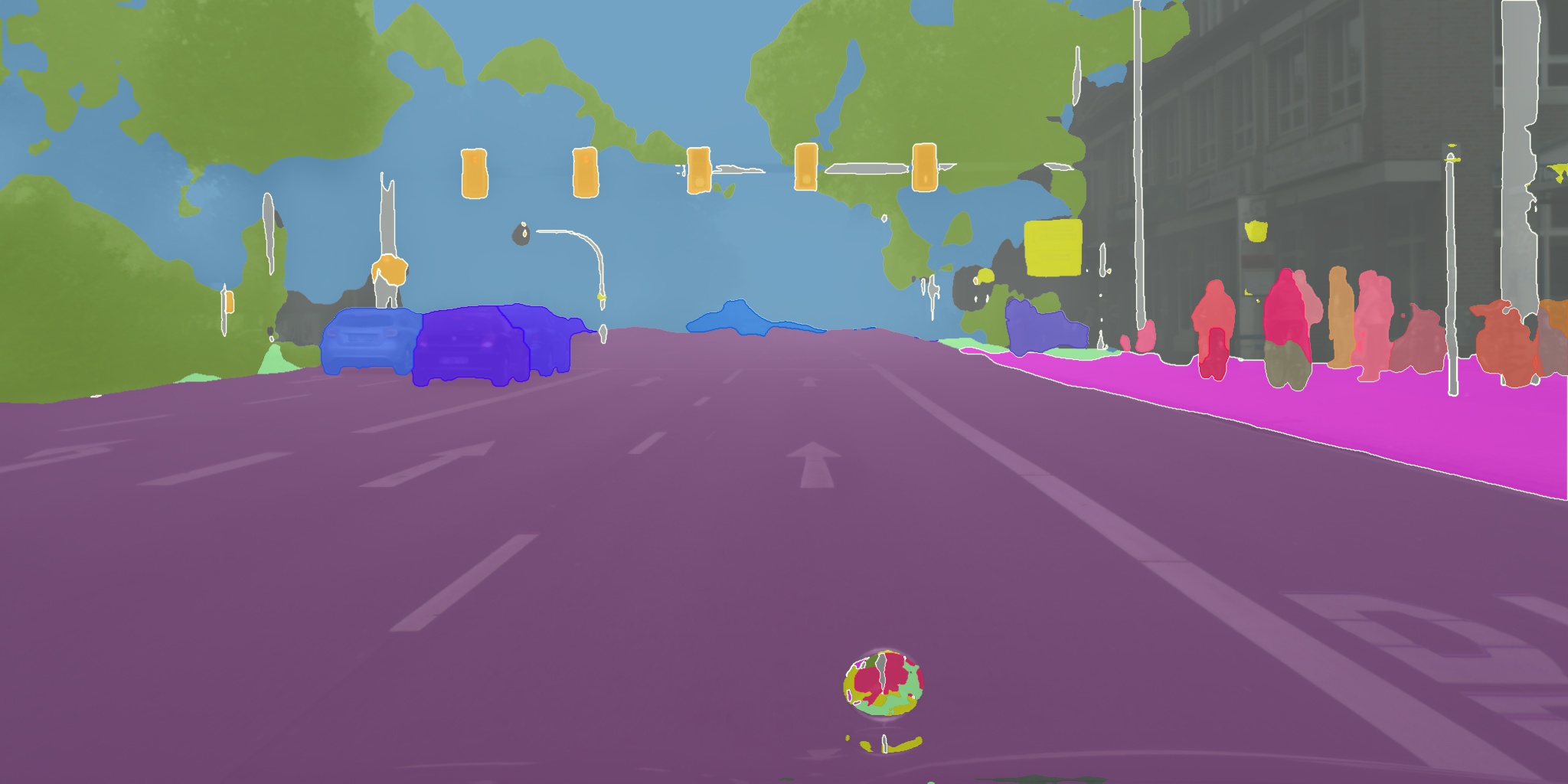}}
\hfill
\mpage{0.22}{\includegraphics[width=1.0\linewidth]{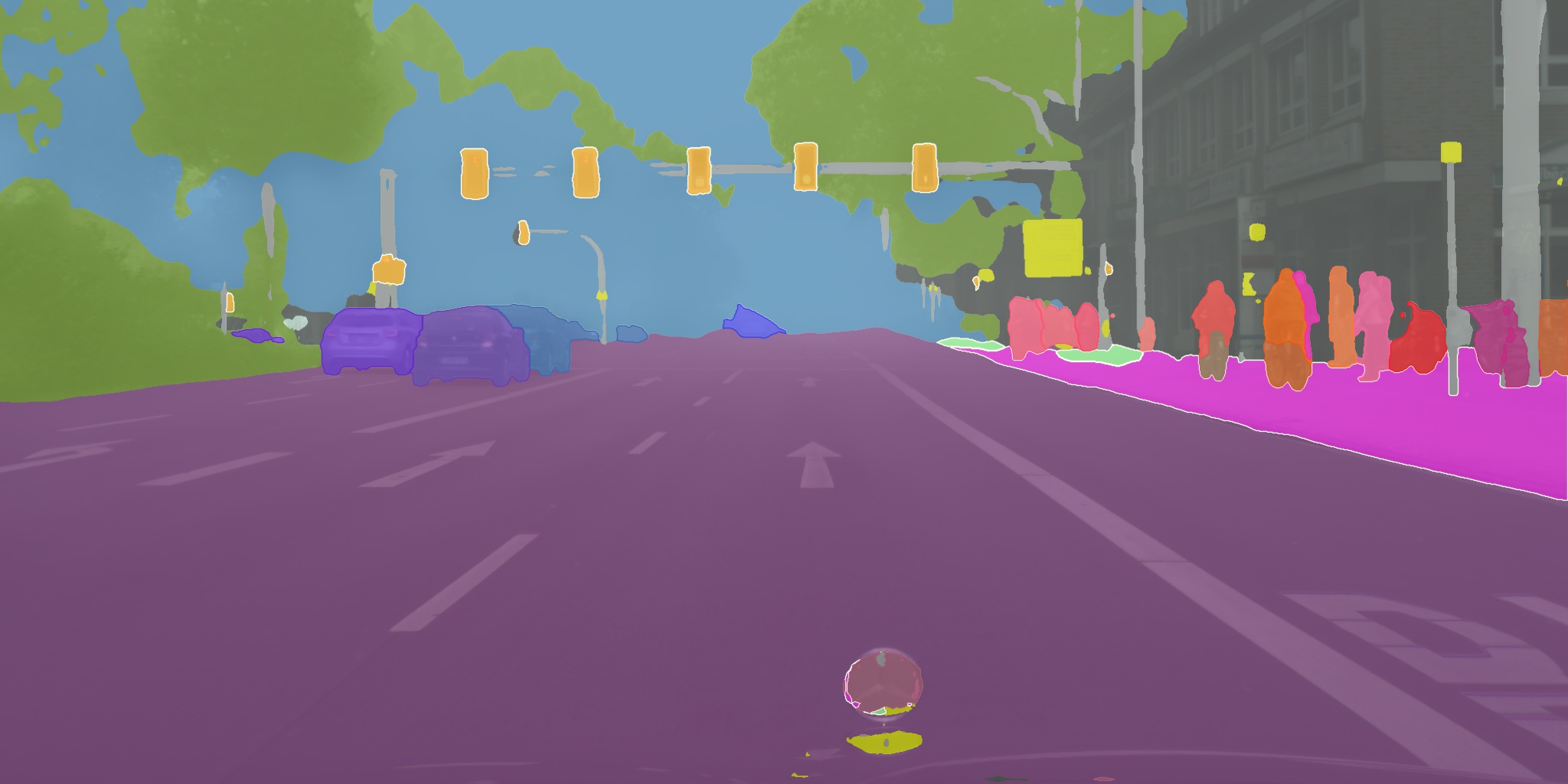}}
\hfill
\\
\mpage{0.22}{\includegraphics[width=1.0\linewidth]{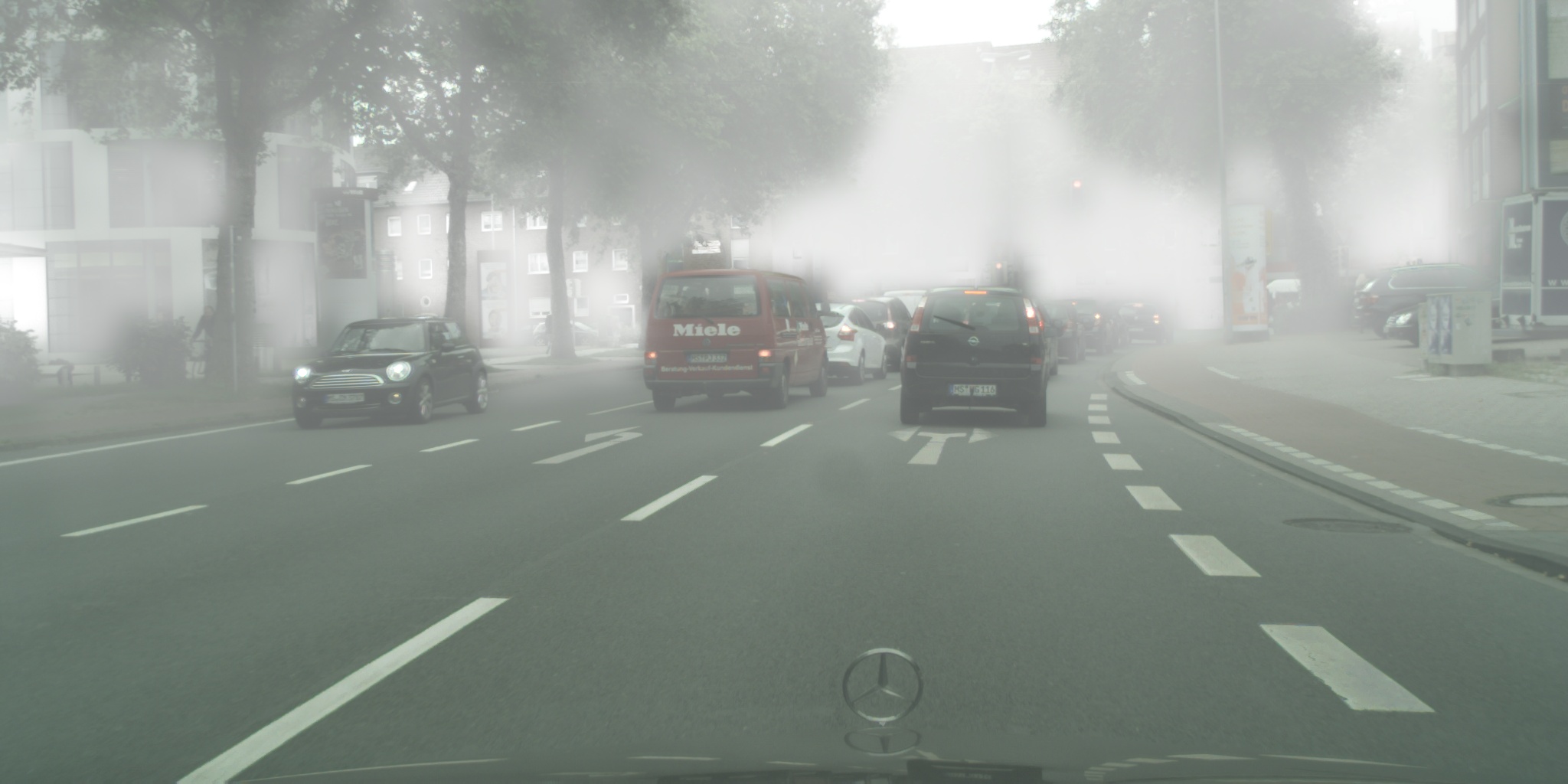}}
\hfill
\mpage{0.22}{\includegraphics[width=1.0\linewidth]{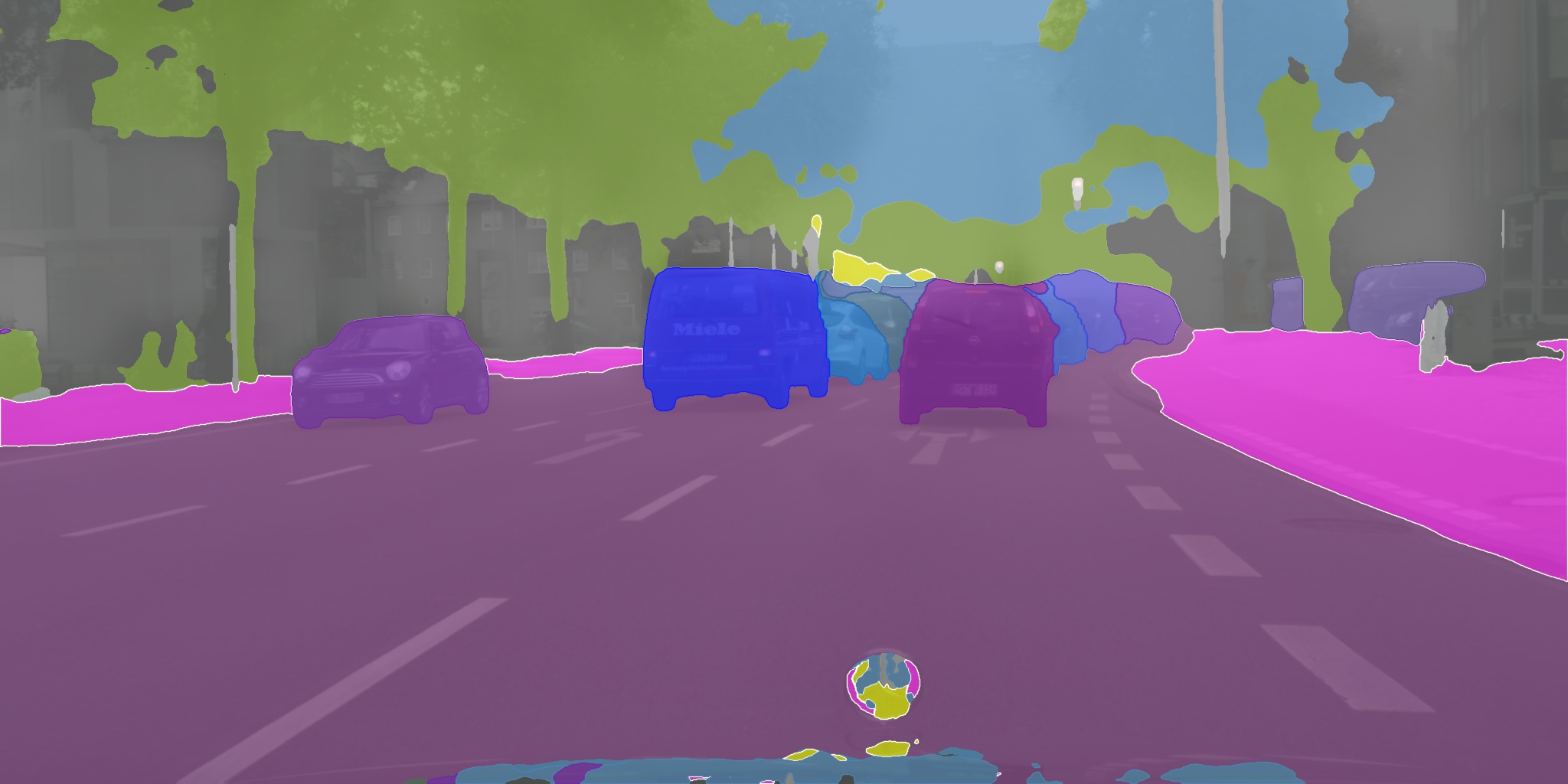}}
\hfill
\mpage{0.22}{\includegraphics[width=1.0\linewidth]{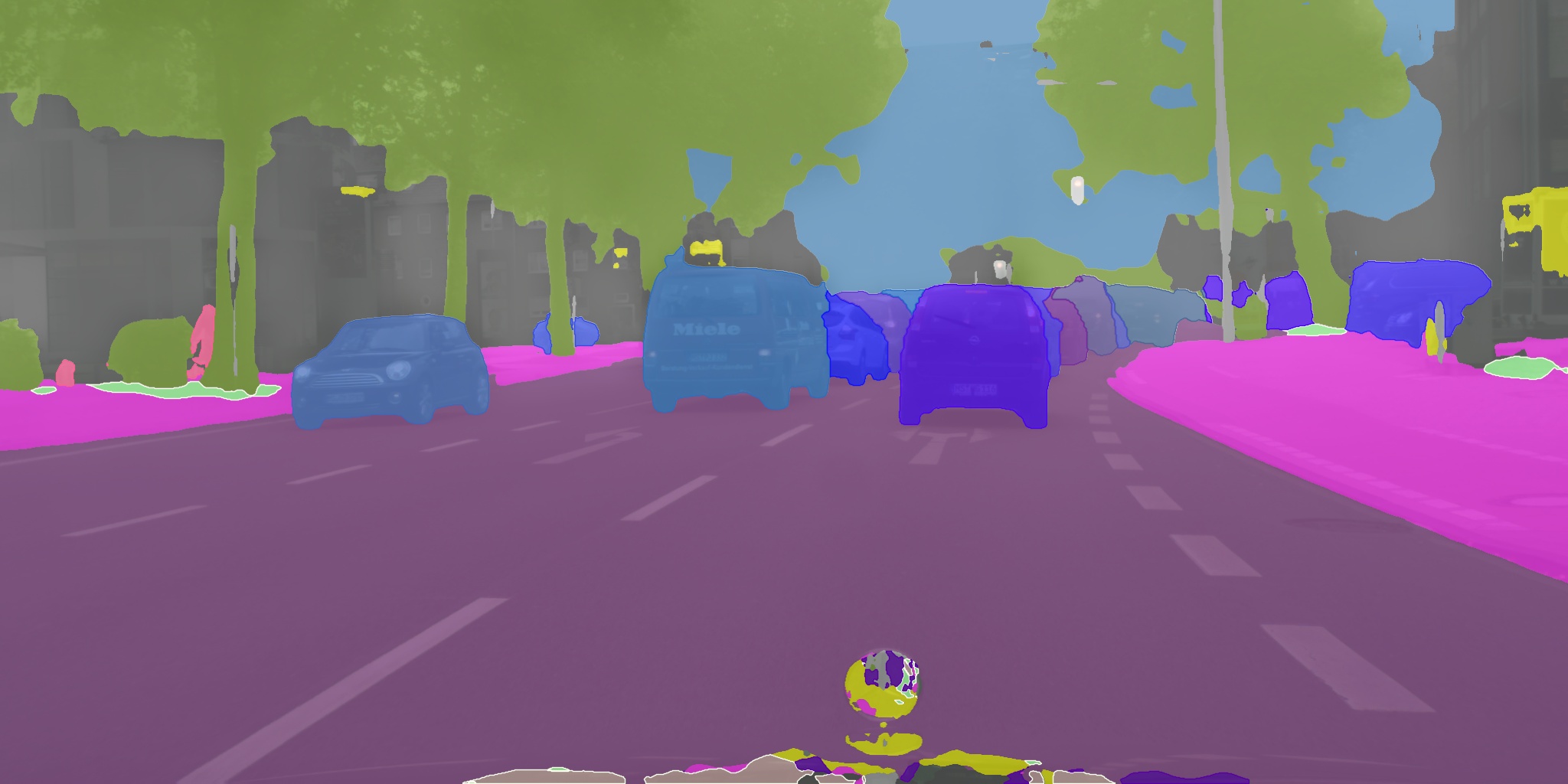}}
\hfill
\mpage{0.22}{\includegraphics[width=1.0\linewidth]{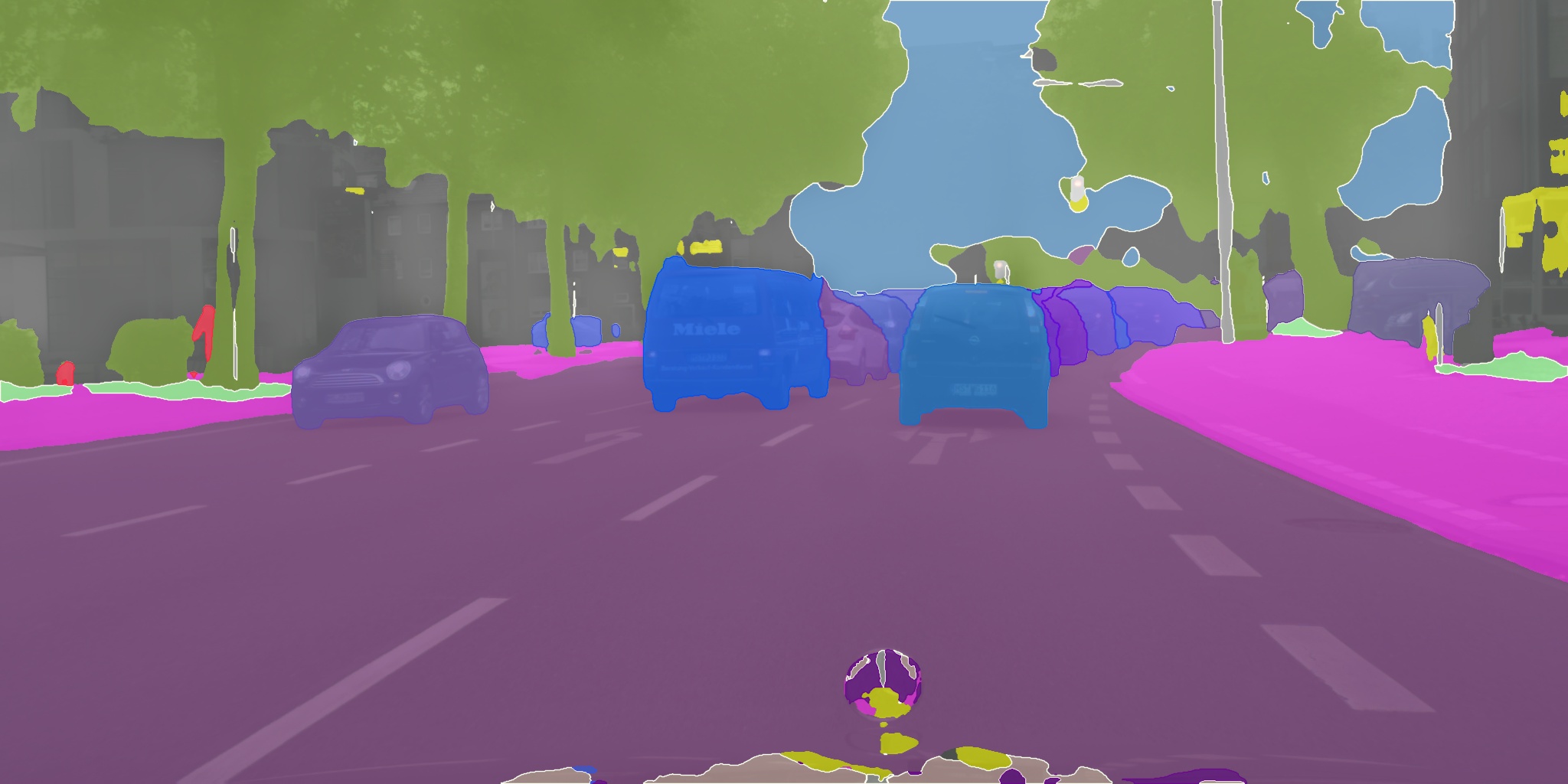}}
\hfill
\\
\mpage{0.22}{\includegraphics[width=1.0\linewidth]{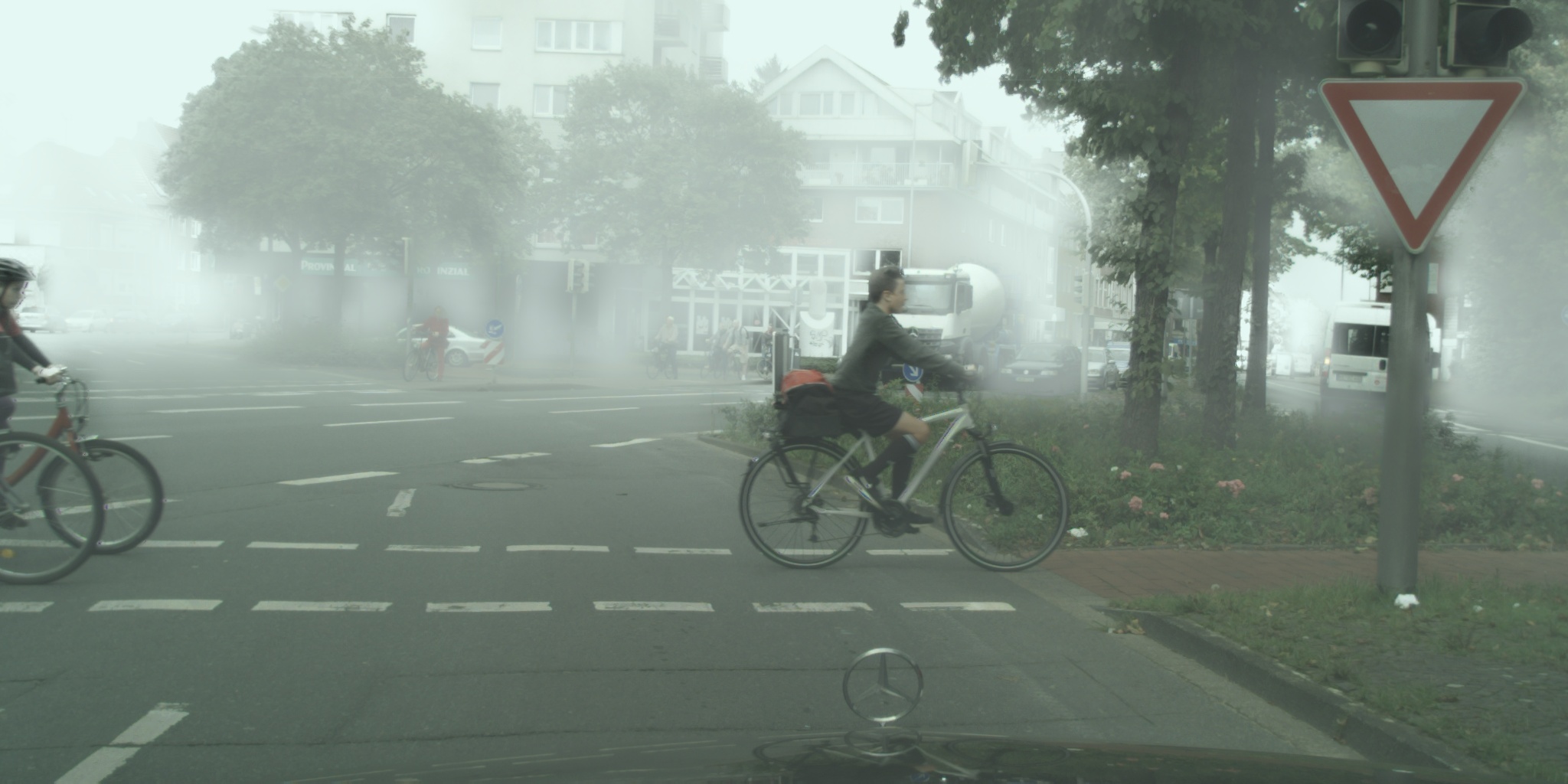}}
\hfill
\mpage{0.22}{\includegraphics[width=1.0\linewidth]{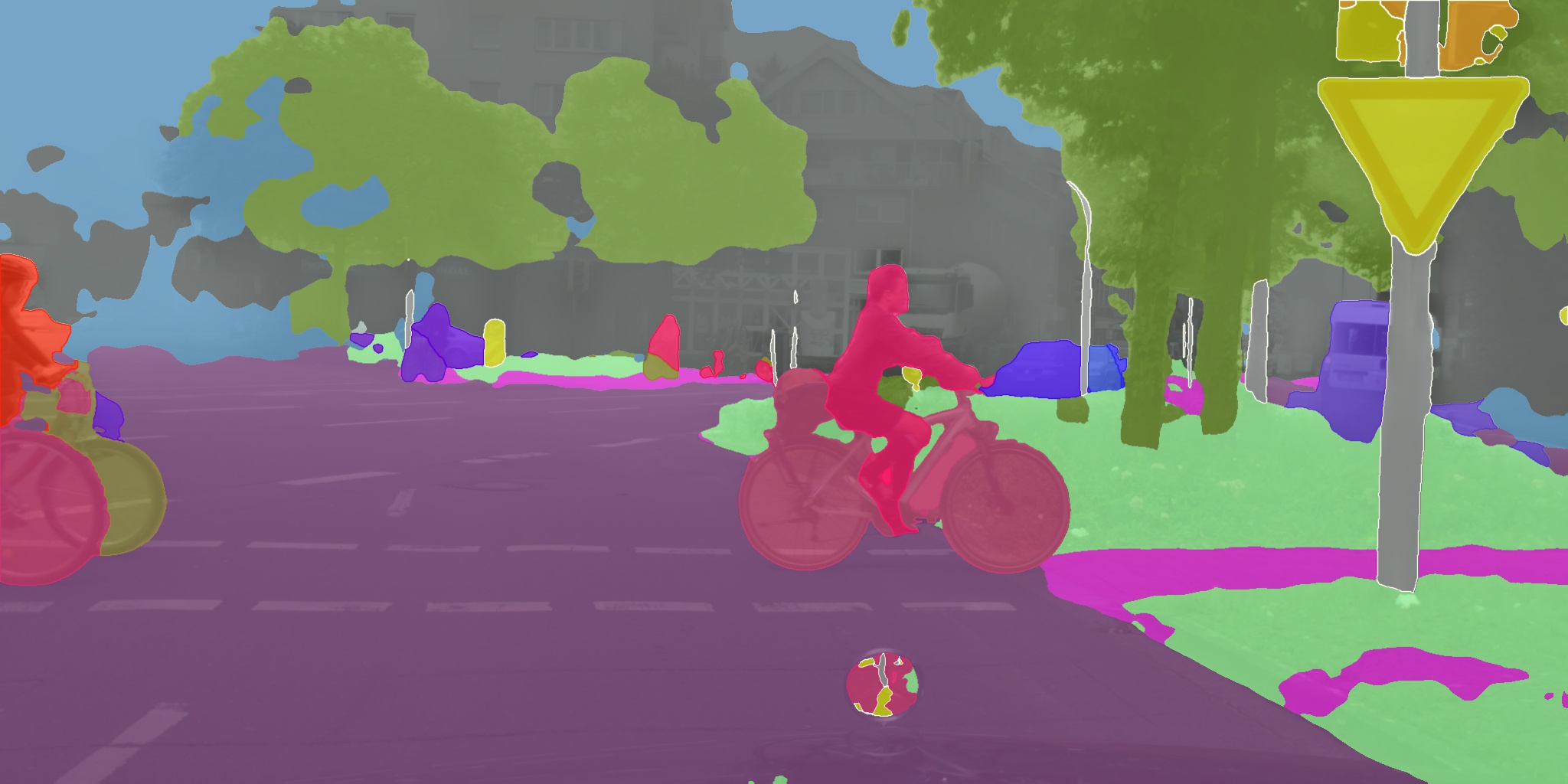}}
\hfill
\mpage{0.22}{\includegraphics[width=1.0\linewidth]{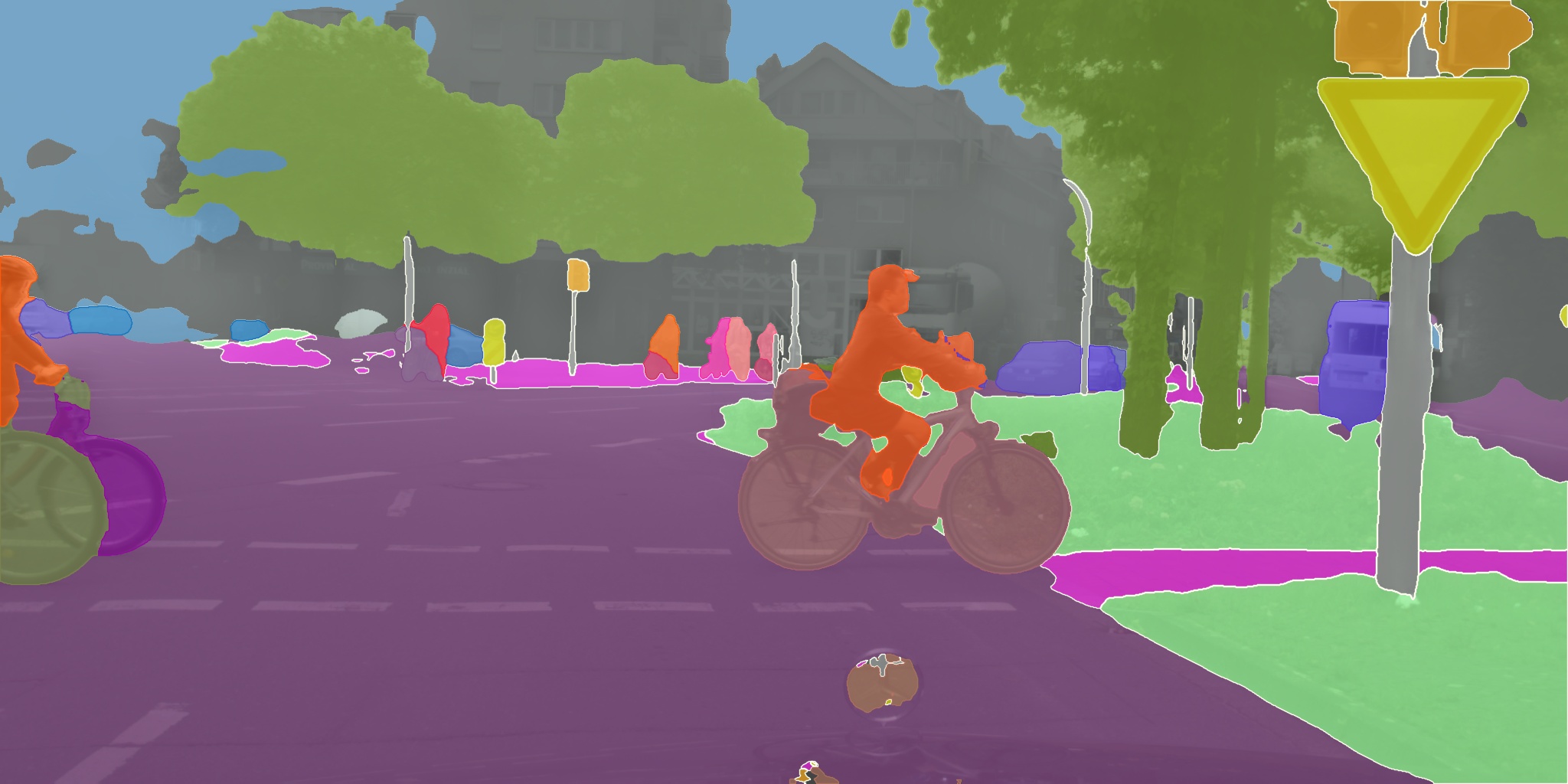}}
\hfill
\mpage{0.22}{\includegraphics[width=1.0\linewidth]{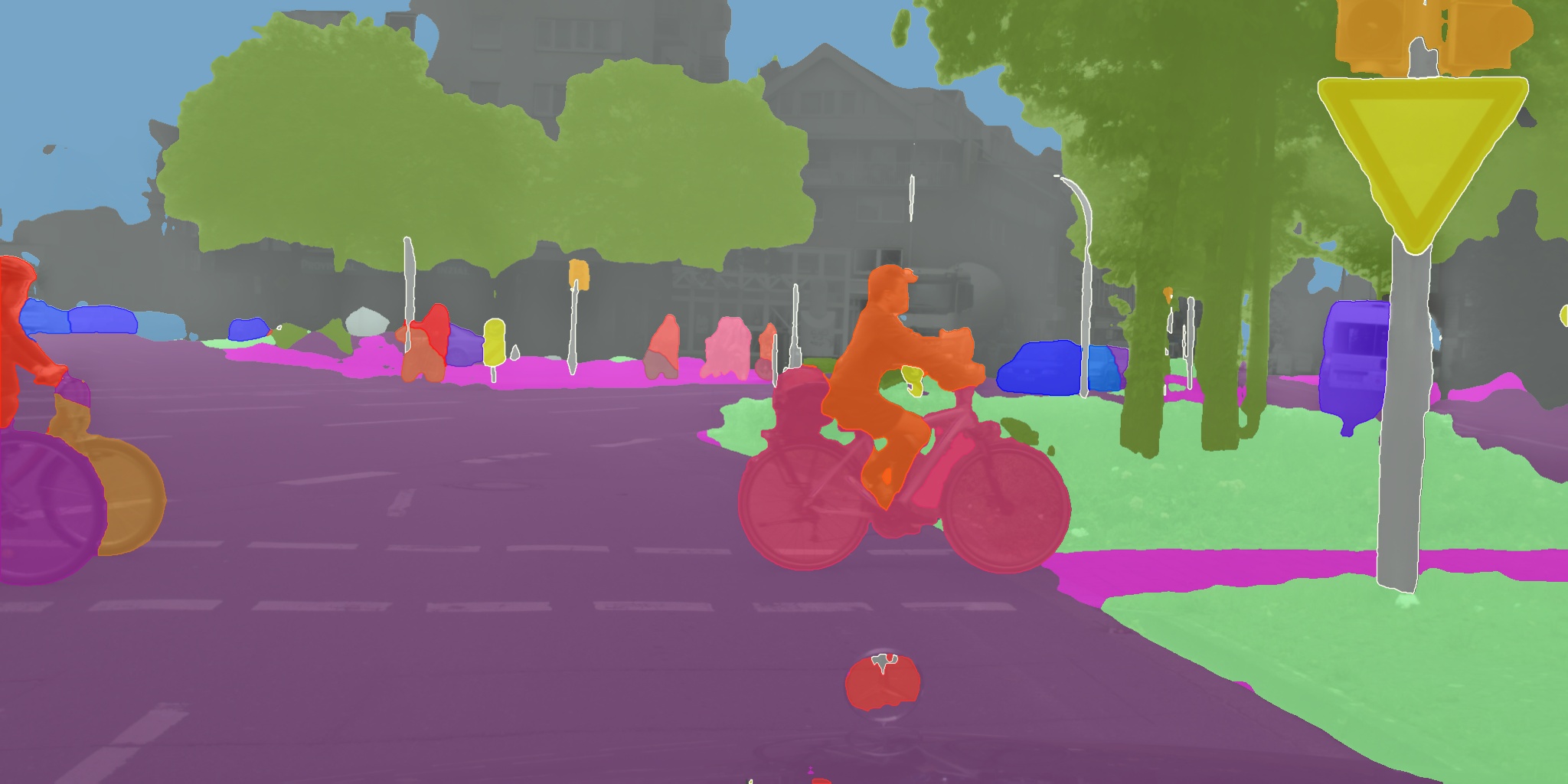}}
\hfill
\\
\mpage{0.22}{Input}
\hfill
\mpage{0.22}{Pre-trained}
\hfill
\mpage{0.22}{InstCal-U}
\hfill
\mpage{0.22}{InstCal-C}
\hfill

\figcapmargin
\figcaption{Qualitative results of panoptic segmentation}
{Models are trained on the clean Cityscapes dataset and tested on the Foggy Cityscapes dataset.
}
\label{fig:supp_pano}
\end{figure*}

We visualize model prediction results in \figref{supp_semantic} and \figref{supp_pano}.
For cross-domain semantic segmentation, the proposed method consistently improves over different settings.
For cross-domain panoptic segmentation, we can see that the proposed method dramatically improves the background segmentation, such as trees.


\end{document}